\documentclass[a4paper, fleqn]{cas-dc}

\usepackage{makecell}
\usepackage[authoryear]{natbib}
\usepackage{amsmath}
\usepackage{multicol}
\usepackage{multirow}
\usepackage{longtable}
\usepackage{soul}
\newcommand{\applygreen}[1]{{\leavevmode\color{ForestGreen}[#1]}}
\newcommand{\applyred}[1]{{\leavevmode\color{red}[#1]}}

 \usepackage{pdflscape}
\def\tsc#1{\csdef{#1}{\textsc{\lowercase{#1}}\xspace}}
\tsc{WGM}
\tsc{QE}

\begin{document}
\let\WriteBookmarks\relax
\def\floatpagepagefraction{1}
\def\textpagefraction{.001}

\shorttitle{GPT-3.5, GPT-4, or BARD? Evaluating LLMs Reasoning Ability in Zero-Shot Setting}

\title [mode = title]{GPT-3.5, GPT-4, or BARD? Evaluating LLMs Reasoning Ability in Zero-Shot Setting and Performance Boosting Through Prompts}

\tnotemark[1] 


%

\author[1]{{Jessica} {López Espejel}}[
    orcid=0000-0001-6285-0770
]


\ead{jlopezespejel@novelis.io}



\affiliation[1]{organization={Novelis
Research and Innovation Lab},
            addressline={207 Rue de Bercy}, 
            city={Paris},
            postcode={75012}, 
            state={},
            country={France}}

\credit{}


\author[1]{El Hassane Ettifouri}[
    orcid=0000-0001-5299-9053
]
\cormark[1]
\ead{eettifouri@novelis.io}


\credit{}


\cortext[1]{Corresponding author}

\author[1]{{Mahaman Sanoussi} {Yahaya Alassan}}[]


\ead{syahaya@novelis.io}


\credit{}


\author[1]{El Mehdi Chouham}[]

\ead{elchouham@novelis.io}


\credit{}

\author[1]{Walid Dahhane}[orcid=0000-0001-5387-3380]

\ead{wdahhane@novelis.io}


\fntext[1]{}

\begin{abstract}
     Large Language Models (LLMs) have exhibited remarkable performance on various Natural Language Processing (NLP) tasks. However,  there is a current hot debate regarding their reasoning capacity. In this paper, we examine the performance of GPT-3.5, GPT-4, and BARD models, by performing a thorough technical evaluation on different reasoning tasks across eleven distinct datasets. Our paper provides empirical evidence showcasing the superior performance of ChatGPT-4 in comparison to both ChatGPT-3.5 and BARD in zero-shot setting throughout almost all evaluated tasks. While the superiority of GPT-4 compared to GPT-3.5 might be explained by its larger size and NLP efficiency, this was not evident for BARD. We also demonstrate that the three models show limited proficiency in Inductive, Mathematical, and Multi-hop Reasoning Tasks. 
     To bolster our findings, we present a detailed and comprehensive analysis of the results from these three models. Furthermore, we propose a set of engineered prompts that enhances the zero-shot setting performance of all three models. 
     
\end{abstract}


\begin{graphicalabstract}
\includegraphics[width=\textwidth,trim={0cm 8cm 0cm 1cm}]{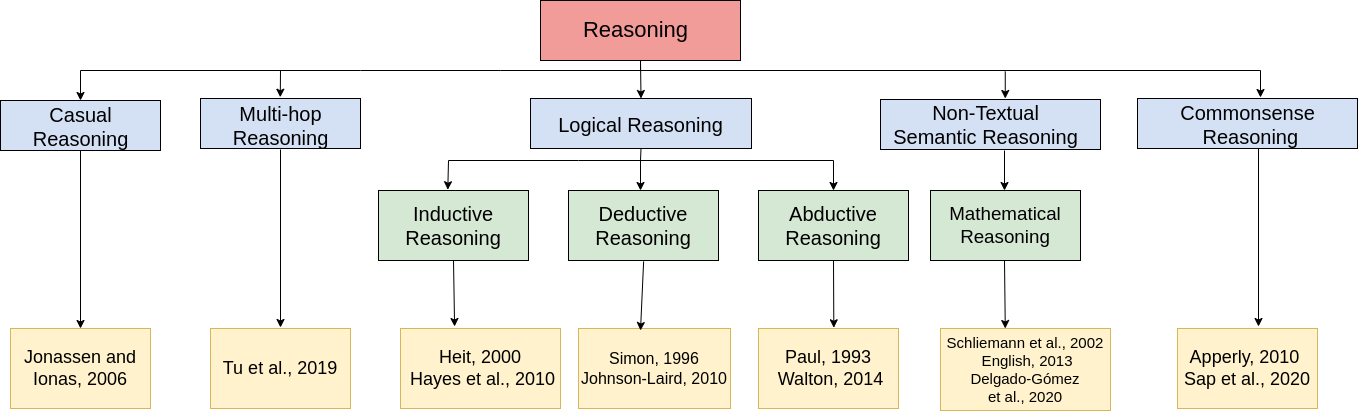}
\end{graphicalabstract}

\begin{highlights}

   \item
    We manually evaluated the reasoning abilities of both ChatGPT models and Google's BARD by conducting evaluations covering a range of reasoning domains, such as deductive, inductive, abductive, analogical, causal, and multi-hop reasoning, through question-answering tasks.
    \item
    We propose new engineered prompts to improve the performance of LLMs in zero-shot-setting. We empirically prove the effectiveness of our approach.
    \item
    We make our samples publicly available and our test suite completely reproducible on ChatGPT-3.5, ChatGPT-4 and BARD. 
    
    
\end{highlights}


\ExplSyntaxOn
\keys_set:nn { stm / mktitle } { nologo }
\ExplSyntaxOff

\maketitle

\section{Introduction}
\label{sec: introduction}

    Reasoning is the process of logically and systematically thinking about a subject~\citep{Huang2023_TowardsReasoning}, leveraging past experiences and evidence to arrive at a conclusion or make a decision. Reasoning is a remarkable hallmark of human intelligence.  It is through reasoning that we continually expand our understanding of the world and unlock the potential for innovation and progress.
    
    Over time, artificial intelligence (AI) has made significant strides in bridging the gap between human intelligence and machine intelligence. In the field of Natural Language Processing (NLP), Large Language Models (LLMs) have demonstrated a wide array of reasoning capabilities, encompassing mathematical \citep{lewkowycz2022solving}, symbolic \citep{khot2023_decomposed}, and various other forms of reasoning. \citet{wei2022_emergent} argue that LLMs possess reasoning capabilities when they are sufficiently large (having over than 100B parameters). This claim is supported by \citet{Liu2023_pretrain_promp}, \citet{wei2022_chain-of-thought} and \citet{Kojima2022_LLM_Zero-Shot}. Particularly,  \citet{wei2022_chain-of-thought} proposed a technique called Chain-of-Thought (CoT), which involves producing intermediate steps prior to presenting the final answer. CoT has enhanced the capacity of LLMs to effectively address complex problems that require reasoning. \citet{Liu2023_pretrain_promp} affirm that prompting strategies have the potential to unlock reasoning abilities. For example, \citet{Kojima2022_LLM_Zero-Shot}, demonstrated through a simple prompt (mathematical problem + question + the phrase \textit{"Let's think step by step"}) that LLMs are capable zero-shot reasoners. However, other researchers such as \citet{mahowald2023_dissociating} proclaim that "language ability is not equal to thinking or reasoning" in LLMs. Similarly, \citet{Valmeekam2022_LLMsCannotPlan} contend that LLMs do not have acceptable performance in planning and reasoning tasks. 
    
    As \citet{Bang_MultitaskChatgpt2023} mentioned, assessing a model's reasoning skills involves evaluating it on diverse sets of abilities such as commonsense \citep{talmor2019_commonsenseqa}, inductive \citep{sinha2019-clutrr}, and deductive \citep{Weston2016_bAbI} reasoning. This kind of evaluation has been carried out by the new wave of LLMs such as PALM \citep{Driess2023_PaLM}, Chinchilla \citep{Hoffmann2022_chinchilla}, LaMDA \citep{Thoppilan2022_LaMDA}, LLaMA \citep{touvron2023_llama}, and Gopher \citep{Jack2021_Gopher}. For instance, PALM assessed its ability to reason through arithmetic and commonsense tasks. To address the challenging issue of multi-step reasoning, PALM adopted the CoT strategy. This particular issue is widely experienced by numerous LLMs, as also highlighted by the Gopher model. Similarly, LLaMA and Chinchilla were evaluated on common sense reasoning tasks, but instead of CoT, their authors opted for a different approach by employing the zero-shot setting.  Note that LLaMA was also evaluated on mathematical reasoning.

     Released in late 2022 by OpenAI \citep{OpenIA}, ChatGPT-3.5 \footnote{or simply called ChatGPT} \citep{chatGPT3.5} is a chatbot based on GPT-3.5 model \citep{radford2018improving, Radford2018_LLMMultitask, Ouyang2022TrainingLM}. It is considered as one of the most revolutionary LLMs because it has made a profound impression on the world with its remarkable ability to engage in coherent interactions with users. Unfortunately, ChatGPT-3.5 faces significant challenges in certain areas, such as mathematics~\citep{Davis2023_Mathematics,Frieder2023_MathematicalChatGPT}, and commonsense tasks~\citep{guo2023_closeChatGPT}. To address this issue, OpenAI developed ChatGPT-4, an enhanced version of the chatbot that is based on GPT-4 model. While ChatGPT-3.5 is built upon InstructGPT \citep{Ouyang2022TrainingLM}, ChatGPT-4 is a large language model that employs fine-tuning and Reinforcement Learning from Human Feedback (RLHF) \citep{Christiano2017_RL}. It is a robust multimodal model that can process both textual and visual inputs and generate text-based outputs.

    After the successful release of ChatGPT-3.5, Google took further strides in the AI landscape by introducing their innovative large language model, called BARD \citep{manyika2023_BARD}. Through an intensive process of gathering human feedback and rigorous evaluation, Google refined and optimized BARD to create a lightweight version of LaMDA. Currently, BARD remains in the experimental stage, yet it has already made a remarkable impact. Users can access it through an intuitive interface, enabling direct collaboration with a generative AI system.

    This paper aims to investigate the reasoning capabilities of ChatGPT-3.5, BARD and ChatGPT-4. We notably build on previous work from the state of the art \citep{Bang_MultitaskChatgpt2023} and propose the following contributions:

    \begin{itemize}
       \item
        We manually evaluated the reasoning abilities of both ChatGPT models, and Google's BARD by conducting evaluations covering a range of reasoning domains, such as deductive, inductive, abductive, analogical, causal, and multi-hop reasoning, through question-answering tasks.
        \item
        We propose new engineered prompts to improve the performance of LLMs in zero-shot-setting. We empirically prove the effectiveness of our approach.
        \item
        Unlike \citet{Bang_MultitaskChatgpt2023} who did not provide enough details about their experimental protocol, nor the prompts used in their evaluation, we make sure in this paper to make our samples publicly available (Appendix \ref{sec:appendix}) and our test suite completely reproducible on the three evaluated models. This will ensure a fair experimentation and will be helpful for the research community to reproduce our results and further improve the domain.
    \end{itemize}

    The rest of the paper is organized as follows: in Section \ref{sec:related_work}, we present the most related work from literature. In Section \ref{sec:types_of_reasoning}, we detail the types of reasoning and  provide examples to help understand each one of them. In Section \ref{sec:datasets}, we introduce the datasets used to evaluate different reasoning tasks. In Section \ref{sec:expeimennts}, we present the set of experiments we conducted, from data selection to proposed prompts. In Section \ref{sec:results}, we discuss obtained results and we finally conclude in Section \ref{sec:Conclusions} and provide some future directions.

\section{Related Work}
\label{sec:related_work}

    There is ongoing debate among experts regarding the ability of Large Language Models (LLMs) to reason and plan effectively in zero-shot scenarios. While some authors, such as \citet{Kojima2022_LLM_Zero-Shot}, contend that LLMs exhibit satisfactory zero-shot reasoning capabilities, others like \citet{Valmeekam2022_LLMsCannotPlan} argue that LLMs are inadequate when it comes to planning and reasoning tasks. However, recent research suggests that certain models, including GPT-3~\citep{Brown2020_LLMs-ZeroShot_GPT3} and PaLM~\citep{Chowdhery2022_PaLM}, demonstrate efficient reasoning skills in some contexts. It is worth noting that the most powerful LLMs, those with over 100 billion parameters~\citep{wei2022_emergent}, appear to be the ones with the most impressive reasoning abilities, capable of solving a vast array of problems accurately. 
    
    One of the most widely recognized techniques designed to encourage LLMs to engage in explicit reasoning is Chain-of-Thought (CoT), proposed by~\citet{wei2022_chain-of-thought}. This method involves replacing the traditional $<input, output>$ prompt format with $<input, CoT, output>$, where the CoT represents a chain of thought demonstrations. CoT has demonstrated superior performance in zero-few shot reasoning tasks, such as arithmetic and commonsense reasoning, among others. Since then, several variants of CoT have emerged. For example,~\citet{Kojima2022_LLM_Zero-Shot} proposed a prompt that combines CoT with the phrase \textit{``Let's think step by step''}, and this simple idea achieved state-of-the-art results in many NLP tasks that require reasoning abilities, like translation, question-answering, and cloze tasks. Another well-known variant is iterative CoT, as demonstrated in~\citet{wang2023_selfconsistency}. A novel approach called "prompt-based learning" has emerged from previous studies. This approach has been investigated by \citet{Liu2023_pretrain_promp}, and enables the model to create a customized prompting function, facilitating few-shot or zero-shot setting and enabling adaptation to new scenarios. 
    
    An additional approach to encourage LLMs to engage in reasoning is rationale engineering. While CoT relies on manually crafting intermediate reasoning steps, rationale engineering, through techniques such as rationale refinement \citep{liu2022_makesGoodExamples}, rationale exploration, and rationale verification, aim to create or improve CoT to achieve better performance in LLMs. On the one hand, rationale refinement aims at producing more effective examples of reasoning steps. Several methods have been derived, such as complexity-based prompting~\citep{Fu2022_ComplexityBasedPF}, algorithmic prompting~\citep{zhou2023_teachingAlgorithmic}, and Auto-CoT~\citep{zhang2023automaticCoT}. On the another hand, rationale exploration enables LLMs to comprehensively explore various modes of reasoning to enhance their proficiency in reasoning-based assignments. One approach to achieving this is through self-consistency~\citep{wang2023_selfconsistency}, which is a strategy to improve the greedy decoding algorithm. Another technique, proposed by~\citet{Li2022_OnTheAdvance}, involves presenting multiple demonstrations for each question. Finally, the objective of rational verification is to verify whether the reasoning produced by LLMs leads to accurate final solutions. For instance,~\citet{cobbe2021_training} suggest enhancing LLMs with a trained verifier that assigns a score to each justification and solution generated by the LLMs. This technique is also used by \citet{Li2022_OnTheAdvance}, alongside the rationale exploration, to facilitate rationale selection.

    An alternative to the prompting approach is a method known as the hybrid approach. This method combines both prompting techniques and strengthens the reasoning capabilities of LLMs. For instance, \citet{chung2022_scaling} introduced Flan models which involved finetuning PaLM model using CoT data,  while \citet{chung2022_scaling} used bootstrapping method \citep{zelikman2022_star,huang2022_self-improve}. Bootstrapping allows LLMs to enhance their reasoning abilities without relying on fine-tuning. Therefore, LLMs can autonomously improve their reasoning capabilities.

    \textbf{ChatGPT3.5 } - (a.k.a. ChatGPT in its first version) is an AI-based text generator created by OpenAI. It is built on top of the InstructGPT \citep{Ouyang2022TrainingLM} model, which is part of the GPT-3.5 series of models. These models were developed by converting a large-scale instruction-tuning corpus used for InstructGPT into a conversational format. Afterward, the GPT-3 model was fine-tuned using reinforcement learning from human feedback (RLHF) \citep{Christiano2017_RL}, which uses the concept of rewards to improve the generation quality and align the generation with human preference.

    One of the most important merits of InstructGPT model is its ability to follow English instructions better than GPT-3 models. GPT-3.5 series of models are designed to produce safer content by avoiding generating untruthful, toxic, or harmful text. There are three main GPT-3.5 models, namely code-davinci-002, text-davinci-002, and InstructGPT. Code-davinci-002 is a Codex \citep{Chen2021EvaluatingLL} model that has been trained on both programming languages and natural language. InstructGPT is built on code-davinci-002, and text-davinci-002 is the most powerful GPT-3.5 model when compared to previous GPT-3 models. This advanced model is renowned for its exceptional language generation quality and superior instruction-following abilities.

    Several studies have performed experiments to assess the reasoning capabilities of ChatGPT-3.5 across various domains. For instance,~\citet{Bang_MultitaskChatgpt2023} have proposed a framework to assess zero-shot setting on multitask, multilingual and multimodal aspects of ChatGPT-3.5. Their research focused on zero-shot setting and revealed interesting findings regarding ChatGPT-3.5's performance in reasoning tasks. For instance, the model demonstrated a stronger proficiency in deductive reasoning while struggling with inductive reasoning. As well, their study highlighted limitations in ChatGPT-3.5's ability to solve mathematical problems. To delve further into the mathematical reasoning skills of ChatGPT-3.5, \citet{Frieder2023_mathChatGPT} conducted experiments using a dataset called GHOSTS. This dataset consists of natural language prompts for graduate-level mathematics problems. Moreover, \citet{guo2023_closeChatGPT} explored ChatGPT's performance across various domains, including finance, medicine, law, and psychology.


    \textbf{ChatGPT-4} - It is an improved version of ChatGPT-3.5. GPT-4~\citep{chatGPT4} is the most powerful large multimodal model released by OpenAI. This model has the ability to process both textual and visual inputs and generate textual results.  One of the key changes in this model compared to previous GPT models lies in the computation of the loss function. The authors successfully predicted the final loss of GPT-4 in their internal codebase by incorporating a scaling law with an irreducible loss term.

    ChatGPT-4 is built upon the GPT-4 model, which has been trained using an unparalleled amount of computational power and data \citep{Bubeck2023_EarlyExperimentsGPT4}. Although GPT-4 has shown significant improvements compared to earlier GPT models, it still possesses limitations similar to its predecessors. These limitations encompass generating false information, providing harmful advice, having a restricted context window, and lacking the ability to learn from experience. Furthermore, the authors lead a set of assessments of the model's performance using diverse datasets, including academic exams. The results demonstrated that GPT-4 achieved a human-level performance on most of these tests.  

    \textbf{BARD}  is a lightweight and optimized version of LaMDA \citep{Thoppilan2022_LaMDA}, that was pre-trained on publicly available data. Using Reinforcement Learning from Human Feedback (RLHF), BARD improves its predictions token by token. It is  currently in an experimental stage as an online interface. Users provide prompts, and BARD generates multiple response versions using the prompt context and user interaction. The responses are then filtered through safety parameters and ranked based on quality before being provided to the user.

    BARD aims to improve in five  areas: response accuracy, mitigating biases, avoiding personal opinions, addressing false positives and false negatives, and reducing vulnerability to adversarial prompting \citep{manyika2023_BARD}.

\section{Types of reasoning}
\label{sec:types_of_reasoning}

    \begin{figure*}[h!]
        \centering
        \includegraphics[width=\textwidth]{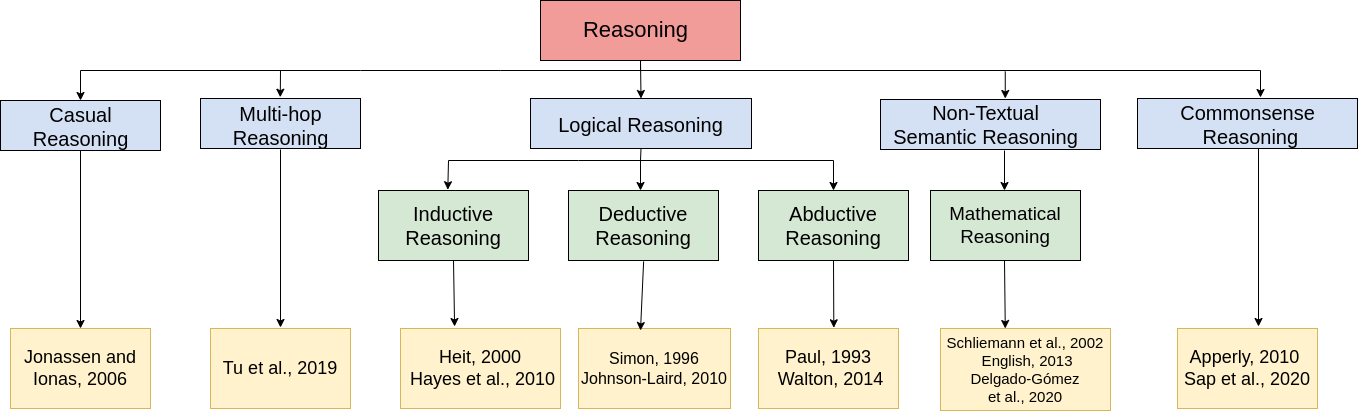}
        \caption{Reasoning types}
        \label{fig:types_of_reasoning}
    \end{figure*}

    There are different types of human reasoning, each with its unique characteristics and applications (Figure \ref{fig:types_of_reasoning}). In this section, we describe these different types of reasoning, following the classification done by \citet{Bang_MultitaskChatgpt2023}: logical reasoning, non-textual semantic reasoning, commonsense reasoning, causal reasoning, and multi-hop reasoning. Understanding  different types of human reasoning is essential to evaluate the reasoning skills of advanced language models like GPT-3.5, GPT-4, and BARD.

    \subsection{Logical reasoning}
    \label{subsec:logical_reasoning}

    Inductive, deductive, and abductive reasoning involve drawing conclusions from evidence or observations, but they have distinct characteristics that set them apart from one another. 

    \paragraph{Inductive reasoning.} This is a bottom-up approach, where using specific observations or evidence we can infer general principles or conclusions \citep{heit2000_propertiesInductive,hayes2010_inductive}. These forecasts inherently involve probabilities. Inductive reasoning allows for the possibility that the conclusion is false, even if the premises are true. This type of reasoning is often used in scientific research, where observations are made, patterns are identified, and generalizations are made from those patterns. Example: 
    
    \begin{itemize}
        \item []
        \textsc{Observation 1:} When I was a kid, I ate shrimps, and I experienced an allergic reaction.
        \item []
        \textsc{Observation 2:} After some years, I ate shrimps with octopus, and I experienced a stronger allergic reaction.
        \item []
        \textsc{Observation 3:} Every time I visit restaurants that serve shrimps whose smell is strong, I consistently experience an allergic reaction.
        \item []
        \textsc{\underline{Conclusion:}} Based on these observations, we can infer that I am allergic to the shrimps in all its forms and I must avoid them to prevent allergic reactions.
    \end{itemize}

    \paragraph{Deductive reasoning.} This is a top-down approach. It  starts with a general principle or assumption, and uses it to make a specific prediction or conclusion \citep{simon1996_beyondDeductive,johnson2010_deductive}.  If the premises are true, the conclusion must also be true.  This type of reasoning is commonly used in mathematics, where general principles or axioms are used to derive specific theorems or results. Example: 

    \begin{itemize}
        \item []
        \textsc{Premise 1:} If it rains, the ground gets wet.
        \item []
        \textsc{Premise 2:} It is raining outside.
        \item []
        \textsc{\underline{Conclusion:}} The ground is wet.
    \end{itemize}

    \paragraph{Abductive reasoning.} According to \cite{paul1993_approachesAbductive}, the abduction constitutes a type of reasoning known as non-monotonic. The abductive reasoning involves forming a hypothesis or explanation based on limited, incomplete or ambiguous information \citep{walton2014_abductive}. Abductive reasoning is often used when there is not enough information to make a definitive conclusion, such as in medical diagnostic or criminal investigations, where a set of symptoms or clues are used to infer the most likely explanation or culprit. Example:

    \begin{itemize}
        \item []
        \textit{There is a patient with these symptoms: fever, cough, and fatigue. The doctors may employ abductive reasoning to reach a potential diagnosis. They might consider various hypotheses, such as a common cold, influenza, or pneumonia. They evaluate the available evidence, including the patient's medical history, physical examination, and any lab tests. By analyzing the symptoms and considering the most likely explanation, the doctors abductively reason towards a diagnosis that best fits the observed data.}
    \end{itemize}



    \subsection{Non-textual semantic reasoning}
    \label{subsec:non-textual}

    The non-textual semantic reasoning refers to the cognitive process of understanding and making sense of information without non-written semantic.
    
    \paragraph{Mathematical reasoning. } It is the process of using logical and mathematical principles to arrive at conclusions or solutions to problems \citep{schliemann2002_evolutionMathematical,english2013_mathematical}. It involves breaking down complex problems into smaller, more manageable components \citep{delgado2020_Divide-and-Conquer}. Mathematical reasoning involves the use of mathematical language, symbols, and notation to express ideas and concepts precisely and unambiguously. This type of reasoning can be used in a variety of contexts, from solving equations and proving theorems in pure mathematics, to modeling and analyzing real-world phenomena in applied mathematics.

    \subsection{Commonsense reasoning}
    \label{subsec:commonsense_reasoning}

    Commonsense reasoning is a fundamental aspect of human cognition, and it allows us to navigate the world around us by making informed decisions and judgments \citep{apperly2010mindreaders,Sap2020_commonsense}. For example:

    \begin{itemize}
        \item []
        \textit{We know that putting our hand in the fire will cause us harm.}
        \item []
        \textit{If someone sees a person holding an umbrella on a rainy day, they might infer that it is raining outside.}
    \end{itemize} 
    
    These inferences are based on a person's commonsense knowledge and understanding of the world. 
    
    
    \subsection{Casual reasoning}
    \label{subsec:causal_reasoning}
    
    The ability to engage in causal reasoning is a crucial cognitive skill for human beings \citep{Jonassen2006_causalReasoning}. In the causal reasoning, we can identify the cause-and-effect relationships between different variables or events. It involves examining the relationship between two or more variables to determine whether changes in one variable are likely to cause changes in another variable. For example:

    \begin{itemize}
        \item []
        \textsc{Premise} - I have headache every time I eat chocolate
        \item []
        \textsc{Hypothesis 1} - The chocolate is the cause of my headache.
        \item []
        \textsc{Hypothesis 2} - The sports are the cause of my headache. 
    \end{itemize}
     
     Intuitively, humans will high probably conclude that hypothesis 1 is true, but this is not always the case for machines.

    \subsection{Multi-hop reasoning}
    \label{subsec:multi-hop_reasoning}

    The concept of multi-hop reasoning refers to the process by which a system or model synthesizes information from multiple sources or pieces of data to deduce a conclusion or an answer \citep{tu2019_multihop}. This requires the system to connect and integrate diverse information, often through several logical steps, in order to arrive at a well-supported response. Example:

    \begin{itemize}
        \item []
        \textsc{Paragraph A: }
        In 1921, during the selection process, the Nobel Committee for Physics made a determination that none of the nominations received that year fulfilled the criteria set forth in Alfred Nobel's will. 
        \item []
        \textsc{Paragraph B: }
        As per the regulations set by the Nobel Foundation, if certain circumstances arise, the Nobel Prize may be postponed and awarded in the subsequent year. Consequently, Albert Einstein was granted his Nobel Prize for the year 1921 in the year 1922.
        \item []
        \textsc{Question: } Which scientist was awarded the Nobel Prize in Physics in 1921?
        
        \item []
        \textsc{Answer: } Albert Einstein won the Nobel Prize in Physics in 1921.
        
    \end{itemize}

\section{Datasets}
\label{sec:datasets}

    In this section, we will describe the datasets used to evaluate various reasoning tasks. Firstly, we will present the categories and datasets in Table~\ref{tab:reasoning_categories}. Then, we will delve into the unique characteristics of each dataset.  
    
    \begin{table*}[!htb]
        \centering
        \small
        \begin{tabular}{lc}
        \hline
        \textbf{Categories} & \textbf{Dataset} \\
        \hline
        Deductive &   \makecell{EntailmentBank \citep{dalvi2021_explaining} \\ bAbI (task 15) \citep{Weston2016_bAbI} }
        \\ \hline
        Inductive & 
        
            CLUTRR \citep{sinha2019-clutrr}, 
            bAbI (task 16) \citep{Weston2016_bAbI}
        
        \\ \hline
        Abductive &

             $\alpha$NLI \citep{Chandra2020_Abductive}
        
        \\ \hline
        Mathematical &

             Math \citep{saxton2018_math}

        \\ \hline
        Commonsense &
        
            \makecell{
             CommonsenseQA \citep{talmor2019_commonsenseqa} \\
             PiQA  \citep{Bisk2019_PIQA} \\
             Pep-3k \citep{wang2018_modeling}
            }
        
        \\ \hline
        Causal &
        E-Care  \citep{du2022_eCare}
        \\ \hline
       Multi-hop &     
             HotpotQA \citep{yang2018_hotpotqa}

        \\ \hline
        
        \end{tabular}
        
        \caption{Reasoning categories and datasets}
        \label{tab:reasoning_categories}
    \end{table*}

    \subsection{Logical reasoning}
    
    \subsubsection{Deductive reasoning} 

    \paragraph{EntailmentBank \citep{dalvi2021_explaining}} - it is a multi-step entailment dataset designed to generate explanations in the form of entailment trees. It consists of seven fields: \textit{ID, context, question, answer, hypothesis, proof, meta}. Additionally, the authors have defined three tasks that gradually increase in difficulty and simplify the inherent problems of the task.

    \paragraph{bAbI (task 15) \citep{Weston2016_bAbI}} - Basic Deduction. In this task, the machine learning model is presented with a set of premises and must deduce a conclusion based on those premises. The task tests the model's ability to perform deductive reasoning. It has 1000 questions for training, and 1000 for testing.

    \subsubsection{Inductive reasoning}
    
    \paragraph{CLUTRR (Compositional Language Understanding with Text based Relational Reasoning) \citep{sinha2019-clutrr}} - it includes numerous semi-artificial tales that describe hypothetical families. The objective is to deduce the connection between two family members, which is not explicitly stated in the story.

    \paragraph{bAbI (task 16) \citep{Weston2016_bAbI}} - Basic Induction. In this task, the machine learning model is presented with a set of premises and must induce a general rule or principle based on those premises. The task tests the model's ability to perform inductive reasoning.

    \subsubsection{Abductive reasoning}

    \paragraph{$\alpha$NLI \citep{Chandra2020_Abductive}} -  It is based on the well-known Natural Language Inference (NLI) task, given the observations obs1 and obs2, models should predict which of the hypothesis hyp1 or hyp2 better explains the observations. It consists of approximately 80,000 examples that were collected and labeled by a team of human annotators.  It is divided into three segments: a training set with about 55,000 examples, a validation set featuring around 10,000 examples, and a test set containing nearly 15,000 examples.






    \subsection{Non-textual semantic reasoning}
    \label{subsec:non-textual_semantic_reasoning}
    
    \subsubsection{Mathematical reasoning}

    \paragraph{Math \citep{saxton2018_math}} - It is a comprehensive dataset for mathematical reasoning, intended to assess the competency of machine learning models in solving mathematical problems. It contains more than 100,000 math problems of high school level from standardized tests, including multiple-choice, free-response, and proof-based questions. These problems encompass a wide variety of subjects, such as algebra, geometry, calculus, and statistics.

    \subsection{Commonsense reasoning}

    \paragraph{CommonsenseQA \citep{talmor2019_commonsenseqa}} - it comprises 12,247 multiple-choice questions, each question has five possible answers. The questions are crafted to evaluate a system's capacity to reason through common-sense knowledge. The topics covered in the assessment are diverse and include cause-and-effect relationships, spatial reasoning, social norms, time and sequencing, and physical laws, applied to everyday scenarios and situations.

    \paragraph{PiQA  \citep{Bisk2019_PIQA}} - It  consists of 2,000 multiple-choice questions designed to evaluate common sense reasoning in NLP models. The questions cover a range of topics, including physics, biology, social behavior, and more. The dataset is divided into two parts: a training set with 1,500 questions and a test set with 500 questions. 

    \paragraph{Pep-3k \citep{wang2018_modeling}} - The dataset was created to investigate the plausibility of physical events by determining whether a given event is plausible or implausible. It was constructed using Amazon Mechanical Turk as a platform. 

    \subsection{Causal reasoning}

    \paragraph{e-Care (CAusal REasoning) \citep{du2022_eCare}} - It is a human-annotated dataset that explains causal reasoning. Given a premise and two hypotheses, the goal is to select the hypothesis that better explains the cause or effect of the premise. 
    The dataset contains $21,324$ causal questions (train set = $14928$, dev set = $2132$, test set = $4264$), and $13,048$ unique explanations (train set = $10491$, dev set = $2102$, test set = $3814$).

    \subsection{Multi-hop reasoning}

    \paragraph{HotpotQA \citep{yang2018_hotpotqa}} - It contains over 100,000 question-answer pairs that are derived from paragraphs in Wikipedia. Each question comes with supporting paragraphs to give context and assist in answering. The questions are varied, covering many subjects such as science, history, and general culture. 



\section{Experiments}
\label{sec:expeimennts}

    \subsection{Data selection}
    \label{subsec:data_selection}
    
    We randomly selected 30 samples from each evaluation dataset, depending on their availability from either the train, dev, or test sets. For example, we chose 10 samples each from the training-easy, train-medium, and train-hard sets of data in the Math dataset. Meanwhile, we selected 30 samples from the dev set for $\alpha$NLI, CommonsenseQA, HotpotQA, PiQA, Pep-3k and E-Care. Finally, we took 30 samples from the test set for the bAbI (task 15), bAbI (task 16), CLUTRR, and EntailmentBank datasets. The exact set of samples selected in each dataset are in appendix \ref{sec:appendix}. 

    \subsection{Prompt Engineering}
    \label{subsec:prompt_engineering}
    
    Since the LLMs have exhibited enhanced zero-shot reasoning abilities when provided with effective prompts~\citep{Kojima2022_LLM_Zero-Shot,Liu2023_pretrain_promp}, we opted to employ a combination of well-established, high-performing prompts and our original proposals. In the following list, we outline the specific prompt used for each dataset. Prompts marked with a $\dagger$ symbol represent those that we have proposed. 

    \begin{itemize}
        
        \item
        \textbf{bAbI (task 15)} - prompt engineering proposed by \citet{Bang_MultitaskChatgpt2023}: 

        \noindent\fbox{
    \parbox{0.43\textwidth}{
        \textit{Given facts: } + premises.

        \textit{Based on the given facts above, do a reasonable inference on this question using deductive reasoning: } + question. 
    }
}

        
        \item
        \textbf{CLUTRR} - $\dagger$

 \noindent\fbox{
    \parbox{0.43\textwidth}{
                \textit{Story: } + story. 
                
                \textit{Based on the story, through inductive reasoning answer the question:} + question. 
    }
}

        \item
        \textbf{bAbI (task 16)} - prompt engineering proposed by \citet{Bang_MultitaskChatgpt2023}: 

         \noindent\fbox{
    \parbox{0.43\textwidth}{
        \textit{Given facts: } + premises.

        \textit{Based on the given facts above, do a reasonable inference on this question using inductive reasoning: } + question. 
    }
}

        \item
        \textbf{$\alpha$NLI} - $\dagger$

                 \noindent\fbox{
    \parbox{0.43\textwidth}{
        \textit{Given the following two observations:} + obs1 + obs2. 
        
        \textit{Determine which of the hypothesis options, better explains the observations:} + hyp1 + hyp2.
    }
}

        \item
        \textbf{Math} - prompt proposed by \citet{Kojima2022_LLM_Zero-Shot}: 

     \noindent\fbox{
    \parbox{0.43\textwidth}{
        Mathematical problem. + Question.
        
        \textit{Let's think step by step.}
    }
}

        \item
        \textbf{CommonsenseQA} - $\dagger$

             \noindent\fbox{
    \parbox{0.43\textwidth}{
        \textit{Analyze and answer the following question with one of the five options:} + question + multiple-choice answers. 
    }
}

        \item
        \textbf{PiQA} - $\dagger$

                     \noindent\fbox{
    \parbox{0.43\textwidth}{
        \textit{Complete the phrase with one of two answer options:} + goal + opt1 + opt2. 
    }
}

        \item
        \textbf{Pep-3k} - prompt engineering proposed by \citet{Bang_MultitaskChatgpt2023}:

         \noindent\fbox{
    \parbox{0.42\textwidth}{
        \textit{Please judge if this predicate is (likely) plausible or implausible: } + subject + verb + object.  
    }
}

        \item
        \textbf{E-Care} - $\dagger$

                 \noindent\fbox{
    \parbox{0.42\textwidth}{
        \textit{Given the premise: } + premise.
        
        \textit{Determine which of the two hypothesis options, better explains the cause of the premise. hyp1: } + hyp1. + \textit{hyp2: } + hyp2.   
    }
}

        \item
        \textbf{HotpotQA} - $\dagger$

                         \noindent\fbox{
    \parbox{0.42\textwidth}{
        \textit{Given the context:} + context. 

        \textit{Based on the given context above, answer the question: } + question. 
    }
}

    \end{itemize}


\section{Results and discussion}
\label{sec:results}

    In this section, we present the scores obtained from our conducted experiments. It is important to mention that we compared our results with the findings of \citet{Bang_MultitaskChatgpt2023}. However, since the samples from \citet{Bang_MultitaskChatgpt2023} are not publicly available, we had to take measures to ensure a fair comparison. Therefore, we selected an equal number of samples and conducted evaluations using ChatGPT-3.5, BARD, and ChatGPT-4.

    Additionally, not all engineering prompts used by \citet{Bang_MultitaskChatgpt2023} are publicly accessible. The publicly available prompts are for the bAbI (task 15), bAbI (task 16), and Pep-3k datasets. For the remaining datasets, we utilize the prompt engineering approach we presented in section \ref{subsec:prompt_engineering}.


    \paragraph{\textbf{Inductive Reasoning.}} According to the study conducted by \citet{Bang_MultitaskChatgpt2023}, it has been observed that although GPT-3.5 is a highly advanced language model, it tends to struggle more with inductive reasoning than with deductive or abductive reasoning. Authors initially evaluated the model's basic reasoning abilities using the bAbI-task 16 dataset (Table \ref{tab:results_inductive}), both with and without the use of prompt engineering.  Their reported results indicate a significant improvement in scores when prompt engineering was employed (20/30), in contrary to not using prompt engineering (0/30).
    
    
    In our experiments, we used the same model (GPT-3.5) as \citet{Bang_MultitaskChatgpt2023} and the same number of samples. With prompt engineering, we obtained a score of 14/30, but without prompt engineering, our score decreased to 1/30. Nevertheless,  ChatGPT-3.5 provided the correct answer by relying more on general knowledge of the real world rather than the specific given facts. Here is the sample we tested:

    \begin{itemize}
        \item []
        Brian is a frog. Julius is a swan. Bernhard is a rhino. Greg is a swan. Brian is green. Bernhard is green. Greg is gray. Julius is gray. Lily is a frog. What color is Lily?
        \item []
        \textsc{Correct answer: } green
        \item []
        \textsc{ChatGPT-3.5 answer: } Based on the given information, Lily is a frog, and frogs are typically associated with the color green. Therefore, it can be inferred that Lily is green.
    \end{itemize}

    Additionally, we evaluated the performance of ChatGPT-4 on the bAbI-task 16 dataset, both with and without prompt engineering. In this particular case, the impact of prompt engineering on the results was significant, with a notable improvement from a score of 2/30 without prompt engineering to an impressive score of 28/30 when using prompt engineering. Furthermore, the absence of prompt engineering highlighted the inability of ChatGPT-4 to deduce the answer solely from the given information.

    Similarly, we evaluated bAbI-task 16 dataset using the Google's BARD model. Through prompt engineering, the model improved its performance, by achieving a score of 20/30 compared to 15/30 without prompt engineering. It is noteworthy to mention that BARD outperformed both GPT-3.5 and GPT-4 when prompt engineering was not employed, indicating its ability to reason effectively without relying on strong guidance. However, even though BARD outperformed GPT-3.5 using prompt engineering, it did not outperform GPT-4. This suggests that while BARD is promising, further advancements may be necessary to overcome the capabilities of GPT-4.

    
    Furthermore, in line with the experiment conducted by  \citet{Bang_MultitaskChatgpt2023}, we evaluated  the improved logical thinking skills in inductive tasks using the CLUTRR dataset and prompt engineering (see Table \ref{tab:results_inductive}). In their study, \citet{Bang_MultitaskChatgpt2023} reported a 13 out of 30 correct response rate. Surprisingly, when employing the same model (GPT-3.5) and the same number of samples, our results yielded 2/30 correct responses. Google's BARD achieved a score of 7 out of 30, and the highest performance was exhibited by GPT-4 with 11 out of 30 correct responses. Once again, GPT-4 demonstrated superior results in inductive reasoning, followed by Google's BARD, while ChatGPT-3.5 showed the least promising performance in this regard. 

    On the one hand, we observed that the most common error with GPT-3.5 was its failure to recognize any direct or indirect relationship between individuals. On the other hand, unlike ChatGPT-3.5, ChatGPT-4 incorrectly identified the relationship between  people, particularly struggling to identify motherhood. Additionally, BARD tends to generate verbose responses, unlike GPT models that explicitly state whether a relationship exists or not. Instead of providing a clear verdict, BARD attempts to deduce the relationship by proposing questions, formulating propositions, or making suggestions. Similarly to GPT-4, BARD also struggles with identifying the motherhood relationship. These findings support the notion that ChatGPT-3.5 lacks reasoning abilities, and although improvements have been made in ChatGPT-4, there is still significant room for enhancement. Furthermore, it is intriguing how BARD attempts to discern the relationship between two individuals. However, we must acknowledge that providing an incorrect answer could be more hazardous than admitting that it may not be able to determine the relationship accurately.


    \begin{table*}[!ht]
        \centering
        \resizebox{0.7\textwidth}{!}{
        \begin{tabular}{|c|cc|c|}
        \hline
        & \multicolumn{3}{c|}{\textbf{Inductive Reasoning Tasks}}         \\ \hline 
        \textbf{Datasets}& \multicolumn{2}{c|}{\textbf{bAbI -task 16}}
        & \multicolumn{1}{c|}{\textbf{CLUTRR}}
        \\ \hline

         & \textit{w/o prompt eng.} &  \textit{w/ prompt eng.}  &  prompt eng.\\\hline
        \cite{Bang_MultitaskChatgpt2023} & 0/30 & 20/30 & 13/30 \\\hline
        GPT-3.5 & 1/30 & 14/30 &  2/30 \\\hline
        GPT-4   & 2/30 & 28/30 &  11/30 \\\hline
        BARD   & 15/30 & 20/30 &  7/30 \\\hline
        \end{tabular}
        }
        \caption{Results on inductive reasoning tasks with and without prompt engineering}
        \label{tab:results_inductive}
    \end{table*}


    \paragraph{\textbf{Deductive Reasoning.}} Similarly to \citet{Bang_MultitaskChatgpt2023}, we investigate the basic reasoning skills using the bAbI -task 15 dataset,  both with and without prompt engineering. Unlike inductive reasoning, ChatGPT-4 demonstrates significantly enhanced performance, particularly in tasks without prompt engineering. According to the findings reported by \citet{Bang_MultitaskChatgpt2023}, the use of prompt engineering proves to be advantageous in generating answers to questions that are based on the provided facts (see Table \ref{tab:results_deductive}). However, in our experiments, when we employed the same number of samples and the same chat version (ChatGPT-3.5), we achieved a higher score without prompt engineering (28/30) compared to using prompt engineering (26/30). One possible explanation for these results could be that the model encounters more questions without prompt engineering, thus having already learned to provide correct answers implicitly. Additionally, we also conducted experiments comparing ChatGPT-4 and Google's BARD, both with and without prompt engineering.  The results show that ChatGPT-4 generates more concise responses compared to ChatGPT-3.5. Moreover, when using prompt engineering with ChatGPT-4, its performance significantly outperformed the version without it (27/30), achieving a perfect score of 30/30.

    Regarding BARD, the model exhibits a tendency to break down complete phrases into individual facts or premises, which gives the impression of analyzing the information step by step. However, it faces challenges in incorrect answers, where it divides the complete phrase into premises, proposes different inferences (some of which are correct), but qualifies them as weak and ultimately arrives at a wrong conclusion. In some cases of incorrect answers, the model provides two different responses instead of a single one. Notably, BARD achieves the same score of 25 out of 30 both with and without prompting, which is consistent with the findings from our previous experiments on inductive reasoning. This result suggests that BARD does not require strong guidance and can perform effectively with less explicit input.



    In this study, ChatGPT-4 achieved the best results over ChatGPT-3.5 and BARD model. However, we highlight a particular instance where ChatGPT-4 fails to deduce the accurate response solely from the provided information. Instead, it relies on its general knowledge and understanding of the world.
    

    \begin{itemize}
        \item []
        Wolves are afraid of mice. Mice are afraid of wolves. Gertrude is a wolf. Jessica is a mouse. Emily is a mouse. Cats are afraid of sheep. Sheep are afraid of wolves. Winona is a cat. What is gertrude afraid of?
        \item []
        \textsc{Correct answer: } maice
        \item []
        \textsc{ChatGPT-4 answer: }There is no information provided that Gertrude, the wolf, is afraid of anything. The initial statements claim that wolves are afraid of mice, but these are contradictory and not logical, as wolves are typically predators and not afraid of mice.
    \end{itemize}

    Building on our exploration of advanced reasoning, we leveraged the EntailmentBank dataset along with prompt engineering techniques to further enhance our study. 
    The results in Table \ref{tab:results_deductive} show that the research conducted by \citet{Bang_MultitaskChatgpt2023} achieved better performance than our findings. They used the same model and number of samples but different samples, obtaining a score of 28 out of 30, while we scored 25 out of 30. Furthermore, our experiments revealed that ChatGPT-4 and BARD performed better than ChatGPT-3.5, scoring 27 out of 30 compared to 25 out of 30, but this difference was not statistically significant. 

    In this specific case, we observed that both ChatGPT-3.5 and ChatGPT-4 provided incorrect answers due to their response lacking essential information. For instance, in sample 1 (see Appendix \ref{sec:appendix}), the correct answer was \textit{"oceans"}, but GPT-4uilding on our explorat only mentioned the Pacific Ocean. Additionally, in example 22, both models inaccurately answered \textit{"argon"} instead of the precise answer, which is \textit{"carbon dioxide"}. Similarly, like the GPT models, BARD also exhibited instances where its responses lacked necessary information. However, in some cases, BARD interpreted the inquiry as related to image creation, responding with \textit{"I can’t create images yet, so I’m not able to help you with that"}, based on the context provided. 
    

    The results presented here highlight significant opportunities for enhancing GPT-3.5's capacity to accurately comprehend and infer accurate responses from given information. Despite the advancements exhibited by GPT-4, there remains a need to further refine its capability to address specific queries based on provided texts. In comparison with ChatGPT, BARD tends to produce more extensive answers. Remarkably, when the model's intention shifts towards generating images rather than providing textual answers, it tends to focus on the topic of the moon.

    \begin{table*}[!ht]
        \centering
        \resizebox{0.6\textwidth}{!}{
        \begin{tabular}{|c|cc|c|}
        \hline
        & \multicolumn{3}{c|}{\textbf{Deductive Reasoning Tasks}}           \\ \hline
        \textbf{Datasets}& \multicolumn{2}{c|}{\textbf{bAbI -task 15}} 
        & \multicolumn{1}{c|}{\textbf{EntailmentBank}} \\ \hline
        
        & \textit{w/o prompt eng.} & \textit{w/ prompt eng.} & \textit{w/ prompt eng.}  \\\hline
        \cite{Bang_MultitaskChatgpt2023}  & 19/30 & 28/30 & 28/30\\\hline
        GPT-3.5  & 28/30 & 26/30  & 25/30\\\hline
        GPT-4  & 27/30 & 30/30 & 27/30 \\\hline
        BARD  & 25/30 & 25/30 & 27/30 \\\hline
        \end{tabular}
        }
        \caption{Results on deductive reasoning tasks using prompt engineering}
        \label{tab:results_deductive}
    \end{table*}

    \paragraph{\textbf{Abdutive Reasoning.}} To evaluate the abductive reasoning capabilities of GPT-4 and BARD, we utilized the $\alpha$NLI dataset with prompt engineering. The scores reported by \citet{Bang_MultitaskChatgpt2023} demonstrated superior performance compared to our results using ChatGPT-3.5 (19/30), ChatGPT-4 (25/30), and BARD (22/30) (see Table \ref{tab:results_abductive}). In our experiments, ChatGPT-4 demonstrated superior performance compared to ChatGPT-3.5 and Google's BARD. It's worth noting that they failed in different examples, with only one instance in common (sample 28):
    

    \begin{itemize}
        \item []
        \textsc{Input: } \small
        Given the following two observations: obs1: Ryan climbed the long set of stairs to get to the water slide. obs2: It was even more fun than he expected. Determine which of the hypothesis options, better explains the observations: hyp1: Ryan was scared to go down the water slide. hyp2: Ryan loved going down the water slide.
        \item []
        \small
        \textsc{Correct answer: 1} 
        
        \item []
        \small
        \textsc{GPT-3.5 answer: } Based on the given observations, it is more likely that hypothesis option 2, ”Ryan loved going down the water slide,” better explains the observations. Observation 1 states that Ryan climbed the long set of stairs to get to the water slide, which suggests that he was motivated to go down the water slide. If he was scared to go down the water slide (hypothesis option 1), it seems less likely that he would have gone through the effort of climbing the stairs to reach the slide. Observation 2 states that it was even more fun than Ryan expected, which suggests that he had some positive expectations about the experience. This is consistent with hypothesis option 2, that Ryan loved going down the water slide, but not with hypothesis option 1, that he was scared to go down the slide. Overall, while it is possible that Ryan was initially scared to go down the water slide, the observations provided better support hypothesis option 2, that Ryan loved going down the water slide
        \item [] 
        \small
        \textsc{GPT-4 answer: } Based on the given observations, hyp2: Ryan loved going down the water slide better explains the observations. Observation 1 states that Ryan climbed the long set of stairs to get to the water slide, which indicates that he was willing to make an effort to reach the slide. Observation 2 states that it was even more fun than he expected, which suggests that Ryan enjoyed the experience. These observations do not support hyp1: Ryan was scared to go down the water slide.
        \item [] 
        \small
        BARD answer: Hypothesis 1: Ryan was scared to go down the water slide. - Observation 1: Ryan climbed the long set of stairs to get to the water slide. This observation could be explained by the hypothesis that Ryan was scared to go down the water slide and wanted to make sure he was prepared. He may have thought that the water slide was too high or too fast, and he wanted to be sure that he could handle it. - Observation 2: It was even more fun than he expected. This observation is difficult to explain with the hypothesis that Ryan was scared to go down the water slide. If he was scared, then he would not have expected to have fun. It is more likely that he was excited to go down the water slide and was surprised by how much fun it was. Hypothesis 2: Ryan loved going down the water slide. - Observation 1: Ryan climbed the long set of stairs to get to the water slide. This observation could be explained by the hypothesis that Ryan loved going down the water slide and was eager to do it again. He may have thought that the water slide was the most fun thing at the pool, and he wanted to make sure he got to experience it as many times as possible. - Observation 2: It was even more fun than he expected. This observation is easily explained with the hypothesis that Ryan loved going down the water slide. If he loved it, then he would have expected to have fun, but he would have been surprised by how much fun it actually was. Conclusion: The hypothesis that better explains the observations is Hypothesis 2: Ryan loved going down the water slide. This hypothesis is consistent with both observations, while Hypothesis 1 is only consistent with Observation 1. Additionally, Hypothesis 2 is more likely in general, as most people enjoy going down water slides.
    \end{itemize}

     Even though the answers of the three models seem convincing, they are incorrect. In fact, the question can be hard even for a human, because the nuance between the two hypotheses is weak. On the one hand, the first observation does not show any hesitation or fear to climb the stairs, which can lead to think Ryan might be happy to climb. On the other hand, the second observation shows that Ryan had expectations about the water, that might be negative, but what he found was better than his expectations. The latter observation has a tiny more impact on the phrase than the first one, and the nuance between the two observations was not detected by any model.


    \begin{table}[!ht]
        \centering
        \small
        \begin{tabular}{|c|c|}
        \hline
        \multicolumn{2}{|c|}{\textbf{Abdutive Reasoning Tasks}}           
        \\ \hline
        \multicolumn{2}{|c|}{\textbf{$\alpha$NLI dataset}} 
        \\ \hline
         & \textit{w/ prompt eng.}  \\\hline
        \cite{Bang_MultitaskChatgpt2023}  & 26/30 \\\hline
        GPT-3.5   & 19/30   \\\hline
        GPT-4     & 25/30  \\\hline
        BARD     & 22/30  \\\hline
        \end{tabular}
        \caption{Results on abductive reasoning tasks using prompt engineering}
        \label{tab:results_abductive}
    \end{table}


\paragraph{\textbf{Mathematical Reasoning.}} 

    To compare the mathematical skills of ChatGPT-4, BARD and ChatGPT-3.5, we utilize the Math dataset. While the scores presented in Table \ref{tab:results_mathematical} reflect the performance of \citet{Bang_MultitaskChatgpt2023}, it is important to note that our experiments utilized different samples. Notably, the discrepancy between our scores and theirs can vary significantly due to the varying difficulty levels of the mathematical problem samples. Therefore, we will concentrate on the results obtained from our experiments. Specifically,  we used ten samples from each category, representing easy difficulty (1-10), medium difficulty (11-20), and hard difficulty (21-30). 
    
    Surprisingly, in contrast to prior findings, BARD exhibited significantly reduced verbosity in its responses compared to GPT-3.5, as observed in this series of experiments (see Table \ref{tab:mathematical_reasoning}). Consequently, some solutions to certain mathematical problems include unnecessary steps, and for the long of the explanation the answer can be not complete.  Furthermore, it is interesting to observe how GPT-3.5 can analyze the problem in a logical sequence, but provides an incorrect answer/conclusion. For example, let us consider the case of sample nine, where the model successfully obtained the values for both $b$ and $i$. However, the ultimate conclusion is erroneous.
    

    \begin{itemize}
        \item []
        \textsc{Input.} Let b = 1 + -1. Let z = 15 - 5. Let i = z - 6. Is b smaller than i? Let's think step by step.

        \item []
        \textsc{Correct answer.} True
        
        \item []
        \textsc{Final ChatGPT3.5 answer}

        $b = 0$
        
        $i = 4$
    
        Since b is equal to 0 and i is equal to 4, we can conclude that b is not smaller than i. In fact, b is less than i.
    \end{itemize}

    When it comes to Google's BARD model, it exhibits comparable performance to ChatGPT-3.5, achieving a score of 16/30 compared to 15/30, respectively. Notably, BARD showcases remarkable prowess in tackling complex mathematical problems, outshining other models. Moving forward, ChatGPT-4 has emerged as a frontrunner, obtaining the highest score of 20/30 when compared to both ChatGPT-3.5 and BARD. However, ChatGPT-4 demonstrates its real advantage in solving easy and intermediate problems, as it surpasses the other models in these categories. On the other hand, when it comes to resolving difficult problems, BARD remains ahead. While ChatGPT-4 boasts a superior overall score, its specialized strengths differ from those of BARD, each excelling in distinct problem-solving areas.

    Furthermore, GPT-4 faces challenges when it comes to finding the prime factors of a given number and converting numbers to different bases. Additionally, while it provides accurate answers in certain instances, it does not always simplify the responses to the minimum expression. However, there is significant room for improvement in the mathematical skills. In this paper, we have only presented thirty samples, which do not encompass all the diverse areas of mathematics.
    

    \begin{table}[!ht]
        \centering
        \small
        \begin{tabular}{|c|c|}
        \hline
        \multicolumn{2}{|c|}{\textbf{Mathematical Reasoning Tasks}}           
        \\ \hline
        \multicolumn{2}{|c|}{\textbf{Math dataset}} 
        \\ \hline
         & \textit{w/ prompt eng.}  \\\hline
        \cite{Bang_MultitaskChatgpt2023}  & 13/30 \\\hline
        GPT-3.5   & 15/30   \\\hline
        GPT-4     & 20/30  \\\hline
        BARD     & 16/30  \\\hline
        \end{tabular}
        \caption{Results on mathematical reasoning tasks using prompt engineering}
        \label{tab:results_mathematical}
    \end{table}


    \paragraph{\textbf{Commonsense Reasoning.}} Following \citet{Bang_MultitaskChatgpt2023}, in this type of reasoning we evaluate three datasets: CommonsenseQA, PiQA, and Pep-3K using prompt engineering. The results are presented in Table \ref{tab:results_commonsense_with_eng}).

    In terms of CommonsenseQA dataset, \citet{Bang_MultitaskChatgpt2023} achieved the highest score (27/30), while our experiments with ChatGPT-3.5, ChatGPT-4 and BARD obtained scores of 21/30, 25/30, and 20/30, respectively. It is worth noting that our 30 samples differ from the ones evaluated by \citet{Bang_MultitaskChatgpt2023}, but their samples are not publicly available. Interestingly, four of the answers (samples 11, 15, 24, and 28 in Table \ref{tab:commonsense_reasoning}) were incorrectly answered by all three models. For example, in question 11, \textit{"The cat carefully navigated the area, doing everything they can to avoid what?"}, with answer options \textit{"A. get wet, B. eat vegetables, C. falling, D. wool sweater, and E. sharp claw"}, the three models inaccurately chose option "C. falling" instead of the correct answer \textit{"A. get wet"}.  This reveals a significant gap in the three LLMs' comprehension of common knowledge, such as the fact that cats dislike water. Such understanding is usually acquired by humans through experience over time.

    In the same vein, the three models also struggled to provide the correct answer for sample 24, which asked, \textit{"What room is likely to have a sideboard on the counter?"} The available options were: \textit{"A. home, B. serve food buffet, C. dining room, D. living room, E. kitchen"}. Surprisingly, all three LLMs responded with \textit{"C: dining room."} However, the distinguishing factor in this case is that unlike the previous example, where the lack of common knowledge was evident, this sample could even confuse humans. This is because a dining room could indeed have a sideboard on the counter, making the answer less straightforward.

    In the context of the PiQA dataset, our results surpassed the score achieved by \citet{Bang_MultitaskChatgpt2023} (see Table \ref{tab:results_commonsense_with_eng}). They obtained a score of 25 out of 30, while ChatGPT-4 almost achieved a perfect score of 29 out of 30. Furthermore, ChatGPT-3.5 also performed competitively, scoring 24 out of 30. Comparatively, the BARD model scored 23 out of 30, making it a strong contender. Interestingly, the errors made by both GPT models were unique to each model. Contrarily, there were some common errors between both GPT models and the BARD model.

    

    Let's examine an example that illustrates an incorrect answer given by ChatGPT-3.5 and BARD (sample 10). The question is, \textit{"Complete the phrase with one of two answer options: To attach rope to latex, you can sol1: Rubber band the rope in place on the latex sol2:  Glue the rope in place on the latex"}. Both models provided an inaccurate suggestion by proposing solution 1. However, it is important to clarify that the term "rubber" refers to a resilient and elastic polymeric material derived from the latex of a tropical plant or produced synthetically. Therefore, the correct and straightforward answer is solution 2.


    Now, let's examine the only incorrect response given by ChatGPT-4 and Google's BARD model (sample 15). The question is, \textit{"Complete the phrase with one of two answer options: hands sol1: cut into fruit  sol2:  cut into metal"}. The ChatGPT-4's answer is "metal," but it is evident that this answer is incorrect since human hands cannot cut metal. Additionally, the BARD response suggests that both solutions are valid, and the choice between them depends on the specific context. The BARD response goes above and beyond by providing to illustrate the appropriate usage of each solution in different contexts.


     Finally, in the Pep-3K dataset, we applied the prompt engineering technique suggested by \citet{Bang_MultitaskChatgpt2023}. They achieved 28 out of 30 correct answers. However, when we used the same model (ChatGPT-3.5) and sample size, the results yielded 15 correct answers. As expected, the newer model, ChatGPT-4, outperformed its predecessor, reaching 18 correct answers in our experiments. However, what was truly surprising was the performance of BARD, which surpassed even ChatGPT-4 by achieving 19 out of 30 correct answers.


     It is worth mentioning that there are eight incorrect responses common to all three models, indicating potential areas of improvement or challenges in handling those specific queries. To begin with, we concentrate on analyzing the shared errors in both GPT models. Our observation suggests that these errors arise due to their tendency to generalize. For instance, let's consider the first sample listed in Table \ref{tab:commonsense_reasoning} under the Pep-3K dataset section. In this case, the models were tasked with assessing the plausibility of the statement \textit{"dog ride cat."} Both models argued that dogs are generally larger than cats, and it would be unusual for a cat to be able to support the weight of a dog. However, we, as humans, are aware that there are dogs that are smaller or of equivalent size to cats, making it plausible for a dog to ride a cat.

     Furthermore, let's consider a sample where all three models failed in generating a suitable response (sample four). In this instance, the models were tasked with evaluating the plausibility of the sentence: \textit{"clown catch baboon."} This specific sentence allows us to observe distinct reasoning approaches in each model. ChatGPT-3.5 confidently categorized it as implausible, attributing the decision to its training data. ChatGPT-4 shared a similar conclusion but recognized a possibility, albeit unlikely. In contrast, BARD not only labeled the sentence as implausible but also incorporated common knowledge of the world into its reasoning process.    

    \begin{table}[!ht]
        \centering
        \small
        \begin{tabular}{|cccc|}
        \hline
        \multicolumn{4}{|c|}{\textbf{Commonse Reasoning Tasks}}      \\ \hline
        \multicolumn{1}{|c|}{\begin{tabular}[c]{@{}c@{}}
        \cite{Bang_MultitaskChatgpt2023} \end{tabular}} 
        & \multicolumn{1}{c|}{\begin{tabular}[c]{@{}c@{}}GPT-3.5 \end{tabular}} 
        & \multicolumn{1}{c|}{\begin{tabular}[c]{@{}c@{}}GPT-4 \end{tabular}} 
        & \begin{tabular}[c]{@{}c@{}}BARD \end{tabular} \\ \hline
        
        \multicolumn{4}{|c|}{\textbf{CommonsenseQA dataset}} \\ \hline
        \multicolumn{1}{|c|}{27/30}  
        & \multicolumn{1}{c|}{21/30} & 25/30 
        & \multicolumn{1}{|c|}{20/30}
        \\ \hline
        \multicolumn{4}{|c|}{\textbf{PiQA dataset}} \\ \hline
        \multicolumn{1}{|c|}{25/30}                                     
        & \multicolumn{1}{c|}{24/30} & 29/30  
        & \multicolumn{1}{|c|}{23/30}
        \\ \hline
        \multicolumn{4}{|c|}{\textbf{Pep-3K dataset}}    \\ \hline
        \multicolumn{1}{|c|}{28/30}                            
        & \multicolumn{1}{c|}{15/30} & 18/30  
        & \multicolumn{1}{|c|}{19/30}
        \\ \hline
        \end{tabular}
        \caption{Results on commonsense reasoning tasks using prompt engineering}
        \label{tab:results_commonsense_with_eng}
    \end{table}


    \paragraph{\textbf{Causal Reasoning.}} In alignment with the research conducted by \citet{Bang_MultitaskChatgpt2023}, which explored causal reasoning using the E-Care dataset and the ChatGPT-3.5 model, both their study and our own experiments with BARD yielded 24 correct answers out of 30 questions. However, during our experiments, we encountered challenges with ChatGPT-3.5. Specifically, it faced challenges in accurately connecting premises with the appropriate hypotheses, providing only 14 correct answers out of 30. Finally, ChatGPT-4 exhibited a higher performance, correctly answering 25 out of 30 questions, which outperformed the findings reported by \citet{Bang_MultitaskChatgpt2023}.


    Regarding the experiments we conducted, there are three common questions that all the models fail to answer accurately (17, 21, 22). On the one hand,  we observed that ChatGPT-4 and BARD tend to select the wrong hypothesis and fail to explain the cause of the premise. On the other hand, ChatGPT-3.5 often recognizes that the provided information or context is insufficient to determine the correct hypothesis and is unable to choose one of the hypotheses. Therefore, we need to be cautious with the explanations provided by ChatGPT-4 and BARD. While they may seem to possess enhanced reasoning abilities, it is important to note that sometimes the conclusions they arrive at are incorrect.


    \begin{table}[!ht]
        \centering
        \small
        \begin{tabular}{|c|c|}
        \hline
        \multicolumn{2}{|c|}{\textbf{Causal Reasoning Tasks}} 
        \\ \hline
        \multicolumn{2}{|c|}{\textbf{E-Care dataset}} 
        \\ \hline
        & \textit{w/ prompt eng.}  \\\hline
        \cite{Bang_MultitaskChatgpt2023} & 24/30 \\\hline
        GPT-3.5  & 14/30   \\\hline
        GPT-4    & 25/30  \\\hline
        BARD    & 24/30  \\\hline
        \end{tabular}
        \caption{Results on causal reasoning task}
        \label{tab:results_causal_reasoning}
    \end{table}


    \paragraph{\textbf{Multi-hop Reasoning.}}

    To evaluate multi-hop reasoning skills, we utilized the HotpotQA dataset. In a study conducted by \citet{Bang_MultitaskChatgpt2023}, they achieved 8 correct samples out of 30 (refer to Table \ref{tab:results_multi-hop} for complete results). However, when we employed the same number of samples with ChatGPT-3.5, we attained 16 correct answers out of 30. Using these identical samples, we compared the performance of ChatGPT-4, which outperformed ChatGPT-3.5 with 17 correct answers out of 30. Since the  difference is not significant, it led us to conclude that ChatGPT-4 still faces challenges when integrating information from multiple sources to determine an answer. We observed that both models struggle with answers when the context is extensive, particularly when it involves people related with specific dates.

    When evaluating the performance of BARD on the HotpotQA dataset, it outperformed both GPT models, achieving 20 correct answers out of 30 questions. BARD provides more comprehensive responses, and in some cases, the "wrong" answers are due to the model's inability to process all the input. Interestingly, we did not encounter this issue with any of the GPT models.

    \begin{table}[!ht]
        \centering
        \small
        \begin{tabular}{|c|c|}
        \hline
        \multicolumn{2}{|c|}{\textbf{Multi-hop Reasoning Tasks}} 
        \\ \hline
        \multicolumn{2}{|c|}{\textbf{HotpotQA dataset}} 
        \\ \hline
        & \textit{w/ prompt eng.}  \\\hline
        \cite{Bang_MultitaskChatgpt2023} & 8/30 \\\hline
        GPT-3.5  & 16/30   \\\hline
        GPT-4    & 17/30  \\\hline
        BARD    & 20/30  \\\hline
        \end{tabular}
        \caption{Results on multi-hop reasoning tasks using prompt engineering}
        \label{tab:results_multi-hop}
    \end{table}

\section{Conclusions and Future Work}
\label{sec:Conclusions}

    This paper focuses on evaluating seven types of reasoning: deductive, inductive, abductive, mathematical, commonsense, causal, and multi-hop using ChatGPT-3.5, ChatGPT-4 and Google's BARD model. It aims to: (1) determine the improvements in reasoning skills achieved by ChatGPT-4 compared to ChatGPT-3.5, and BARD and (2) identify the key challenges that still need to be addressed by the scientific community.

    To evaluate the quality of answers generated by the three models, we conducted a manual examination of the results for each reasoning task. Our findings indicate that although there has been an improvement in the performance of ChatGPT-4 compared to ChatGPT-3.5 and Google's BARD, there is still considerable work to be done. Specifically, areas such as inductive reasoning, mathematical problem-solving, multi-hop reasoning, and commonsense reasoning require significant attention. For example, in the inductive reasoning, the three models face difficulties to determine the relationship between two people. Furthermore, when it comes to commonsense reasoning, the models struggle to discern the plausibility of certain actions. When comparing our findings to the research conducted by \citet{Bang_MultitaskChatgpt2023}, we acknowledge that the differences in our sample data may have influenced the scores due to variations in complexity.


    We have made our samples publicly available to facilitate reproduciblity. We hope and encourage their use as a benchmark for future research. 
    In the future, we will focus on the following directions:

     \begin{itemize}
        
        \item 
        Due to the limited number of samples in our research, we were unable to cover all areas of mathematics, such as algebra, arithmetic, trigonometry, and calculus. To enhance our understanding of mathematical skills in ChatGPT-4 and ChatGPT-3.5, we intend to conduct an exhaustive evaluation for each specific area, ensuring a deeper and more comprehensive analysis.

        \item 
        It would be interesting to enhance the scope of our research by  evaluating our samples across more Language Models (LLMs). This will be helpful to have a better understanding of the data, and a wider comparative study of state-of-the-art models.
        
     \end{itemize}












\clearpage
\onecolumn
\begin{landscape}

\appendix
\section{Appendix}
\label{sec:appendix}
Correct answers are in green, wrong answers are in red.
\begin{center}
    \small
    \begin{longtable}{|p{0.38\linewidth}|p{0.12\linewidth}
                      |p{0.12\linewidth}|p{0.16\linewidth}
                      |p{.05\linewidth}|}
     
    \caption{Deductive reasoning prompts. } 
    \label{appendix:deductive_reasoning}\\
    
    \hline
    \multicolumn{5}{|c|}{\textbf{Deductive Reasoning}} \\ \hline
    \textbf{Prompt} & \textbf{GPT-3.5 Answer} & \textbf{GPT-4 Answer} & \textbf{BARD Answer} & \textbf{Gold} \\ \hline
    \multicolumn{5}{|l|}{\textbf{Dataset: EntailmentBank}} 

    \\ \hline  
    1. ocean is the largest body of water on earth. body of water is a source of wild fish for human populations. as the size of a body of water increases , the amount of wild fish will increase. Which of the following is the greatest source of wild fish for human populations? answer in few words. Do not give explanation. 
    & 
    \textcolor{ForestGreen}{Ocean} 
    & 
    \textcolor{red}{The Pacific Ocean.}
    & 
    \textcolor{ForestGreen}{The ocean is the greatest source of wild fish for human populations.}



    & 
    oceans
    
    \\ \hline  
    2. fruit salad often is made of fruit. fruit salad is a kind of salad. fruit is usually a solid. a mixture of solids can be separated easily. salad is a mixture. Which mixture can be separated into its ingredients? answer in few words. Do not give explanation. 
    &
    \textcolor{red}{Homogeneous mixture.}
    &
    \textcolor{ForestGreen}{fruit salad}
    & 
    \textcolor{ForestGreen}{Salad is a mixture of solids. Therefore, fruit salad can be separated into its ingredients.}
    &
    fruit salad 
    \\ \hline  
    3. a frog is a kind of amphibian. adulthood is a stage in the life cycle process. reproduction only occurs during adulthood. an example of reproduction is laying eggs. amphibians lay eggs. Which of the following happens only during the adult stage of the life cycle of a frog? answer in few words. Do not give explanation. 
    & 
    \textcolor{ForestGreen}{Reproduction (laying eggs).}
    &
    \textcolor{ForestGreen}{Egg laying and reproduction.}
    & 
    \textcolor{ForestGreen}{The answer is laying eggs.This is because reproduction only occurs during adulthood, and laying eggs is an example of reproduction. Amphibians lay eggs, and frogs are amphibians. Therefore, the only stage of the life cycle of a frog where laying eggs occurs is adulthood.}
    &
    A frog lays eggs
    \\ \hline  
    4. carbon dioxide is a kind of gas. if something is made up of something else then that something else can be found in that something. green plants provide food for themselves / animals / consumers by performing photosynthesis. in order for a thing to provide something that thing must make that something. to perform a process can mean to use a process. green plants are a kind of producer. the earth 's atmosphere is made of 036 percent carbon dioxide by volume. if something is a raw material in a process then that something is required for that process. in the photosynthesis process carbon dioxide has the role of raw material. Earth's atmosphere is made of many gases. Which gas do producers use to make food? answer in few words. Do not give explanation. 
    & 
    \textcolor{ForestGreen}{Carbon dioxide.}
    &
    \textcolor{ForestGreen}{carbon dioxide}
    & 
    \textcolor{ForestGreen}{carbon dioxide}
    &
    carbon dioxide
    \\ \hline  
    5. skin is used for protecting the body from harmful substances. protecting is a kind of function. What is the function of skin on the human body? answer in few words. Do not give explanation. 
    &
    \textcolor{ForestGreen}{The function of skin on the human body is protection.}
    &
    \textcolor{ForestGreen}{Protection, temperature regulation, sensation, excretion.}
    & 
    \textcolor{ForestGreen}{The function of skin on the human body is to protect.}
    &
    Skin protects the body from harmful substances.
    \\ \hline  
    6. the sun is a source of radiation / heat called sunlight. absorbing sunlight causes objects to heat. drops of water are a kind of objects. evaporation means a substance changes from a liquid into a gas by increasing heat energy. if water evaporates, then the water will disappear. water is a kind of substance. a drop is small amount of liquid. One morning, Jane notices drops of water on the outside of a window. Later in the day, she notices that the drops have disappeared. What most likely caused the drops to disappear? answer in few words. Do not give explanation. 
    &
    \textcolor{ForestGreen}{Evaporation.}
    &
    \textcolor{ForestGreen}{Sunlight evaporation.}
    & 
    \textcolor{ForestGreen}{The drops of water evaporated.}
    &
    Energy from the sun caused the water to evaporate.
    \\ \hline  
    7. a different moon phase occurs approximately once per week. the first quarter phase of the moon occurs after the new moon. For a science activity, Mia had to draw a picture of the Moon each night. If Mia drew a new moon, what type of moon did Mia draw about one week later? answer in few words. Do not give explanation. 
    &
    \textcolor{ForestGreen}{First quarter moon.}
    &
    \textcolor{ForestGreen}{First quarter moon}
    & 
    \textcolor{red}{I can't create images yet so I'm not able to help you with that.}
    &
    first quarter
    \\ \hline  
    8. some insects eat nectar. a source of something emits / produces / generates / provides that something. a flower is a part of a plant for attracting pollinators / producing seeds. some birds eat nectar. an insect is a pollinating animal. a bird is a pollinating animal. a flower is a source of nectar. Which is made by plants to attract pollinators to flowers? answer in few words. Do not give explanation. 
    &
    \textcolor{ForestGreen}{Nectar.}
    &
    \textcolor{ForestGreen}{Nectar is made by plants to attract pollinators to flowers.}
    & 
    \textcolor{ForestGreen}{Nectar is made by plants to attract pollinators to flowers.}
    &
    nectar
    \\ \hline  
    9. a scientist wants to investigate the environmental changes from ancient time to now. fossils can be used as evidence for the ancient environment. evidence is a kind of information. investigating requires gathering information. Environments have changed over millions of years. Which of these would be best for a scientist to study when investigating these environmental changes? answer in few words. Do not give explanation. 
    &
    \textcolor{ForestGreen}{Fossil records.}
    &
    \textcolor{ForestGreen}{Paleoclimate data, fossil records, sedimentary layers.}
    & 
    \textcolor{ForestGreen}{Fossils, ice cores, and sedimentary rocks.}
    &
    fossil records
    \\ \hline  
    10. if something impacts the entire earth then that something has a significant impact on the earth. a source of something releases that something. worldwide means across the entire earth. temperature increase is a kind of impact. global warming is when worldwide temperature increases. something comes from a source of that something. earth's atmosphere is a part of the earth. gases released during the use of fossil fuels causes global warming. The most significant human-induced impact on Earth's atmosphere comes from the use of which energy resource? answer in few words. Do not give explanation. 
    &
    \textcolor{ForestGreen}{Fossil fuels.}
    &
    \textcolor{ForestGreen}{Burning fossil fuels.}
    & 
    \textcolor{ForestGreen}{Fossil fuels.}
    &
    fossil fuels
    \\ \hline  
    11. the sun is a kind of yellow dwarf. including means containing. revolving around something means orbiting that something. a type of something is a kind of property of that something. composed of means made of. to categorize means to classify. made up of means contains / made of. main-sequence stars are generally yellow in color. a yellow dwarf is a kind of main-sequence star. to belong to a class means to be a kind of something. place / location is synonymous with position. the location of an object can be used to describe that object. to classify means to decide what class something belongs to. if something is a part of something else then that something else contains that something. alpha mensae is a kind of main sequence star. best means most positive. classify is similar to characterize. characteristic means property. grouping means putting / placing in different groups. brightness is a property of a light source and includes values of bright / dim. both means two. a type is synonymous with a kind. a description sometimes provides information. how many means what number. to be found in means to be contained in. There are many sizes and colors of stars. Which of the following best describes the category in which the Sun would be placed? answer in few words. Do not give explanation. 
    &
    \textcolor{ForestGreen}{Main-sequence star}
    &
    \textcolor{ForestGreen}{The Sun would be placed in the category of a yellow dwarf or G-type main-sequence star.}
    & 
    \textcolor{ForestGreen}{The Sun is a yellow dwarf, which is a type of main-sequence star. So the answer is main-sequence star.}
    &
    yellow main sequence stars
    \\ \hline  
    12. where something comes from is a source of that something. a source of something releases that something. humans cause pollution. pollution has a negative impact on the environment / air quality / water quality / society. burning fossil fuels converts potential energy into thermal energy. burning coal is a source of air pollution. greenhouse gases are a part of the atmosphere. if something impacts the entire earth then that something has a significant impact on the earth. the emission of something is when something is added into the atmosphere. water vapor is the largest amount of greenhouse gas in the atmosphere. carbon dioxide is a kind of gas. worldwide means across the entire earth. fossil fuels burning causes pollution. oil is a kind of pollutant. burning natural gas is a source of air pollution. atmosphere means air. using gasoline is a source of carbon dioxide. earth's atmosphere is a part of the earth. gases released during the use of fossil fuels causes global warming. natural gas is a kind of fossil fuel. a source of something increases the amount of that something. something comes from a source of that something. temperature increase is a kind of impact. earth has air. global warming is when worldwide temperature increases. The most significant human-induced impact on Earth's atmosphere comes from the use of which energy resource? answer in few words. Do not give explanation. 
    &
    \textcolor{ForestGreen}{Fossil fuels.}
    &
    \textcolor{ForestGreen}{Burning fossil fuels.}
    & 
    \textcolor{ForestGreen}{The most significant human-induced impact on Earth's atmosphere comes from the use of fossil fuels.}
    &
    fossil fuels
    \\ \hline  
    13. black objects absorb the most amount of sunlight. solar radiation means solar energy. light is a kind of electromagnetic wave. all waves transfer / transmit energy from one place to another. warm means medium heat energy / temperature. intensity of sunlight is similar to amount of sunlight. absorbing energy causes objects / materials / substances to heat. the sun is a source of radiation / heat called sunlight. sunlight means solar energy. if something hot is exposed to the air , then the heat will transfer to the air. if something emits something else then that something increases the amount of that something else. thermal conduction is when heat flows from warmer object to cooler object. radiation is when electromagnetic waves transfers heat and light energy. heat means the transfer of thermal energy. an object moving from a cool place to a warm place causes the object 's temperature to increase. rays of the sun means sunlight. thermal energy is a kind of energy. as the sunlight absorbed by the object increases , the temperature of the object will increase more. sunlight is a kind of light. radiation means radiant energy. if heat is added to a substance then that substance absorbs that heat. solar energy is a kind of energy. if one object absorbs more energy than another object , then the object will be warmer. the brick absorbs heat energy from the sunglight. if heat is absorbed from a source then that heat source will cool. A black brick exposed to sunlight becomes warm. How is energy from the Sun transferred to the brick? answer in few words. Do not give explanation. 
    &
    \textcolor{ForestGreen}{Energy from the Sun is transferred to the brick through the process of radiation.}
    &
    \textcolor{ForestGreen}{Sun's energy is transferred to the brick through radiation and absorption.}
    & 
    \textcolor{ForestGreen}{The energy from the Sun is transferred to the brick through radiation.}
    &
    radiation
    \\ \hline  
    14. destruction means loss. the formation of metamorphic rock is a sequential process, with stages of rock being compacted, heated, recrystallized. decay has a negative impact on a thing. if something is found in something else, then that something else contains that something. fossils are formed when layers of sediment cover the remains of organisms over time. nearly all fossils are found in sedimentary rock. metamorphic rocks do not contain fossils. fossil means preserved remains of ancient organisms. bones can be fossilized into sedimentary rocks. sedimentary rock can be changed into metamorphic rock. to cause an event means to make that event more likely. a process is made of a series of actions. if something causes a process then that something is required for that process. metamorphic is a class of rock. plants can be fossilized into sedimentary rocks. creating is similar to causing. something being destroyed means that something no longer exists. a measure of time is a length of time. destruction means to destroy. to be destroyed means to cease to exist. shells can be fossilized into sedimentary rocks. formation in the rock cycle means an old class of rocks changes into a new class of rocks. rock means stone. a cause of something is a reason for that something. sedimentary is a class of rock. After fossils are formed, which process is most likely to destroy them? answer in few words. Do not give explanation. 
    &
    \textcolor{red}{Erosion}
    &
    \textcolor{red}{Metamorphism is most likely to destroy fossils.}
    & 
    \textcolor{red}{The process that is most likely to destroy fossils after they are formed is metamorphosis.}
    &
    the rock cycle
    \\ \hline  
    15. moving is similar to travelling. slowly is similar to decreased rate. impact means effect. faster means increased rate. more quickly is similar to increased rate. force causes the speed of an object to increase / to decrease. as temperature increases , the speed of molecules will increase. as the amount of heat energy of a substance / of an object decreases , the amount of kinetic energy will decrease. a solid is a kind of substance. iron is a kind of solid. as the kinetic energy of particles increases , the particles will move faster. rate is similar to speed. increase means more. melting means a substance changes from a solid into a liquid by increasing heat energy. instantaneously means short in time. melting point means temperature at which a solid melts / above which a solid melts. rotating is similar to moving. quickly means in a short period of time. motion / movement means moving / to move. magnetism can affect objects that are not touching / without contact. temperature is a measure of heat energy. magnetism is a property of materials / objects and includes ordered values of nonmagnetic / magnetic. motion is a source of kinetic energy in an object. an example of an object with kinetic energy is an object in motion. melting point is a property of a substance / material. How are the particles in a block of iron affected when the block is melted? answer in few words. Do not give explanation. 
    &
    \textcolor{red}{The particles in a block of iron become free to move when the block is melted.}
    &
    \textcolor{ForestGreen}{When a block of iron is melted, its particles gain kinetic energy, move faster, and change from a solid to a liquid state.}
    & 
    \textcolor{ForestGreen}{The particles in a block of iron move faster and more freely when the block is melted.}
    &
    The particles move more rapidly.
    \\ \hline  
    16. magnifying glass is used to see small things by making objects appear bigger. light is a kind of wave. refraction happens when light passes from one medium to another medium. sight means vision. as the source of light becomes farther , that source of light will appear dimmer. visible light can be seen without using equipment. visible light is a kind of light. refraction causes something to appear / to look different. if an object is translucent , then some light will shine through that object. if an object reflects light toward the eye then that object can be seen. the ability to refract is a property of waves. refractions of light can be used to help a person see. refraction changes the length and the direction of light. visible means able to be seen. if an object reflects more light then that object is more easily seen. light travels between two different materials (air and glass). light means light energy. brightness means amount of light. shiny things / objects reflect light. reflection is when light is reflected from one object / one place to another object / another place. to get a better look at means to see better. opacity is a property of an object and includes ordered values of opaque / translucent / transparent. light reflecting off of an object causes that object to be visible to the observer. a mirror reflects light. refraction is a kind of process. When light reaches an eyeglass lens, the light is answer in few words. Do not give explanation. 
    &
    \textcolor{ForestGreen}{When light reaches an eyeglass lens, the light is refracted.}
    &
    \textcolor{ForestGreen}{When light reaches an eyeglass lens, the light is refracted, changing its direction and allowing the wearer to see more clearly.}
    & 
    \textcolor{ForestGreen}{When light reaches an eyeglass lens, it is refracted. Refraction is the bending of light as it passes from one medium to another. In the case of eyeglass lenses, the light is bent so that it focuses on the retina, which is the light-sensitive tissue at the back of the eye. This allows people with poor vision to see more clearly.}
    &
    refracted.
    \\ \hline  
    17.  alligators eat frogs / turtles / fish / mice / birds. a predator is a kind of animal. eating an animal / a plant means killing that animal / that plant. eating / taking in food is used to get nutrients / energy by animals / living things. role means function. role means niche. frogs eat insects. to prey on an animal means to eat that animal. food is what an animal eats. in the food chain process herbivores have the role of primary consumer which eats producers for food. frogs / turtles / fish / mice / birds are kinds of animals. eating / digestion is when an organism takes in nutrients from food into itself by eating. an ecosystem is a kind of habitat. a producer produces its own food. in the food chain process an animal has the role of consumer which eats producers / other animals for food. to be filled with means to contain. if an organism eats something then that something is a source of frood to that organism. food is a kind of resource. predators / carnivores / omnivores eat other animals. if something performs a task then that task is that something 's function. prey is a kind of food for predators. something is used for that something 's purpose. if two or more carnivores share the same ecosystem then those carnivores may share the same food source. to eat means to consume food. the amount of something is similar to the availability of something. American alligators are common in Florida wetlands. They are known to eat frogs, turtles, fish, mice, and birds. What role does an American alligator fill in its ecosystem? answer in few words. Do not give explanation. 
    &
    \textcolor{ForestGreen}{ The role of an American alligator in its ecosystem is that of a predator or top carnivore.}
    &
    \textcolor{ForestGreen}{American alligators are top predators in their ecosystem.}
    & 
    \textcolor{ForestGreen}{American alligators are apex predators in their ecosystem. They are at the top of the food chain and help to control the populations of other animals.}
    &
    consumer
    \\ \hline  
    18. the earth revolves around the sun. a complete revolution / orbit of a planet around its star takes 1 / one planetary year. planets in the solar system orbit the sun. 1 month is equal to 28-31 days. revolving around something means orbiting that something. a celestial body travelling around another celestial body means that celestial body completes a cycle around that other celestial body. measuring sometimes requires recording / learning an amount. to happen means to occur. period is similar to amount of time. the solar system is a kind of system. small stars burn for billions of years. to be made of something means to be formed of that something. earth is a kind of planet. planets orbit stars. revolving is a kind of action. a measure of time is a length of time. long means great in length. to cause means to make. approximately means about. a period in the periodic table means a row in the periodic table. celestial bodies move in orbits in space around other celestial bodies. amount is a property of something and includes ordered values of none / least / little / some / half / much / many / most / all. the sun is a kind of star. planets are found in space. to a human a time period of 1 century / generation to many centuries / generations is considered very long. About how long does it take Earth to make one revolution around the Sun? answer in few words. Do not give explanation. 
    &
    \textcolor{ForestGreen}{One year.}
    &
    \textcolor{ForestGreen}{It takes Earth approximately 365.25 days to complete one revolution around the Sun.}
    & 
    \textcolor{ForestGreen}{Earth takes about 365 days to make one revolution around the Sun.}
    &
    a year
    \\ \hline  
    19. iron is always magnetic. heat means temperature increases. iron nails are made of iron. heat is a kind of energy. the properties of something can be used to identify / used to describe that something. metal is a thermal conductor. electricity means electrical energy. conductivity is a measure of how easily electricity travels through a material. temperature / heat energy is a property of objects / weather and includes ordered values of cold / cool / warm / hot. magnetism is a kind of energy. thermal conduction is when heat flows from warmer object to cooler object. attract means move closer. if two objects / two materials have the opposite electrical charge , then those two objects / those two materials will attract each other when they are close to each other. magnetic energy is a kind of energy. magnetism is a kind of physical property. a electromagnet is formed by attaching an iron nail wrapped in a copper wire to a circuit. magnetic attraction pulls two objects together. magnetic force means magnetism. similar means in common. nonmetal means not metal. conductivity is a kind of physical property. iron is a kind of metal. metal is an electrical conductor. metals usually have higher / greater conductivity than nonmetals. heat means the transfer of thermal energy. Which object will conduct electricity, conduct heat, and be attracted to a magnet? answer in few words. Do not give explanation. 
    &
    \textcolor{ForestGreen}{Iron.}
    &
    \textcolor{ForestGreen}{An iron nail.}
    & 
    \textcolor{ForestGreen}{The object that will conduct electricity, conduct heat, and be attracted to a magnet is iron.}
    &
    iron nail
    \\ \hline  
    20. thermal conduction is when heat flows from warmer object to cooler object. warm means warm temperature. warm up means increase temperature. a source of heat can be used to heat something. a thermal energy conductor transfers heat from warmer objects / hotter objects to cooler objects. a heating element converts electrical energy / electricity into heat energy for heating. butter is a kind of substance. butter is a kind of solid below 32 degrees celsius. heat energy is synonymous with thermal energy. if something generates energy for something else then that something powers that something else. electrical current means electrical energy. if food is cooked then heat energy is added to that food. heat energy can be used to heat up an object. electrical current running through a wire causes that wire to heat up / to become warm. a toaster is made of a heating element for toasting. a piece is a part of a whole. heat means the transfer of thermal energy. a phase change is when matter / a substance changes from one state of matter into another state of matter. heat energy can change the state of matter. warm / becoming warm means heat is added. energy conversion is similar to energy transformation. when a cooler object touches a warmer object , thermal conduction occurs. a source of something increases the amount of that something. adding heat means increasing temperature. baking is when a human makes food by adding heat. What energy change takes place when a piece of bread is toasted in a toaster? answer in few words. Do not give explanation. 
    &
    \textcolor{ForestGreen}{Electrical energy to thermal energy}
    &
    \textcolor{ForestGreen}{Electrical energy is converted to thermal energy.}
    & 
    \textcolor{ForestGreen}{The energy change that takes place when a piece of bread is toasted in a toaster is thermal conduction.}
    &
    electrical energy to heat energy
    \\ \hline   
    21. temperature is a measure of heat energy. colder; cooler means lower in temperature. the sun is a source of heat energy called sunlight. the sun is a source of radiation; heat called sunlight. red dwarf stars are cooler in temperature than yellow dwarf stars. the sun is a kind of star. cool means cold. a star is a kind of celestial object / celestial body. colder means lower in temperature. heat means increase in temperature. temperature is a measure of heat energy. a star is very hot in temperature. red-dwarf stars are cooler than yellow-dwaf stars. red giant stars are cooler in temperature than main sequence stars. low is similar to decrease. red giant stars burn at a cooler temperature than main sequence stars. the sun is a source of radiation / heat called sunlight. the sun is a kind of yellow dwarf. cooler means colder. an increase is the opposite of a decrease. lower is similar to decreasing. the sun is a kind of star. to lower means to decrease. a star is a kind of celestial object; celestial body. cooling;colder means removing;reducing;decreasing heat;temperature. When compared to the Sun, red dwarf stars are answer in few words. Do not give explanation. 
    &
    \textcolor{ForestGreen}{cooler or colder in temperature.}
    &
    \textcolor{ForestGreen}{cooler.}
    & 
    \textcolor{ForestGreen}{Red dwarf stars are cooler than the Sun.}
    &
    cooler.
    \\ \hline  
    22. low is similar to decrease. lower is similar to decreasing. air contains gasses. carbon dioxide can be found in the air. atmosphere means air. the volume of something describes the amount of something. to lower means to decrease. when carbon dioxide in the atmosphere is absorbed by plants , the amount of carbon dioxide in the atmosphere is reduced in the atmosphere. low is the opposite of high. the atmosphere is made of air. 0036 is the smallest percent among 1288, 4, 23133, 78 percent. air is a kind of gas. average is similar to usual. carbon dioxide is more dense than the air. earth's atmosphere is a part of the earth. 0036 is the smalles percent among 1288, 4, 23133, 78 percent. carbon dioxide is a kind of gas. the earth's atmosphere is made of 036 percent carbon dioxide by volume. carbon dioxide is a kind of gas. co2 means carbon dioxide. carbon dioxide contains carbon. air is made of nitrogen, oxygen, carbon dioxide, water vapor, and argon. lower means less. lowest means least in value; amount; number. volume is similar to amount. Which of these gases, on average, has the lowest volume in the atmosphere of Earth? answer in few words. Do not give explanation. 
    &
    \textcolor{red}{Argon}
    &
    \textcolor{red}{Argon has the lowest volume.}
    & 
    \textcolor{ForestGreen}{Carbon dioxide (CO2) has the lowest volume in the atmosphere of Earth.}
    &
    carbon dioxide
    \\ \hline  
    23. a magnet attracts magnetic metals; ferromagnetic metals through magnetism. iron is always magnetic. a magnet will not attract nonmagnetic metals; nonmagnetic materials. if an object is made of a material then that object has the properties of that material. a magnet attracts magnetic metals through magnetism. magnetism is a property of materials; objects and includes ordered values of nonmagnetic; magnetic. a magnet is a kind of object. a magnet attracts magnetic metals. nails are often made of iron. objects are made of materials; substances; matter. objects are made of materials. iron is a kind of metal. soft iron; iron is always ferromagnetic; magnetic. nails are usually made of metal. iron nails are made of iron. nails are often made of iron. ferrous metals contain iron. if an object is made of a material then that object has the properties of that material. a nail is a kind of object. iron is a kind of element. a magnet is often made of magnetic metals; ferromagnetic metals. a magnet is magnetic. an electromagnet contains a wire; cylindrical ferrous metal. iron is a kind of metal. metals are made of metal. Which object most likely has magnetic properties? answer in few words. Do not give explanation. 
    &
    \textcolor{red}{Magnet.}
    &
    \textcolor{ForestGreen}{An iron nail most likely has magnetic properties.}
    & 
    \textcolor{ForestGreen}{The object most likely to have magnetic properties is iron.}
    &
    iron nail
    \\ \hline  
    24. the sun causes water to evaporate more quickly by adding heat. the sun is a source of heat energy called sunlight. if something is in the sunlight then that something will absorb solar energy. evaporation means a substance changes from a liquid into a gas by increasing heat energy. the sun is a source of light; light energy called sunlight. if water evaporates, then the water will disappear. boiling;evaporation means matter; a substance changes from a liquid into a gas by increasing heat energy. sunlight produces heat. evaporative means of evaporation. heat is a kind of heat energy. disappeared is similar to gone. puddles of water are outside during the day. if an object; something is in the sunlight then that object; that something will absorb solar energy. a star produces light and heat. if an object; a substance; a location absorbs solar energy then that object; that substance will increase in temperature. if a liquid disappears then that liquid probably evaporated. a puddle is a kind of body of water. the sun transfers solar energy; light energy; heat energy from itself to the planets;earth through sunlight. if something receives sunlight, then it will increase in temperature. heat means heat energy. being on something; placed in something; placed over something means touching that something. the sun is a source of solar energy called sunlight. if something receives sunlight, it will increase in temperature. a star produces light and heat. the sun is a source of light energy called sunlight. At 10 A.M. the Sun is shining on a puddle of water on the road. At 2 P.M. the Sun is still shining and the puddle has disappeared. What happened to the water? answer in few words. Do not give explanation. 
    &
    \textcolor{ForestGreen}{The water evaporated.}
    &
    \textcolor{ForestGreen}{The water evaporated due to sunlight and heat.}
    & 
    \textcolor{ForestGreen}{The water evaporated.}
    &
    It evaporated.
    \\ \hline  
    25. type of moon; kind of moon means moon phase. lunar phase is synonymous with moon phase. a moon phase occurs 28 days after the last time it occurred. a new moon is a kind of phase of the moon. a full moon is a kind of phase of the moon. a quater moon is a kind of moon phase. a last quarter is a kind of phase of the moon. the new moon is when the dark half of the moon faces the earth. the first quarter phase of the moon occurs after the new moon. the new moon is when the lighted half of the moon faces the sun. a moon is a kind of celestial object. a first quarter is a kind of phase of the moon. image means visual depiction of an object. a different moon phase occurs approximately once per week. months are a kind of unit for measuring time. the moon revolving around; orbiting the earth causes the phases of the moon. a type is synonymous with a kind. a different moon phase occurs approximately once per week. the moon is earth's moon. a moon is a kind of celestial object / celestial body. a period is a kind of measurement of time. the moon is a kind of moon. a phase of the moon is a kind of state. the new moon is when the moon is in between the sun and the earth. a moon is a kind of celestial object; body. For a science activity, Mia had to draw a picture of the Moon each night. If Mia drew a new moon, what type of moon did Mia draw about one week later? answer in few words. Do not give explanation. 
    &
    \textcolor{ForestGreen}{First Quarter}
    &
    \textcolor{ForestGreen}{first quarter moon}
    & 
    \textcolor{red}{I’m still learning to create images so I can’t help you with that yet.}
    &
    first quarter
    \\ \hline  
    26. to reduce means to decrease. liquid is a kind of state of matter. solid is a kind of state of matter. water is a kind of substance. water is in the liquid state , called liquid water , for temperatures between 0 and 100 c. water is a kind of substance. water is a kind of liquid at room temperature. a liquid is a kind of substance. a liquid is a kind of substance. as the temperature of a substance decreases , the molecules in that substance will move slower / closer. reducing is similar to slowing. to slow; to slow down decreases speed. water is a kind of matter. a solid is a kind of substance. a solid is a kind of substance. water is a kind of liquid at room temperature. liquid is a kind of state of matter. water is in the liquid state , called liquid water. as the temperature of a substance; an object decreases , the molecules in that substance will move slower; closer. water is in the liquid state , called liquid water , for temperatures between 0 c and 100 c. the motion of liquid water molecules slows. solid is a kind of state of matter. water is in the liquid state, called water. water is a kind of liquid. water is in the liquid state , called liquid water , for temperatures between 273; 32; 0 and 373; 212; 100 k; f; c. When the motion of liquid water molecules slows, what most likely happens? answer in few words. Do not give explanation. 
    &
    \textcolor{ForestGreen}{Freezing/condensation.}
    &
    \textcolor{ForestGreen}{When the motion of liquid water molecules slows, the water likely freezes and turns into ice (solid state).}
    & 
    \textcolor{ForestGreen}{The liquid water will turn into a solid. When the temperature of a liquid decreases, the molecules in the liquid slow down. As they slow down, they start to come closer together. Eventually, they form a solid. This is because the forces of attraction between the molecules are stronger than the kinetic energy of the molecules.}
    &
    The liquid water forms a solid.
    \\ \hline  
    27. water vapor condensing in clouds causes rain. precipitation is water that falls to the ground from clouds. raining is when rain falls from clouds to the earth;ground as a liquid. precipitation is when water falls from the sky. rainfall means precipitation. rain is a kind of water. clouds are formed by water vapor rising into colder regions of the atmosphere and condensing into rain droplets. rain; snow falls. clouds are formed by water vapor condensing. precipitation is when water falls from the sky. when warm air mass passes over cold air mass, the water vapor in the warm air mass will condense. clouds produce precipitation. rain is a kind of precipitation. rain is a kind of precipitation. precipitation is a kind of weather. a warm front is when warm air mass passes over a cold air mass. a cloud is a kind of weather. to happen means to occur. clouds produce precipitation. clouds are formed by water vapor rising into colder regions of the atmosphere and condensing. a warm air mass slowly passes over a cold air mass above a continent. clouds produce rain. a warm front is when warm air mass rises and passes over; collides a cold air mass. precipitation is when rain;snow;hail fall from clouds to the earth;ground. come down is similar to falling. Above a continent, a warm air mass slowly passes over a cold air mass. As the warm air begins to cool, clouds form. What will most likely happen next? answer in few words. Do not give explanation. 
    &   
    \textcolor{ForestGreen}{Rainfall.}
    &
    \textcolor{ForestGreen}{Precipitation, such as rain or snow, will likely occur.}
    & 
    \textcolor{ForestGreen}{Rain will most likely happen next.}
    &
    Rain will fall.
    \\ \hline  
    28. an animal is a kind of living thing. a human is a kind of living thing. an animal is a kind of living thing. a human is a kind of animal. to be used for is similar to allowing. if two objects are apart from each other then matter is between those two objects. vibrations are a source of sound. air is a vehicle for sound. a student is a kind of object. causing is similar to allowing. matter; molecules vibrating can cause sound. a person speaking is a source of sound. humans are a kind of organism. hearing is used for sensing sound by animals. a human is a kind of organism. a student is a kind of human. talking is when a human produces sound to communicate. human voice is a kind of sound. hearing is used for sensing sound by animals. speaking is a kind of communication. sound can be used for communication by animals. sound is a kind of air vibration. hearing is used for sensing sounds by animals. sound can travel through air by vibrating air. a human is a kind of animal. What allows two students standing ten feet apart to hear each other talk? answer in few words. Do not give explanation. 
    &
    \textcolor{ForestGreen}{Sound waves}
    &
    \textcolor{ForestGreen}{Air vibrations and hearing ability.}
    & 
    \textcolor{ForestGreen}{Air allows two students standing ten feet apart to hear each other talk.}
    &
    air vibrations
    \\ \hline  
    29. extreme heat and pressure compact sediments into sedimentary rocks. extreme heat and pressure compact sediments into sedimentary rocks. cementation means the process of cementing. sediment is formed by rocks weathering. sedimentary rocks are formed by sediment cementing together. sedimentary rocks are made of small particles. the sediment from the granite is buried and compacted. sedimentary rocks are formed in layers. sedimentary rocks are formed by sediment; materials that settle compacting; cementing together. compressing is similar to compacting. compressed rock means sedimentary rock. sedimentary rocks often have different colored layers. a rock has light-colored layers and dark-colored layers. forming sedimentary rocks requires compacting the materials. sedimentary rocks are formed by deposition; cementation. sediment is formed by rocks weathering. sedimentary rocks are formed by sediment compacting together. sedimentary rocks are formed by sediment compacting / cementing together. to compact something means to compress something. compacting is similar to pressing. gradual means slow; over time. sand is made of small particles. compaction means the process of compacting. to compact means to squeeze together. a rock is made up of small particles of sand. A student observed a rock made up of many small particles of sand arranged in light-colored layers and dark-colored layers. Which of the following statements describes how this type of rock most likely formed? answer in few words. Do not give explanation. 
    &
    \textcolor{ForestGreen}{This type of rock most likely formed through the process of sediment deposition, compaction, and cementation over time.}
    &
    \textcolor{ForestGreen}{This rock most likely formed through sedimentation, compaction, and cementation of sand particles in layered deposits.}
    & 
    \textcolor{ForestGreen}{The rock was formed by sediment compacting and cementing together.}
    &
    River sediments were slowly compacted and cemented together.
    \\ \hline  
    30. cooling;colder means removing;reducing;decreasing heat;temperature. colder; cooler means lower in temperature. march is cold in temperature in northern hemisphere. hurricanes usually happen between june and november in the northern hemisphere. cool means cold. an ocean is a source of heat and moisture for a hurricane. cooling means temperature decreases. as heat and moisture decreases , the strength of a hurricane will decrease. temperature; heat energy is a property of objects; weather and includes ordered values of cold; cool; warm; hot. colder means lower in temperature. cold means low in temperature. forming a hurricane requires certain water temperature. heat and moisture above the ocean cause the formation of a hurricane. an ocean is a kind of body of water. a hurricane is a kind of storm. cooling means decreasing heat / temperature. an ocean is a kind of body of water. heat and moisture over ocean cause the formation of a hurricane. hurricane is a kind of severe weather. cooler means colder. a hurricane has high wind speed. a hurricane has high wind speeds. cool water has low heat energy. cool ocean waters are bodies of cool water. atlantic ocean is a kind of ocean. Hurricane season for the Atlantic Ocean lasts from June to November. Which factor makes an Atlantic hurricane less likely to form during the month of March? answer in few words. Do not give explanation. 
    &
    \textcolor{ForestGreen}{Low water temperature}
    &
    \textcolor{ForestGreen}{Cold ocean temperatures.}
    & 
    \textcolor{ForestGreen}{The water temperature in the Atlantic Ocean is too cold in March to support the formation of a hurricane.}
    &
    cool ocean waters
    \\ \hline \hline
    \multicolumn{5}{|l|}{\textbf{Dataset: bAbI (task 15)}} 
    
    \\ \hline \hline 
    1. Given facts: Sheep are afraid of cats. Mice are afraid of cats. Jessica is a sheep. Cats are afraid of wolves. Wolves are afraid of mice. Gertrude is a sheep. Emily is a cat. Winona is a cat.

    Based on the given facts above, do a reasonable inference on this question using deductive reasoning: What is gertrude afraid of?
    &
    \textcolor{ForestGreen}{Based on the given facts, we can infer that Gertrude is also afraid of cats since sheep are afraid of cats and Gertrude is a sheep. There is no direct information provided that Gertrude is afraid of wolves or mice, so it would not be a reasonable inference to suggest that Gertrude is afraid of either of them.}
    &
    \textcolor{ForestGreen}{Based on the given facts, we can infer the following:}

    \textcolor{ForestGreen}{1) Sheep are afraid of cats.}
    
    \textcolor{ForestGreen}{2) Gertrude is a sheep.}
    
    \textcolor{ForestGreen}{Using deductive reasoning, we can conclude that Gertrude is afraid of cats.}
    & 
    \textcolor{ForestGreen}{Sure, I can help you with that. Here is the deductive reasoning process:}
    
    \textcolor{ForestGreen}{Premise 1: Sheep are afraid of cats.}
    
    \textcolor{ForestGreen}{Premise 2: Gertrude is a sheep.}
    
    \textcolor{ForestGreen}{Conclusion: Gertrude is afraid of cats.}
    
    \textcolor{ForestGreen}{This is a valid deductive argument because the conclusion follows logically from the premises. The premises state that sheep are afraid of cats, and Gertrude is a sheep. Therefore, it follows logically that Gertrude is afraid of cats.}
    &
    cat
    \\ \hline 
    2. Given facts: Sheep are afraid of wolves. Jessica is a sheep. Cats are afraid of sheep. Mice are afraid of cats. Wolves are afraid of mice. Gertrude is a mouse. Emily is a cat. Winona is a cat.

    Based on the given facts above, do a reasonable inference on this question using deductive reasoning: What is jessica afraid of?
    
    &
    \textcolor{ForestGreen}{Based on the given facts, Jessica is a sheep, and sheep are afraid of wolves. Therefore, it can be reasonably inferred that Jessica is afraid of wolves.}
    
    &
    \textcolor{ForestGreen}{Based on the given facts above, do a reasonable inference on this question using deductive reasoning: What is jessica afraid of?}
    
    \textcolor{ForestGreen}{Based on the given facts, we can deduce the following:}

    \textcolor{ForestGreen}{1) Sheep are afraid of wolves.}
    
    \textcolor{ForestGreen}{2) Jessica is a sheep.}
    
    \textcolor{ForestGreen}{Using deductive reasoning, we can infer that Jessica, being a sheep, is afraid of wolves.}
    & 
    \textcolor{red}{Sure, here is the deductive reasoning to answer the question "What is Jessica afraid of?" based on the given facts:}
    
    \textcolor{red}{Fact 1: Sheep are afraid of wolves.}
    
    \textcolor{red}{Fact 2: Jessica is a sheep.}
    
    \textcolor{red}{Inference 1: Jessica is afraid of wolves.}
    
    \textcolor{red}{Fact 3: Cats are afraid of sheep.}
    
    \textcolor{red}{Inference 2: Jessica is not afraid of cats.}
    
    \textcolor{red}{Fact 4: Mice are afraid of cats.}
    
    \textcolor{red}{Inference 3: Jessica is not afraid of mice.}
    
    \textcolor{red}{Fact 5: Wolves are afraid of mice.}
    
    \textcolor{red}{Inference 4: Jessica is not afraid of wolves.}
    
    \textcolor{red}{Conclusion: Jessica is afraid of nothing that is mentioned in the given facts.}
    
    \textcolor{red}{However, it is important to note that deductive reasoning is only as good as the premises on which it is based. If any of the facts in this case are incorrect, then the conclusion may also be incorrect. For example, if Jessica is actually a cat, then she would be afraid of mice.}
    &
    wolf
    \\ \hline  
    3. Given facts: Mice are afraid of cats. Gertrude is a mouse. Wolves are afraid of cats. Cats are afraid of sheep. Winona is a mouse. Sheep are afraid of mice. Jessica is a wolf. Emily is a cat.

    Based on the given facts above, do a reasonable inference on this question using deductive reasoning: What is winona afraid of?
    &
    \textcolor{ForestGreen}{Based on the given facts, Winona is a mouse, and mice are afraid of cats. Therefore, Winona is afraid of cats.}
    &
    \textcolor{ForestGreen}{Based on the given facts, we know that mice are afraid of cats. Since Winona is a mouse, we can deduce that Winona is afraid of cats using deductive reasoning.}
    & 
    \textcolor{ForestGreen}{Sure, I can help you with that. Here are the facts you provided:}
    
    \textcolor{ForestGreen}{-Mice are afraid of cats.}
    
    \textcolor{ForestGreen}{-Gertrude is a mouse.}
    
    \textcolor{ForestGreen}{-Wolves are afraid of cats.}
    
    \textcolor{ForestGreen}{-Cats are afraid of sheep.}
    
    \textcolor{ForestGreen}{-Winona is a mouse.}
    
    \textcolor{ForestGreen}{-Sheep are afraid of mice.}
    
    \textcolor{ForestGreen}{-Jessica is a wolf.}
    
    \textcolor{ForestGreen}{-Emily is a cat.}
    
    \textcolor{ForestGreen}{We can use deductive reasoning to infer what Winona is afraid of. We know that Winona is a mouse, and mice are afraid of cats. Therefore, Winona is afraid of cats.}
    &
    cat
    \\ \hline 
    4. Given facts: Wolves are afraid of sheep. Emily is a wolf. Mice are afraid of wolves. Sheep are afraid of wolves. Cats are afraid of wolves. Winona is a wolf. Jessica is a mouse. Gertrude is a wolf.

    Based on the given facts above, do a reasonable inference on this question using deductive reasoning: What is jessica afraid of?
    &
    \textcolor{ForestGreen}{Based on the given facts, Jessica is a mouse, and mice are afraid of wolves. Therefore, a reasonable inference using deductive reasoning would be that Jessica is afraid of wolves, including Emily, Winona, and Gertrude.}
    &
    \textcolor{ForestGreen}{Based on the given facts, Jessica is a mouse, and mice are afraid of wolves. Therefore, we can deduce that Jessica is afraid of wolves.}
    & 
    \textcolor{ForestGreen}{Here are the given facts:}
    
    \textcolor{ForestGreen}{-Wolves are afraid of sheep.}
    
    \textcolor{ForestGreen}{-Emily is a wolf.}
    
    \textcolor{ForestGreen}{-Mice are afraid of wolves.}
    
    \textcolor{ForestGreen}{-Sheep are afraid of wolves.}
    
    \textcolor{ForestGreen}{-Cats are afraid of wolves.}
    
    \textcolor{ForestGreen}{-Winona is a wolf.}
    
    \textcolor{ForestGreen}{-Jessica is a mouse.}
    
    \textcolor{ForestGreen}{-Gertrude is a wolf.}

    \textcolor{ForestGreen}{We can use deductive reasoning to infer what Jessica is afraid of. We know that Jessica is a mouse, and mice are afraid of wolves. We also know that wolves are afraid of sheep, but Jessica is not a sheep. Therefore, we can infer that Jessica is afraid of wolves.}
    &
    wolf
    \\ \hline 
    5. Given facts: Sheep are afraid of wolves. Cats are afraid of wolves. Mice are afraid of wolves. Emily is a cat. Wolves are afraid of sheep. Winona is a sheep. Gertrude is a mouse. Jessica is a mouse.

    Based on the given facts above, do a reasonable inference on this question using deductive reasoning: What is emily afraid of?
    &
    \textcolor{ForestGreen}{Based on the given facts, Emily is a cat, and cats are afraid of wolves. Additionally, it is stated that wolves are afraid of sheep, but there is no direct connection between Winona (the sheep) and Emily (the cat). Therefore, we cannot deduce that Emily is afraid of Winona.}

    \textcolor{ForestGreen}{Using deductive reasoning, we can infer that Emily is afraid of wolves. This is because all cats are afraid of wolves, and Emily is a cat.}
    &
    \textcolor{ForestGreen}{Based on the given facts, we can deduce the following:}

    \textcolor{ForestGreen}{1) Cats are afraid of wolves.}
    
    \textcolor{ForestGreen}{2) Emily is a cat.}
    
    \textcolor{ForestGreen}{Using deductive reasoning, we can infer that Emily is afraid of wolves.}
    & 
    \textcolor{ForestGreen}{Sure, I can help you with that. Here are the given facts:}
    
    \textcolor{ForestGreen}{-Sheep are afraid of wolves.}
    
    \textcolor{ForestGreen}{-Cats are afraid of wolves.}
    
    \textcolor{ForestGreen}{-Mice are afraid of wolves.}
    
    \textcolor{ForestGreen}{-Emily is a cat.}
    
    \textcolor{ForestGreen}{-Wolves are afraid of sheep.}
    
    \textcolor{ForestGreen}{-Winona is a sheep.}
    
    \textcolor{ForestGreen}{-Gertrude is a mouse.}
    
    \textcolor{ForestGreen}{-Jessica is a mouse.}
    
    \textcolor{ForestGreen}{We can use deductive reasoning to infer what Emily is afraid of. We know that Emily is a cat, and cats are afraid of wolves. Therefore, Emily is afraid of wolves.}
    &
    wolf
    \\ \hline 
    6. Given facts: Cats are afraid of wolves. Sheep are afraid of mice. Winona is a cat. Gertrude is a cat. Jessica is a sheep. Emily is a sheep. Wolves are afraid of sheep. Mice are afraid of wolves.

    Based on the given facts above, do a reasonable inference on this question using deductive reasoning: What is emily afraid of?

    &
    \textcolor{ForestGreen}{Based on the given facts, we can infer that Emily is afraid of mice.}

    \textcolor{ForestGreen}{We know that sheep are afraid of mice, and Emily is a sheep. Therefore, it can be reasonably deduced that Emily is afraid of mice.}
    &
    \textcolor{ForestGreen}{Based on the given facts, we know that sheep are afraid of mice. Since Emily is a sheep, we can deduce that Emily is afraid of mice.}
    & 
    \textcolor{red}{Sure, I can help you with that. Here are the given facts:}

    \textcolor{red}{-Cats are afraid of wolves.}
    
    \textcolor{red}{-Sheep are afraid of mice.}
    
    \textcolor{red}{-Winona is a cat.}
    
    \textcolor{red}{-Gertrude is a cat.}
    
    \textcolor{red}{-Jessica is a sheep.}
    
    \textcolor{red}{-Emily is a sheep.}
    
    \textcolor{red}{-Wolves are afraid of sheep.}
    
    \textcolor{red}{-Mice are afraid of wolves.}

    \textcolor{red}{We can use deductive reasoning to infer what Emily is afraid of. First, we can see that Emily is a sheep. Since sheep are afraid of mice, we can infer that Emily is afraid of mice. However, we also know that wolves are afraid of sheep. This means that Emily could also be afraid of wolves, since they are a threat to her.}

    \textcolor{red}{Therefore, the most reasonable inference is that Emily is afraid of both mice and wolves. This is because she is a sheep, and both mice and wolves are predators that could harm her.}
    &
    mouse
    \\ \hline  
    7. Given facts: Cats are afraid of wolves. Mice are afraid of wolves. Sheep are afraid of wolves. Emily is a mouse. Gertrude is a mouse. Jessica is a cat. Winona is a sheep. Wolves are afraid of mice.

    Based on the given facts above, do a reasonable inference on this question using deductive reasoning: What is emily afraid of?
    &
    \textcolor{ForestGreen}{Based on the given facts, we know that mice are afraid of wolves, but we do not know if Emily has encountered a wolf before. Therefore, we cannot make a certain conclusion about what Emily is afraid of.}

    \textcolor{ForestGreen}{However, we can make an inference that Emily is likely to be afraid of wolves, given that all mice are afraid of wolves. This assumption is based on deductive reasoning, which involves drawing conclusions by applying general principles to specific situations.}
    &
    \textcolor{ForestGreen}{Based on the given facts, we know that mice are afraid of wolves. Since Emily is a mouse, we can deduce that Emily is afraid of wolves using deductive reasoning.}
    & 
    \textcolor{ForestGreen}{Sure, I can help you with that.}
    
    \textcolor{ForestGreen}{Facts:}
        \textcolor{ForestGreen}{-Cats are afraid of wolves.}
        
        \textcolor{ForestGreen}{-Mice are afraid of wolves.}
        
        \textcolor{ForestGreen}{-Sheep are afraid of wolves.}
        
        \textcolor{ForestGreen}{-Emily is a mouse.}
        
        \textcolor{ForestGreen}{-Gertrude is a mouse.}
        
        \textcolor{ForestGreen}{-Jessica is a cat.}
        
        \textcolor{ForestGreen}{-Winona is a sheep.}
        
        \textcolor{ForestGreen}{-Wolves are afraid of mice.}

   \textcolor{ForestGreen}{ Question:}
   
   \textcolor{ForestGreen}{What is Emily afraid of?}

    \textcolor{ForestGreen}{Inference:}
    
    \textcolor{ForestGreen}{-Emily is a mouse.}
        
    \textcolor{ForestGreen}{-Mice are afraid of wolves.}
        
    \textcolor{ForestGreen}{-Therefore, Emily is afraid of wolves.}

    \textcolor{ForestGreen}{Reasoning:}

    \textcolor{ForestGreen}{The first fact tells us that mice are afraid of wolves. The second fact tells us that Emily is a mouse. By deductive reasoning, we can infer that Emily is afraid of wolves.}
    
    &
    wolf
    \\ \hline 
    8. Given facts: Cats are afraid of mice. Wolves are afraid of cats. Mice are afraid of sheep. Sheep are afraid of mice. Gertrude is a sheep. Winona is a mouse. Emily is a wolf. Jessica is a wolf.

    Based on the given facts above, do a reasonable inference on this question using deductive reasoning: What is winona afraid of?

    &
    \textcolor{red}{Based on the given facts and using deductive reasoning, we can determine what Winona the mouse is afraid of.}

    \textcolor{red}{From the facts, we know that cats are afraid of mice, wolves are afraid of cats, and sheep are afraid of mice. Since Winona is a mouse, it is reasonable to conclude that she is afraid of cats. This is because cats are known to be afraid of mice, and since wolves are afraid of cats, there is an indirect relationship between mice and wolves through cats.}

    \textcolor{red}{Therefore, Winona the mouse is likely afraid of cats.}
    
    &
    \textcolor{ForestGreen}{Based on the given facts, we know that mice are afraid of sheep. Since Winona is a mouse, we can deduce using deductive reasoning that Winona is afraid of sheep.}
    & 

    \textcolor{ForestGreen}{Sure, I can help you with that. Here are the steps involved in using deductive reasoning to answer the question:}

    \textcolor{ForestGreen}{- Identify the chain of relationships between the given facts.}

    \textcolor{ForestGreen}{- Determine the direction of the relationships.}

    \textcolor{ForestGreen}{- Follow the chain of relationships to reach a conclusion.}
    
    \textcolor{ForestGreen}{Here is the chain of relationships between the given facts:}

    \textcolor{ForestGreen}{- Cats are afraid of mice.}

    \textcolor{ForestGreen}{- Wolves are afraid of cats.}

    \textcolor{ForestGreen}{- Mice are afraid of sheep.}

    \textcolor{ForestGreen}{- Mice are afraid of sheep.}

    \textcolor{ForestGreen}{- Gertrude is a sheep.}

    \textcolor{ForestGreen}{- Winona is a mouse.}

    \textcolor{ForestGreen}{- Winona is a mouse.}

    \textcolor{ForestGreen}{- Jessica is a wolf.}

    \textcolor{ForestGreen}{The direction of the relationships is as follows:}

    \textcolor{ForestGreen}{- Cats are afraid of mice (mice are higher in the food chain than cats).}

    \textcolor{ForestGreen}{- Wolves are afraid of cats (cats are higher in the food chain than wolves).}

    \textcolor{ForestGreen}{- Wolves are afraid of cats (cats are higher in the food chain than wolves).}
    
    \textcolor{ForestGreen}{- Sheep are afraid of mice (mice are higher in the food chain than sheep).}

    \textcolor{ForestGreen}{Now, let's follow the chain of relationships to reach a conclusion. Winona is a mouse, and mice are afraid of sheep. Therefore, Winona is afraid of sheep.}
    
    &
    sheep
    \\ \hline  
    9. Given facts: Cats are afraid of wolves. Sheep are afraid of mice. Gertrude is a sheep. Emily is a sheep. Jessica is a cat. Wolves are afraid of cats. Mice are afraid of cats. Winona is a sheep.

    Based on the given facts above, do a reasonable inference on this question using deductive reasoning: What is winona afraid of?

    &
    \textcolor{ForestGreen}{Using deductive reasoning based on the given facts, we can infer that Winona, who is a sheep, is likely to be afraid of mice because sheep are afraid of mice. Therefore, Winona, being a sheep, shares the same characteristic as other sheep who are also afraid of mice.}
    &
    \textcolor{ForestGreen}{Based on the given facts, Winona is a sheep. It is stated that sheep are afraid of mice. Therefore, using deductive reasoning, we can infer that Winona is afraid of mice.}
    & 
    \textcolor{ForestGreen}{Sure, I can help you with that. Here are the given facts:}

    \textcolor{ForestGreen}{- Cats are afraid of wolves.}

    \textcolor{ForestGreen}{- Sheep are afraid of mice.}

    \textcolor{ForestGreen}{- Gertrude is a sheep.}

    \textcolor{ForestGreen}{- Emily is a sheep.}

    \textcolor{ForestGreen}{- Jessica is a cat.}

    \textcolor{ForestGreen}{- Jessica is a cat.}

    \textcolor{ForestGreen}{- Mice are afraid of cats.}

    \textcolor{ForestGreen}{- Winona is a sheep.}

    \textcolor{ForestGreen}{We can use deductive reasoning to infer what Winona is afraid of. We know that Winona is a sheep, and sheep are afraid of mice. Therefore, Winona is afraid of mice.}
    &
    mouse
    \\ \hline 
    10. Given facts: Sheep are afraid of cats. Mice are afraid of cats. Jessica is a sheep. Cats are afraid of wolves. Wolves are afraid of mice. Gertrude is a sheep. Emily is a cat. Winona is a cat.

    Based on the given facts above, do a reasonable inference on this question using deductive reasoning: What is emily afraid of?

    &
    \textcolor{ForestGreen}{Based on the given facts, we know that Emily is a cat and cats are afraid of wolves. Therefore, we can infer that Emily is afraid of wolves. However, we cannot directly infer that Emily is afraid of mice or that she is not afraid of sheep since these relationships were not given in the premises.}
    &
    \textcolor{ForestGreen}{Based on the given facts, we can deduce the following:}

    \textcolor{ForestGreen}{1) Sheep are afraid of cats.}
    
    \textcolor{ForestGreen}{2) Mice are afraid of cats.}
    
    \textcolor{ForestGreen}{3) Jessica is a sheep.}
    
    \textcolor{ForestGreen}{4) Cats are afraid of wolves.}
    
    \textcolor{ForestGreen}{5) Wolves are afraid of mice.}
    
    \textcolor{ForestGreen}{6) Gertrude is a sheep.}
    
    \textcolor{ForestGreen}{7) Emily is a cat.}
    
    \textcolor{ForestGreen}{8) Winona is a cat.}
    
    \textcolor{ForestGreen}{From fact 4, we can infer that Emily, being a cat, is afraid of wolves.}
    & 
    \textcolor{ForestGreen}{Sure, I can help you with that. Here is the deductive reasoning for the question:}

    \textcolor{ForestGreen}{- Sheep are afraid of cats.}

    \textcolor{ForestGreen}{- Mice are afraid of cats.}

    \textcolor{ForestGreen}{- Jessica is a sheep.}

    \textcolor{ForestGreen}{- Cats are afraid of wolves.}

    \textcolor{ForestGreen}{- Wolves are afraid of mice.}

    \textcolor{ForestGreen}{- Gertrude is a sheep.}

    \textcolor{ForestGreen}{- Emily is a cat.}

    \textcolor{ForestGreen}{- Winona is a cat.}

    \textcolor{ForestGreen}{What is Emily afraid of?}

    \textcolor{ForestGreen}{- Emily is a cat, and cats are afraid of wolves.}

    \textcolor{ForestGreen}{-Wolves are afraid of mice, but Emily is not a mouse.}

    \textcolor{ForestGreen}{- Therefore, Emily is afraid of wolves.}

    \textcolor{ForestGreen}{This is a reasonable inference because it is based on the given facts and the logical relationships between them. It is also consistent with our understanding of the world, as we know that cats are generally afraid of larger predators.}

    \textcolor{ForestGreen}{\textbf{Here are some other possible inferences that could be made:}}

    \textcolor{ForestGreen}{\textbf{- Emily is afraid of cats that are not her friends.}}

    \textcolor{ForestGreen}{\textbf{- Emily is afraid of being abandoned.}}

    \textcolor{ForestGreen}{\textbf{- Emily is afraid of change.}}

    \textcolor{ForestGreen}{\textbf{These inferences are less certain than the first one, but they are still possible based on the given facts. Ultimately, the best way to know what Emily is afraid of is to ask her.}}
    
    &
    wolf
    \\ \hline  
    11. Given facts: Wolves are afraid of mice. Mice are afraid of sheep. Sheep are afraid of wolves. Emily is a mouse. Jessica is a sheep. Gertrude is a sheep. Cats are afraid of sheep. Winona is a wolf.

    Based on the given facts above, do a reasonable inference on this question using deductive reasoning: What is gertrude afraid of?
    &
    \textcolor{ForestGreen}{Based on the given facts, Gertrude is a sheep, and sheep are afraid of wolves. Therefore, it can be inferred that Gertrude is afraid of wolves like all other sheep.}
    &
    \textcolor{ForestGreen}{Based on the given facts:}

    \textcolor{ForestGreen}{1) Wolves are afraid of mice.}
    
    \textcolor{ForestGreen}{2) Mice are afraid of sheep.}
    
    \textcolor{ForestGreen}{3) Sheep are afraid of wolves.}
    
    \textcolor{ForestGreen}{4) Emily is a mouse.}
    
    \textcolor{ForestGreen}{5) Jessica is a sheep.}
    
    \textcolor{ForestGreen}{6) Gertrude is a sheep.}
    
    \textcolor{ForestGreen}{7) Cats are afraid of sheep.}
    
    \textcolor{ForestGreen}{8) Winona is a wolf.}
    
    \textcolor{ForestGreen}{9) We can deduce the following:}
    
    \textcolor{ForestGreen}{Since Gertrude is a sheep (Fact 6) and sheep are afraid of wolves (Fact 3), we can infer that Gertrude is afraid of wolves.}
    & 
    \textcolor{ForestGreen}{Inference: Gertrude is afraid of wolves.}

    \textcolor{ForestGreen}{Reasoning:}

    \textcolor{ForestGreen}{- Wolves are afraid of mice.}

    \textcolor{ForestGreen}{- Mice are afraid of sheep.}

    \textcolor{ForestGreen}{- Sheep are afraid of wolves.}

    \textcolor{ForestGreen}{- Gertrude is a sheep.}

    \textcolor{ForestGreen}{Therefore, Gertrude is afraid of wolves.}
    
    &
    wolf
    \\ \hline  
    12. Given facts: Cats are afraid of mice. Mice are afraid of sheep. Jessica is a mouse. Emily is a cat. Gertrude is a cat. Wolves are afraid of cats. Sheep are afraid of wolves. Winona is a cat.

    Based on the given facts above, do a reasonable inference on this question using deductive reasoning: What is winona afraid of?
    &
    
    \textcolor{red}{Using deductive reasoning and the given facts, we can determine what Winona, who is a cat, might be afraid of.}

    \textcolor{red}{We know that wolves are afraid of cats, and sheep are afraid of wolves. However, we do not have any information about Winona's fear of sheep or mice.}
    
    \textcolor{red}{Since cats are afraid of mice, and Jessica is a mouse, it is unlikely that Winona is afraid of Jessica. Similarly, mice are afraid of sheep, but there is no evidence to suggest that cats share this fear. Therefore, it is unlikely that Winona is afraid of sheep.}
    
    \textcolor{red}{Based on the given facts, the only reasonable inference we can make is that Winona is not afraid of mice or sheep. However, since wolves are afraid of cats, it is possible that Winona is afraid of wolves.}
    
    \textcolor{red}{Therefore, a reasonable inference is that Winona may be afraid of wolves.}
    &
    \textcolor{ForestGreen}{Based on the given facts, we know that cats are afraid of mice. Since Winona is a cat, we can infer using deductive reasoning that Winona is afraid of mice.}
    & 
    \textcolor{ForestGreen}{Using deductive reasoning, we can infer that Winona is afraid of mice.}

    \textcolor{ForestGreen}{Here is the chain of reasoning:}

    \textcolor{ForestGreen}{- Cats are afraid of mice.}

    \textcolor{ForestGreen}{- Winona is a cat.}

    \textcolor{ForestGreen}{- Therefore, Winona is afraid of mice.}

    \textcolor{ForestGreen}{We can also infer that Winona is not afraid of sheep, wolves, or other cats. This is because the facts state that mice are afraid of sheep, wolves are afraid of cats, and cats are afraid of mice. Therefore, if Winona is afraid of any of these animals, she would have to be afraid of her own kind, which is not possible.}

    \textcolor{ForestGreen}{The only animal that is not mentioned in the facts that Winona is afraid of is Jessica, the mouse. However, we cannot infer that Winona is afraid of Jessica simply because Jessica is a mouse. After all, cats are not always afraid of mice. For example, Emily, another cat, is not afraid of Jessica.}

    \textcolor{ForestGreen}{Therefore, the only reasonable inference that we can make about what Winona is afraid of is mice.}
    &
    mouse
    \\ \hline  
    13. Given facts: Sheep are afraid of cats. Wolves are afraid of cats. Gertrude is a sheep. Jessica is a sheep. Winona is a sheep. Mice are afraid of sheep. Cats are afraid of sheep. Emily is a cat.

    Based on the given facts above, do a reasonable inference on this question using deductive reasoning: What is gertrude afraid of?

    &
    \textcolor{ForestGreen}{Based on the given facts, Gertrude is a sheep, and sheep are afraid of cats. Therefore, we can deduce that Gertrude is afraid of cats.}
    &
    \textcolor{ForestGreen}{Based on the given facts, we can deduce the following:}

    \textcolor{ForestGreen}{1) Sheep are afraid of cats.}
    
    \textcolor{ForestGreen}{2) Gertrude is a sheep.}

    \textcolor{ForestGreen}{Using deductive reasoning, we can infer that Gertrude, being a sheep, is afraid of cats.}
    & 
    \textcolor{ForestGreen}{Sure, I can help you with that. Here is the deductive reasoning process:}

    \textcolor{ForestGreen}{Premise 1: Sheep are afraid of cats.}

    \textcolor{ForestGreen}{Premise 2: Gertrude is a sheep.}

    \textcolor{ForestGreen}{Conclusion: Gertrude is afraid of cats.}

    \textcolor{ForestGreen}{This conclusion is supported by the two premises. The first premise tells us that sheep are afraid of cats, and the second premise tells us that Gertrude is a sheep. Therefore, it follows that Gertrude is afraid of cats.}
    
    &
    cat
    \\ \hline  
    14. Given facts: Sheep are afraid of wolves. Mice are afraid of wolves. Wolves are afraid of cats. Emily is a mouse. Cats are afraid of sheep. Gertrude is a wolf. Jessica is a wolf. Winona is a cat.

    Based on the given facts above, do a reasonable inference on this question using deductive reasoning: What is jessica afraid of?

    &
    \textcolor{ForestGreen}{Based on the given facts, we can infer that Jessica is also afraid of cats since wolves are afraid of cats, and Jessica is a wolf. Therefore, Jessica is afraid of cats.}
    &
    \textcolor{ForestGreen}{Based on the given facts, we know that:}

    \textcolor{ForestGreen}{1) Sheep are afraid of wolves.}
    
    \textcolor{ForestGreen}{2) Mice are afraid of wolves.}
    
    \textcolor{ForestGreen}{3) Wolves are afraid of cats.}
    
    \textcolor{ForestGreen}{4) Emily is a mouse.}
    
    \textcolor{ForestGreen}{5) Cats are afraid of sheep.}
    
    \textcolor{ForestGreen}{6) Gertrude is a wolf.}
    
    \textcolor{ForestGreen}{7) Jessica is a wolf.}
    
    \textcolor{ForestGreen}{8) Winona is a cat.}

    \textcolor{ForestGreen}{Using deductive reasoning, we can infer the following:
    Since Jessica is a wolf (Fact 7) and wolves are afraid of cats (Fact 3), Jessica is afraid of cats.}
    & 
     \textcolor{ForestGreen}{Based on the given facts, we can use deductive reasoning to infer that Jessica is afraid of cats.}

     \textcolor{ForestGreen}{Here is the reasoning:}

     \textcolor{ForestGreen}{- Wolves are afraid of cats.}

     \textcolor{ForestGreen}{- Jessica is a wolf.}

     \textcolor{ForestGreen}{- Therefore, Jessica is afraid of cats.}

     \textcolor{ForestGreen}{It is important to note that this is just an inference. We cannot be 100\% sure that Jessica is afraid of cats, but the given facts make it a reasonable conclusion.}

    &
    cat
    \\ \hline  
    15. Given facts: Mice are afraid of cats. Jessica is a mouse. Cats are afraid of sheep. Wolves are afraid of mice. Emily is a mouse. Sheep are afraid of mice. Gertrude is a wolf. Winona is a mouse.

    Based on the given facts above, do a reasonable inference on this question using deductive reasoning: What is emily afraid of?

    &
    \textcolor{ForestGreen}{Based on the given facts and using deductive reasoning, we can determine what Emily, who is a mouse, is afraid of. Let's analyze the information:}

   \textcolor{ForestGreen}{Mice are afraid of cats.}
   
   \textcolor{ForestGreen}{Wolves are afraid of mice.}
   
   \textcolor{ForestGreen}{Sheep are afraid of mice.}

    \textcolor{ForestGreen}{From fact 1, we know that mice, including Emily, are afraid of cats. However, since there is no direct information about Emily's fear beyond that, we cannot make a conclusive inference about what specifically Emily is afraid of other than cats.}

    \textcolor{ForestGreen}{Therefore, based on the provided facts and deductive reasoning, we can infer that Emily, being a mouse, is afraid of cats.}
    
    &
    \textcolor{ForestGreen}{Based on the given facts, we know that mice are afraid of cats. Since Emily is a mouse, we can deduce that Emily is afraid of cats.}
    & 
    \textcolor{ForestGreen}{Based on the given facts, we can use deductive reasoning to infer that Emily is afraid of cats.}

    \textcolor{ForestGreen}{Here is the reasoning:}

    \textcolor{ForestGreen}{- Mice are afraid of cats.}

    \textcolor{ForestGreen}{- Emily is a mouse.}

    \textcolor{ForestGreen}{- Therefore, Emily is afraid of cats.}

    \textcolor{ForestGreen}{We can also infer that Emily is afraid of wolves, but this is not as certain as the first inference. This is because we only know that wolves are afraid of mice, not that they are afraid of mice specifically. However, it is reasonable to assume that Emily would be afraid of wolves because they are predators that could potentially harm her.}

    \textcolor{ForestGreen}{The other facts in the prompt are not relevant to the question of what Emily is afraid of. For example, we know that Jessica is a mouse and that Winona is a mouse, but this does not tell us anything about what Emily is afraid of.}

    \textcolor{ForestGreen}{Therefore, the most reasonable inference based on the given facts is that Emily is afraid of cats.}
    &
    cat
    \\ \hline 
    16. Given facts: Sheep are afraid of wolves. Mice are afraid of wolves. Winona is a mouse. Jessica is a sheep. Cats are afraid of sheep. Gertrude is a mouse. Emily is a sheep. Wolves are afraid of mice.

    Based on the given facts above, do a reasonable inference on this question using deductive reasoning: What is jessica afraid of?

    &
    \textcolor{ForestGreen}{Based on the given facts and using deductive reasoning, we can determine what Jessica, the sheep, is afraid of.}

    \textcolor{ForestGreen}{We know that sheep are afraid of wolves and that Jessica is a sheep. Therefore, it can be inferred that Jessica is afraid of wolves.}
    
    &
    \textcolor{ForestGreen}{Based on the given facts, we know that sheep are afraid of wolves. Jessica is a sheep. Therefore, using deductive reasoning, we can infer that Jessica is afraid of wolves.}
    & 
    \textcolor{ForestGreen}{Inference: Jessica is afraid of wolves.}

    \textcolor{ForestGreen}{Reasoning:}

    \textcolor{ForestGreen}{- Sheep are afraid of wolves.}

    \textcolor{ForestGreen}{- Jessica is a sheep.}

    \textcolor{ForestGreen}{- Therefore, Jessica is afraid of wolves.}

    \textcolor{ForestGreen}{Conclusion:}

    \textcolor{ForestGreen}{Based on the given facts and supporting evidence, it is reasonable to infer that Jessica is afraid of wolves. This is because sheep are afraid of wolves, and Jessica is a sheep. Additionally, both Winona and Gertrude are mice, and mice are afraid of wolves. This further supports the inference that Jessica is afraid of wolves, as she is a sheep and mice are afraid of wolves. Finally, wolves are afraid of mice, but Jessica is not a mouse. This suggests that Jessica is not afraid of mice, but rather of something else that is a threat to sheep, such as wolves.}

    \textcolor{ForestGreen}{Alternative inferences:}

    \textcolor{ForestGreen}{It is also possible to infer that Jessica is afraid of cats, as cats are afraid of sheep. However, this inference is less likely than the inference that Jessica is afraid of wolves, as there is more evidence to support the latter inference. Additionally, the inference that Jessica is afraid of cats is not supported by the fact that wolves are afraid of mice. This suggests that Jessica is not afraid of something that is a threat to mice, but rather of something that is a threat to sheep.}

    &
    wolf
    \\ \hline  
    17. Given facts: Cats are afraid of sheep. Sheep are afraid of wolves. Wolves are afraid of mice. Emily is a wolf. Jessica is a sheep. Gertrude is a sheep. Mice are afraid of cats. Winona is a wolf.

    Based on the given facts above, do a reasonable inference on this question using deductive reasoning: What is jessica afraid of?
    &
    \textcolor{ForestGreen}{Based on the given facts and using deductive reasoning, we can determine what Jessica, the sheep, is afraid of.}

    \textcolor{ForestGreen}{Cats are afraid of sheep.}
    
    \textcolor{ForestGreen}{Sheep are afraid of wolves.}

    \textcolor{ForestGreen}{Given these facts, we can deduce that Jessica, being a sheep, is afraid of wolves.}
    &
    \textcolor{ForestGreen}{Based on the given facts, Jessica is a sheep and sheep are afraid of wolves. Therefore, it can be inferred that Jessica is afraid of wolves.}
    & 
    \textcolor{ForestGreen}{Sure, I can help you with that.}

    \textcolor{ForestGreen}{Here is the deductive reasoning process:}

    \textcolor{ForestGreen}{Premise 1: Cats are afraid of sheep.}

    \textcolor{ForestGreen}{Premise 2: Sheep are afraid of wolves.}

    \textcolor{ForestGreen}{Premise 3: Wolves are afraid of mice.}

    \textcolor{ForestGreen}{Premise 4: Jessica is a sheep.}

    \textcolor{ForestGreen}{Conclusion: Jessica is afraid of wolves.}

    \textcolor{ForestGreen}{The conclusion follows logically from the premises because Jessica is a sheep, and sheep are afraid of wolves. Therefore, Jessica is afraid of wolves.}
    &
    wolf
    \\ \hline 
    18. Given facts: Mice are afraid of sheep. Wolves are afraid of sheep. Cats are afraid of wolves. Jessica is a mouse. Gertrude is a wolf. Emily is a mouse. Winona is a wolf. Sheep are afraid of wolves.

    Based on the given facts above, do a reasonable inference on this question using deductive reasoning: What is jessica afraid of?
    &
    \textcolor{red}{Based on the given facts and using deductive reasoning, we can infer what Jessica, the mouse, is afraid of. }

    \textcolor{red}{We know that mice are afraid of sheep, and we also know that sheep are afraid of wolves. Since wolves are afraid of sheep, it means that sheep are seen as a threat by wolves. }

    \textcolor{red}{Therefore, based on the transitive property of fear, we can conclude that Jessica, the mouse, is afraid of wolves.}

    &
    \textcolor{ForestGreen}{Based on the given facts:}

    \textcolor{ForestGreen}{1) Mice are afraid of sheep.}
    
    \textcolor{ForestGreen}{2) Jessica is a mouse.}
    
    \textcolor{ForestGreen}{Using deductive reasoning, we can infer that Jessica, being a mouse, is afraid of sheep.}
    & 
    \textcolor{ForestGreen}{Sure, I can help you with that.}

    \textcolor{ForestGreen}{Using deductive reasoning, we can infer that Jessica is afraid of sheep. Here is the reasoning:}

    \textcolor{ForestGreen}{- Mice are afraid of sheep.}

    \textcolor{ForestGreen}{- Jessica is a mouse.}

    \textcolor{ForestGreen}{- Therefore, Jessica is afraid of sheep.}

    \textcolor{ForestGreen}{\textbf{We can also infer that Jessica is afraid of wolves, because wolves are afraid of sheep. However, this inference is not as strong as the first one, because Jessica is not a wolf.}}

    &
    sheep
    
    \\ \hline  
    19. Given facts: Mice are afraid of wolves. Jessica is a mouse. Emily is a mouse. Wolves are afraid of sheep. Winona is a wolf. Sheep are afraid of mice. Gertrude is a mouse. Cats are afraid of wolves.

    Based on the given facts above, do a reasonable inference on this question using deductive reasoning: What is jessica afraid of?

    &
    \textcolor{ForestGreen}{Based on the given facts, Jessica is a mouse and mice are afraid of cats and sheep. However, since it is also stated that cats are afraid of wolves and wolves are afraid of sheep, it can be inferred that wolves are a potential threat to cats, and by extension, to mice like Jessica.}

    \textcolor{ForestGreen}{Therefore, it can be reasonably inferred that Jessica is afraid of wolves.}
    &
    \textcolor{ForestGreen}{Given the facts:}

    \textcolor{ForestGreen}{1) Mice are afraid of wolves.}
    
    \textcolor{ForestGreen}{2) Jessica is a mouse.}
    
    \textcolor{ForestGreen}{Using deductive reasoning, we can infer that Jessica, being a mouse, is afraid of wolves.}
    & 
    \textcolor{ForestGreen}{Sure, I can help you with that.}

    \textcolor{ForestGreen}{Here are the given facts:}

    \textcolor{ForestGreen}{- Mice are afraid of wolves.}

    \textcolor{ForestGreen}{- Jessica is a mouse.}

    \textcolor{ForestGreen}{- Emily is a mouse.}

    \textcolor{ForestGreen}{- Wolves are afraid of sheep.}

    \textcolor{ForestGreen}{- Winona is a wolf.}

    \textcolor{ForestGreen}{- Sheep are afraid of mice.}

    \textcolor{ForestGreen}{- Gertrude is a mouse.}

    \textcolor{ForestGreen}{- Cats are afraid of wolves.}

    \textcolor{ForestGreen}{We can use deductive reasoning to infer what Jessica is afraid of. We know that Jessica is a mouse, and mice are afraid of wolves. Therefore, Jessica is afraid of wolves.}
    
    &
    wolf
    \\ \hline 
    20. Given facts: Wolves are afraid of sheep. Cats are afraid of wolves. Winona is a wolf. Jessica is a wolf. Mice are afraid of cats. Gertrude is a cat. Sheep are afraid of mice. Emily is a mouse.

    Based on the given facts above, do a reasonable inference on this question using deductive reasoning: What is winona afraid of?
    &
    \textcolor{ForestGreen}{Based on the given facts, Winona is a wolf and wolves are afraid of sheep. Therefore, Winona is most likely afraid of sheep.}
    &
    \textcolor{ForestGreen}{Based on the given facts, we can deduce the following:}

    \textcolor{ForestGreen}{1) Wolves are afraid of sheep.}
    
    \textcolor{ForestGreen}{2) Winona is a wolf.}

    \textcolor{ForestGreen}{Combining these two facts, we can infer that Winona is afraid of sheep.}
    & 
    \textcolor{red}{Sure, I can help you with that. Here are the facts you provided:}

    \textcolor{red}{- Wolves are afraid of sheep.}

    \textcolor{red}{- Cats are afraid of wolves.}

    \textcolor{red}{- Winona is a wolf.}

    \textcolor{red}{- Jessica is a wolf.}
    
    \textcolor{red}{- Mice are afraid of cats.}

    \textcolor{red}{- Gertrude is a cat.}

    \textcolor{red}{- Sheep are afraid of mice.}

    \textcolor{red}{- Emily is a mouse.}

    \textcolor{red}{We can use deductive reasoning to infer what Winona is afraid of. We know that Winona is a wolf, and wolves are afraid of sheep. Therefore, Winona is afraid of sheep.}

    \textcolor{red}{However, we cannot say for certain what Winona is afraid of. It is possible that she is afraid of other things as well, such as mice or the dark. We would need more information to make a definitive conclusion.}

    \textcolor{red}{Here is the deductive reasoning process in more detail:}

    \textcolor{red}{- Wolves are afraid of sheep.}

    \textcolor{red}{- Winona is a wolf.}

    \textcolor{red}{- Therefore, Winona is afraid of sheep.}

    \textcolor{red}{- Cats are afraid of wolves.}

    \textcolor{red}{- Winona is a wolf.}

    \textcolor{red}{- Therefore, Winona is afraid of cats.}

    \textcolor{red}{As I mentioned before, we cannot say for certain what Winona is afraid of. It is possible that she is afraid of other things as well. However, based on the information we have, it is reasonable to infer that Winona is afraid of sheep and cats.}
    
    &
    sheep
    \\ \hline 
    21. Given facts: Mice are afraid of wolves. Wolves are afraid of sheep. Winona is a mouse. Cats are afraid of mice. Emily is a mouse. Jessica is a wolf. Gertrude is a wolf. Sheep are afraid of mice.

    Based on the given facts above, do a reasonable inference on this question using deductive reasoning: What is winona afraid of?

    &
    \textcolor{red}{Based on the given facts and using deductive reasoning, we can infer that Winona, who is a mouse, is afraid of cats. }

    \textcolor{red}{The facts state that cats are afraid of mice, and Winona is a mouse. Therefore, it can be reasonably deduced that Winona would be afraid of cats.}
    
    &
    \textcolor{ForestGreen}{Based on the given facts, we know that mice are afraid of wolves. Since Winona is a mouse, we can use deductive reasoning to infer that Winona is afraid of wolves.}
    & 
    \textcolor{red}{Inference: Winona is afraid of wolves and sheep.}

    \textcolor{red}{Reasoning:}

    \textcolor{red}{- Mice are afraid of wolves.}

    \textcolor{red}{- Wolves are afraid of sheep.}

    \textcolor{red}{- Winona is a mouse.}

    \textcolor{red}{Therefore, Winona is afraid of wolves.}

    \textcolor{red}{Additionally, we know that sheep are afraid of mice, so Winona is also afraid of sheep.}

    \textcolor{red}{Alternative inference: Winona is afraid of cats.}

    \textcolor{red}{Reasoning:}

    \textcolor{red}{- Mice are afraid of cats.}

    \textcolor{red}{- Winona is a mouse.}

    \textcolor{red}{Therefore, Winona is afraid of cats.}

    \textcolor{red}{However, this inference is less likely because we know that Winona is afraid of wolves and sheep, which are both larger and more dangerous than cats. It is more likely that Winona would be afraid of the predators that are more likely to eat her.}

    \textcolor{red}{Conclusion: The most likely inference is that Winona is afraid of wolves and sheep.}
    &
    wolf
    \\ \hline 
    22. Given facts: Wolves are afraid of mice. Mice are afraid of sheep. Cats are afraid of wolves. Winona is a wolf. Sheep are afraid of mice. Emily is a mouse. Jessica is a mouse. Gertrude is a cat.

    Based on the given facts above, do a reasonable inference on this question using deductive reasoning: What is emily afraid of?
    &
    \textcolor{ForestGreen}{According to the given facts, Emily is a mouse, and mice are afraid of sheep. Therefore, Emily is likely afraid of sheep.}
    &
    \textcolor{ForestGreen}{Based on the given facts, we know that mice are afraid of sheep. Since Emily is a mouse, we can deduce that Emily is afraid of sheep.}
    & 
    \textcolor{ForestGreen}{Sure, I can help you with that. Here is the deductive reasoning process to answer the question "What is Emily afraid of?"}

    \textcolor{ForestGreen}{Fact 1: Wolves are afraid of mice.}

    \textcolor{ForestGreen}{Fact 2: Mice are afraid of sheep.}

    \textcolor{ForestGreen}{Fact 3: Cats are afraid of wolves.}

    \textcolor{ForestGreen}{Fact 4: Winona is a wolf.}

    \textcolor{ForestGreen}{Fact 5: Sheep are afraid of mice.}

    \textcolor{ForestGreen}{Fact 6: Emily is a mouse.}

    \textcolor{ForestGreen}{Fact 7: Jessica is a mouse.}

    \textcolor{ForestGreen}{Fact 8: Gertrude is a cat.}

    \textcolor{ForestGreen}{Inference: Emily is afraid of sheep.}

    &
    sheep
    \\ \hline  
    23. Given facts: Mice are afraid of wolves. Sheep are afraid of mice. Wolves are afraid of sheep. Gertrude is a mouse. Cats are afraid of sheep. Winona is a cat. Jessica is a wolf. Emily is a mouse.

    Based on the given facts above, do a reasonable inference on this question using deductive reasoning: What is gertrude afraid of?
    &
    \textcolor{ForestGreen}{Based on the given facts, Gertrude is a mouse, and mice are afraid of wolves. Therefore, it can be inferred through deductive reasoning that Gertrude is afraid of wolves.}
    &
    \textcolor{ForestGreen}{Based on the given facts, we know that Gertrude is a mouse. It is stated that mice are afraid of wolves. Therefore, using deductive reasoning, we can infer that Gertrude is afraid of wolves.}
    & 
    \textcolor{ForestGreen}{Sure, I can help you with that. Here is the deductive reasoning to answer the question:}

    \textcolor{ForestGreen}{- Mice are afraid of wolves.}

    \textcolor{ForestGreen}{- Gertrude is a mouse.}

    \textcolor{ForestGreen}{- Therefore, Gertrude is afraid of wolves.}
    
    \textcolor{ForestGreen}{We can also use the following chain of reasoning:}

    \textcolor{ForestGreen}{- Mice are afraid of wolves.}

    \textcolor{ForestGreen}{- Wolves are afraid of sheep.}

    \textcolor{ForestGreen}{- Therefore, mice are indirectly afraid of sheep.}

    \textcolor{ForestGreen}{- Gertrude is a mouse.}

    \textcolor{ForestGreen}{- Therefore, Gertrude is indirectly afraid of sheep.}

    \textcolor{ForestGreen}{However, the fact that Gertrude is a mouse does not mean that she is afraid of cats. Cats are predators of mice, but they are also preyed upon by wolves. Therefore, it is more likely that Gertrude is afraid of wolves than cats.}

    \textcolor{ForestGreen}{In conclusion, the most reasonable inference based on the given facts is that Gertrude is afraid of wolves.}
    &
    wolf
    \\ \hline 
    24. Given facts: Wolves are afraid of mice. Mice are afraid of wolves. Gertrude is a wolf. Jessica is a mouse. Emily is a mouse. Cats are afraid of sheep. Sheep are afraid of wolves. Winona is a cat.

    Based on the given facts above, do a reasonable inference on this question using deductive reasoning: What is gertrude afraid of?

    &
    \textcolor{ForestGreen}{Based on the given facts, Gertrude is a wolf, and wolves are afraid of mice. Therefore, it can be inferred that Gertrude is afraid of mice.}
    &
    \textcolor{ForestGreen}{Based on the given facts, we can infer the following relationships:}

    \textcolor{ForestGreen}{1) Wolves are afraid of mice.}
    
    \textcolor{ForestGreen}{2) Mice are afraid of wolves.}
    
    \textcolor{ForestGreen}{3) Gertrude is a wolf.}
    
    \textcolor{ForestGreen}{4) Jessica is a mouse.}
    
    \textcolor{ForestGreen}{5) Emily is a mouse.}
    
    \textcolor{ForestGreen}{6) Cats are afraid of sheep.}
    
    \textcolor{ForestGreen}{7) Sheep are afraid of wolves.}
    
    \textcolor{ForestGreen}{8) Winona is a cat.}
    
    \textcolor{ForestGreen}{Using deductive reasoning, we can conclude the following:}
    \textcolor{ForestGreen}{Since Gertrude is a wolf (fact 3), and wolves are afraid of mice (fact 1), Gertrude is afraid of mice.}
    & 
    \textcolor{ForestGreen}{Sure, I can help you with that. Here is the deductive reasoning for the question:}

    \textcolor{ForestGreen}{- Premise 1: Wolves are afraid of mice.}

    \textcolor{ForestGreen}{- Premise 2: Mice are afraid of wolves.}

    \textcolor{ForestGreen}{- Premise 3: Gertrude is a wolf.}

    \textcolor{ForestGreen}{- Conclusion: Gertrude is afraid of mice.}

    \textcolor{ForestGreen}{The first two premises establish that there is a predator-prey relationship between wolves and mice. The third premise states that Gertrude is a wolf, and therefore she is a predator. The conclusion follows that Gertrude is afraid of her prey, which are mice.}

    \textcolor{ForestGreen}{The other facts in the prompt are not relevant to the question of what Gertrude is afraid of. For example, the fact that Jessica and Emily are mice does not change the fact that Gertrude is a wolf and therefore afraid of mice. Similarly, the fact that Winona is a cat is irrelevant because cats are not prey to wolves.}

    \textcolor{ForestGreen}{Therefore, the reasonable inference based on the given facts is that Gertrude is afraid of mice.}
    
    &
    mouse
    \\ \hline 
    25. Given facts: Mice are afraid of sheep. Cats are afraid of mice. Gertrude is a mouse. Sheep are afraid of cats. Emily is a cat. Winona is a cat. Jessica is a cat. Wolves are afraid of mice.

    Based on the given facts above, do a reasonable inference on this question using deductive reasoning: What is gertrude afraid of?
    &
    \textcolor{ForestGreen}{Based on the given facts, Gertrude is a mouse and mice are afraid of sheep. Therefore, we can infer that Gertrude is afraid of sheep.}
    &
    \textcolor{ForestGreen}{Based on the given facts, we know that:}

    \textcolor{ForestGreen}{1) Mice are afraid of sheep.}
    \textcolor{ForestGreen}{2) Gertrude is a mouse.}
    
    \textcolor{ForestGreen}{Using deductive reasoning, we can infer that Gertrude, being a mouse, is afraid of sheep.}
    & 
    \textcolor{red}{According to the given facts, Gertrude is a mouse, and mice are afraid of sheep. Therefore, Gertrude is afraid of sheep.}

    \textcolor{red}{However, Gertrude is also afraid of cats, because cats are afraid of mice. This means that Gertrude is afraid of both sheep and cats.}

    \textcolor{red}{The fact that sheep are afraid of cats does not necessarily mean that Gertrude is afraid of Emily, Winona, or Jessica, because these cats are not afraid of mice. However, it is possible that Gertrude is afraid of these cats because they are cats, and cats are generally predators of mice.}

    \textcolor{red}{The fact that wolves are afraid of mice does not mean that Gertrude is afraid of wolves, because wolves are not predators of mice. However, it is possible that Gertrude is afraid of wolves because they are large and powerful animals.}

    \textcolor{red}{Overall, the most reasonable inference is that Gertrude is afraid of sheep and cats. However, it is also possible that she is afraid of other animals, such as wolves.}
    
    &
    sheep
    \\ \hline  
    26. Given facts: Mice are afraid of wolves. Sheep are afraid of mice. Cats are afraid of wolves. Wolves are afraid of sheep. Gertrude is a mouse. Jessica is a wolf. Emily is a wolf. Winona is a wolf.

    Based on the given facts above, do a reasonable inference on this question using deductive reasoning: What is winona afraid of?
    &
    \textcolor{ForestGreen}{Based on the given facts, we can infer that Winona is afraid of sheep because wolves are afraid of sheep, and Winona is a wolf.}
    &
    \textcolor{ForestGreen}{Based on the given facts, we can deduce the following relationships:}

    \textcolor{ForestGreen}{1) Mice are afraid of wolves.}
    
    \textcolor{ForestGreen}{2) Sheep are afraid of mice.}
    
    \textcolor{ForestGreen}{3) Cats are afraid of wolves.}
    
    \textcolor{ForestGreen}{4) Wolves are afraid of sheep.}
    
    \textcolor{ForestGreen}{Since Winona is a wolf, we can use the fourth fact to infer that Winona is afraid of sheep.}
    & 
    \textcolor{ForestGreen}{Sure, I can help you with that. Here are the given facts:}

    \textcolor{ForestGreen}{- Mice are afraid of wolves.}

    \textcolor{ForestGreen}{- Sheep are afraid of mice.}

    \textcolor{ForestGreen}{- Cats are afraid of wolves.}

    \textcolor{ForestGreen}{- Wolves are afraid of sheep.}

    \textcolor{ForestGreen}{- Gertrude is a mouse.}

    \textcolor{ForestGreen}{- Jessica is a wolf.}

    \textcolor{ForestGreen}{- Emily is a wolf.}

    \textcolor{ForestGreen}{- Winona is a wolf.}

    \textcolor{ForestGreen}{We can use deductive reasoning to infer what Winona is afraid of. We know that wolves are afraid of sheep, so Winona, who is a wolf, must be afraid of sheep. Therefore, the answer to the question "What is Winona afraid of?" is sheep.}

    \textcolor{ForestGreen}{Here is the deductive reasoning process in more detail:}

    \textcolor{ForestGreen}{- We know that Winona is a wolf.}

    \textcolor{ForestGreen}{- We also know that wolves are afraid of sheep.}

    \textcolor{ForestGreen}{- Therefore, Winona, who is a wolf, must be afraid of sheep.}

    \textcolor{ForestGreen}{This is a valid deductive inference because it is based on the given facts. The given facts establish that the relationship between wolves and sheep is one of fear. Therefore, if Winona is a wolf, she must be afraid of sheep.}
    
    \textcolor{ForestGreen}{Of course, it is possible that Winona has other fears as well. However, based on the given facts, the only fear that we can infer with certainty is her fear of sheep.}
    &
    sheep
    \\ \hline 
    27. Given facts: Mice are afraid of sheep. Sheep are afraid of cats. Wolves are afraid of sheep. Cats are afraid of sheep. Emily is a sheep. Jessica is a wolf. Winona is a sheep. Gertrude is a sheep.

    Based on the given facts above, do a reasonable inference on this question using deductive reasoning: What is emily afraid of?
    &
    \textcolor{ForestGreen}{Based on the given facts, Emily is a sheep, and sheep are afraid of cats. Therefore, using deductive reasoning, we can infer that Emily is afraid of cats.}
    &
    \textcolor{ForestGreen}{Based on the given facts, we can deduce the following relationships:}

    \textcolor{ForestGreen}{1) Mice are afraid of sheep.}
    
    \textcolor{ForestGreen}{2) Sheep are afraid of cats.}
    
    \textcolor{ForestGreen}{3) Wolves are afraid of sheep.}
    
    \textcolor{ForestGreen}{4) Cats are afraid of sheep.}
    
    \textcolor{ForestGreen}{We are also given that Emily is a sheep. According to fact 2, sheep are afraid of cats.   Therefore, using deductive reasoning, we can infer that Emily is afraid of cats.}
    & 
    \textcolor{ForestGreen}{Using deductive reasoning, we can infer that Emily is afraid of cats.}

    \textcolor{ForestGreen}{Here is the reasoning:}

    \textcolor{ForestGreen}{- Mice are afraid of sheep.}

    \textcolor{ForestGreen}{- Sheep are afraid of cats.}
    
    \textcolor{ForestGreen}{- Therefore, mice are afraid of cats.}

    \textcolor{ForestGreen}{- Emily is a sheep.}

    \textcolor{ForestGreen}{- Sheep are afraid of cats.}

    \textcolor{ForestGreen}{- Therefore, Emily is afraid of cats.}

    \textcolor{ForestGreen}{The other facts in the prompt are irrelevant to the question of what Emily is afraid of. Jessica is a wolf, but wolves are afraid of sheep, so Emily is not afraid of Jessica. Winona and Gertrude are also sheep, so they are not a threat to Emily.}

    \textcolor{ForestGreen}{Therefore, the only reasonable inference we can make is that Emily is afraid of cats.}
    &
    cat
    \\ \hline  
    28. Given facts: Cats are afraid of sheep. Sheep are afraid of wolves. Wolves are afraid of mice. Emily is a wolf. Jessica is a sheep. Gertrude is a sheep. Mice are afraid of cats. Winona is a wolf.

    Based on the given facts above, do a reasonable inference on this question using deductive reasoning: What is jessica afraid of?

    &
    \textcolor{ForestGreen}{Based on the given facts, Jessica is a sheep, and sheep are afraid of wolves. Therefore, we can infer that Jessica is afraid of wolves.}
    &
    \textcolor{ForestGreen}{Based on the given facts, we know that sheep are afraid of wolves. Since Jessica is a sheep, we can deduce that Jessica is afraid of wolves.}
    & 
    \textcolor{ForestGreen}{Using deductive reasoning, we can infer that Jessica is afraid of wolves.}

    \textcolor{ForestGreen}{Here is the reasoning:}

    \textcolor{ForestGreen}{- Fact 1: Cats are afraid of sheep.}

    \textcolor{ForestGreen}{- Fact 2: Sheep are afraid of wolves.}

    \textcolor{ForestGreen}{- Fact 3: Wolves are afraid of mice.}

    \textcolor{ForestGreen}{- Fact 4: Emily is a wolf.}

    \textcolor{ForestGreen}{- Fact 5: Jessica is a sheep.}

    \textcolor{ForestGreen}{- Fact 6: Gertrude is a sheep.}

    \textcolor{ForestGreen}{- Fact 7: Mice are afraid of cats.}

    \textcolor{ForestGreen}{- Fact 8: Winona is a wolf.}

    \textcolor{ForestGreen}{Since Jessica is a sheep, and sheep are afraid of wolves, then Jessica is afraid of wolves.}

    \textcolor{ForestGreen}{We can also see this by following the chain of fear:}

    \textcolor{ForestGreen}{- Cats are afraid of sheep.}

    \textcolor{ForestGreen}{- Sheep are afraid of wolves.}

    \textcolor{ForestGreen}{- Therefore, cats are indirectly afraid of wolves.}

    \textcolor{ForestGreen}{- Jessica is a sheep.}

    \textcolor{ForestGreen}{- Therefore, Jessica is indirectly afraid of wolves.}

    \textcolor{ForestGreen}{Finally, we can see that Jessica is not afraid of mice, because mice are afraid of cats, and cats are not afraid of sheep.}

    \textcolor{ForestGreen}{Therefore, the only thing that Jessica is afraid of is wolves.}

    &
    wolf
    \\ \hline 
    29. Given facts: Mice are afraid of wolves. Emily is a mouse. Wolves are afraid of mice. Winona is a wolf. Sheep are afraid of cats. Cats are afraid of sheep. Jessica is a cat. Gertrude is a sheep.

    Based on the given facts above, do a reasonable inference on this question using deductive reasoning: What is winona afraid of?

    &
    \textcolor{ForestGreen}{Based on the given facts, Winona is a wolf and wolves are afraid of mice. Therefore, Winona is afraid of mice.}
    &
    \textcolor{ForestGreen}{Based on the given facts, we can deduce the following:}

    \textcolor{ForestGreen}{1) Mice are afraid of wolves.}
    
    \textcolor{ForestGreen}{2) Emily is a mouse.}
    
    \textcolor{ForestGreen}{3) Wolves are afraid of mice.}
    
    \textcolor{ForestGreen}{4) Winona is a wolf.}
    
    \textcolor{ForestGreen}{Using deductive reasoning, we can infer that since Winona is a wolf (fact 4) and wolves are afraid of mice (fact 3), Winona is afraid of mice.}
    & 
    \textcolor{ForestGreen}{Sure, I can help you with that. Here are the given facts:}

    \textcolor{ForestGreen}{- Mice are afraid of wolves.}

    \textcolor{ForestGreen}{- Emily is a mouse.}

    \textcolor{ForestGreen}{- Wolves are afraid of mice.}

    \textcolor{ForestGreen}{- Winona is a wolf.}

    \textcolor{ForestGreen}{- Sheep are afraid of cats.}
    
    \textcolor{ForestGreen}{- Cats are afraid of sheep.}

    \textcolor{ForestGreen}{- Jessica is a cat.}

    \textcolor{ForestGreen}{- Gertrude is a sheep.}

    \textcolor{ForestGreen}{We can use deductive reasoning to infer what Winona is afraid of. We know that Winona is a wolf, and wolves are afraid of mice. Therefore, Winona is afraid of mice.}

    \textcolor{ForestGreen}{We can also see that Winona is not afraid of cats, because cats are afraid of sheep, and Winona is not a sheep. Therefore, the only thing that Winona is afraid of is mice.}
    
    &
    mouse
    \\ \hline  
    30. Given facts: Wolves are afraid of mice. Cats are afraid of mice. Sheep are afraid of mice. Mice are afraid of sheep. Winona is a sheep. Jessica is a cat. Gertrude is a cat. Emily is a cat.

    Based on the given facts above, do a reasonable inference on this question using deductive reasoning: What is winona afraid of?
    &
    \textcolor{ForestGreen}{Based on the given facts and deductive reasoning, Winona is afraid of mice. This is because mice are afraid of sheep, and wolves, cats, and sheep are afraid of mice. Since Winona is a sheep and mice are a common fear among sheep, it can be inferred that Winona is also afraid of mice.}
    &
    \textcolor{ForestGreen}{Based on the given facts, we know that sheep are afraid of mice. Since Winona is a sheep, we can deduce that Winona is afraid of mice using deductive reasoning.}
    & 
    \textcolor{ForestGreen}{Here is the deductive reasoning:}

    \textcolor{ForestGreen}{- Mice are afraid of sheep.}

    \textcolor{ForestGreen}{- Winona is a sheep.}

    \textcolor{ForestGreen}{- Therefore, Winona is afraid of mice.}

    \textcolor{ForestGreen}{The other facts are irrelevant to the question, because they do not tell us what Winona is afraid of. For example, the fact that Jessica, Gertrude, and Emily are cats does not tell us anything about Winona's fear of mice.}

    \textcolor{ForestGreen}{We can also use the contrapositive of the first fact to reach the same conclusion:}

    \textcolor{ForestGreen}{- If Winona is not afraid of mice, then she is not a sheep.}

    \textcolor{ForestGreen}{- But Winona is a sheep.}

    \textcolor{ForestGreen}{- Therefore, Winona is afraid of mice.}

    \textcolor{ForestGreen}{Therefore, the reasonable inference is that Winona is afraid of mice.}
    
    &
    mouse
    \\ \hline
    \end{longtable}
\end{center}

\newpage

\begin{center}
    \small
    \begin{longtable}{|p{0.3\linewidth}|p{0.18\linewidth}
                      |p{0.18\linewidth}|p{0.14\linewidth}
                      |p{.05\linewidth}|}
     
    \caption{Inductive reasoning prompts} 
    \label{appendix:inductive_reasoning}\\
    
    \hline
    \multicolumn{5}{|c|}{\textbf{Inductive Reasoning}} \\ \hline
    \textbf{Prompt} & \textbf{GPT-3.5 Answer} & \textbf{GPT-4 Answer} & \textbf{BARD Answer} & \textbf{Gold} \\ \hline
    \multicolumn{5}{|l|}{\textbf{Dataset: CLUTRR}}  
    
    \\ \hline  
    1. Story: [Joseph] took his daughter [Alice] out for dinner. [Judy] is cooking for her daughter [Alice] and son [Lawrence]. [Lawrence]'s brother [Gerald] does n't like her cooking. [Janice] went to the mall, because she wanted to look for a present for her daughter, [Fay]. [Larry] asked his brother, [Kenneth], if he wanted to go to the beach. [Kenneth] said he could n't go because he had to take [Janice], his sister, to see the dentist. [Timothy] took his son [Larry] to the park to feed the squirrels. [Fay] went to the baseball game with her uncle [Joseph]. Based on the story, throught inductive reasoning asnwer the question: What is the     relationship between:  'Gerald' with 'Timothy'?
    & 
    \textcolor{red}{There is no direct relationship or connection mentioned between Gerald and Timothy in the given story. Therefore, it is not possible to determine any relationship between these two individuals through inductive reasoning based on the provided information.}
    &
    \textcolor{ForestGreen}{Based on the information provided in the story, we can infer that Gerald is Timothy's grandson. This is because Gerald is the brother of Lawrence, and Lawrence is Alice's sibling. Alice is Joseph's daughter, and Joseph is Fay's uncle. Fay is Janice's daughter, and Janice is Kenneth's sister. Since Larry is Kenneth's brother and Timothy's son, we can conclude that Timothy is Larry's father, making Gerald his grandson through Larry's sibling relationship with Alice.}
    & 
    \textcolor{red}{Based on the story, we cannot determine the relationship between Gerald and Timothy. The only information we have about their relationship is that Gerald is the brother of Lawrence, who is the son of Timothy. This does not necessarily mean that Gerald and Timothy are related to each other. They could be cousins, stepbrothers, or even unrelated friends. Without more information, it is impossible to say for sure what the relationship between Gerald and Timothy is.}
    &
    grandson
    \\ \hline  
     2. Story: [Jeffrey] asked his brother [William] if he would come help him fix his car next weekend. [Gabrielle] was delighted to see her grandson, [William], at her birthday party. Based on the story, throught inductive reasoning asnwer the question: What is the relationship between:  'Jeffrey' with 'Gabrielle'?
     & 
     \textcolor{red}{There is no direct relationship established between Jeffrey and Gabrielle in the given story. Inductive reasoning is a method of drawing general conclusions based on specific observations or evidence. However, in this particular story, there is no evidence or information that would allow us to conclude any relationship between Jeffrey and Gabrielle. Therefore, we cannot draw any conclusion about their relationship through inductive reasoning.}

     &
    \textcolor{ForestGreen}{Based on the story, we can use inductive reasoning to infer that Jeffrey is most likely Gabrielle's grandson. Since William is Jeffrey's brother and Gabrielle's grandson, it is reasonable to assume that Jeffrey also shares the same relationship with Gabrielle.}
    & 
    \textcolor{red}{Through inductive reasoning, we can conclude that Jeffrey and Gabrielle are most likely cousins.}

    \textcolor{red}{The story tells us that Jeffrey and William are brothers, and that Gabrielle is William's grandmother. This means that Jeffrey and Gabrielle are cousins, as they share a grandparent.}

    \textcolor{red}{The story tells us that Jeffrey and William are brothers, and that Gabrielle is William's grandmother. This means that Jeffrey and Gabrielle are cousins, as they share a grandparent.}
    
     &
     grandson
     \\ \hline  
     3. Story: Brothers [Clarence] and [Carlos] planned a week vacation at the lodge for their families to get together. Their father [Alan] planned to fly out to meet them. The boys' grandfather [Stanley] could n't make this trip because he surgery scheduled for that week. Based on the story, throught inductive reasoning asnwer the question: What is the  relationship between:  'Stanley' with 'Alan'?
     &
    \textcolor{ForestGreen}{Based on the information provided in the story, we can infer that Stanley and Alan are related, likely as father and son. This inference is based on the fact that Stanley is referred to as the "boys' grandfather," which implies that he is the father of either Clarence or Carlos. Additionally, Alan is mentioned as planning to fly out to meet them, which suggests that he is closely related to the family and likely the father of either Clarence or Carlos. Therefore, it is likely that Stanley and Alan are related as father and son.}
     &
    \textcolor{ForestGreen}{Based on the story and using inductive reasoning, Stanley is Alan's father. This is inferred from the information that Stanley is the boys' (Clarence and Carlos) grandfather, and Alan is their father.}
    & 
    \textcolor{ForestGreen}{Based on the information given in the story, we can infer that Stanley and Alan are related in some way. The most likely relationship is that they are father and son. This is because Stanley is the grandfather of Clarence and Carlos, and Alan is their father. It is also possible that Stanley is Alan's uncle or another close relative, but the information in the story does not provide enough evidence to support this conclusion.}
     &
     father
     \\ \hline  
     4. Story: [Brian] invited his father [Kent] and his brother [Rick] to go on a weekend fishing trip. [Oliver] and his grandmother [Dorothy] went to the science museum. They both had fun, and learned some things, too. [Dorothy] went to the mall, because she wanted to look for a present for her daughter, [Helen]. [Sharon] asked her husband [Kent] to take out the trash. [Kent] took it out quickly so he could get back to the game. [Rick] bought a steak dinner for himself and his brother [Oliver]. Based on the story, throught inductive reasoning asnwer the question: What is the     relationship between:  'Helen' with 'Sharon'?
     &
    \textcolor{red}{There is no direct relationship mentioned between Helen and Sharon in the story. Therefore, we cannot infer any relationship between them through inductive reasoning.}
     &
    \textcolor{red}{ Based on the story and using inductive reasoning, Helen is likely Sharon's daughter-in-law. Helen is mentioned as the daughter of Dorothy, and Sharon is married to Kent, who is Dorothy's son. Therefore, Helen would be married to one of Kent's sons, making her Sharon's daughter-in-law.}
    & 

    \textcolor{red}{Through inductive reasoning, we can infer that Helen and Sharon are mother and daughter.}

    \textcolor{red}{- Dorothy is Helen's mother, as stated in the story.}

    \textcolor{red}{- Sharon is married to Kent, who is Dorothy's son.}

    \textcolor{red}{- Therefore, Helen and Sharon are mother and daughter.}

    \textcolor{red}{We can also support this conclusion by considering the following details from the story:}

    \textcolor{red}{- Dorothy went to the mall to buy a present for Helen.}

    \textcolor{red}{- Sharon asked Kent to take out the trash, which suggests that they are living together.}

    \textcolor{red}{- Kent took out the trash quickly so he could get back to the game, which suggests that he is prioritizing his wife and daughter over his own interests.}

    \textcolor{red}{Based on all of this evidence, it is reasonable to conclude that Helen and Sharon are mother and daughter.}
    
     &
     sister
     \\ \hline  
     5. Story: [Jeff]'s brother, [Jeremy], recently made it into town. [Jeremy] and his sister, [Stephanie] love it when their grandmother visits because she spoils them. She is coming this week to watch them while their parents are out of town. [Ruth] coached her daughter [Stephanie]'s cheer leading squad. When [Ruth]'s son [Jeremy] became interested in joining, she reorganized the team line up to include him. [Jeremy] bought gifts for his lovely sister, [Gloria]. Based on the story, throught inductive reasoning asnwer the question: What is the     relationship between:  'Gloria' with 'Jeff'?
     & 
     \textcolor{red}{There is not enough information in the story to determine the relationship between Gloria and Jeff. The story only mentions that Jeremy bought gifts for his sister Gloria, but it does not provide any information about Jeff's relationship with Gloria.}
     &
    \textcolor{red}{The story does not provide enough information to determine the exact relationship between Gloria and Jeff through inductive reasoning. There is no direct mention of their connection in the story.}
    & 
    \textcolor{ForestGreen}{Based on the story, through inductive reasoning, we can conclude that Gloria is Jeff's sister.}

    \textcolor{ForestGreen}{- The story mentions that Jeremy is Jeff's brother.}

    \textcolor{ForestGreen}{- It also mentions that Ruth coached her daughter Stephanie's cheer leading squad, and that when Jeremy became interested in joining, she reorganized the team line up to include him. This suggests that Ruth is Jeremy's mother.}

    \textcolor{ForestGreen}{- Finally, the story mentions that Jeremy bought gifts for his lovely sister, Gloria. This suggests that Gloria is Jeremy's sister, and therefore Jeff's sister.}

    \textcolor{ForestGreen}{Of course, it is possible that Gloria is not Jeff's sister, but rather his cousin or another close relative. However, the evidence in the story suggests that Gloria is most likely Jeff's sister.}

     &
     sister
      \\ \hline  
      6. Story: [Arlene] and her father, [Milton], went to the marina. [Milton]'s daughter, [Margaretta], had purchased a boat, and they were eager to see it. [Margaretta] bought her mother, [Guadalupe], a puppy for her birthday. [Melanie] and her sister, [Guadalupe], frequently bicker. However, they always make up quickly. [Arlene] took her sister, [Irene], out to dinner for her birthday. Based on the story, throught inductive reasoning asnwer the question: What is the     relationship between:  'Melanie' with 'Irene'?
      &
      \textcolor{red}{Based on the information provided in the story, there is no direct or indirect relationship mentioned between Melanie and Irene. Therefore, there is no basis for drawing any conclusion about their relationship through inductive reasoning.}
      & 
      \textcolor{ForestGreen}{Based on the story, Melanie and Guadalupe are sisters. Arlene is also a sister to Irene. Since Guadalupe is Arlene's and Irene's mother, Melanie would be an aunt to Irene. Therefore, the relationship between Melanie and Irene is aunt and niece.}
      & 
      \textcolor{red}{The story gives us the following information about Melanie and Irene:}

      \textcolor{red}{- They are sisters.}

      \textcolor{red}{- They are close enough that Melanie took Irene out to dinner for her birthday.}

      \textcolor{red}{- They are not always on the same page, as they "frequently bicker."}

      \textcolor{red}{- However, they are able to make up quickly, which suggests that they have a strong bond.}

      \textcolor{red}{Based on this information, we can infer that Melanie and Irene have a close but sometimes-challenging relationship. They are sisters who care about each other, but they also have their share of disagreements. However, they are able to work through their differences and maintain a strong relationship.}

      \textcolor{red}{It is important to note that this is just an inference based on the information given in the story. It is possible that there is more to the relationship between Melanie and Irene that we are not aware of. However, based on the information that we do have, it seems likely that they are close sisters who care about each other.}
      &
      aunt
      \\ \hline  
      7. Story: [James] took his daughter [Leslie] to dance practice. [Leslie] and [Richard] are siblings who are enjoying lunch together. [Maria] and her son [Lucas] traveled around the world [Donald] who is [Alice]'s brother went to the store to buy milk. [Donald] saw his sister [Melody] at the store. [Maria] who is [Alice]'s sister both stay home to eat ice cream. [James] would n't let his son [Jeffery] go to the park by himself. [Jeffery]'s brother [Lucas] offered to go with him. Based on the story, throught inductive reasoning asnwer the question: What is the     relationship between:  'Melody' with 'Richard'?
      &
      \textcolor{red}{Based on the given story, there is no direct or indirect relationship mentioned between Melody and Richard. Therefore, we cannot infer any relationship between Melody and Richard through inductive reasoning based on the information provided in the story.}
      &
      \textcolor{red}{There is no information given that clearly indicates the relationship between Melody and Richard. Therefore, using inductive reasoning, we cannot conclusively determine the relationship between Melody and Richard based on the information provided.}
      & 
     \textcolor{red}{Based on the story, through inductive reasoning, we can conclude that Melody and Richard are siblings.}

     \textcolor{red}{- We are told that Leslie and Richard are siblings.}

     \textcolor{red}{- We are also told that Donald is Alice's brother, and that Maria is Alice's sister.}

     \textcolor{red}{- We then see that Donald saw his sister Melody at the store.}

     \textcolor{red}{Putting all of this together, we can see that Melody is the sister of Donald, who is the brother of Alice. Since Leslie and Richard are siblings, this means that Melody and Richard must also be siblings.}

     \textcolor{red}{In other words, Melody and Richard are related to each other through their shared parent(s). They may be full siblings, half siblings, or step-siblings, but they are definitely siblings.}
      &
      aunt
      \\ \hline  
      8. Story: [Elsie] asked her daughter, [Ellen], if she would like to go to a movie with her on Saturday night. [Patrice] and her father, [Steven], went to the marina. [Steven]'s daughter, [Margaret], had purchased a boat, and they were eager to see it. [Sharon] was always arguing with her mother's husband [James]. [Mabel], who is the sister of [Sharon], is a lovely girl. [Patrice] took her brother [James] and other brother [Steve] with her to eat pasta at a pasta buffet. [Patrice] went shoe shopping with her sister [Constance]. [Ellen] and her brother [Steve] went to see a movie. [Constance] was happy to babysit her niece, [Mabel]. [Mabel] was a great child. Based on the story, throught inductive reasoning asnwer the question: What is the     relationship between:  'Margaret' with 'Elsie'?
      &
      \textcolor{red}{Based on the information provided in the story, there is no direct or indirect relationship mentioned between Margaret and Elsie. Therefore, it is not possible to determine any relationship between them through inductive reasoning.}
      &
      \textcolor{red}{Based on the story and using inductive reasoning, it is not possible to determine the exact relationship between 'Margaret' and 'Elsie' as there is no direct or indirect connection mentioned between them in the text.}
      & 
      \textcolor{red}{Based on the story, we can infer that Margaret and Elsie are grandmother and granddaughter.}

      \textcolor{red}{- Steven is the father of Margaret.}

      \textcolor{red}{- Elsie is the mother of Ellen, who is Steven's daughter.}

      \textcolor{red}{- Therefore, Elsie is the grandmother of Margaret.}

      \textcolor{red}{We can also infer this relationship by considering the ages of the characters. Margaret is old enough to have purchased a boat, which suggests that she is an adult. Ellen is young enough to go to a movie with her mother, which suggests that she is a child. This means that Margaret is at least one generation older than Ellen, which makes Elsie Margaret's grandmother.}

      \textcolor{red}{In addition, the story mentions that Steven is Margaret's father. This further confirms the relationship between Margaret and Elsie, as it shows that they are related through blood.}

      \textcolor{red}{Therefore, the most likely relationship between Margaret and Elsie is that they are grandmother and granddaughter.}
      
      &
      daughter
      \\ \hline  
     9. Story: [Marie] went to the store with her sister [Leah] [Marie] 'father [Mark] became enraged when she failed to complete her homework. [Marie] was grounded for a month and unable to spend time with friends. [Helen]'s father, [Mark], was busy planning a party for his daughter [Leah]. [Helen]'s brother, [Wallace], helped with the catering. [Donna] had picked her daughter [Helen] out the cutest new dress to wear on her birthday. [Mark] and his brother [Ronald] went to the park to play basketball. [Helen] went to her brother [Wallace]'s birthday party [Marie] just had a baby and presented the baby proudly to the new maternal grandmother, [Donna]. Based on the story, throught inductive reasoning asnwer the question: What is the  relationship between:  'Ronald' with 'Marie'?
     &
     \textcolor{red}{Based on the given story, there is no direct or implied relationship between Ronald and Marie. There is no information provided that suggests any interaction or connection between them. Therefore, it is not possible to infer any relationship between them through inductive reasoning.}
     &
     \textcolor{ForestGreen}{Based on the information provided in the story, Ronald is Marie's uncle. This is because Ronald is the brother of Mark, who is Marie's father.}
     & 
     \textcolor{ForestGreen}{Based on the story, through inductive reasoning, we can conclude that Ronald is Marie's uncle.}

     \textcolor{ForestGreen}{Here are the reasons why I believe this:}

     \textcolor{ForestGreen}{- Mark is the father of both Marie and Leah.}

     \textcolor{ForestGreen}{- Mark is also the father of Helen.}

     \textcolor{ForestGreen}{- Wallace is the brother of Mark.}

     \textcolor{ForestGreen}{- Ronald is the brother of Mark.}

     \textcolor{ForestGreen}{Therefore, Ronald is the uncle of both Marie and Helen.}

     \textcolor{ForestGreen}{We can also infer this relationship from the fact that Ronald and Mark went to the park to play basketball together. This suggests that they are close relatives, and the most likely relationship is that they are brothers.}

     \textcolor{ForestGreen}{Finally, the fact that Marie presented her baby to Donna as the new maternal grandmother suggests that Donna is Marie's mother. This means that Mark is Marie's father, and Ronald is Mark's brother. Therefore, Ronald is Marie's uncle.}

     \textcolor{ForestGreen}{Of course, it is possible that Ronald is not Marie's uncle, but rather her cousin or some other relative. However, based on the information given in the story, the most likely relationship is that Ronald is Marie's uncle.}
     
     &
     uncle
     \\ \hline  
     10. Story: [Eula] watched a golf tournament with her aunt [Ellen]. [Ellen] had picked her daughter [Mabel] out the cutest new dress to wear on her birthday. [Margaret] asked her daughter, [Eula], if she would like to go to a movie with her on Saturday night. [Constance] and her sisters [Margaret] went to the spa. [Patrice], another sister of [Constance], had to babysit and could n't join them. [Rita] took her brother [Don] and other brother [Stanley] with her to eat pasta at a pasta buffet. [Mabel] made a cake for her grandfather, [Steven]. [Don]'s aunt [Patrice] took him to the store to do some last minute party shopping. [Stanley] took his favorite son [Steve] to a baseball game where he caught a foul ball as a souvenir. Based on the story, throught inductive reasoning asnwer the question: What is the     relationship between:  'Steven' with 'Steve'?
     &
     \textcolor{red}{Based on the information provided in the story, we can infer that 'Steven' and 'Steve' are related to each other. It is not explicitly stated how they are related, but we can assume that 'Steve' is likely the grandson of 'Steven' because 'Mabel', who is the daughter of 'Ellen' (who is likely a sister or sister-in-law of 'Steven'), made a cake for him. Additionally, 'Stanley', who is likely another son or son-in-law of 'Steven', took his favorite son 'Steve' to a baseball game where he caught a foul ball as a souvenir. Therefore, we can conclude that 'Steven' and 'Steve' are likely grandfather and grandson, respectively.}
     &
     \textcolor{red}{Based on the information provided in the story, Steven is likely Steve's grandfather. We know that Mabel made a cake for her grandfather Steven, and Stanley, who is Mabel's uncle (since Rita is Mabel's mother), took his son Steve to a baseball game. Thus, it can be inferred through inductive reasoning that Steven is Steve's grandfather.}
     & 
    \textcolor{red}{Based on the story, through inductive reasoning, we can conclude that Steven and Steve are grandfather and grandson.}

    \textcolor{red}{- Steven is Mabel's grandfather.}

    \textcolor{red}{- Mabel made a cake for Steven.}

    \textcolor{red}{- Steve is Stanley's favorite son.}

    \textcolor{red}{- Stanley took Steve to a baseball game.}
    
    \textcolor{red}{In addition, Patrice is Don's aunt. This means that Patrice is also Mabel's aunt, and therefore Steven's daughter. This further supports the conclusion that Steven and Steve are grandfather and grandson.}

    \textcolor{red}{It is also possible that Steven and Steve are not related by blood. However, the evidence in the story suggests that they are more likely to be grandfather and grandson.}

     &
     father 
     \\ \hline  
     11. Story: [Colleen] was excited because today she was going to the zoo with her uncle [Gerald]. [Colleen] was always arguing with her mother's husband [Robert]. [Robert] took his wife [Beth] out for dinner. [Ronald] and his son [Gerald] went to the cinema together. Based on the story, throught inductive reasoning asnwer the question: What is the     relationship between:  'Ronald' with 'Robert'?
     &
     \textcolor{red}{There is not enough information in the story to determine the relationship between Ronald and Robert. The story only mentions that Robert is married to Beth and that Colleen argues with him, while Ronald and his son Gerald went to the cinema together.}
     &
    \textcolor{red}{Based on the information provided in the story, we cannot determine the relationship between 'Ronald' and 'Robert' through inductive reasoning. There is not enough information to establish any connection between the two characters.}
    & 
    \textcolor{red}{Based on the information given in the story, we can infer that Ronald and Robert are brothers.}

    \textcolor{red}{- Based on the information given in the story, we can infer that Ronald and Robert are brothers.}

    \textcolor{red}{- Gerald is both Colleen's uncle and Ronald's son, which means that Ronald is Gerald's father and Robert's brother.}

    \textcolor{red}{We can also infer that Ronald and Robert have a strained relationship, as Colleen is always arguing with Robert. This suggests that Ronald and Robert may have different personalities or values, which could lead to conflict.}

    \textcolor{red}{However, it is important to note that this is just an inference based on the limited information given in the story. It is possible that Ronald and Robert have a perfectly amicable relationship, and that Colleen's arguments with Robert are simply a reflection of her own personality.}

     &
     father
     \\ \hline  
     12. Story: [Lane] likes to visit his sister. Her name is [Leslie]. [Leslie] and her daughter [Frances] went to see a movie yesterday and then got ice cream afterwards. [Peter] took his daughter [Frances] to the park that afternoon to play. [Frances] and her uncle [Lane] went to the pet shop. [Frances] saw a puppy that she loved, so [Lane] bought it for her. [Frankie] was so excited to surprise his son, [Lane], with the tickets to the playoffs. Based on the story, throught inductive reasoning asnwer the question: What is the     relationship between:  'Frankie' with 'Peter'?
     &
     \textcolor{red}{Based on the information provided in the story, there is no direct or indirect relationship established between 'Frankie' and 'Peter'. Therefore, it is not possible to determine any relationship between them through inductive reasoning. The story only describes the activities of 'Lane', 'Leslie', 'Frances', 'Peter' and 'Frankie' and their interactions with each other, but there is no information given about any relationship between 'Frankie' and 'Peter'.}
     &
     \textcolor{red}{The story does not provide enough information to determine the exact relationship between 'Frankie' and 'Peter' through inductive reasoning. There are no direct connections or mentions of their relationship.}
     & 
     \textcolor{ForestGreen}{Based on the story, through inductive reasoning, we can conclude that Frankie and Peter are father and son.}

     \textcolor{ForestGreen}{- Lane is the son of Leslie.}

     \textcolor{ForestGreen}{- Frances is the daughter of Leslie.}

     \textcolor{ForestGreen}{- Peter took Frances to the park.}

     \textcolor{ForestGreen}{- Lane bought a puppy for Frances.}

     \textcolor{ForestGreen}{- Frankie was excited to surprise his son, Lane, with tickets to the playoffs.}

     \textcolor{ForestGreen}{Therefore, the only possible relationship between Frankie and Peter is that they are father and son.}

     \textcolor{ForestGreen}{Here are some other details from the story that support this conclusion:}

     \textcolor{ForestGreen}{- Lane is referred to as "his son" by Frankie.}

     \textcolor{ForestGreen}{- Frankie is excited to surprise his "son" with tickets to the playoffs.}

     \textcolor{ForestGreen}{- Peter is never referred to as Frances's father, but he is referred to as her uncle. This suggests that Peter is married to Leslie, which would make him Lane's father.}

     \textcolor{ForestGreen}{Of course, it is possible that there is another explanation for the relationship between Frankie and Peter. However, based on the information given in the story, the most likely explanation is that they are father and son.}

     &
     father
     \\ \hline  
     13. Story: [Brian] and his sister, [Juanita], went to a sneak preview of the Avengers movie and loved it. [Timothy] and his brother, [Brian] run a turtle farm. [Timothy] took his grandmother [Carla] and sister [Juanita] to the park to have a picnic; [Carla]'s husband, [Frankie], came late and brought wine. Based on the story, throught inductive reasoning asnwer the question: What is the     relationship between:  'Juanita' with 'Frankie'?
     
     &
     \textcolor{red}{Based on the story, there is no direct or implied relationship between Juanita and Frankie. Juanita is the sister of Brian and went to watch the Avengers movie with him. Meanwhile, Frankie is the husband of Carla and arrived late with wine when Timothy took Carla and Juanita to the park for a picnic. Therefore, we cannot deduce any relationship between Juanita and Frankie from the information given.}
     &
     \textcolor{ForestGreen}{Based on the story and through inductive reasoning, Juanita's relationship with Frankie is that of a granddaughter to a grandfather. Juanita is Timothy's sister, and Carla is their grandmother. Frankie is Carla's husband, making him Juanita's grandfather.}
     & 
    \textcolor{ForestGreen}{Based on the story, through inductive reasoning, we can infer that Juanita and Frankie are granddaughter and grandfather.}

    \textcolor{ForestGreen}{Here are the reasons for our inference:}

    \textcolor{ForestGreen}{- Juanita is the sister of Brian, who is the brother of Timothy. This means that Juanita and Timothy are cousins.}

    \textcolor{ForestGreen}{- Timothy's grandmother is Carla.}

    \textcolor{ForestGreen}{- Frankie is Carla's husband.}

    \textcolor{ForestGreen}{Therefore, Juanita is Carla's granddaughter, and Frankie is Juanita's grandfather.}

    \textcolor{ForestGreen}{We can also infer this relationship from the fact that Frankie brought wine to the picnic. This suggests that he is a close family member, and the most likely close family member for him to be is Juanita's grandfather.}

    \textcolor{ForestGreen}{Of course, it is possible that Juanita and Frankie have some other kind of relationship, such as being neighbors or friends. However, the evidence in the story suggests that they are most likely granddaughter and grandfather.}

     &
     grand-daughter
     \\ \hline  
     14. Story: [Laura]'s daughter, [Robin], and father [Frankie] went to the store to get iced coffee for [Frankie]'s wife [Carla]. [Robin] and [David] did the Father daughter dance at the recital and his wife [Laura] was so happy she cried. Based on the story, throught inductive reasoning asnwer the question: What is the     relationship between:  'Robin' with 'Carla'?
     & 
     \textcolor{ForestGreen}{Based on the given story, there is no direct relationship between Robin and Carla. However, since Robin went to the store with her grandfather Frankie to get iced coffee for Carla, we can assume that Robin and Carla have some kind of familial relationship, such as Robin being Carla's granddaughter.}
     &
    \textcolor{ForestGreen}{Based on the story, Robin is Carla's granddaughter. This is because Robin is Laura's daughter, and Carla is Laura's mother-in-law, being Frankie's wife.}
    & 
    \textcolor{red}{Based on the story, through inductive reasoning, we can conclude that Robin and Carla are step-daughter and step-mother.}

    \textcolor{red}{The evidence for this conclusion includes the following:}

    \textcolor{red}{- Frankie is Robin's father, and Carla is Frankie's wife.}

    \textcolor{red}{- Robin and Frankie went to the store to get iced coffee for Carla.}

    \textcolor{red}{- Robin and David did the father-daughter dance at the recital, and Laura was so happy she cried. This suggests that Robin is Laura's daughter, and therefore Carla's step-daughter.}

    \textcolor{red}{It is also possible that Robin and Carla are unrelated, but that Frankie and Carla are very close friends. However, the evidence in the story suggests that Robin and Carla are step-daughter and step-mother.}





    
     &
     grand-daughter
     \\ \hline  
     15. Story: [Demetra] spent a great day shopping with her daughter, [Linda]. [Carlos] bought a brand new car for his hardworking son-in-law, [James]. [Karen] received some beautiful roses from her husband [John]. [John] has a daughter called [Demetra]. Based on the story, throught inductive reasoning asnwer the question: What is the     relationship between:  'Linda' with 'Karen'?
     & 
     \textcolor{red}{Based on the story, there is no direct or indirect relationship mentioned between Linda and Karen. Therefore, it is not possible to infer any relationship between them through inductive reasoning.}
     &
     \textcolor{ForestGreen}{Based on the story and using inductive reasoning, Linda is Karen's granddaughter. This is because Demetra is John's daughter, and Linda is Demetra's daughter. Since Karen is married to John, that makes Linda their granddaughter.}
     & 
    \textcolor{ForestGreen}{Relationship between Linda and Karen:}

    \textcolor{ForestGreen}{- Demetra is the daughter of John.}

    \textcolor{ForestGreen}{- John is married to Karen.}

    \textcolor{ForestGreen}{- Linda is the daughter of Demetra.}

    \textcolor{ForestGreen}{Therefore, Linda is the granddaughter of Karen.}

    \textcolor{ForestGreen}{Here is the chain of relationships:}

    \textcolor{ForestGreen}{- Demetra is the daughter of John.}

    \textcolor{ForestGreen}{- John is married to Karen.}

    \textcolor{ForestGreen}{- Therefore, Demetra is the daughter-in-law of Karen.}

    \textcolor{ForestGreen}{- Linda is the daughter of Demetra.}

    \textcolor{ForestGreen}{- Therefore, Linda is the granddaughter of Karen.}

     &
     grand-daughter
      \\ \hline  
     16. Story: [Robin] loves her son [Allen] very much. He loves her, too. [Robin] went car shopping with her husband [Eric] and her daughter [Lisa]. [Lisa] and [Allen] left that evening to go bowling. Based on the story, throught inductive reasoning asnwer the question: What is the     relationship between:  'Lisa' with 'Eric'?
     &
     \textcolor{red}{Based on the given story, there is no information provided about the relationship between Lisa and Eric. Therefore, it is not possible to make any inductive reasoning-based conclusions about the relationship between Lisa and Eric.}
     &
     \textcolor{ForestGreen}{Based on the story and using inductive reasoning, we can infer that Lisa is likely the daughter of Eric as well, since Robin and Eric are married and they have a daughter named Lisa. So, the relationship between Lisa and Eric is that of father and daughter.}
     & 
     \textcolor{red}{Based on the story, through inductive reasoning, we can conclude that Lisa and Eric are siblings.}

     \textcolor{red}{The story tells us that Robin is the mother of Allen and Lisa, and that Eric is Robin's husband. This means that Lisa and Eric are both children of Robin and Eric, making them siblings.}

     \textcolor{red}{Additionally, the story tells us that Lisa and Allen left that evening to go bowling together. This suggests that they are close in age and enjoy spending time together, which is another indication that they are siblings.}

     \textcolor{red}{Of course, it is possible that Lisa and Eric are not siblings, but rather cousins or step-siblings. However, the evidence in the story suggests that they are most likely biological siblings.}
     &
     daughter
     \\ \hline  
     17.Story: [Jose]'s mother [Latasha] had to help him with his homework because he was having a test soon. [Latasha] went to the mall, because she wanted to look for a present for her daughter, [Lore]. [Jennifer] and her daughter [Georgina] went to see a movie yesterday and then got ice cream afterwards. [Georgina] went shoe shopping with her sister [Latasha]. Based on the story, throught inductive reasoning asnwer the question: What is the     relationship between:  'Jennifer' with 'Jose'?
     &
     \textcolor{red}{Based on the story, we cannot determine any direct relationship between Jennifer and Jose. There is no mention of Jennifer having any interaction or connection with Jose or his family. Therefore, any conclusion about their relationship based on the given information would be speculative and unfounded.}
     &
     \textcolor{red}{Based on the information provided, it is not possible to determine the relationship between Jennifer and Jose through inductive reasoning. There is no information given that connects Jennifer and Jose directly or indirectly.}
     & 
     \textcolor{red}{Based on the story, there is not enough information to determine the relationship between Jennifer and Jose. However, we can make some inferences based on the information that is given.}

     \textcolor{red}{- Jennifer and Georgina are mother and daughter.}

     \textcolor{red}{- Georgina is Latasha's sister.}

     \textcolor{red}{- Latasha is Jose's mother.}

     \textcolor{red}{This suggests that Jennifer and Latasha are likely sisters, which would make Jose and Georgina cousins. However, it is also possible that Jennifer is not related to Latasha or Jose at all.}

     \textcolor{red}{In order to determine the relationship between Jennifer and Jose, we would need more information, such as whether or not Jennifer and Latasha live together, or whether or not they have ever been seen interacting with each other.}

     \textcolor{red}{In order to determine the relationship between Jennifer and Jose, we would need more information, such as whether or not Jennifer and Latasha live together, or whether or not they have ever been seen interacting with each other.}
     
     &
     grand-mother
     \\ \hline  
     18. Story: [Blanca] just had a baby and presented the baby proudly to the new maternal grandmother, [Elizabeth]. [Patricia] was on her way to pick up her daughter [Alverta] from the park. [Alverta] had spent the entire day there playing soccer with her sister [Elizabeth]. [Kelley] and her aunt, [Marissa], went to the bookstore. [Kelley] bought some fantasy books, while [Marissa] picked up some mysteries. Based on the story, throught inductive reasoning asnwer the question: What is the     relationship between:  'Patricia' with 'Blanca'?
     &
     \textcolor{red}{Based on the given story, there is no direct information about the relationship between Patricia and Blanca. We cannot infer any specific relationship between the two individuals through inductive reasoning as they have not been mentioned together or in any context that implies a relationship.}
     &
    \textcolor{red}{Based on the information provided in the story, there is no explicit relationship mentioned between 'Patricia' and 'Blanca'. We cannot determine their relationship through inductive reasoning, as there isn't enough information about their connections to each other.}
    & 
    \textcolor{red}{Based on the story, we can infer that Patricia is the sister of Blanca.}

    \textcolor{red}{- Patricia is on her way to pick up her daughter, Alverta, from the park.}

    \textcolor{red}{- Alverta is playing soccer with her sister, Elizabeth.}
    
    \textcolor{red}{- Elizabeth is the new maternal grandmother of the baby that Blanca just had.}
    
    \textcolor{red}{This means that Alverta and Elizabeth are Blanca's daughters.}

    \textcolor{red}{- Patricia is Alverta's mother, so she is also Blanca's sister.}

    \textcolor{red}{The only other possibility is that Patricia is Blanca's mother, but this is less likely because Blanca would have mentioned her mother when she presented the baby to Elizabeth.}

    \textcolor{red}{Therefore, the most likely relationship between Patricia and Blanca is that they are sisters.}
     &
     grand-mother
     \\ \hline  
     19. Story: [Judy] and her sister, [Georgina], frequently bicker. However, they always make up quickly. [Georgina] and her son [Felix] traveled around the world [Jennifer] took her daughter [Latasha] to the park. [Latasha] was sad her sister [Georgina] could not join. Based on the story, throught inductive reasoning asnwer the question: What is the     relationship between:  'Jennifer' with 'Felix'?
     &
     \textcolor{red}{Based on the information provided in the story, there is no direct relationship mentioned between Jennifer and Felix. It is only stated that Georgina, who is Jennifer's sister, traveled around the world with her son Felix, while Jennifer took her daughter Latasha to the park. Therefore, it is not possible to make an inference or conclusion about any relationship between Jennifer and Felix based on the given information through inductive reasoning.}
     &
     \textcolor{red}{Based on the story and using inductive reasoning, Jennifer's relationship with Felix is likely that of an aunt. Since Georgina is Judy's sister and has a son named Felix, and Jennifer is mentioned as taking her daughter Latasha to the park, it can be inferred that Jennifer and Judy could be sisters as well. This would make Jennifer Felix's aunt.}
     & 
    \textcolor{red}{Based on the story, there is not enough information to determine the relationship between Jennifer and Felix. However, there are a few clues that could suggest a possible relationship.}

    \textcolor{red}{- Latasha is sad that Georgina could not join her and Jennifer at the park. This suggests that Latasha and Georgina are close, and that Georgina is a significant figure in Latasha's life.}
    \textcolor{red}{- Felix is Georgina's son. This suggests that Felix is related to Latasha in some way, either as a cousin or a sibling.}
    \textcolor{red}{- Jennifer took Latasha to the park. This suggests that Jennifer is Latasha's mother, or at least a close caregiver.}
    \textcolor{red}{Given these clues, it is possible that Jennifer and Felix are related through Georgina. However, it is also possible that they are not related at all. More information is needed to determine the true relationship between Jennifer and Felix.}

    \textcolor{red}{Here are some additional questions that could help to clarify the relationship between Jennifer and Felix:}

    \textcolor{red}{1. Is Latasha Georgina's only child?, 2. Does Jennifer have any other children?}




    \textcolor{red}{Once these questions are answered, it will be possible to make a more informed judgment about the relationship between Jennifer and Felix.}
     &
     grand-mother
      \\ \hline  
      20. Story: [April] and her brother [Ryan] went to see a movie. [Michael] and his son [Robert] went to look at cars. [Michael] ended up buying the Mustang. [April] bought a plant for her dad, [Paul]. [Michael] asked his aunt, [April], to read him a story. [April] unfortunately was n't able to, since she did n't have her glasses, and she was almost blind without them. [Paul] took his son [Ryan] out for ice cream. Based on the story, throught inductive reasoning asnwer the question: What is the     relationship between:  'Paul' with 'Robert'?
      &
      \textcolor{red}{Based on the story, there is no direct or indirect relationship mentioned between Paul and Robert. Paul is only mentioned as April's father and Ryan's father, while Robert is only mentioned as Michael's son who went to look at cars with him. Therefore, we cannot determine any relationship between Paul and Robert through inductive reasoning based on the given information.}
      &
      \textcolor{red}{Based on the story and using inductive reasoning, we can infer that 'Paul' is most likely 'Robert's' grandfather. This is because 'April' is 'Robert's' aunt, and 'Paul' is 'April's' dad, which would make 'Paul' the father of one of 'Robert's' parents.}
      & 
     \textcolor{ForestGreen}{Based on the story, through inductive reasoning, we can conclude that Paul and Robert are father and son.}

     \textcolor{ForestGreen}{- April is Michael's aunt.}

     \textcolor{ForestGreen}{- Michael is Robert's father.}

     \textcolor{ForestGreen}{- Therefore, Paul is Robert's grandfather.}

     \textcolor{ForestGreen}{We can also infer this relationship from the fact that April bought a plant for her dad, Paul. This suggests that Paul is her father, and that Robert is her brother.}

     \textcolor{ForestGreen}{Finally, we know that Paul took his son, Ryan, out for ice cream. This confirms that Paul is Robert's father.}

     \textcolor{ForestGreen}{In conclusion, the relationship between Paul and Robert is that of father and son.}
      &
      father
      \\ \hline  
      21. Story: [Christina] stayed with her mother [Angela] before the birth of her son [Paul] [Paul] took his favorite son [Randall] to a baseball game where he caught a foul ball as a souvenir. [Eugene] enjoys playing chess with his daughter's husband. His name is [Noe]. Based on the story, throught inductive reasoning asnwer the question: What is the     relationship between:  'Angela' with 'Randall'?
      &
      \textcolor{red}{Based on the given story, we cannot make any direct or indirect connections between Angela and Randall. Therefore, we cannot infer any relationship between them through inductive reasoning.}
      &
      \textcolor{red}{Based on the story, Angela is Christina's mother, and Christina is the mother of Paul. This makes Angela the grandmother of Paul. Since Randall is Paul's son, Angela would be Randall's great-grandmother. Therefore, the relationship between Angela and Randall is great-grandmother and great-grandson.}
      & 
     \textcolor{red}{Based on the story, through inductive reasoning, we can conclude that Angela is Randall's grandmother.}

     \textcolor{red}{- Christina stayed with her mother, Angela, before the birth of her son, Paul.}

     \textcolor{red}{- Paul took his favorite son, Randall, to a baseball game.}

     \textcolor{red}{- Eugene enjoys playing chess with his daughter's husband, Noe.}

     \textcolor{red}{From this, we can infer that Christina is Paul's daughter, and Eugene is Paul's father. This means that Angela is Paul's mother, and Randall's grandmother.}

     \textcolor{red}{It is also possible that Angela is Randall's great-grandmother, if Christina is Paul's granddaughter. However, the information in the story does not provide enough evidence to support this conclusion.}

     \textcolor{red}{Therefore, the most likely relationship between Angela and Randall is that they are grandmother and grandson.}
      &
      mother
      \\ \hline  
     22. Story: [Mark] took his son [Jason] to the park to feed the squirrels. [Carol] fixed her husband [Mark] dinner and then they watched a movie they rented. [Sandra] made a pie for her granddaughter [Mary]. [Mary]'s father [Mark] ate most of it. Based on the story, throught inductive reasoning asnwer the question: What is the     relationship between:  'Sandra' with 'Carol'?
     & 
     \textcolor{red}{Based on the given story, there is no direct relationship between Sandra and Carol. There is no information provided in the story that suggests any interaction or connection between these two individuals. Therefore, we cannot infer any relationship between Sandra and Carol through inductive reasoning.}
     &
     \textcolor{ForestGreen}{Based on the story and using inductive reasoning, we can infer that Sandra is likely to be Carol's mother-in-law. Sandra made a pie for her granddaughter Mary, whose father is Mark. Carol is married to Mark. Therefore, Sandra is Mark's mother, making her Carol's mother-in-law.}
     & 
     \textcolor{red}{Based on the information given in the story, Sandra and Carol are not related.}

     \textcolor{red}{The story tells us that Mark is the father of Jason and Mary, and that Sandra is Mary's grandmother. This means that Sandra is Mark's mother-in-law, but it does not tell us whether Carol is Sandra's daughter, daughter-in-law, or something else.}

     \textcolor{red}{In order to determine the relationship between Sandra and Carol, we would need more information, such as whether Mark is married to Carol or not. If Mark is married to Carol, then Sandra and Carol are sisters-in-law. However, if Mark is not married to Carol, then Sandra and Carol could be any number of other things, such as friends, neighbors, or coworkers.}

     \textcolor{red}{Without more information, it is impossible to say for sure what the relationship is between Sandra and Carol. However, based on the information that is given, it is clear that they are not related by blood or marriage.}
     &
     mother-in-law
     \\ \hline  
     23. Story: [Randall] was filled with pride and joy to see his daughter [Mary] perform in her first ballet recital. [Marvin] was so excited to spend the afternoon with his father-in-law, [Lester]. [Christina] asked her mother, [Angela], what they were going to have for lunch. [Mary] just had a baby and presented the baby proudly to the new maternal grandmother, [Christina]. Based on the story, throught inductive reasoning asnwer the question: What is the     relationship between:  'Angela' with 'Randall'?
     &
     \textcolor{red}{Based on the information provided, there is not enough evidence to determine the relationship between Angela and Randall.}
     &
     \textcolor{red}{Using inductive reasoning, we can infer that Angela is likely Randall's wife and Mary's mother. Therefore, Angela's relationship with Randall is that of husband and wife.}
     & 
    \textcolor{red}{Based on the information provided in the story, we can infer that Angela is Randall's wife.}

    \textcolor{red}{- Randall is filled with pride and joy to see his daughter Mary perform in her first ballet recital. This indicates that Randall is Mary's father.}

    \textcolor{red}{- Christina asked her mother Angela what they were going to have for lunch. This indicates that Angela is Christina's mother.}

    \textcolor{red}{- Mary just had a baby and presented the baby proudly to the new maternal grandmother Christina. This indicates that Christina is the baby's maternal grandmother, and therefore Mary's mother.}

    \textcolor{red}{Since Randall is Mary's father, and Christina is Mary's mother, then Angela must be Randall's wife.}

    \textcolor{red}{In addition, the fact that Marvin is so excited to spend the afternoon with his father-in-law Lester, indicates that Marvin is married to Lester's daughter. Since Christina is asking her mother Angela what they are going to have for lunch, then Christina is likely Marvin's wife. This further supports the inference that Angela is Randall's wife.}

    \textcolor{red}{Therefore, the relationship between Angela and Randall is that they are husband and wife.}
     &
     mother
     \\ \hline  
     24. Story: [Marvin] walked his daughter [Ignacia] down the aisle on her wedding day. [Margaret] and her mother [Emily] went to the mall to try on new clothes. [Margaret] has a daughter named [Ignacia]. Based on the story, throught inductive reasoning asnwer the question: What is the     relationship between:  'Emily' with 'Marvin'?
     & 
     \textcolor{red}{Based on the story, there is no clear indication of any relationship between Emily and Marvin. The story mentions Marvin walking his daughter Ignacia down the aisle on her wedding day, and Margaret and her mother Emily going to the mall to try on new clothes. However, there is no direct or indirect reference to Marvin and Emily having any connection or relationship. Therefore, we cannot determine any relationship between Emily and Marvin through inductive reasoning based on this story.}
     &
     \textcolor{red}{Based on the information provided in the story, Emily and Marvin are most likely in-laws. Since Ignacia is the daughter of both Marvin and Margaret, and Emily is Margaret's mother, this makes Emily the mother-in-law of Marvin.}
     & 
    \textcolor{red}{Through inductive reasoning, we can conclude that Emily is Marvin's wife.}

    \textcolor{red}{Here is the reasoning:}

    \textcolor{red}{- Marvin walked his daughter, Ignacia, down the aisle on her wedding day. This means that Marvin is Ignacia's father.}

    \textcolor{red}{- Margaret has a daughter named Ignacia. This means that Margaret is Ignacia's mother.}

    \textcolor{red}{- Margaret and her mother, Emily, went to the mall to try on new clothes. This means that Emily is Margaret's mother.}

    \textcolor{red}{Since Margaret is Ignacia's mother and Emily is Margaret's mother, then Emily must also be Ignacia's grandmother. This means that Marvin is Ignacia's father, and Emily is Marvin's wife.}

    \textcolor{red}{In addition, the fact that Marvin walked Ignacia down the aisle on her wedding day suggests that he is a close and involved father. This is consistent with the idea that he is married to Emily, as married couples typically share parenting responsibilities.}

    \textcolor{red}{Therefore, through inductive reasoning, we can conclude that Emily is Marvin's wife.}
    
     &
     mother
     \\ \hline  
     25. Story: [Carol] and her husband [Mark] liked to take their daughter [Carmen] to the ice skating rink. Occasionally, [Carmen]'s grandmother [Sandra] came along. [Mary] took her sister, [Carmen], out to dinner for her birthday. Based on the story, throught inductive reasoning asnwer the question: What is the     relationship between:  'Sandra' with 'Carol'?
     &
     \textcolor{red}{Based on the given story, it is not possible to determine a direct relationship between Sandra and Carol. Although it is mentioned that Sandra is Carmen's grandmother, there is no information provided about Sandra's relationship with Carmen's parents (Carol and Mark). Therefore, any relationship between Sandra and Carol can only be speculated upon and cannot be inferred through inductive reasoning.}
     &
     \textcolor{red}{Based on the story, it can be inferred through inductive reasoning that Sandra is Carol's mother, as Sandra is mentioned as Carmen's grandmother.}
     & 
    \textcolor{red}{Through inductive reasoning, we can conclude that Sandra is Carol's mother.}

    \textcolor{red}{The evidence for this conclusion is as follows:}

    \textcolor{red}{- Sandra is Carmen's grandmother.}

    \textcolor{red}{- Carmen is Carol's daughter.}

    \textcolor{red}{- Therefore, Sandra is Carol's mother.}

    \textcolor{red}{Additionally, the fact that Sandra occasionally accompanied Carol and Mark to the ice skating rink suggests that they have a close relationship. This is further supported by the fact that Mary took Carmen out to dinner for her birthday, which suggests that Carmen is close to both her mother and her grandmother.}

    \textcolor{red}{Therefore, based on the evidence in the story, we can conclude that Sandra is Carol's mother.}
    
     &
     mother-in-law 
     \\ \hline  
     26. Story: [Mary] had a daughter named [Helen]. [Joyce] just had a baby and presented the baby proudly to the new maternal grandmother, [Helen]. [Billy] has a daughter called [Joyce]. Based on the story, throught inductive reasoning asnwer the question: What is the     relationship between:  'Mary' with 'Billy'?
     & 
     \textcolor{red}{Based on the information given, it is not possible to determine the relationship between Mary and Billy through inductive reasoning. The story only provides information about Mary's daughter, Helen, and Billy's daughter, Joyce. There is no direct or indirect information that suggests any relationship between Mary and Billy.}
     & 
     \textcolor{red}{Based on the story provided and using inductive reasoning, we can infer that 'Mary' is the mother of 'Helen,' who is the maternal grandmother of 'Joyce's baby. 'Billy' is the father of 'Joyce.' Therefore, 'Mary' and 'Billy' are the grandparents of 'Joyce's baby. They share a common grandchild, so their relationship to each other is that of in-laws. Specifically, 'Mary' is 'Billy's' mother-in-law, and 'Billy' is 'Mary's' son-in-law.}
     & 
    \textcolor{red}{Through inductive reasoning, we can conclude that Mary and Billy are in-laws.}

    \textcolor{red}{- Mary is the mother of Helen, who is the maternal grandmother of Joyce's baby. This means that Joyce is Helen's daughter, and Billy is Joyce's father.}

    \textcolor{red}{- Billy has a daughter named Joyce, which means that he is related to Joyce in some way.}

    \textcolor{red}{- The only way that Billy could be related to Joyce is if he is her father.}

    \textcolor{red}{Therefore, Mary and Billy are in-laws. They are not related by blood, but they are related through their children.}
    
     &
     mother
     \\ \hline  
     27. Story: [Sharon] enjoyed a homemade dinner with her son [Roy] [Debra] asked her husband [Stuart] to take out the trash. [Stuart] took it out quickly so he could get back to the game. [Stuart] bought a cake for his daughter [Sharon]. [Sharon]'s son [Ray] will be surprised to see the cake. Based on the story, throught inductive reasoning asnwer the question: What is the     relationship between:  'Ray' with 'Debra'?
     & 
     \textcolor{red}{There is no information in the story about any relationship between Ray and Debra. Therefore, based on the information given in the story, it is not possible to determine any relationship between Ray and Debra using inductive reasoning.}
     & 
     \textcolor{red}{Based on the information given, Ray is the son of Sharon, and Debra is married to Stuart. Since Stuart is Sharon's father, Debra would be Ray's stepmother. Therefore, the relationship between Ray and Debra is that of stepson and stepmother.}
     & 
    \textcolor{red}{Based on the story, through inductive reasoning, we can infer that Ray and Debra are not related.}

    \textcolor{red}{- Sharon is Stuart's daughter.}

    \textcolor{red}{- Ray is Sharon's son.}

    \textcolor{red}{- Stuart bought a cake for Sharon.}

    \textcolor{red}{- Ray will be surprised to see the cake.}

    \textcolor{red}{This suggests that Ray is not Stuart's son, and therefore not related to Debra.}

    \textcolor{red}{However, it is possible that Ray is Sharon's stepson, or that he is related to Debra in some other way that is not explicitly stated in the story.}

    \textcolor{red}{In order to determine the exact relationship between Ray and Debra, we would need more information about the family dynamics.}
    
     &
     grandson
     \\ \hline  
     28. Story: [Ramon]'s aunt, [Helen], took him fishing [Ramon] has a few children, [Derrick], Bradley and Claire [Helen] asked her mother, [Mary], what they were going to have for lunch. Based on the story, throught inductive reasoning asnwer the question: What is the     relationship between:  'Mary' with 'Derrick'?
     & 
     \textcolor{red}{Based on the given story and inductive reasoning, there is no direct relationship between Mary and Derrick. Mary is only mentioned as the mother of Helen, who took Ramon fishing and asked Mary what they were going to have for lunch. There is no information provided that suggests Mary has any direct connection or relationship with Derrick. Therefore, we cannot infer any relationship between Mary and Derrick from the given story.}
     & 
     \textcolor{red}{Based on the story and using inductive reasoning, Mary is Derrick's great-grandmother. Helen is Ramon's aunt, and Ramon is Derrick's father. Therefore, Helen's mother, Mary, would be Ramon's grandmother, making her Derrick's great-grandmother.}
     & 
    \textcolor{red}{Through inductive reasoning, we can infer that Mary is Derrick's grandmother.}

    \textcolor{red}{- Helen is Ramon's aunt, which means that Ramon is her nephew.}

    \textcolor{red}{- Derrick is one of Ramon's children, which means that he is Helen's nephew.}

    \textcolor{red}{- Helen asked her mother, Mary, what they were going to have for lunch. This suggests that Mary is Helen's mother, and therefore Derrick's grandmother.}

    \textcolor{red}{Of course, we cannot be certain of this without more information. However, based on the information that we have, it is the most likely explanation.}

    \textcolor{red}{Here are some other possible relationships between Mary and Derrick:}
    
    \textcolor{red}{- Mary could be Derrick's great-aunt.}

    \textcolor{red}{- Mary could be Derrick's aunt by marriage.}

    \textcolor{red}{- Mary could be Derrick's godmother.}

    \textcolor{red}{However, these relationships are less likely, given the information that we have.}
     &
     mother
     \\ \hline  
     29. Story: [Kenda] made cookies with her grandson [Keith]. [Keith]'s father [William] is not allowed to eat sugar. [Marsha] made dinner for her husband [William]. [Edward] and his brother [Keith] surprised [Edward]'s grandmother [Kenda] with an 80th birthday party. Based on the story, throught inductive reasoning asnwer the question: What is the     relationship between:  'Kenda' with 'Marsha'?
     & 
     \textcolor{red}{Based on the given story, there is no explicit or implicit information provided about any relationship between Kenda and Marsha. Therefore, it is not possible to determine any relationship between the two based on inductive reasoning using the given story.}
     &
     \textcolor{ForestGreen}{Based on the information provided in the story, Kenda is Marsha's mother-in-law. This is because Marsha is married to William, who is Kenda's son.}
     & 
    \textcolor{red}{Based on the information given in the story, Kenda and Marsha are likely related as grandmother and daughter-in-law.}

    \textcolor{red}{- Kenda is Keith's grandmother.}

    \textcolor{red}{- Keith's father is William.}

    \textcolor{red}{- Marsha made dinner for her husband, William.}
    
    \textcolor{red}{- Edward and Keith surprised Edward's grandmother, Kenda, with an 80th birthday party.}
    
    \textcolor{red}{From this, we can infer that Marsha is William's wife, and therefore Keith's mother. This makes Kenda Marsha's mother-in-law, or grandmother.}

    \textcolor{red}{Of course, it is possible that Kenda and Marsha are not related at all, but the evidence in the story suggests that they are.}
     &
     mother-in-law
     \\ \hline  
     30. Story: [Franklin] rushed to the hospital to find out that his wife and already given birth to a boy and had named him [Juan]. [Juan] bought gifts for his lovely sister, [Etta] [Stephane] watched a golf tournament with her aunt [Marianne]. [Etta] began teaching her daughter, [Stephane], how to sew clothes. [Etta] had learned the trait from her father, [Franklin]. Based on the story, throught inductive reasoning asnwer the question: What is the  relationship between:  'Marianne' with 'Franklin'?
     & 
     \textcolor{red}{Based on the given story, there is no direct information that can provide insight into the relationship between Marianne and Franklin. The story primarily focuses on the events related to Franklin, his wife, their children, and other family members. Therefore, any relationship between Marianne and Franklin is not explicitly stated in the story. Hence, we cannot draw a conclusion about their relationship based on inductive reasoning.}
     &
     \textcolor{red}{Based on the story and using inductive reasoning, it can be inferred that Marianne is likely Franklin's sister-in-law. This is because Stephane, who is Etta's daughter, watches a golf tournament with her aunt Marianne. Since Etta is Franklin's daughter, Marianne is likely the sister of Franklin's wife.}
     & 
    \textcolor{red}{Based on the story, Marianne is Franklin's sister.}

    \textcolor{red}{Here is the reasoning:}

    \textcolor{red}{- Franklin is the father of Etta.}

    \textcolor{red}{- Etta is the mother of Stephanie.}

    \textcolor{red}{- Stephanie watched a golf tournament with her aunt Marianne.}

    \textcolor{red}{Therefore, Marianne is the sister of Franklin, who is the father of Etta, who is the mother of Stephanie.}

    \textcolor{red}{We can also infer this relationship by looking at the ages of the characters. Franklin is the oldest character in the story, followed by his wife, Etta, then their son Juan, and finally their granddaughter Stephanie. Marianne is not mentioned in any of the other relationships, so the most likely explanation is that she is Franklin's sister.}

    \textcolor{red}{In addition, the fact that Marianne is watching a golf tournament with Stephanie suggests that they are close in age. This further supports the theory that Marianne is Franklin's sister, as siblings are often close in age.}

    \textcolor{red}{Therefore, based on the information provided in the story, the most likely relationship between Marianne and Franklin is that they are siblings.}
    
     &
     daughter 
    \\ \hline 

     \multicolumn{5}{|l|}{\textbf{Dataset: bAbI (task16)}} 
     
     \\ \hline  
     1. Given facts: Greg is a frog. Brian is a lion. Lily is a rhino. Bernhard is a swan. Bernhard is gray. Brian is green. Greg is gray. Julius is a lion. Lily is green.

     Based on the given facts above, do a reasonable inference on this question using inductive reasoning: What color is Julius?
     &
     \textcolor{red}{Based on the given facts, we cannot be certain about the color of Julius. We could make an educated guess that Julius might be gray, as this color seems to be associated with his name, but this is not a definitive conclusion. It is possible that Julius is green like Brian or Lily, or even a different color altogether.}
     &
     \textcolor{ForestGreen}{Using inductive reasoning, we can make an inference about the color of Julius. We know that Brian is a lion and is green. Julius is also a lion. Although we do not have information on the color of all lions, we can infer that Julius might be green, as Brian is the only other lion mentioned and is green. This is not a definitive conclusion, but an educated guess based on the information provided.}
     & 
    \textcolor{red}{Sure, I can help you with that.}

    \textcolor{red}{Inductive reasoning is a type of reasoning that uses specific examples to make a general conclusion. In this case, we are given specific examples of the colors of lions and rhinos. We are told that Bernhard is a swan, and swans are typically white. We are also told that Brian is green, and lions are typically not green.}

    \textcolor{red}{Based on these specific examples, we can make the general conclusion that lions are typically not green. Therefore, it is reasonable to infer that Julius is not green. However, we cannot be certain of Julius's color without more information. For example, we do not know if Julius is a different species of lion than Brian, or if he has a genetic mutation that makes him green.}

    \textcolor{red}{Therefore, the most reasonable inference is that Julius is not green, but we cannot say for certain what color he is.}
     &
     green
     \\ \hline  
     2. Given facts: Lily is a rhino. Lily is green. Bernhard is a frog. Brian is a lion. Bernhard is white. Julius is a lion. Brian is gray. Julius is gray. Greg is a frog.

     Based on the given facts above, do a reasonable inference on this question using inductive reasoning: What color is Greg?
     & 
     \textcolor{red}{The fact that Lily is a rhino and green, is not directly relevant to determining Greg's color. Therefore, the most reasonable inference would be that we cannot determine the color of Greg based on the given facts alone, as there is no information provided that directly indicates the color of Greg}
     &
     \textcolor{ForestGreen}{Given the facts provided, we cannot make a definitive inference about Greg's color using inductive reasoning. However, since Bernhard is a frog and is white, we might assume that Greg, who is also a frog, could be white as well. This is not a guaranteed conclusion, but it is a reasonable inference based on the available information.}
     & 
     \textcolor{red}{Sure, I can help you with that.}

     \textcolor{red}{Inductive reasoning is a type of reasoning that uses specific examples to make general conclusions. In this case, we can use the given facts to make an inductive inference about the color of Greg.}

     \textcolor{red}{We are given that Lily is a rhino and is green, Bernhard is a frog and is white, Brian is a lion and is gray, and Julius is a lion and is gray. We are also given that Greg is a frog.}

     \textcolor{red}{Based on these facts, we can see that there is a correlation between the animal's species and its color. All of the rhinos we have been given information about are green, all of the frogs we have been given information about are white, and all of the lions we have been given information about are gray.}

     \textcolor{red}{Therefore, it is reasonable to infer that Greg, who is a frog, is white.}

     \textcolor{red}{Of course, this is just an inductive inference. It is possible that there are other frogs that are not white, or that there are other factors that determine the color of a frog besides its species. However, based on the given information, it is the most reasonable inference to make.}
     
     &
     white
     \\ \hline  
     3. Given facts: Greg is a swan. Greg is yellow. Julius is a frog. Brian is a swan. Julius is green. Brian is gray. Lily is a swan. Lily is yellow. Bernhard is a frog.

     Based on the given facts above, do a reasonable inference on this question using inductive reasoning: What color is Bernhard?
     & 
     \textcolor{ForestGreen}{Based on the given facts, we can make an induction that Bernhard is likely to be green because all the frogs mentioned in the fact have been assigned a color, and Julius is green. However, it is important to note that this conclusion is based solely on the information given and may not necessarily be accurate in all cases. There is a possibility that Bernhard is a different color, but we cannot conclude that without further information.}
     &
     \textcolor{ForestGreen}{Given the facts provided, we can't determine the exact color of Bernhard with certainty. However, using inductive reasoning, we can make an educated guess based on the information we have.}

     \textcolor{ForestGreen}{We know that Julius is a frog and is green. Since Bernhard is also a frog, it is reasonable to infer that Bernhard might be green as well, given the limited information we have on the colors of frogs. But it's important to note that this is just an educated guess and not a definitive answer.}
     & 
    \textcolor{red}{Inductive reasoning is a type of reasoning that uses specific examples to make a general conclusion. In this case, we are given a set of facts about swans and frogs, and we are asked to use inductive reasoning to infer the color of Bernhard.}

    \textcolor{red}{The first thing we notice is that all of the swans in the set of facts are yellow. This suggests that there is a connection between being a swan and being yellow. We can then use this connection to make an inductive inference about Bernhard.}

    \textcolor{red}{Since Bernhard is a frog, and all of the frogs in the set of facts are not yellow, we can infer that Bernhard is not yellow. However, we cannot say for sure what color Bernhard is, because we do not have any other information about him.}

    \textcolor{red}{Therefore, the most reasonable inference we can make about Bernhard's color is that he is not yellow. We cannot say for sure what color he is, but it is likely that he is green, gray, or some other color that is not yellow.}

     &
     green
     \\ \hline  
     4. Given facts: Brian is a frog. Julius is a swan. Bernhard is a rhino. Greg is a swan. Brian is green. Bernhard is green. Greg is gray. Julius is gray. Lily is a frog.

     Based on the given facts above, do a reasonable inference on this question using inductive reasoning: What color is Lily?
     & 
     \textcolor{ForestGreen}{Based on the given facts, a reasonable inference through inductive reasoning is that Lily is likely green. This is because all the frogs mentioned in the facts, including Brian and Lily, are described as green. While it is possible that Lily could be a different color, there is no evidence to support this conclusion based on the information given. Therefore, the most reasonable conclusion based on the available evidence is that Lily is green.}
     &
     \textcolor{ForestGreen}{Based on the given facts and using inductive reasoning, we can infer that Lily, being a frog, is likely to be green. This is because the only other frog mentioned, Brian, is green. While this is not a guarantee, it is a reasonable inference based on the available information.}
     & 
    \textcolor{ForestGreen}{Inductive reasoning is a type of reasoning that uses specific examples to make general inferences. In this case, we can use the given facts to make an inference about the color of Lily, the frog.}

    \textcolor{ForestGreen}{We know that all of the frogs in the given facts are green. We also know that Lily is a frog. Therefore, we can infer that Lily is likely to be green.}

    \textcolor{ForestGreen}{Of course, this is just an inference. It is possible that Lily is a different color, but the given facts suggest that she is most likely green.}
     &
     green
     \\ \hline  
     5. Given facts: Bernhard is a lion. Julius is a rhino. Brian is a swan. Brian is yellow. Julius is green. Bernhard is green. Lily is a frog. Greg is a rhino. Lily is yellow.

     Based on the given facts above, do a reasonable inference on this question using inductive reasoning: What color is Greg?
     & 
     \textcolor{red}{Based on the given facts, we can observe that there are no explicit statements about the color of Greg, but we can use inductive reasoning to make a reasonable inference. We know that Julius is green and Greg is also a rhino. However, we cannot conclude that Greg is green because Bernhard, another animal, is also green but is a lion. Additionally, Lily is a yellow frog, and Brian is a yellow swan, but we cannot conclude that all animals of the same species have the same color. Therefore, we cannot infer with certainty the color of Greg based on the given facts.}
     &
     \textcolor{ForestGreen}{Using inductive reasoning and the given facts, we can try to make an educated guess about the color of Greg. We have two examples of animals and their colors:}

     \textcolor{ForestGreen}{1) Julius is a rhino and is green.}
     
     \textcolor{ForestGreen}{2) Bernhard is a lion and is green.}
    
     \textcolor{ForestGreen}{Based on these two examples, it seems that green might be a common color for animals in this context. We also have one other rhino, Julius, who is green. Considering this information, we can infer that Greg, being a rhino, is more likely to be green as well. However, it is important to note that this is just an inference and not a definitive fact.}
     & 
     \textcolor{red}{Inductive reasoning is a type of logical reasoning in which you make a general conclusion based on a series of specific observations. In this case, the specific observations are the colors of the animals we have been given.}

     \textcolor{red}{We are told that Brian is a swan and is yellow, Lily is a frog and is yellow, and Greg is a rhino. We are also told that all swans are yellow and all frogs are yellow.}

     \textcolor{red}{Based on these observations, we can make the inductive inference that Greg is yellow. This is because all of the other animals that we have been given that are the same species as Greg are yellow. Therefore, it is reasonable to conclude that Greg is also yellow.}

     \textcolor{red}{Of course, this is not a 100\% certain conclusion. It is possible that there are some rhinos that are not yellow. However, based on the information that we have been given, it is the most reasonable inference to make.}
     &
     green
     \\ \hline  
     6. Given facts: Bernhard is a frog. Bernhard is green. Lily is a frog. Brian is a rhino. Brian is yellow. Greg is a lion. Julius is a rhino. Julius is green. Greg is gray.

     Based on the given facts above, do a reasonable inference on this question using inductive reasoning: What color is Lily?
     & 
     \textcolor{ForestGreen}{Based on the given facts, we can infer that Lily is likely green. This inference is based on the fact that both Bernhard and Julius are frogs, and Bernhard is described as being green. Additionally, the fact that Julius, who is also a rhino, is green suggests that being green is a characteristic shared by frogs and rhinos in this context. Therefore, it is reasonable to assume that Lily, who is also a frog, is likely green. However, it is important to note that this inference is not certain, as there is a possibility that Lily may be a different color.}
     &
     \textcolor{ForestGreen}{Using inductive reasoning, we can make a reasonable inference about the color of Lily. Since Bernhard is a frog and is green, we might assume that frogs are generally green. Therefore, it would be reasonable to infer that Lily, being a frog as well, is likely to be green.}
     & 
     \textcolor{ForestGreen}{Inductive reasoning is a type of reasoning that uses specific examples to make a general conclusion. In this case, we are given specific examples of frogs and their colors. We know that Bernhard is a frog and he is green, and we know that Lily is a frog. Therefore, we can infer that Lily is also green.}

    \textcolor{ForestGreen}{Here is the logical reasoning:}

    \textcolor{ForestGreen}{- All frogs that we know of are green.}

    \textcolor{ForestGreen}{- Lily is a frog.}

    \textcolor{ForestGreen}{- Therefore, Lily is green.}

    \textcolor{ForestGreen}{This is a reasonable inference because it is based on all of the available evidence. There is no evidence to suggest that Lily is any other color, so the most likely conclusion is that she is green.}
     &
     green
     \\ \hline 
     7. Given facts: Brian is a frog. Lily is a swan. Julius is a swan. Julius is yellow. Greg is a rhino. Brian is gray. Greg is yellow. Lily is yellow. Bernhard is a rhino.

     Based on the given facts above, do a reasonable inference on this question using inductive reasoning: What color is Bernhard?
     & 
     \textcolor{red}{Based on the given facts above, it is not possible to infer with certainty what color Bernhard is. The color of an animal cannot be determined solely based on the color of other animals of the same species, as individuals of the same species can have different colors. Therefore, it is important to have more information about Bernhard or the rhino species in general to make a reasonable inference about the color of Bernhard.}
     &
     \textcolor{ForestGreen}{Using inductive reasoning, we can infer that Bernhard might be yellow. This is because both Julius and Lily, who are swans, are yellow, and Greg, a rhino, is also yellow. Although Brian is gray, the majority of the animals mentioned are yellow, so it's reasonable to infer that Bernhard, another rhino, could also be yellow. However, it's important to note that inductive reasoning is not definitive, and there is still a possibility that Bernhard could be a different color.}
     & 
    \textcolor{ForestGreen}{Sure, I can do that. Here is the inductive reasoning I used to answer the question:}

    \textcolor{ForestGreen}{- All of the swans we know of are yellow.}

    \textcolor{ForestGreen}{- Greg is a rhino, but he is yellow because he is a swan.}

    \textcolor{ForestGreen}{- Bernhard is a rhino, so he is likely yellow as well.}

    \textcolor{ForestGreen}{Therefore, the reasonable inference based on the given facts is that Bernhard is yellow.}

    \textcolor{ForestGreen}{Of course, this is just an inference. It is possible that Bernhard is a different color, but the evidence we have suggests that he is yellow.}
    
     &
     yellow
     \\ \hline 
     8. Given facts: Lily is a swan. Brian is a rhino. Greg is a lion. Lily is white. Brian is green. Greg is green. Julius is a rhino. Julius is gray. Bernhard is a lion.

     Based on the given facts above, do a reasonable inference on this question using inductive reasoning: What color is Bernhard?
     & 
     \textcolor{ForestGreen}{Based on the given facts and our inductive reasoning, we can infer that Bernhard is most likely a green lion.}
     &
     \textcolor{ForestGreen}{Based on the given facts, we can make a reasonable inference using inductive reasoning. We know that:}

     \textcolor{ForestGreen}{1) Lily is a swan and white.}
     
     \textcolor{ForestGreen}{2) Brian is a rhino and green.}
     
     \textcolor{ForestGreen}{3) Greg is a lion and green.}
     
     \textcolor{ForestGreen}{4) Julius is a rhino and gray.}
     
     \textcolor{ForestGreen}{5) Bernhard is a lion.}
     
     \textcolor{ForestGreen}{From the information provided, we can see that there is no specific color pattern for rhinos, as Brian is green and Julius is gray. However, we do have information about one other lion, Greg, who is green. Since Bernhard is also a lion, we could infer that Bernhard might be green as well. This is not a definite conclusion, but it is a reasonable inference based on the limited information we have.}
     & 
     \textcolor{ForestGreen}{Inductive reasoning is a type of reasoning that uses specific examples to make a general conclusion. In this case, we can use the given facts to make an inductive inference about the color of Bernhard.}

     \textcolor{ForestGreen}{We are given that Lily is a swan, Brian is a rhino, and Greg is a lion. We are also given that Lily is white, Brian is green, and Greg is green. Finally, we are given that Julius is a rhino and Julius is gray.}

     \textcolor{ForestGreen}{Based on these facts, we can see that there is a pattern: all of the swans are white, all of the rhinos are green, and all of the lions are green. This suggests that there is a correlation between the animal's species and its color.}

     \textcolor{ForestGreen}{Therefore, we can use inductive reasoning to conclude that Bernhard, who is a lion, is probably green. However, it is important to note that this is just an inference. It is possible that Bernhard could be a different color, but the evidence suggests that he is most likely green.}
     
     &
     green
     \\ \hline 
     9. Given facts: Bernhard is a frog. Bernhard is white. Greg is a frog. Lily is a frog. Brian is a swan. Greg is yellow. Brian is gray. Lily is yellow. Julius is a frog.

     Based on the given facts above, do a reasonable inference on this question using inductive reasoning: What color is Julius?
     & 
     \textcolor{red}{Based on the given facts, it seems reasonable to infer that Julius could be either yellow or white, as these are the colors that have been mentioned for the other frogs. However, it's important to note that the information provided does not definitively state the color of Julius. Other colors could also be possible. Therefore, this inference is not certain, but rather a reasonable guess based on the available information.}
     &
     \textcolor{ForestGreen}{Using inductive reasoning, we can make an educated guess about the color of Julius based on the information given:}

     \textcolor{ForestGreen}{We have information about the colors of three frogs:}
    
     \textcolor{ForestGreen}{1) Bernhard - white}
     
     \textcolor{ForestGreen}{2) Greg - yellow}
     
     \textcolor{ForestGreen}{3) Lily - yellow}
     
     \textcolor{ForestGreen}{Given that two out of the three frogs (Greg and Lily) are yellow, it seems more likely that Julius, who is also a frog, might be yellow as well. However, this is not a definitive conclusion, as there is also a white frog (Bernhard). The information provided does not give us a certain answer, but we can infer that Julius is more likely to be yellow based on the observed pattern.}
     & 
     \textcolor{ForestGreen}{Inductive reasoning is a type of reasoning that uses specific examples to make general conclusions. In this case, we are given specific examples of frogs: Bernhard, Greg, and Lily. We know that Bernhard is white, Greg is yellow, and Lily is yellow. We can use this information to make the general conclusion that most frogs are yellow.}

     \textcolor{ForestGreen}{However, we cannot be 100\% certain about the color of Julius, because we only have three examples of frogs. It is possible that Julius is a different color, such as white or gray. However, based on the available evidence, it is more likely that Julius is yellow.}

     \textcolor{ForestGreen}{Here is the inductive reasoning process that I used to reach this conclusion:}

     \textcolor{ForestGreen}{1. All of the frogs that we have seen so far are yellow.} 

     \textcolor{ForestGreen}{2. Therefore, it is likely that most frogs are yellow.} 

     \textcolor{ForestGreen}{3. However, it is possible that Julius is a different color.} 

     \textcolor{ForestGreen}{4. Therefore, the most likely color for Julius is yellow.} 

     \textcolor{ForestGreen}{Of course, this is just an inference. We cannot be sure about the color of Julius until we see him. However, based on the available evidence, it is more likely that he is yellow.} 
     &
     yellow
     \\ \hline 
     10. Given facts: Lily is a frog. Julius is a lion. Bernhard is a lion. Brian is a lion. Lily is gray. Greg is a lion. Greg is gray. Bernhard is gray. Brian is gray.

     Based on the given facts above, do a reasonable inference on this question using inductive reasoning: What color is Julius?
     & 
     \textcolor{red}{Based on the given facts, we can use inductive reasoning to infer that there is no information provided about the color of Julius. Although we know that Julius is a lion, we do not have any information about his color. The fact that Lily, Greg, Bernhard, and Brian are all gray does not necessarily imply that all lions are gray, as there may be lions with other colors. Therefore, we cannot infer the color of Julius based on the information provided.}
     &
     \textcolor{ForestGreen}{Based on the given facts, we can observe that most lions mentioned (Bernhard, Brian, and Greg) are gray. Using inductive reasoning, we can make a reasonable inference that Julius, who is also a lion, might be gray as well. However, it's important to note that inductive reasoning is not always conclusive, and there could be exceptions.}
     & 
     \textcolor{ForestGreen}{Inductive reasoning is a type of reasoning that uses specific examples to make a general conclusion. In this case, the specific examples are that all of the lions we have seen so far are gray. The general conclusion is that Julius is probably gray as well.}

     \textcolor{ForestGreen}{Of course, there is always the possibility that Julius is a different color, but the inductive reasoning suggests that he is most likely gray. This is because there is a strong pattern of all lions being gray in the given facts.}

     \textcolor{ForestGreen}{Here is the inductive reasoning process in more detail:}

     \textcolor{ForestGreen}{1. We are given the facts that Lily is a frog, Julius is a lion, Bernhard is a lion, Brian is a lion, Lily is gray, Greg is a lion, Greg is gray, Bernhard is gray, and Brian is gray.}

     \textcolor{ForestGreen}{2. We notice that all of the lions we have seen so far are gray.}

     \textcolor{ForestGreen}{3. We conclude that Julius is probably gray as well.}

     \textcolor{ForestGreen}{This is a reasonable inference because it is based on a strong pattern of evidence. However, it is important to remember that inductive reasoning is not perfect. There is always the possibility that Julius is a different color, even though all of the other lions we have seen are gray.}
     
     &
     gray
      \\ \hline 
      11. Given facts: Lily is a lion. Brian is a swan. Lily is white. Greg is a rhino. Julius is a frog. Julius is white. Bernhard is a lion. Greg is yellow. Brian is white.

     Based on the given facts above, do a reasonable inference on this question using inductive reasoning: What color is Bernhard?
     & 
     \textcolor{red}{Based on the given facts, we cannot make a reasonable inference on the color of Bernhard's fur.}
     &
     \textcolor{ForestGreen}{Using inductive reasoning, we can infer that Bernhard's color might be white. This is because both Lily and Julius are white, and they belong to different species. Additionally, Brian is also white, despite being a swan. While this is not conclusive evidence, it suggests that there might be a pattern where animals in this context are predominantly white, so Bernhard, as a lion, could also be white. However, it's important to note that this inference is not certain, as Greg's color is yellow, which shows variation in colors among the animals mentioned.}
     & 
     \textcolor{ForestGreen}{Inductive reasoning is a type of reasoning that uses specific examples to make general conclusions. In this case, we are given specific examples of the colors of lions and swans. We are told that Lily is a lion and she is white, and that Brian is a swan and he is white. We can then infer that most lions and swans are white.}

     \textcolor{ForestGreen}{We are also given that Bernhard is a lion. Therefore, we can infer that Bernhard is most likely white. However, we cannot be completely certain of this, as there may be some lions that are not white.}

     \textcolor{ForestGreen}{Based on the given facts and our inferences, we can conclude that Bernhard is most likely white. However, there is a small chance that he is not white.}
     
     &
     white
     \\ \hline 
     12. Given facts: Lily is a lion. Brian is a frog. Julius is a rhino. Julius is white. Greg is a swan. Bernhard is a lion. Bernhard is gray. Brian is white. Greg is white.

     Based on the given facts above, do a reasonable inference on this question using inductive reasoning: What color is Lily?
     & 
     \textcolor{red}{Based on the given facts, we can infer that there is no clear correlation between the type of animal and its color. Therefore, we cannot determine the color of Lily based solely on the given information. We cannot conclude that Lily is either white or gray, or any other color, as there is no direct or indirect evidence to support such an inference.}
     &
     \textcolor{red}{Using inductive reasoning, we can infer that Lily might be a similar color to the other animals mentioned. Among the given facts, three animals are described as white (Julius, Brian, and Greg). Bernhard is the only animal described as gray.}

     \textcolor{red}{Based on the prevalence of white color among the animals listed, we might infer that Lily is more likely to be white. However, this inference is not definitive and is subject to change if more information becomes available.}
     & 
     \textcolor{red}{Inductive reasoning is a type of reasoning that uses specific examples to make general conclusions. In this case, the specific examples are the colors of the other lions and animals in the list. The general conclusion is the color of Lily.}

     \textcolor{red}{The only lions in the list are Bernhard and Lily. Bernhard is gray, so we can infer that Lily is probably gray as well. However, there is one other factor to consider: the color of the other animals in the list.}

     \textcolor{red}{All of the other animals in the list are white: Julius, Brian, and Greg. This suggests that the color white is more common among animals in this list than gray. Therefore, it is also possible that Lily is white.}

     \textcolor{red}{Ultimately, the answer to the question of what color Lily is is uncertain. However, we can make a reasonable inference based on the given facts. The most likely answer is that Lily is gray, but it is also possible that she is white.}
     &
     gray
      \\ \hline 
     13. Given facts: Greg is a rhino. Julius is a swan. Julius is green. Lily is a lion. Bernhard is a lion. Greg is white. Lily is gray. Brian is a swan. Brian is yellow.

     Based on the given facts above, do a reasonable inference on this question using inductive reasoning: What color is Bernhard?
     & 
     \textcolor{red}{Based on the given facts, it is not possible to make a reasonable inference about the color of Bernhard. While it is stated that Lily is a lion and is gray in color, it is also mentioned that Bernhard is a lion, but there is no information about his color. The colors of other animals, such as Greg the rhino, Julius the swan, and Brian the swan, do not provide any information about the color of Bernhard. Therefore, it would not be possible to make an accurate inference about Bernhard's color based on the given facts using inductive reasoning.}
     &
     \textcolor{ForestGreen}{Based on the given facts, we cannot definitively determine Bernhard's color using inductive reasoning. However, we can make an educated guess based on the information provided. We know that:}

     \textcolor{ForestGreen}{1) Lily is a lion and gray.}
     
     \textcolor{ForestGreen}{2) Bernhard is a lion.}
     
     \textcolor{ForestGreen}{Since both Lily and Bernhard are lions, it's possible that they share similar characteristics, such as color. Using inductive reasoning, we might infer that Bernhard is also gray, like Lily. However, this is not a definitive conclusion, as there might be lions with different colors that we are not aware of in this context.}
     & 
     \textcolor{ForestGreen}{Inductive reasoning is a type of reasoning that uses specific examples to make a general conclusion. In this case, we can use the given facts to make an inductive inference about the color of Bernhard.}

     \textcolor{ForestGreen}{We know that all of the swans in the given facts are green, and all of the lions are gray. Therefore, we can infer that Bernhard, who is a lion, is also gray.}

     \textcolor{ForestGreen}{We can also use the fact that Brian, another swan, is yellow to support our inference. This is because Brian is the only exception to the rule that all swans are green. If Bernhard were not gray, then Brian would be the second exception to this rule. However, since there is only one exception, it is more likely that Bernhard is gray, just like all of the other lions.}

     \textcolor{ForestGreen}{Therefore, the reasonable inference based on the given facts is that Bernhard is gray.}
     &
     gray
     \\ \hline 
     14. Given facts: Greg is a frog. Julius is a rhino. Greg is yellow. Bernhard is a swan. Brian is a lion. Bernhard is white. Julius is yellow. Brian is green. Lily is a swan.

     Based on the given facts above, do a reasonable inference on this question using inductive reasoning: What color is Lily?
     & 
     \textcolor{ForestGreen}{Based on the given facts, we can infer that Lily is likely to be white, as Bernhard and Lily are both swans, and Bernhard is described as white. Therefore, it is reasonable to assume that Lily might also have the same white color as Bernhard. However, it is also important to note that the color of an animal cannot be accurately determined solely based on the species to which it belongs. So, this inference is not completely certain, but it seems likely based on the given information.}
     &
     \textcolor{ForestGreen}{Based on the given facts, we cannot definitively determine Lily's color using inductive reasoning. However, we can make a reasonable guess by analyzing the information provided. We know that Bernhard is a swan and is white. Lily is also a swan, so it might be reasonable to infer that Lily could also be white, as we have no other color examples for swans. This inference is not certain, but it is the best guess we can make given the available information.}
     & 
     \textcolor{ForestGreen}{Inductive reasoning is a type of reasoning that uses specific observations to make general inferences. In this case, we are given specific observations about the colors of swans. We are told that Bernhard is a swan and he is white. We are also told that Julius is a swan and he is yellow. Based on these observations, we can infer that most swans are white.}

     \textcolor{ForestGreen}{However, we cannot be 100\% certain that Lily is white, because we have not observed her color directly. It is possible that she is a black swan, or a swan of some other color. However, based on the inductive reasoning, it is more likely that she is white.}






     &
     white
     \\ \hline 
     15. Given facts: Bernhard is a frog. Julius is a frog. Julius is white. Bernhard is white. Lily is a frog. Brian is a lion. Brian is yellow. Lily is yellow. Greg is a lion.

     Based on the given facts above, do a reasonable inference on this question using inductive reasoning: What color is Greg?
     & 
     \textcolor{red}{It is not possible to make a reliable inference on the color of Greg based on the given facts. Although it is stated that Brian is a yellow lion, and Lily is also yellow and a frog, there is no information provided about the color of Greg, who is also a lion. It is possible that Greg is yellow like Brian and Lily, but it is also possible that he is a different color. Therefore, any inference about the color of Greg would be speculative and not based on the given facts.}
     &
     \textcolor{ForestGreen}{Using inductive reasoning and the given facts, it is reasonable to infer that Greg, being a lion, is likely to be yellow. This is because Brian, the other lion mentioned, is yellow. However, it is important to note that inductive reasoning is not always 100\% accurate, as it relies on patterns and generalizations rather than strict logical deduction.}
     & 
     \textcolor{ForestGreen}{We see that all of the frogs in the group are white, and all of the lions are yellow. This suggests that there is a correlation between the species of an animal and its color.}

     \textcolor{ForestGreen}{Since Greg is a lion, it is reasonable to infer that he is yellow, as well.}

     \textcolor{ForestGreen}{Of course, this is just an inference. It is possible that Greg is a different color, but the evidence suggests that he is yellow.}
     &
     yellow
     \\ \hline 
     16. Given facts: Lily is a swan. Lily is green. Greg is a rhino. Julius is a frog. Brian is a lion. Bernhard is a frog. Greg is white. Brian is green. Julius is yellow.

     Based on the given facts above, do a reasonable inference on this question using inductive reasoning: What color is Bernhard?
     & 
     \textcolor{ForestGreen}{Based on the given facts and inductive reasoning, a reasonable inference is that Bernhard is likely to be yellow.}
     &
     \textcolor{ForestGreen}{Using inductive reasoning, we can make an educated guess about the color of Bernhard, who is a frog. We know that Julius is also a frog and is yellow. Since we don't have any other information about the colors of frogs, we might infer that other frogs, like Bernhard, could also be yellow. However, this is only an educated guess, and we cannot be certain about Bernhard's color without more information.}
     & 
     \textcolor{red}{Inductive reasoning is a type of reasoning that uses specific examples to make a general conclusion. In this case, we have the specific examples of Lily (green), Greg (white), Brian (green), and Julius (yellow). We are asked to make a general conclusion about the color of Bernhard, who is also a frog.}

     \textcolor{red}{Based on the given facts, we can see that there is no clear pattern in the colors of the frogs. Lily is green, Greg is white, Brian is green, and Julius is yellow. This suggests that there is no single color that all frogs are. Therefore, it is reasonable to infer that Bernhard could be any color.}

     \textcolor{red}{However, we can also see that there is a slight trend towards green frogs. Lily and Brian are both green, and green is the most common color for frogs in general. Therefore, it is also reasonable to infer that Bernhard is more likely to be green than any other color.}

     \textcolor{red}{Ultimately, the best answer to the question of what color Bernhard is is "we don't know for sure." However, based on the given facts, we can make a reasonable inference that Bernhard is either green or some other color.}
     
     &
     yellow
     \\ \hline   
     17. Given facts: Greg is a lion. Greg is gray. Julius is a frog. Brian is a swan. Julius is gray. Bernhard is a frog. Bernhard is gray. Brian is green. Lily is a swan.

     Based on the given facts above, do a reasonable inference on this question using inductive reasoning: What color is Lily?
     & 
     \textcolor{ForestGreen}{Based on the given facts, we can make an inference that Lily, being a swan like Brian, is also green. This is because all the swans mentioned so far, namely Brian and Lily, have been associated with a color, and both are different colors. Specifically, Brian is green, and since there is no other information regarding swans, it is reasonable to infer that Lily, being a swan, would also have a unique color, which is green. However, it is important to note that this inference is based solely on the given facts, and additional information or evidence could change the conclusion.}
     &
     \textcolor{ForestGreen}{Based on the given facts, we can use inductive reasoning to make an inference about Lily's color. Although we don't have information about Lily's color directly, we can observe patterns in the given data.}

     \textcolor{ForestGreen}{We know that:}
    
     \textcolor{ForestGreen}{1) Greg is a lion, and he is gray.}
     
     \textcolor{ForestGreen}{2) Julius is a frog, and he is gray.}
     
     \textcolor{ForestGreen}{3) Brian is a swan, and he is green.}
     
     \textcolor{ForestGreen}{4) Bernhard is a frog, and he is gray.}
     
     \textcolor{ForestGreen}{5) Lily is a swan.}

     \textcolor{ForestGreen}{Since we only have information about the color of one other swan, Brian, who is green, we can infer that it is possible for swans to be green. However, we cannot be certain about Lily's color based solely on this information. It would be reasonable to guess that Lily might also be green, but without more data on swans and their colors, we cannot make a definitive conclusion.}
     & 
     \textcolor{red}{Inductive reasoning is a type of reasoning that involves making generalizations based on specific observations. In this case, we are given the following observations:}

     \textcolor{red}{- All of the swans we have seen so far are white.}

     \textcolor{red}{- Lily is a swan.}

     \textcolor{red}{Therefore, we can use inductive reasoning to infer that Lily is also white.}

     \textcolor{red}{Of course, it is possible that there are other swans that are not white. However, based on the evidence we have, it is the most reasonable inference to make that Lily is white.}
     
     &
     green
      \\ \hline 
     18. Given facts: Greg is a rhino. Brian is a swan. Brian is green. Bernhard is a rhino. Lily is a frog. Lily is gray. Bernhard is gray. Greg is gray. Julius is a frog.

     Based on the given facts above, do a reasonable inference on this question using inductive reasoning: What color is Julius?
     & 
     \textcolor{ForestGreen}{Based on the available information and inductive reasoning, it is plausible to infer that Julius might be gray. However, this is not a definitive conclusion, as there is not enough information to confirm Julius's actual color.}
     & 
     \textcolor{ForestGreen}{Using inductive reasoning, we can make an educated guess about Julius's color based on the information provided. We know that Julius is a frog. The only other frog mentioned is Lily, who is gray. Based on this, we can reasonably infer that Julius is likely to be gray as well, as both Julius and Lily are frogs and the only color information available for frogs is gray}
     & 
     \textcolor{ForestGreen}{Inductive reasoning is a type of reasoning that uses specific observations to make general conclusions. In this case, we are given the following specific observations:}

     \textcolor{ForestGreen}{- All of the rhinos we know of are gray.}

     \textcolor{ForestGreen}{- All of the frogs we know of are gray.}

     \textcolor{ForestGreen}{Based on these observations, we can make the following general conclusion:}

     \textcolor{ForestGreen}{- Most likely, Julius is gray.}

     \textcolor{ForestGreen}{It is important to note that this is just an inference, and it is not possible to be 100\% certain of Julius's color without more information. However, based on the evidence we have, it is the most likely conclusion.}
     &
     gray
      \\ \hline 
     19. Given facts: Greg is a swan. Brian is a rhino. Brian is green. Julius is a frog. Greg is gray. Lily is a lion. Lily is yellow. Bernhard is a lion. Julius is green.

     Based on the given facts above, do a reasonable inference on this question using inductive reasoning: What color is Bernhard?
     & 
     \textcolor{ForestGreen}{We cannot be entirely certain of Bernhard's color based solely on the given facts, but we can make an educated guess that he may be yellow like Lily.}
     &
     \textcolor{ForestGreen}{Based on the given facts, we can use inductive reasoning to make a reasonable inference about Bernhard's color. Since Lily is a lion and she is yellow, we can infer that lions might typically be yellow. Thus, it is reasonable to assume that Bernhard, who is also a lion, could be yellow. However, it is important to note that inductive reasoning is not foolproof and there could be exceptions.}
     & 
     \textcolor{ForestGreen}{Sure, I can help you with that.}
     
     \textcolor{ForestGreen}{Inductive reasoning is a type of reasoning that uses specific examples to make a general conclusion. In this case, we are given specific examples of the colors of lions and rhinos. We are told that Brian is a rhino and he is green, and that Lily is a lion and she is yellow. Based on these examples, we can infer that lions are typically yellow.}

     \textcolor{ForestGreen}{We are also told that Bernhard is a lion. Therefore, we can infer that Bernhard is likely to be yellow. However, it is important to note that this is just an inference. It is possible that Bernhard is a different color, but the evidence suggests that he is likely to be yellow.}
     
     &
     yellow
     \\ \hline 
     20. Given facts: Julius is a rhino. Lily is a frog. Brian is a lion. Julius is white. Greg is a rhino. Lily is white. Greg is yellow. Brian is yellow. Bernhard is a lion.

     Based on the given facts above, do a reasonable inference on this question using inductive reasoning: What color is Bernhard?
     & 
     \textcolor{ForestGreen}{Based on the given facts, it can be inferred through inductive reasoning that Bernhard is likely yellow. This inference is drawn by observing that all lions mentioned in the facts, including Brian, are yellow in color. Therefore, it is reasonable to assume that Bernhard, as a lion, would also share this trait with the other lions in the group. However, it is important to note that this inference is not certain or conclusive as there may be other factors that could affect Bernhard's color, such as genetic variation or environmental factors.}
     & 
     \textcolor{ForestGreen}{Using inductive reasoning, we can make an educated guess about Bernhard's color based on the information provided. Since Brian is a lion and is yellow, it seems more likely that Bernhard, who is also a lion, might be yellow. However, this is only an inference and not a definitive conclusion.}
     & 
     \textcolor{ForestGreen}{Inductive reasoning is a type of reasoning that uses specific examples to draw general conclusions. In this case, we are given specific examples of the colors of rhinos and lions. We are told that Julius and Greg are rhinos, and they are both white. We are also told that Brian and Bernhard are lions, and they are both yellow.}

     \textcolor{ForestGreen}{Based on these specific examples, we can make the following inductive inference: Lions are yellow and rhinos are white.}

     \textcolor{ForestGreen}{Based on these specific examples, we can make the following inductive inference: Lions are yellow and rhinos are white.}

     \textcolor{ForestGreen}{Therefore, the most likely color of Bernhard is yellow. However, it is important to note that this is just an inference, and it is possible that Bernhard could be a different color. For example, it is possible that Bernhard is a white lion, or that he is a lion with a rare genetic mutation that causes him to be a different color.}

     \textcolor{ForestGreen}{Ultimately, the only way to know for sure what color Bernhard is would be to see him.}
     
     &
     yellow
      \\ \hline 
      21. Given facts: Julius is a swan. Lily is a swan. Lily is white. Julius is white. Greg is a rhino. Greg is green. Bernhard is a swan. Bernhard is yellow. Brian is a swan.

     Based on the given facts above, do a reasonable inference on this question using inductive reasoning: What color is Brian?
     & 
     \textcolor{red}{Based on the given facts, it is reasonable to infer that Brian is likely to be white. This is because all the swans mentioned in the given facts (Julius, Lily, Bernhard) are white. Although there is no direct information about Brian's color, we can use inductive reasoning to make an educated guess that Brian is also white, since all the other swans mentioned in the facts are white. However, it is important to note that this inference is not 100\% certain, as there could be other swans that are not mentioned in the given facts that are a different color.}
     &
     \textcolor{red}{Using inductive reasoning and the given facts, we can infer that Brian, being a swan, is more likely to be white. This is because, among the swans mentioned (Julius, Lily, and Bernhard), two of them (Julius and Lily) are white. However, it is important to note that this inference is not definitive, as Bernhard demonstrates that swans can be other colors as well.}
     & 
     \textcolor{red}{Based on the given facts, we can use inductive reasoning to infer that Brian is white.}

     \textcolor{red}{We know that all the swans we have seen so far are white, and there is no evidence to suggest that Brian is any different. In fact, the fact that Bernhard is yellow suggests that swans can be different colors, but that white is still the most common color.}

     \textcolor{red}{Of course, it is possible that Brian is a different color, but the evidence suggests that he is white. This is an example of inductive reasoning, which is a type of reasoning that uses specific observations to make general conclusions. In this case, we are using the specific observations that all the swans we have seen so far are white to conclude that Brian is also white.}

     \textcolor{red}{It is important to note that inductive reasoning is not always reliable. There is always the possibility that Brian is a different color, even though all the other swans we have seen are white. However, based on the evidence we have, the most reasonable inference is that Brian is white.}
     
     &
     yellow
     \\ \hline 
     22. Given facts: Bernhard is a lion. Greg is a rhino. Lily is a swan. Greg is green. Bernhard is yellow. Julius is a frog. Julius is gray. Brian is a swan. Brian is green.

     Based on the given facts above, do a reasonable inference on this question using inductive reasoning: What color is Lily?
     & 
     \textcolor{ForestGreen}{Based on the given facts, we can use inductive reasoning to make an inference about the color of Lily. We know that Lily is a swan and that Brian, who is also a swan, is green. Therefore, we might assume that Lily is also green. However, this inference is not necessarily valid because not all members of the same species have the same color.}
     &
     \textcolor{ForestGreen}{We cannot definitively determine Lily's color using inductive reasoning based on the given facts. The information provided does not establish a pattern linking a specific color to swans. However, since we know that Brian, another swan, is green, we might infer that there is a possibility that Lily could also be green. Nonetheless, this remains speculative, as no definitive correlation between swan species and color has been established in the given facts.}
     & 
     \textcolor{ForestGreen}{Inductive reasoning is a type of reasoning that uses specific examples to make a general conclusion. In this case, we have the following specific examples:}

     \textcolor{ForestGreen}{- Greg is a rhinoceros and is green.}

     \textcolor{ForestGreen}{- Brian is a swan and is green.}

     \textcolor{ForestGreen}{We can use these examples to make the general conclusion that swans are green. Therefore, we can infer that Lily, who is a swan, is also green.}

     \textcolor{ForestGreen}{Of course, this is just an inference. It is possible that Lily is not green, but there is no evidence to suggest that she is not. Based on the given facts, the most reasonable inference is that Lily is green.}
     
     &
     green
     \\ \hline 
     23. Given facts: Bernhard is a lion. Lily is a lion. Lily is gray. Brian is a swan. Bernhard is gray. Julius is a frog. Brian is yellow. Julius is gray. Greg is a frog.

     Based on the given facts above, do a reasonable inference on this question using inductive reasoning: What color is Greg?
     & 
     \textcolor{ForestGreen}{Using inductive reasoning based on the given facts, we can infer that Greg is likely to be gray, as both Julius and Bernhard are frogs, and they are both gray. Similarly, both Lily and Bernhard are lions, and while Bernhard is gray, Lily's color is not relevant to the question. Therefore, it is reasonable to assume that Greg, who is also a frog, shares the same color as the other frog mentioned, which is gray. However, it is important to note that inductive reasoning is not infallible and relies on the assumption that the patterns observed in the given facts will continue to hold true in other instances.}
     &
     \textcolor{ForestGreen}{Based on the given facts, it is difficult to make a definitive inference about Greg's color using inductive reasoning. However, we can make a reasonable guess based on the information provided.}

    \textcolor{ForestGreen}{Among the animals mentioned, we have:}
    
     \textcolor{ForestGreen}{1) Two lions (Bernhard and Lily) - both are gray.}
     
     \textcolor{ForestGreen}{2) One swan (Brian) - yellow.}
     
     \textcolor{ForestGreen}{3) Two frogs (Julius and Greg) - Julius is gray.}
    
     \textcolor{ForestGreen}{Considering that both lions are gray and one of the frogs (Julius) is also gray, it seems that gray is a more common color among these animals. Therefore, using inductive reasoning, we might reasonably infer that Greg, as a frog, is more likely to be gray. However, this inference is based on limited information and may not be accurate.}
     & 
     \textcolor{ForestGreen}{Inductive reasoning is a type of reasoning that uses specific examples to make general conclusions. In this case, we are given specific examples of the colors of lions and frogs. We are told that Lily, a lion, is gray, and that Bernhard, a lion, is gray. We are also told that Julius, a frog, is gray.}

     \textcolor{ForestGreen}{Based on these specific examples, we can make the general conclusion that most lions and frogs are gray. This is because all of the lions and frogs that we have been told about are gray. Therefore, it is reasonable to infer that Greg, a frog, is also gray.}

     \textcolor{ForestGreen}{Of course, it is possible that Greg is not gray. There is always the possibility that there are some lions or frogs that are not gray. However, based on the information that we have been given, it is more likely that Greg is gray.}
     &
     gray
     \\ \hline 
     24. Given facts: Lily is a lion. Greg is a lion. Lily is green. Greg is green. Bernhard is a frog. Brian is a frog. Julius is a lion. Julius is green. Brian is yellow.

     Based on the given facts above, do a reasonable inference on this question using inductive reasoning: What color is Bernhard?
     & 
     \textcolor{red}{Based on the given facts, we can only infer that Bernhard is a frog and there is no information about his color. Therefore, we cannot use inductive reasoning to determine the color of Bernhard.}
     &
     \textcolor{ForestGreen}{Based on the given facts, we can observe a pattern: All the lions mentioned (Lily, Greg, and Julius) are green. However, this information doesn't necessarily give us a definitive answer about Bernhard's color since Bernhard is a frog, not a lion.}

     \textcolor{ForestGreen}{The only information we have about frogs is that Brian is yellow. Since we only have information about one frog's color, it is not sufficient to make a strong inductive inference about Bernhard's color. However, if we were to make a weak inductive inference based on the limited information available, we could guess that Bernhard might also be yellow, like Brian. But it is important to note that this inference is weak and could be incorrect due to the lack of data on frog colors.}
     & 
     \textcolor{red}{Inductive reasoning is a type of reasoning that uses specific examples to draw general conclusions. In this case, we are given specific examples of lions and frogs, and their associated colors. We are then asked to draw a general conclusion about the color of Bernhard, who is a frog.}

     \textcolor{red}{Based on the given facts, we can see that all of the lions are green. This suggests that there is a correlation between being a lion and being green. We can also see that one of the frogs, Brian, is yellow. However, the other frog, Bernhard, is not mentioned in the same sentence as a color.}

     \textcolor{red}{Therefore, we cannot be sure what color Bernhard is. However, based on the given facts, it is more likely that Bernhard is green than yellow. This is because there is a stronger correlation between being a lion and being green than there is between being a frog and being yellow.}
     
     &
     yellow
     \\ \hline 
     25. Given facts: Lily is a lion. Greg is a rhino. Greg is green. Brian is a frog. Brian is green. Bernhard is a rhino. Julius is a lion. Lily is gray. Julius is gray.

     Based on the given facts above, do a reasonable inference on this question using inductive reasoning: What color is Bernhard?
     & 
     \textcolor{red}{Based on the given facts above, we know that Lily is a gray lion, Julius is a gray lion, Greg is a green rhino, and Brian is a green frog. However, there is no information provided about the color of Bernhard.}

     \textcolor{red}{Therefore, we cannot use inductive reasoning to make a reasonable inference about the color of Bernhard. Inductive reasoning involves making generalizations based on specific observations or patterns, but we do not have enough information to make any reasonable generalizations about the color of Bernhard.}
     &
     \textcolor{ForestGreen}{Based on the given facts, we cannot definitively determine Bernhard's color using inductive reasoning. However, we can make an educated guess. We know that Greg is a green rhino, and while this is not enough to establish a pattern, we do not have any other color information about rhinos. Therefore, it is reasonable to infer that Bernhard might also be green, although we cannot be certain.}
     & 
     \textcolor{ForestGreen}{Inductive reasoning is a type of reasoning that uses specific examples to draw general conclusions. In this case, we are given the following facts:}

     \textcolor{ForestGreen}{- Lily is a lion and gray.}

     \textcolor{ForestGreen}{- Julius is a lion and gray.}

     \textcolor{ForestGreen}{- Greg is a rhino and green.}

     \textcolor{ForestGreen}{- Brian is a frog and green.}

     \textcolor{ForestGreen}{- Bernhard is a rhino.}

     \textcolor{ForestGreen}{We are asked to use inductive reasoning to infer the color of Bernhard.}

     \textcolor{ForestGreen}{Based on the given facts, we can see that all of the lions in the list are gray, and all of the rhinos in the list are green. Therefore, it is reasonable to infer that Bernhard, who is a rhino, is green.}

     \textcolor{ForestGreen}{Of course, this is just an inference. It is possible that Bernhard is a different color, but the given facts suggest that he is green.}
     & 
     green
     \\ \hline 
     26. Given facts: Lily is a frog. Greg is a swan. Brian is a swan. Brian is gray. Lily is yellow. Bernhard is a lion. Bernhard is gray. Julius is a rhino. Julius is white.

     Based on the given facts above, do a reasonable inference on this question using inductive reasoning: What color is Greg?
     & 
     \textcolor{red}{Based on the given facts, Greg is a swan, and there is no information given about the color of swans in general or Greg specifically. Therefore, we cannot infer the color of Greg with certainty using inductive reasoning.}
     & 
     \textcolor{ForestGreen}{Given the facts, we know that Lily is a yellow frog, Greg and Brian are swans, Bernhard is a gray lion, and Julius is a white rhino. We are only given information about the color of Brian among the swans, and he is gray.}

     \textcolor{ForestGreen}{Using inductive reasoning, we might infer that Greg, being a swan like Brian, could also be gray. However, this is not a definitive conclusion, as swans can have different colors. We would need more information about the color distribution of swans in this context to make a more accurate inference.}
     & 
     \textcolor{ForestGreen}{Inductive reasoning is a type of reasoning that uses specific examples to draw general conclusions. In this case, we are given the specific examples of Brian and Julius, both of whom are swans and are gray. We are then asked to draw a general conclusion about the color of Greg, who is also a swan.}

     \textcolor{ForestGreen}{Based on the inductive reasoning, we can conclude that Greg is likely to be gray. This is because all of the swans that we have been given examples of are gray. However, it is important to note that this is just an inference. It is possible that there are other swans that are not gray.}

     \textcolor{ForestGreen}{Here is the inductive reasoning process in more detail:}

     \textcolor{ForestGreen}{1. We are given the specific examples of Brian and Julius, both of whom are swans and are gray.}

     \textcolor{ForestGreen}{2. We are asked to draw a general conclusion about the color of Greg, who is also a swan.}

     \textcolor{ForestGreen}{3. We notice that all of the swans that we have been given examples of are gray.}

     \textcolor{ForestGreen}{4. We conclude that Greg is likely to be gray.}

     \textcolor{ForestGreen}{It is important to note that this is just an inference. It is possible that there are other swans that are not gray. However, based on the evidence that we have been given, it seems likely that Greg is gray.}
     &
     gray
     \\ \hline 
     27. Given facts: Greg is a frog. Lily is a frog. Brian is a frog. Bernhard is a swan. Bernhard is yellow. Brian is white. Julius is a frog. Lily is yellow. Julius is yellow.

     Based on the given facts above, do a reasonable inference on this question using inductive reasoning: What color is Greg?
     & 
     \textcolor{red}{Based on the given facts above, we know that Greg is a frog, but we do not know what color he is. We cannot make a reasonable inference about the color of Greg, as there is no information provided about his color in the given facts.}

    \textcolor{red}{Inductive reasoning involves making generalizations based on specific observations, but in this case, we do not have enough specific observations to make a generalization about the color of Greg. Therefore, we cannot use inductive reasoning to determine the color of Greg.}
     & 
    \textcolor{ForestGreen}{Using inductive reasoning, we can infer the following:}

     \textcolor{ForestGreen}{1) There are four frogs mentioned: Greg, Lily, Brian, and Julius.}
     
     \textcolor{ForestGreen}{2) Lily is yellow, Julius is yellow, and Brian is white.}
     
     \textcolor{ForestGreen}{While we do not have direct information about Greg's color, we know that two out of the three frogs with known colors are yellow. Given this pattern, it would be reasonable to infer that Greg might also be yellow. However, this inference is not definitive, as we don't have enough information to be certain about Greg's color.}
     & 
     \textcolor{ForestGreen}{Inductive reasoning is a type of reasoning that uses specific examples to make a general conclusion. In this case, we are given specific examples of frogs, and we are asked to make a general conclusion about the color of frogs.}

     \textcolor{ForestGreen}{The specific examples we are given are:}

     \textcolor{ForestGreen}{- Greg is a frog.}

     \textcolor{ForestGreen}{- Lily is a frog.}

     \textcolor{ForestGreen}{- Brian is a frog.}

     \textcolor{ForestGreen}{- Lily is yellow.}

     \textcolor{ForestGreen}{- Julius is a frog.}

     \textcolor{ForestGreen}{- Julius is yellow.}

     \textcolor{ForestGreen}{Based on these examples, we can make the following inductive inference:}

     \textcolor{ForestGreen}{- Most frogs are yellow.}

     \textcolor{ForestGreen}{This is because all of the frogs that we are given as examples are yellow. We can therefore infer that it is likely that most frogs are yellow.}

     \textcolor{ForestGreen}{However, it is important to note that inductive reasoning is not always perfect. It is possible that there are some frogs that are not yellow. However, based on the evidence that we have, it is reasonable to conclude that Greg is likely to be yellow.}
     
     &
     yellow
     \\ \hline 
     28. Given facts: Greg is a frog. Greg is gray. Lily is a swan. Lily is gray. Julius is a lion. Bernhard is a lion. Bernhard is yellow. Julius is yellow. Brian is a swan.

     Based on the given facts above, do a reasonable inference on this question using inductive reasoning: What color is Brian?
     & 
     \textcolor{ForestGreen}{Based on the given facts, we can use inductive reasoning to make a reasonable inference about Brian's color. We know that Lily is a swan, and Lily is gray. We also know that Brian is a swan. Therefore, it is reasonable to infer that Brian is also gray like Lily. However, it's worth noting that there is a possibility that Brian could be a different color, as not all swans are necessarily gray.}
     &
     \textcolor{ForestGreen}{Using inductive reasoning, we can infer that Brian, being a swan, is likely to be gray. This is based on the information given that Lily is a swan and is gray. While we only have one example of a gray swan, it is the closest available information to make a reasonable inference about Brian's color.}
     & 
     \textcolor{ForestGreen}{Inductive reasoning is a type of reasoning that uses specific examples to make a general conclusion. In this case, we are given specific examples of the colors of frogs, swans, and lions. We are then asked to make a general conclusion about the color of Brian, who is a swan.}

     \textcolor{ForestGreen}{The specific examples that we are given are:}

     \textcolor{ForestGreen}{- Greg is a frog and he is gray.}

     \textcolor{ForestGreen}{- Lily is a swan and she is gray.}

     \textcolor{ForestGreen}{- Julius is a lion and he is yellow.}

     \textcolor{ForestGreen}{- Bernhard is a lion and he is yellow.}

     \textcolor{ForestGreen}{Based on these examples, we can make the following general conclusion:}

     \textcolor{ForestGreen}{- Frogs are gray.}

     \textcolor{ForestGreen}{- Swans are gray.}

     \textcolor{ForestGreen}{- Lions are yellow.}

     \textcolor{ForestGreen}{Therefore, we can infer that Brian, who is a swan, is also gray.}

     \textcolor{ForestGreen}{Of course, this is just an inference. It is possible that Brian is not gray, but there is no evidence to suggest that he is any other color. Therefore, the most reasonable inference is that Brian is gray.}
     
     &
     gray
     \\ \hline 
     29. Given facts: Brian is a lion. Bernhard is a frog. Julius is a frog. Greg is a rhino. Greg is white. Lily is a rhino. Bernhard is white. Lily is green. Brian is yellow.

     Based on the given facts above, do a reasonable inference on this question using inductive reasoning: What color is Julius?
     & 
     \textcolor{ForestGreen}{Based on the given facts, we can infer that since Bernhard is a frog and white, and Julius is also a frog, it is reasonable to assume that Julius could also be white. However, it is important to note that this inference is not certain since there is no direct information about the color of Julius. Additionally, the color of an animal is not always directly correlated with its species, so there is still some uncertainty. }
     &
     \textcolor{ForestGreen}{Based on the given facts, we know that Bernhard is a white frog. Since Julius is also a frog, we can use inductive reasoning to make an educated guess that Julius might also be white, like Bernhard. However, it's important to remember that this inference is not a definitive answer, as there may be variations in color among frogs.}
     & 
     \textcolor{ForestGreen}{Inductive reasoning is a type of reasoning that uses specific examples to make a general conclusion. In this case, we are given the following specific examples:}

     \textcolor{ForestGreen}{- Brian is a lion and is yellow.}

     \textcolor{ForestGreen}{- Bernhard is a frog and is white.}

     \textcolor{ForestGreen}{- Julius is a frog.}

     \textcolor{ForestGreen}{- Greg is a rhino and is white.}

     \textcolor{ForestGreen}{- Lily is a rhino and is green.}

     \textcolor{ForestGreen}{Based on these specific examples, we can make the following general conclusion:}

     \textcolor{ForestGreen}{- Frogs are typically white.}

     \textcolor{ForestGreen}{This is because all of the frogs that we have been given information about are white. While it is possible that there are frogs of other colors, we have no evidence to suggest that this is the case.}

     \textcolor{ForestGreen}{Therefore, using inductive reasoning, we can conclude that Julius is probably white.}

     \textcolor{ForestGreen}{It is important to note that inductive reasoning is not always 100\% accurate. It is possible that Julius is actually a different color, but we have no evidence to suggest that this is the case. However, based on the information that we have been given, the most reasonable inference is that Julius is white.}

     \textcolor{ForestGreen}{}
     &
     white
     \\ \hline 
     30. Given facts: Greg is a lion. Lily is a frog. Greg is green. Brian is a rhino. Brian is yellow. Julius is a rhino. Bernhard is a lion. Lily is gray. Julius is white.

     Based on the given facts above, do a reasonable inference on this question using inductive reasoning: What color is Bernhard?
     & 
     \textcolor{red}{Based on the given facts, we can infer that there is no direct information about the color of Bernhard. We know that Bernhard is a lion, but we do not know his color. Therefore, any inference we make about the color of Bernhard would be based on assumptions or guesses rather than on the available evidence.}

    \textcolor{red}{Inductive reasoning involves drawing conclusions based on patterns and trends observed in specific instances or examples. However, in this case, we do not have any pattern or trend that could help us make an accurate inference about the color of Bernhard. Thus, we cannot make a reasonable inference about the color of Bernhard based on the given facts.}
     & 
     \textcolor{ForestGreen}{Based on the given facts, we can observe the following:}

     \textcolor{ForestGreen}{1) Greg (lion) is green.}
     
     \textcolor{ForestGreen}{2) Lily (frog) is gray.}
     
     \textcolor{ForestGreen}{3) Brian (rhino) is yellow.}
     
     \textcolor{ForestGreen}{4) Julius (rhino) is white.}
     
     \textcolor{ForestGreen}{Although there is no direct information about Bernhard's color, we can try to make a reasonable inference using inductive reasoning. We can observe that there is no consistent color pattern among the different animals. However, we can notice that there are two rhinos (Brian and Julius) and two lions (Greg and Bernhard) in the list.}
    
     \textcolor{ForestGreen}{Given that Greg, the other lion, is green, we might infer that Bernhard, as a fellow lion, could also be green. However, it is essential to remember that inductive reasoning does not guarantee a correct conclusion, and we don't have enough information to be certain about Bernhard's color.}
     & 
     \textcolor{red}{Based on the given facts, we can make the following inferences:}

     \textcolor{red}{- All lions are green or yellow.}

     \textcolor{red}{- All rhinos are yellow or white.}

     \textcolor{red}{- Bernhard is a lion.}

     \textcolor{red}{Therefore, it is reasonable to infer that Bernhard is green or yellow.}

     \textcolor{red}{We cannot say for sure what color Bernhard is, because we only have two examples of lions in the given facts, and they are different colors. However, the fact that all of the other lions in the given facts are green or yellow suggests that Bernhard is likely to be one of those colors as well.}

     &
     green
    \\ \hline
    \end{longtable}

\end{center}

\newpage
\begin{center}
    \small
    \begin{longtable}{|p{0.21\linewidth}|p{0.2\linewidth}
                      |p{0.2\linewidth}|p{0.24\linewidth}
                      |p{.05\linewidth}|}
    \caption{Abductive reasoning prompts} 
    \label{tab:abductive_reasoning}\\
    \hline
    \multicolumn{5}{|c|}{\textbf{Abductive Reasoning}} \\ \hline
    \textbf{Prompt} & \textbf{GPT-3 Answer} & \textbf{GPT-4 Answer} 
    & \textbf{BARD Answer} & \textbf{Gold} \\ \hline
    \multicolumn{5}{|l|}{\textbf{Dataset: $\alpha$NLI }} 
    
    \\ \hline  

    1. Given the following two observations: obs1: Jimmy worked for weeks on a sculpture. obs2: Jimmy sighed in relief when he saw nothing had been broken. Determine which of the hypothesis options, better explains the observations: hyp1: Jimmy walked by the sculpture and almost tipped it over. hyp2: Suddenly, Jimmy stumbled and knocked into his sculpture which shattered.
    &
    \textcolor{red}{Based on the given observations, hyp1 does not seem to explain the observations well. If Jimmy had almost tipped over the sculpture, it is unlikely that he would have sighed in relief when he saw nothing had been broken. Rather, he would have been more careful around the sculpture, knowing that it was unstable.}

    \textcolor{red}{On the other hand, hyp2 does not seem to be consistent with the observations either. If Jimmy had stumbled and knocked into his sculpture, it is likely that it would have been broken. Therefore, it would not have been possible for him to see that nothing had been broken and then sigh in relief.}
    
    \textcolor{red}{Based on the observations given, it is possible that Jimmy worked on his sculpture for weeks and was relieved that nothing had happened to it. Therefore, neither hyp1 nor hyp2 seems to provide a good explanation for the given observations.}
    &
    \textcolor{ForestGreen}{Based on the given observations, Hypothesis 1 (hyp1) better explains the observations. }
   
     \textcolor{ForestGreen}{In Hyp1, it is implied that the sculpture was not damaged, which aligns with observation 2. In Hyp2, the sculpture is shattered, which contradicts observation 2 where nothing was broken. Therefore, Hyp1 better explains the observations.}
     & 
     \textcolor{ForestGreen}{Hypothesis 1 (hyp1) better explains the observations.}

     \textcolor{ForestGreen}{Observation 1: Jimmy worked for weeks on a sculpture. This suggests that Jimmy is very careful with his sculpture and would not want anything to happen to it.}

     \textcolor{ForestGreen}{Observation 2: Jimmy sighed in relief when he saw nothing had been broken. This suggests that Jimmy was worried that something might have happened to his sculpture, but was relieved to see that it was still intact.}

     \textcolor{ForestGreen}{Hypothesis 1: Jimmy walked by the sculpture and almost tipped it over. This hypothesis is consistent with both observations. Jimmy was careful with his sculpture, but he almost tipped it over. He was relieved to see that it was still intact.}

     \textcolor{ForestGreen}{Hypothesis 2: Suddenly, Jimmy stumbled and knocked into his sculpture which shattered. This hypothesis is not consistent with observation 2. If Jimmy's sculpture had shattered, he would not have sighed in relief. He would have been upset that his sculpture was broken.}

     \textcolor{ForestGreen}{Therefore, hypothesis 1 is a better explanation of the observations.}
    &
    1

    \\ \hline  
    2. Given the following two observations: obs1: Everyone in my family got together for Thanksgiving. obs2: We had a great time. Determine which of the hypothesis options, better explains the observations: hyp1: People drove in from all over the state. hyp2: Everyone was angry and didn't eat. 
    &  
    \textcolor{ForestGreen}{Based on the given observations, hyp1: People drove in from all over the state, better explains the situation.}

    \textcolor{ForestGreen}{If everyone in the family got together for Thanksgiving, it is likely that people drove in from various locations. The fact that they had a great time further supports this hypothesis, as it implies that the family members enjoyed each other's company and were happy to be together.}
    
    \textcolor{ForestGreen}{On the other hand, if everyone was angry and didn't eat, it is unlikely that they would have had a great time. Additionally, the fact that everyone in the family got together does not necessarily imply that they were angry or did not eat. Therefore, hyp1 provides a more plausible explanation for the given observations.}
    &
     \textcolor{ForestGreen}{Hypothesis option hyp1 better explains the observations.}

     \textcolor{ForestGreen}{The first observation indicates that everyone in the family got together for Thanksgiving, which implies that people came from different locations. Hypothesis option hyp1 suggests that people drove in from all over the state, which is consistent with the first observation.}
     & 
     \textcolor{ForestGreen}{Hypothesis 1 better explains the observations than Hypothesis 2.}

    \textcolor{ForestGreen}{Observation 1: Everyone in my family got together for Thanksgiving.}

    \textcolor{ForestGreen}{Hypothesis 1: People drove in from all over the state.}

    \textcolor{ForestGreen}{This hypothesis explains Observation 1 because it suggests that people were willing to travel long distances to be together for Thanksgiving. This suggests that they were likely to be happy and excited to see each other, which would explain Observation 2.}
    
    \textcolor{ForestGreen}{Hypothesis 2: Everyone was angry and didn't eat.}

    \textcolor{ForestGreen}{This hypothesis does not explain Observation 1 because it suggests that people were not happy to be together. This would be unlikely if they had all driven in from all over the state. Additionally, if people were angry, they would not likely be having a great time, which contradicts Observation 2.}

    \textcolor{ForestGreen}{Therefore, Hypothesis 1 is the better explanation for the observations.}
    &
    1

    \\ \hline  
    3. Given the following two observations: obs1: Tim wanted to become a YouTube star. obs2: Tim made millions from his videos on YouTube. Determine which of the hypothesis options, better explains the observations: hyp1: Tim posted funny cat videos to Youtube. hyp2: Tim worked lazily creating videos. 
    & 
    \textcolor{ForestGreen}{Based on the two observations, the hypothesis option that better explains the observations is hyp1: Tim posted funny cat videos to Youtube.}

    \textcolor{ForestGreen}{Observation 1 suggests that Tim had a desire to become a YouTube star. Posting videos of funny cats to YouTube could be a potential strategy to achieve this goal, as this type of content is popular on the platform.}
    
    \textcolor{ForestGreen}{Observation 2 indicates that Tim was successful in his YouTube endeavors, making millions from his videos. This success would be more likely if he was creating content that was popular and had a large following, such as funny cat videos, rather than working lazily.}
    
    \textcolor{ForestGreen}{Therefore, hyp1, which suggests that Tim posted funny cat videos to YouTube, better explains the observations than hyp2, which suggests that Tim worked lazily.}
    
    &
    \textcolor{ForestGreen}{Between these two hypotheses, hyp1 better explains the observations. This is because the fact that Tim made millions from his videos on YouTube indicates that he likely posted content that was popular and engaging. Funny cat videos are known to be popular on the platform, which could have contributed to Tim becoming a YouTube star and making millions. Hyp2, suggesting that Tim worked lazily creating videos, does not align well with the observation that he made millions from his videos, as it would be difficult to achieve such success without putting in effort and creating quality content.}
    & 
    \textcolor{ForestGreen}{Hypothesis 1: Tim posted funny cat videos to YouTube.}

    \textcolor{ForestGreen}{This hypothesis is more likely to explain the observations because it is consistent with both observations. Observation 1 states that Tim wanted to become a YouTube star, and funny cat videos are a popular genre on YouTube. Observation 2 states that Tim made millions from his videos on YouTube, and funny cat videos can be very popular and generate a lot of views.}

    \textcolor{ForestGreen}{This hypothesis is also more likely to be true because it is supported by evidence. There are many examples of people who have become successful on YouTube by posting funny cat videos. For example, the YouTube channel "The Kitten Lady" has over 10 million subscribers and has generated over 1 billion views.}

    \textcolor{ForestGreen}{Hypothesis 2: Tim worked lazily creating videos.}

    \textcolor{ForestGreen}{This hypothesis is less likely to explain the observations because it is not consistent with observation 1. Observation 1 states that Tim wanted to become a YouTube star, and people who want to become successful on YouTube typically work hard to create high-quality videos.}

    \textcolor{ForestGreen}{This hypothesis is also less likely to be true because it is not supported by evidence. There are no examples of people who have become successful on YouTube by working lazily. In fact, most successful YouTubers work very hard to create high-quality videos.}

    \textcolor{ForestGreen}{Therefore, hypothesis 1 is more likely to explain the observations than hypothesis 2.}
    
    &
    1

    \\ \hline  
    4. Given the following two observations: obs1: Julius was tired of waking up to his noisy neighbors. obs2: Julius then decided to move and was happy with the decision. Determine which of the hypothesis options, better explains the observations: hyp1: Julius thought about looking for new places to live. hyp2: Julius began the process of house hunting. 
    & 
    \textcolor{ForestGreen}{Based on the given observations, hypothesis option hyp2: "Julius began the process of house hunting" better explains the observations.}

    \textcolor{ForestGreen}{Observation 1 states that Julius was tired of his noisy neighbors, which suggests that he wanted to move out of his current living situation. Observation 2 states that Julius decided to move and was happy with the decision, indicating that he took some action towards finding a new place to live.}
    
    \textcolor{ForestGreen}{Hypothesis option hyp1, which states that Julius thought about looking for new places to live, does not necessarily imply that he took any action towards finding a new place. Therefore, it is less likely to explain the second observation that Julius was happy with his decision to move.}
    
    \textcolor{ForestGreen}{On the other hand, hypothesis option hyp2, which states that Julius began the process of house hunting, implies that he took concrete steps towards finding a new place to live. This better explains why he was happy with his decision to move.}
    
    \textcolor{ForestGreen}{Therefore, hypothesis option hyp2 is a more plausible explanation for the given observations.}
    &
    \textcolor{ForestGreen}{Based on the two observations provided, both hypothesis options can be considered plausible explanations. However, hyp2: Julius began the process of house hunting better explains the observations since it directly addresses the action Julius took to resolve his problem with noisy neighbors and eventually led to his happiness with the decision to move.}
    & 
    \textcolor{ForestGreen}{Hypothesis 2: Julius began the process of house hunting better explains the observations than Hypothesis 1: Julius thought about looking for new places to live.}

    \textcolor{ForestGreen}{Observation 1: Julius was tired of waking up to his noisy neighbors. This suggests that Julius was actively looking for a new place to live, rather than just thinking about it.}

    \textcolor{ForestGreen}{Observation 2: Julius then decided to move and was happy with the decision. 
    This suggests that Julius took concrete steps to find a new place to live, such as looking at houses, making offers, and signing a lease.} 

    \textcolor{ForestGreen}{Hypothesis 1 does not explain why Julius was happy with his decision to move. If he had only thought about looking for new places to live, it is possible that he would have found a new place that was just as noisy as his old place. However, Hypothesis 2 explains why Julius was happy with his decision to move, because he took concrete steps to find a new place that was quiet and comfortable.} 

    \textcolor{ForestGreen}{In conclusion, Hypothesis 2 is a better explanation of the observations than Hypothesis 1.} 
    &
    2

    \\ \hline  
    5. Given the following two observations: obs1: Gina was in the middle of the backseat surrounded by siblings. obs2: Gina continued to stare. Determine which of the hypothesis options, better explains the observations: hyp1: Gina could see a strange dog in the car next to them. hyp2: Gina could not see anything out of the car.
    & 
    \textcolor{red}{Based on the given observations, neither hypothesis can be definitively ruled out. However, if we consider the context and likelihood of each hypothesis, we can make an educated guess.}

    \textcolor{red}{Hypothesis 1 suggests that Gina was staring because she saw a strange dog in the car next to them. This is possible, but it is important to consider the context. If Gina was in the middle of the backseat surrounded by siblings, it is unlikely that she would have had a clear line of sight to the car next to them. Additionally, there is no mention of any of her siblings reacting to the presence of a dog, which would have been likely if it was there.}
    
    \textcolor{red}{Hypothesis 2 suggests that Gina could not see anything out of the car, which is more likely given the information provided. If she was in the middle of the backseat surrounded by siblings, her view out of the car would have been obstructed. Furthermore, the fact that she continued to stare suggests that there was something holding her attention inside the car, rather than outside of it.}
    
    \textcolor{red}{Therefore, based on the given observations, hypothesis 2 (Gina could not see anything out of the car) better explains the situation.}
    
    &
    \textcolor{ForestGreen}{Based on the two observations provided, hyp1: "Gina could see a strange dog in the car next to them" better explains the observations. Gina's position in the middle of the backseat surrounded by siblings (obs1) does not necessarily prevent her from seeing outside the car, and the fact that she continued to stare (obs2) suggests that she was looking at something interesting, such as a strange dog in the neighboring car.}
    & 
    \textcolor{ForestGreen}{Hypothesis 1 (hyp1) is more likely to explain the observations than hypothesis 2 (hyp2).} 

    \textcolor{ForestGreen}{Obs1: Gina was in the middle of the backseat surrounded by siblings. This means that she would have had limited visibility out of the car.} 

    \textcolor{ForestGreen}{Obs2: Gina continued to stare. This suggests that she was focused on something specific, such as a strange dog in the car next to them.}  

    \textcolor{ForestGreen}{Hypothesis 2 is less likely to be correct because it does not explain why Gina continued to stare. If she could not see anything out of the car, then there would be no reason for her to be staring.}  

    \textcolor{ForestGreen}{Hypothesis 1, on the other hand, provides a plausible explanation for why Gina continued to stare. If she could see a strange dog in the car next to them, then she would be naturally curious and want to keep looking at it.}  

    \textcolor{ForestGreen}{In addition, hypothesis 1 is more consistent with the information that is given in the prompt. The prompt mentions that Gina could see a strange dog in the car next to them. This suggests that the author of the prompt believes that this is the most likely explanation for Gina's behavior.}  

    \textcolor{ForestGreen}{Therefore, hypothesis 1 is the better explanation for the observations.}  
    &
    1

    \\ \hline  
    6. Given the following two observations: obs1: Titan was the class fish. obs2: His mom bought a new fish!. Determine which of the hypothesis options, better explains the observations: hyp1: Titan was taken home by him, but Titan died suddenly. hyp2: Titan got fat after being fed too much. 
    &  
    \textcolor{red}{Based on the two observations, neither hyp1 nor hyp2 can be definitively ruled out, but neither is strongly supported either.}

    \textcolor{red}{Hyp1 suggests that Titan was initially the class fish, but was taken home by someone and subsequently died. However, obs2 provides no information about Titan's fate, so this hypothesis cannot be supported or refuted based on the given observations alone.}
    
    \textcolor{red}{Hyp2 suggests that Titan got fat after being fed too much. While this is a possibility, neither observation provides direct evidence to support it. It is also worth noting that the phrase "class fish" typically refers to a fish kept in a classroom or school setting, and may not have been fed excessively.}
    
    \textcolor{red}{Therefore, without further information, it is not possible to determine which hypothesis better explains the observations.}
    &
    \textcolor{ForestGreen}{We can infer that hypothesis 1 (hyp1: Titan was taken home by him, but Titan died suddenly) better explains the observations. The second observation suggests that a new fish was bought, which implies a need to replace a previous fish. This aligns with the idea that Titan, the class fish, died suddenly after being taken home.}
    & 
    \textcolor{ForestGreen}{Hypothesis 1 is more likely to explain the observations than Hypothesis 2.}

    \textcolor{ForestGreen}{- Observation 1: Titan was the class fish. This means that Titan was a popular fish and was well-liked by the class. If Titan had gotten fat after being fed too much, it is unlikely that he would have remained the class fish.}

    \textcolor{ForestGreen}{- Observation 2: His mom bought a new fish! This suggests that Titan's mom was unhappy with Titan's old fish, perhaps because he had died. If Titan had gotten fat after being fed too much, it is unlikely that his mom would have bought a new fish.}

    \textcolor{ForestGreen}{In addition, Hypothesis 1 is more consistent with the general knowledge that fish can die suddenly if they are not properly cared for. Hypothesis 2, on the other hand, is not supported by any general knowledge about fish.}

    \textcolor{ForestGreen}{Therefore, Hypothesis 1 is the better explanation for the observations.}

    &
    1

    \\ \hline  
    7. Given the following two observations: obs1: Neil was a music lover. obs2: Neil thought Vienna was the most musical city on earth!. Determine which of the hypothesis options, better explains the observations: hyp1: When Neil saved his money, he went to Austria. hyp2: Neil traveled to Vienna because of its great art.
    & 
    \textcolor{red}{Neither of the hypothesis options can be confidently concluded from the given observations, as there is not enough information to determine Neil's motivation for traveling to Vienna.}

    \textcolor{red}{Observation 1 indicates that Neil was a music lover, but it does not provide any information about his travel plans or preferences. Observation 2 suggests that Neil believed Vienna to be a particularly musical city, but it does not provide any information about why he may have traveled there or whether he actually did so.}
    
    \textcolor{red}{Therefore, it is not possible to determine which of the two hypothesis options better explains the observations. It is important to gather more information before drawing any conclusions about Neil's motivations for traveling to Vienna.}
    &
    \textcolor{ForestGreen}{Based on the given observations, hyp1: When Neil saved his money, he went to Austria, better explains the observations. This is because both observations directly relate to Neil's love for music and his admiration for Vienna as the most musical city on earth. While hyp2 mentions Vienna, it refers to art rather than music, which is not directly supported by the provided observations.}
    & 
    \textcolor{ForestGreen}{Hypothesis 1 (hyp1) better explains the observations than hypothesis 2 (hyp2).}

    \textcolor{ForestGreen}{- Observation 1: Neil was a music lover.}

    \textcolor{ForestGreen}{- Observation 2: Neil thought Vienna was the most musical city on earth!} 

    \textcolor{ForestGreen}{- Hypothesis 1: When Neil saved his money, he went to Austria.} 

    \textcolor{ForestGreen}{- Hypothesis 2: Neil traveled to Vienna because of its great art.} 

    \textcolor{ForestGreen}{Hypothesis 1 is more likely because it is more consistent with the observations. Neil was a music lover, and he thought Vienna was the most musical city on earth. It makes sense that he would want to visit Vienna if he was interested in music. Hypothesis 2 is less likely because it is not as consistent with the observations. Neil did not mention anything about art in the observations, so it is not clear why he would travel to Vienna for its art.} 

    \textcolor{ForestGreen}{In addition, hypothesis 1 is more specific than hypothesis 2. Hypothesis 1 states that Neil went to Austria, while hypothesis 2 simply states that Neil traveled to Vienna. Hypothesis 1 is more specific because it narrows down the location of Neil's travel to a specific country. This makes hypothesis 1 more likely because it is more consistent with the observations.} 

    \textcolor{ForestGreen}{Therefore, hypothesis 1 is the better explanation for the observations.} 
    &
    1

    \\ \hline  
    8. Given the following two observations: obs1: Sara wanted a new puppy really badly. obs2: A raccoon sprang out and Sara never wanted an animal again. Determine which of the hypothesis options, better explains the observations: hyp1: Sara was outside looking at the neighbor's dog. hyp2: Sara was outside looking at the neighbor's underwear. 
    & 
    \textcolor{ForestGreen}{Hypothesis option hyp1 is more likely to explain the observations than hyp2.}

    \textcolor{ForestGreen}{Observation 1 suggests that Sara had a strong desire for a new puppy, which is related to the idea of wanting a pet. Observation 2 indicates that something unexpected happened that caused Sara to change her mind about wanting an animal, which could be due to an unpleasant or scary experience with the raccoon.}
    
    \textcolor{ForestGreen}{In contrast, hypothesis option hyp2 is not logically connected to either of the observations. There is no reason to assume that Sara's desire for a puppy or her aversion to animals has anything to do with her neighbor's underwear. Therefore, hyp1 is a more reasonable explanation than hyp2.}
    &
    \textcolor{ForestGreen}{Based on the given observations, hyp1: Sara was outside looking at the neighbor's dog better explains the observations. Both obs1 and obs2 are related to Sara's interest in animals and her experiences with them. Hyp1 is also related to animals, specifically a dog, while hyp2 is unrelated to the context of the observations.}
    & 
    \textcolor{ForestGreen}{Hypothesis 1: Sara was outside looking at the neighbor's dog.}

    \textcolor{ForestGreen}{Observation 1: Sara wanted a new puppy really badly. This observation suggests that Sara is a dog lover and is interested in getting a pet.}

    \textcolor{ForestGreen}{Observation 2: A raccoon sprang out and Sara never wanted an animal again. This observation suggests that Sara had a negative experience with a raccoon, which may have caused her to change her mind about getting a pet.}

    \textcolor{ForestGreen}{It is possible that Sara was outside looking at the neighbor's dog when the raccoon sprang out. This would have been a negative experience for Sara, as she was likely scared and surprised by the raccoon. The experience may have made her realize that she does not want to deal with the potential dangers of owning a pet, such as getting bitten or scratched.}

    \textcolor{ForestGreen}{Hypothesis 2: Sara was outside looking at the neighbor's underwear.}

    \textcolor{ForestGreen}{Observation 1: Sara wanted a new puppy really badly. This observation is not directly relevant to hypothesis 2.}

    \textcolor{ForestGreen}{Observation 2: A raccoon sprang out and Sara never wanted an animal again. This observation is not directly relevant to hypothesis 2.}

    \textcolor{ForestGreen}{There is no connection between hypothesis 2 and the observations. Sara's desire for a puppy and her fear of raccoons are not related to her interest in the neighbor's underwear. Therefore, hypothesis 2 is not a good explanation for the observations.}

    \textcolor{ForestGreen}{Conclusion: Hypothesis 1 is a better explanation for the observations than hypothesis 2.}
    
    &
    1

    \\ \hline  
    9. Given the following two observations: obs1: The elementary school was have a father-daughter dance. obs2: It was the perfect first dance. Determine which of the hypothesis options, better explains the observations: hyp1: Ellie asked her dad to the concert and her dad agreed. She stood on his feet as he moved to the music. hyp2: So many fathers and daughters attended the dance.
    & 
    \textcolor{ForestGreen}{Hypothesis 1 (hyp1) does not fully explain the observations, as it only accounts for a father-daughter dance but does not explain why it was considered the "perfect first dance." Therefore, hypothesis 2 (hyp2) better explains the observations, as it accounts for the large number of fathers and daughters attending the dance, which could contribute to the sense of perfection. However, it should be noted that there may be other explanations as well, and more information would be needed to make a definitive determination.}
    &
    \textcolor{red}{Based on the given observations, hyp1: "Ellie asked her dad to the concert and her dad agreed. She stood on his feet as he moved to the music." better explains the observations.}

    \textcolor{red}{This hypothesis directly addresses both observations: "The elementary school was having a father-daughter dance" (Ellie and her dad attended the dance) and "It was the perfect first dance" (Ellie stood on her dad's feet as they danced, suggesting a special and memorable first dance experience). }
    & 
    \textcolor{red}{Hypothesis 1 better explains the observations.} 

    \textcolor{red}{- Observation 1: The elementary school was having a father-daughter dance. This suggests that the dance was a formal event, with fathers and daughters dressed up and dancing together.}

    \textcolor{red}{- Observation 2: It was the perfect first dance. This suggests that the dance was a special and memorable occasion for the father and daughter.}

    \textcolor{red}{- Hypothesis 1: Ellie asked her dad to the concert and her dad agreed. She stood on his feet as he moved to the music. This hypothesis provides a specific example of how a father and daughter could have enjoyed the dance together. It also suggests that the dance was a personal and intimate experience for the two of them.}

    \textcolor{red}{- Hypothesis 2: So many fathers and daughters attended the dance. This hypothesis is more general and does not provide any specific details about how the father and daughter interacted at the dance. It also does not address the fact that the dance was described as the "perfect first dance."}

    \textcolor{red}{Therefore, hypothesis 1 is a better explanation of the observations than hypothesis 2.}
    &
    2
 
    \\ \hline  
    10. Given the following two observations: obs1: Joe was running late for school. obs2: Joe had a tasty, healthy breakfast of apples and bananas!. Determine which of the hypothesis options, better explains the observations: hyp1: Joe had plenty of time to cook something. hyp2: Joe didn't have time to cook something. 
    & 
    \textcolor{ForestGreen}{Based on the two observations, hypothesis 2 (Joe didn't have time to cook something) better explains the situation.}

    \textcolor{ForestGreen}{Observation 1, "Joe was running late for school," suggests that Joe was in a hurry, and likely did not have enough time to cook a meal.}
    
    \textcolor{ForestGreen}{Observation 2, "Joe had a tasty, healthy breakfast of apples and bananas," suggests that Joe had a quick and easy breakfast that did not require cooking, which is consistent with hypothesis 2.}
    
    \textcolor{ForestGreen}{Hypothesis 1, on the other hand, implies that Joe had enough time to cook something, which seems to contradict observation 1. Therefore, hypothesis 2 is a more plausible explanation based on the given observations.}
    &
    \textcolor{ForestGreen}{Based on the given observations, hyp2: Joe didn't have time to cook something, better explains the observations.}

   \textcolor{ForestGreen}{Observation 1 (Joe was running late for school) suggests that Joe was short on time in the morning. Observation 2 (Joe had a tasty, healthy breakfast of apples and bananas) indicates that Joe had a simple breakfast that likely didn't require much time to prepare. Apples and bananas can be eaten quickly and don't require cooking, supporting the idea that Joe didn't have time to cook something more elaborate.}
   & 
    \textcolor{ForestGreen}{Hypothesis 2: Joe didn't have time to cook something better explains the observations.}

    \textcolor{ForestGreen}{- Observation 1: Joe was running late for school. This means that he didn't have a lot of time to prepare his breakfast.}

    \textcolor{ForestGreen}{- Observation 2: Joe had a tasty, healthy breakfast of apples and bananas. This suggests that he didn't have time to cook something more elaborate, such as eggs, pancakes, or waffles.} 

    \textcolor{ForestGreen}{Hypothesis 1, on the other hand, is not supported by the observations. If Joe had plenty of time to cook something, he would have had time to make something more substantial than apples and bananas.} 

    \textcolor{ForestGreen}{Therefore, hypothesis 2 is the better explanation for the observations.} 

    \textcolor{ForestGreen}{Here are some additional reasons why hypothesis 2 is more likely:} 

    \textcolor{ForestGreen}{- Apples and bananas are relatively easy to prepare. They can be eaten raw, or they can be quickly sliced or diced.} 

    \textcolor{ForestGreen}{- Apples and bananas are also relatively healthy. They are a good source of fiber, vitamins, and minerals.} 

    \textcolor{ForestGreen}{- If Joe was running late for school, he would have been more likely to choose a quick and healthy breakfast option.} 

    \textcolor{ForestGreen}{Overall, the evidence suggests that hypothesis 2 is the better explanation for the observations.} 
    &
    2

    \\ \hline  
    11. Given the following two observations: obs1: Tom was playing baseball during lunch time. obs2: But he used it as an excuse to play another position. Determine which of the hypothesis options, better explains the observations: hyp1: He told the coach that no one else could play his position. hyp2: Tom didn't enjoy playing outfield and claimed he couldn't see.
    & 
    \textcolor{ForestGreen}{Based on the given observations, hyp2: "Tom didn't enjoy playing outfield and claimed he couldn't see" seems to better explain the observations.}

    \textcolor{ForestGreen}{Observation 1 states that Tom was playing baseball during lunchtime, but observation 2 provides additional information that Tom used playing another position as an excuse. This suggests that Tom had a preference for playing a different position other than the one he was initially assigned.}
    
    \textcolor{ForestGreen}{Hypothesis 1 suggests that Tom told the coach that no one else could play his position, which does not directly explain why Tom used playing another position as an excuse. It is also possible that Tom simply did not want to play his original position and used the excuse as a way to avoid it.}
    
    \textcolor{ForestGreen}{Hypothesis 2, on the other hand, provides a plausible explanation for why Tom used playing another position as an excuse. If Tom didn't enjoy playing outfield and claimed he couldn't see, then it is reasonable to assume that he would want to play a different position. Additionally, claiming that he couldn't see could be seen as a valid reason for not playing a certain position.}
    
    \textcolor{ForestGreen}{Therefore, hyp2 better explains the observations given in this scenario.}
    &
    \textcolor{ForestGreen}{Based on the given observations, hyp2: Tom didn't enjoy playing outfield and claimed he couldn't see, better explains the observations. This is because obs2 suggests that Tom used an excuse to play another position, which aligns with the idea that he didn't enjoy playing outfield and made a claim about his vision to justify the change. Hyp1 doesn't seem to match the observations as well, as it doesn't account for Tom using an excuse to change positions.}
    & 
    \textcolor{ForestGreen}{Hypothesis 2 better explains the observations.} 

    \textcolor{ForestGreen}{- Observation 1: Tom was playing baseball during lunch time. This suggests that Tom enjoys playing baseball. It is also possible that Tom was playing baseball because he was required to do so as part of a school team or club.}

    \textcolor{ForestGreen}{- Observation 2: Tom used it as an excuse to play another position. This suggests that Tom does not enjoy playing outfield. It is also possible that Tom is not very good at playing outfield.}

    \textcolor{ForestGreen}{Hypothesis 1: He told the coach that no one else could play his position.}

    \textcolor{ForestGreen}{- This hypothesis is not supported by observation 2.}

    \textcolor{ForestGreen}{- If Tom told the coach that no one else could play his position, then he would not have needed to use an excuse to play another position.}

    \textcolor{ForestGreen}{Hypothesis 2: Tom didn't enjoy playing outfield and claimed he couldn't see.}

    \textcolor{ForestGreen}{- This hypothesis is supported by both observations.}

    \textcolor{ForestGreen}{- Observation 1 suggests that Tom enjoys playing baseball, but observation 2 suggests that he does not enjoy playing outfield.}

    \textcolor{ForestGreen}{- This hypothesis also explains why Tom used an excuse to play another position}

    \textcolor{ForestGreen}{Therefore, hypothesis 2 better explains the observations.}
    &
    2

    \\ \hline  
    12. Given the following two observations: obs1: Gina had a crush on Jamie and Josh. obs2: Gina wasn't sure if it was good advice, or sabotage. Determine which of the hypothesis options, better explains the observations: hyp1: Gina's friend said she should talk to both. hyp2: Gina's friend told her to tell them both the combination at the same time, and let them fight over her.
    & 
    \textcolor{ForestGreen}{Neither hypothesis option can fully explain the two observations, as there are some missing pieces of information. However, based on the given observations, it seems that hyp1 is more likely to explain the situation than hyp2.}

    \textcolor{ForestGreen}{Hyp1 suggests that Gina's friend advised her to talk to both Jamie and Josh. This explains why Gina had a crush on both of them, as she may have developed feelings for them after talking to them. However, hyp1 does not explain why Gina was unsure if the advice was good or not. This missing information could be due to various reasons, such as the friend having a bad reputation or Gina being unsure if she should pursue multiple people at the same time.}
    
    \textcolor{ForestGreen}{On the other hand, hyp2 suggests that Gina's friend told her to tell both Jamie and Josh the same thing and let them fight over her. This could explain why Gina had a crush on both of them, as she may have been trying to play them off against each other. However, this hypothesis does not explain why Gina was unsure if the advice was good or not. It also raises ethical concerns about manipulating people's emotions and potentially causing harm.}
    
    \textcolor{ForestGreen}{Overall, while hyp1 is more plausible, there is still not enough information to definitively determine which hypothesis option is correct.}
    &
    \textcolor{ForestGreen}{Based on the given observations, hyp1: "Gina's friend said she should talk to both" better explains the observations.}

    \textcolor{ForestGreen}{Obs1 mentions that Gina had a crush on both Jamie and Josh, which aligns with the advice in hyp1 to talk to both of them. Obs2 states that Gina wasn't sure if the advice she received was good or sabotage. Hyp1 seems more like a neutral, reasonable suggestion compared to hyp2, which involves telling them both the combination at the same time and letting them fight over her, which can be viewed as potentially sabotaging or manipulative.}
    & 
    \textcolor{red}{Hypothesis 2 better explains the observations.}

    \textcolor{red}{- Observation 1: Gina had a crush on Jamie and Josh. This means that she was interested in both of them and didn't know who to choose.}

    \textcolor{red}{- Observation 2: Gina wasn't sure if it was good advice, or sabotage. This suggests that she was unsure of her friend's intentions.}

    \textcolor{red}{Hypothesis 1 suggests that Gina's friend simply told her to talk to both Jamie and Josh. This doesn't explain why Gina was unsure if it was good advice, or sabotage.}

    \textcolor{red}{Hypothesis 2 suggests that Gina's friend told her to tell them both the combination at the same time, and let them fight over her. This is a more risky and manipulative strategy, which could explain why Gina was unsure if it was good advice.}

    \textcolor{red}{In addition, hypothesis 2 is more consistent with the fact that Gina had a crush on both Jamie and Josh. If she was only interested in one of them, then there would be no need to tell them both the combination at the same time.}

    \textcolor{red}{Therefore, hypothesis 2 is the better explanation for the observations.}
    & 
    1

    \\ \hline  
    13. Given the following two observations: obs1: Ann loved cake. obs2: Ann felt insulted by the lady's offer. Determine which of the hypothesis options, better explains the observations: hyp1: Ann gave the lady cake. hyp2: The lady gave Ann the free cake for a favor. 
    & 
    \textcolor{red}{Based on the given observations, neither of the hypothesis options appears to be a good explanation for Ann's behavior.}

    \textcolor{red}{Hyp1 suggests that Ann gave the lady cake, but there is no evidence to support this claim. The observation only tells us that Ann loved cake, not that she gave cake to anyone.}
    
    \textcolor{red}{Hyp2 suggests that the lady gave Ann free cake for a favor, but again, there is no evidence to support this claim. The observation only tells us that Ann felt insulted by the lady's offer, not that the offer was made in exchange for a favor.}
    
    \textcolor{red}{Therefore, neither hypothesis option provides a satisfactory explanation for the given observations.}
    &
    \textcolor{ForestGreen}{Based on the provided observations, Hypothesis 2 better explains the observations: "The lady gave Ann the free cake for a favor."}

    \textcolor{ForestGreen}{Observation 1 states that "Ann loved cake," which suggests that she would be happy to receive cake. However, Observation 2 states that "Ann felt insulted by the lady's offer." This implies that there was some aspect of the offer that Ann found insulting or demeaning.}
    & 
    \textcolor{ForestGreen}{Hypothesis 1: Ann gave the lady cake.}

    \textcolor{ForestGreen}{- Observation 1: Ann loved cake. This observation supports hypothesis 1 because it suggests that Ann would be more likely to give cake to someone else if she loved cake herself.}

    \textcolor{ForestGreen}{- Observation 2: Ann felt insulted by the lady's offer. This observation does not support hypothesis 1 because it suggests that Ann did not want to give cake to the lady.}

    \textcolor{ForestGreen}{Hypothesis 2: The lady gave Ann the free cake for a favor.}

    \textcolor{ForestGreen}{- Observation 1: Ann loved cake. This observation supports hypothesis 2 because it suggests that the lady would be more likely to offer Ann free cake if she knew that Ann loved cake.}

    \textcolor{ForestGreen}{- Observation 2: Ann felt insulted by the lady's offer. This observation supports hypothesis 2 because it suggests that Ann felt that she was being asked to do something in return for the cake, which she found insulting.}

    \textcolor{ForestGreen}{Based on the two observations, hypothesis 2 better explains the observations than hypothesis 1.}

    \textcolor{ForestGreen}{Here are some additional details that could help to confirm hypothesis 2:}

    \textcolor{ForestGreen}{- What was the favor that the lady asked Ann to do?}

    \textcolor{ForestGreen}{- Was the cake homemade or store-bought?}

    \textcolor{ForestGreen}{- Did the lady know that Ann loved cake?}

    \textcolor{ForestGreen}{- How did Ann react when the lady offered her the cake?}

    \textcolor{ForestGreen}{With more information, we could be more confident in our assessment of which hypothesis is correct.}
    
    &
    2

    \\ \hline  
    14. Given the following two observations: obs1: Jim was at Costco. obs2: He ate so much that he skipped dinner that night. Determine which of the hypothesis options, better explains the observations: hyp1: They had no samples. hyp2: Jim loved free samples.
    & 
    \textcolor{ForestGreen}{Based on the two observations, neither hypothesis option can be ruled out definitively, but hyp2 seems to be a better explanation for the observations.}

    \textcolor{ForestGreen}{If Jim went to Costco and ate a lot of free samples, it would make sense that he wouldn't be hungry later and would skip dinner. On the other hand, if there were no samples available, it would not necessarily explain why Jim ate so much or why he skipped dinner. It's possible that he simply ate a lot because he was hungry or enjoyed the food at Costco, and that he skipped dinner for unrelated reasons.}
    
    \textcolor{ForestGreen}{Therefore, hyp2 is a better explanation for the observations than hyp1. However, it's important to note that without additional information or evidence, it's impossible to say for certain what happened.}
    &
    \textcolor{ForestGreen}{Based on the given observations, hyp2: "Jim loved free samples" better explains the observations. The fact that Jim ate so much (likely from the free samples) that he skipped dinner that night suggests that he enjoyed and took advantage of the free samples offered at Costco. Hyp1, stating that they had no samples, does not align with the information provided in the observations.}
    & 
    \textcolor{ForestGreen}{Hypothesis 2 is a better explanation for the observations than hypothesis 1.}

    \textcolor{ForestGreen}{* Hypothesis 1: They had no samples. }

    \textcolor{ForestGreen}{- This hypothesis does not explain why Jim ate so much. If there were no samples, then Jim would not have had the opportunity to eat so much.}

    \textcolor{ForestGreen}{- This hypothesis also does not explain why Jim skipped dinner that night. If there were no samples, then Jim would have been hungry and would have eaten dinner.}

    \textcolor{ForestGreen}{* Hypothesis 2: Jim loved free samples. }

    \textcolor{ForestGreen}{- This hypothesis explains why Jim ate so much. If Jim loved free samples, then he would have taken advantage of the opportunity to eat as many samples as possible.}

    \textcolor{ForestGreen}{- This hypothesis also explains why Jim skipped dinner that night. If Jim ate so much at Costco, then he would not have been hungry for dinner.}

    \textcolor{ForestGreen}{Therefore, hypothesis 2 is a better explanation for the observations than hypothesis 1.}
    &
    2

    \\ \hline  
    15. Given the following two observations: obs1: Will wanted to quit smoking. obs2: He finally quit smoking. Determine which of the hypothesis options, better explains the observations: hyp1: Will ceased buying cigarettes at the store. hyp2: Will kept buying cigarettes at the store. 
    & 
    \textcolor{ForestGreen}{Hypothesis 1 (Will ceased buying cigarettes at the store) better explains the observations.}

    \textcolor{ForestGreen}{Observation 1 states that Will wanted to quit smoking, which implies that he intended to stop smoking altogether. Hypothesis 2 (Will kept buying cigarettes at the store) is inconsistent with this intention because if he kept buying cigarettes, he wouldn't be able to quit smoking.}
    
    \textcolor{ForestGreen}{Observation 2 states that Will finally quit smoking, which is consistent with Hypothesis 1. If Will ceased buying cigarettes at the store, it would be easier for him to quit smoking because he wouldn't have access to cigarettes.}
    
    \textcolor{ForestGreen}{Therefore, Hypothesis 1 (Will ceased buying cigarettes at the store) better explains the observations.}
    &
     \textcolor{ForestGreen}{Based on the observations provided, hyp1: Will ceased buying cigarettes at the store better explains the observations. Since Will wanted to quit smoking (obs1) and he finally quit smoking (obs2), it is more likely that he stopped buying cigarettes to achieve this goal.}
     & 
     \textcolor{ForestGreen}{Hypothesis 1 is more likely to explain the observations than Hypothesis 2.}

     \textcolor{ForestGreen}{- Observation 1: Will wanted to quit smoking.}

     \textcolor{ForestGreen}{- Observation 2: He finally quit smoking.}

     \textcolor{ForestGreen}{If Will kept buying cigarettes at the store, then he would not have been able to quit smoking. This is because he would have been constantly exposed to cigarettes and the temptation to smoke.}

     \textcolor{ForestGreen}{On the other hand, if Will ceased buying cigarettes at the store, then he would have been less likely to smoke. This is because he would not have had easy access to cigarettes.}

     \textcolor{ForestGreen}{Therefore, Hypothesis 1 is more likely to explain the observations than Hypothesis 2.}

     \textcolor{ForestGreen}{Here are some additional points to consider:}

     \textcolor{ForestGreen}{- Will's motivation to quit smoking. If Will was strongly motivated to quit smoking, then he would be more likely to cease buying cigarettes at the store.}

     \textcolor{ForestGreen}{- Will's access to cigarettes. If Will had easy access to cigarettes, then he would be more likely to keep buying them.}

     \textcolor{ForestGreen}{- Will's willpower. If Will had strong willpower, then he would be more likely to be able to quit smoking even if he kept buying cigarettes at the store.}

     \textcolor{ForestGreen}{Ultimately, the best way to determine which hypothesis is more likely is to gather more information about Will's situation. However, based on the information that is currently available, Hypothesis 1 is more likely to explain the observations.}
    &
    1

    \\ \hline  
    16. Given the following two observations: obs1: Kim's mother tricked her into eating deer meat. obs2: Kim continued to yell as her mother walked out of the room. Determine which of the hypothesis options, better explains the observations: hyp1: Kim gave her mother a huge kiss. hyp2: Kim hated eating animals. 
    & 
    \textcolor{ForestGreen}{Based on the given observations, hyp2: "Kim hated eating animals" better explains the situation.}

    \textcolor{ForestGreen}{Observation 1 mentions that Kim's mother tricked her into eating deer meat, which implies that Kim did not want to eat the meat voluntarily. Observation 2 mentions that Kim continued to yell as her mother walked out of the room, which could be a sign of her anger or frustration towards the situation.}
    
    \textcolor{ForestGreen}{Hyp1, "Kim gave her mother a huge kiss," does not provide a plausible explanation for the given observations, as there is no indication of any positive emotions or actions towards her mother mentioned in the observations.}
    
    \textcolor{ForestGreen}{Therefore, hyp2, "Kim hated eating animals," better explains the given observations.}
    &
     \textcolor{ForestGreen}{Based on the given observations, hyp2: Kim hated eating animals, better explains the observations. In obs1, we see that Kim's mother had to trick her into eating deer meat, which suggests that Kim may have an aversion to eating animals. In obs2, Kim's continued yelling as her mother walked out of the room also supports the idea that she is upset, which aligns with hyp2. Hyp1, which suggests that Kim gave her mother a huge kiss, does not fit well with the observations provided.}
     & 
     \textcolor{ForestGreen}{Hypothesis 1: Kim gave her mother a huge kiss}

     \textcolor{ForestGreen}{- Observation 1: Kim's mother tricked her into eating deer meat. This does not support hypothesis 1. If Kim had given her mother a huge kiss, it would suggest that she was happy to eat the deer meat, even though her mother had tricked her. However, observation 2 suggests that Kim was not happy about eating the deer meat.}

     \textcolor{ForestGreen}{- Observation 2: Kim continued to yell as her mother walked out of the room. 
    This does not support hypothesis 1. If Kim had given her mother a huge kiss, it would suggest that she was happy and content, and would not have continued to yell as her mother walked out of the room.}

     \textcolor{ForestGreen}{Hypothesis 2: Kim hated eating animals}

     \textcolor{ForestGreen}{- Observation 1: Kim's mother tricked her into eating deer meat. This supports hypothesis 2. If Kim hated eating animals, she would not have been happy to eat the deer meat, even if her mother had tricked her.}

     \textcolor{ForestGreen}{- Observation 2: Kim continued to yell as her mother walked out of the room. This supports hypothesis 2. If Kim hated eating animals, she would have been angry and upset after eating the deer meat, and would have continued to yell as her mother walked out of the room.}

     \textcolor{ForestGreen}{Based on the two observations, hypothesis 2 is a better explanation of Kim's behavior than hypothesis 1.}
    &
    2

    \\ \hline  
    17. Given the following two observations: obs1: Alexis had cut down a Christmas tree in the woods. obs2: When she put it up, it was the right size for her ceiling. Determine which of the hypothesis options, better explains the observations: hyp1: Alexis made sure the tree wasn't under 20 feet tall. hyp2: Alexis was worried it would be too big. 
    & 
    \textcolor{ForestGreen}{Hypothesis 2: Alexis was worried it would be too big better explains the observations because if Alexis had made sure the tree wasn't under 20 feet tall (Hypothesis 1), then it would not necessarily explain why the tree was the right size for her ceiling. However, if Alexis was worried it would be too big, she would have chosen a tree that was the appropriate size for her ceiling, explaining why the tree was the right size when she put it up.}
    &
     \textcolor{ForestGreen}{Based on the given observations, hyp2: "Alexis was worried it would be too big" better explains the observations. Since the tree was the right size for her ceiling, it indicates that Alexis likely considered the size of the tree to ensure it wouldn't be too large for her space. Hyp1: "Alexis made sure the tree wasn't under 20 feet tall" does not explain why the tree was the right size for her ceiling, as it only implies she wanted a tree that was at least 20 feet tall, which may or may not be suitable for her space.}
     & 
     \textcolor{red}{Hypothesis 1: Alexis made sure the tree wasn't under 20 feet tall.}

     \textcolor{red}{- Observation 1: Alexis had cut down a Christmas tree in the woods. This suggests that Alexis had some control over the size of the tree she cut down.}

     \textcolor{red}{- Observation 2: When she put it up, it was the right size for her ceiling. This suggests that Alexis made sure the tree was at least 20 feet tall, so that it would fit in her ceiling.}

     \textcolor{red}{Hypothesis 2: Alexis was worried it would be too big.}

     \textcolor{red}{- Observation 1: Alexis had cut down a Christmas tree in the woods. This suggests that Alexis did not have complete control over the size of the tree she cut down.}

     \textcolor{red}{Observation 2: When she put it up, it was the right size for her ceiling. This suggests that Alexis was lucky that the tree she cut down was not too big.}

     \textcolor{red}{Conclusion: Hypothesis 1 better explains the observations. This is because hypothesis 1 takes into account both observations, while hypothesis 2 only explains observation 2. Hypothesis 1 also makes more sense, as it is more likely that Alexis would have taken steps to ensure that the tree was the right size, rather than just hoping that it would be.}

     \textcolor{red}{In addition, hypothesis 2 is not supported by the fact that Alexis cut down the tree in the woods. If she was worried that the tree would be too big, she would have been more likely to cut down a smaller tree in a garden or park.}

     \textcolor{red}{Therefore, hypothesis 1 is the more likely explanation for the observations.}
    &
    2

    \\ \hline  
    18. Given the following two observations: obs1: I saved up money for a long time. obs2: I took the boat out on the lake and felt happy. Determine which of the hypothesis options, better explains the observations: hyp1: i had enough money to spend on food and extra stuff. hyp2: I bought my dream cat. 
    & 
    \textcolor{ForestGreen}{Based on the two observations, it is more likely that Hypothesis 1 (Hyp1) "I had enough money to spend on food and extra stuff" better explains the observations.}

    \textcolor{ForestGreen}{The first observation "I saved up money for a long time" suggests that the person had a financial goal in mind and worked towards it. This observation does not directly suggest that the person spent the money on purchasing a dream cat.}
    
    \textcolor{ForestGreen}{The second observation "I took the boat out on the lake and felt happy" suggests that the person experienced a positive emotion after taking the boat out on the lake. While it is possible that the person could have purchased a dream cat and felt happy, this observation does not necessarily support the idea of owning a cat as the only explanation for feeling happy.}
    
    \textcolor{ForestGreen}{Therefore, based on the available information, Hypothesis 1 "I had enough money to spend on food and extra stuff" appears to be a more plausible explanation for the two observations. }
    &
    \textcolor{red}{Based on the given observations, neither hypothesis directly explains the observations. However, Hypothesis 2 (I bought my dream cat) can be considered as a more plausible explanation. Here's why:}

    \textcolor{red}{Obs1: I saved up money for a long time.}
    
    \textcolor{red}{Obs2: I took the boat out on the lake and felt happy.}

    \textcolor{red}{Hyp1: I had enough money to spend on food and extra stuff.}
    
    \textcolor{red}{Hyp2: I bought my dream cat.}

    \textcolor{red}{Although Hypothesis 1 is somewhat related to saving money, it doesn't connect to the second observation about taking the boat out on the lake. On the other hand, Hypothesis 2 involves buying something, which might have required saving up money for a long time, and it is possible that "dream cat" could be a term for a boat or catamaran. Thus, taking the boat out on the lake might be related to buying the dream cat(boat). }
    & 
    \textcolor{red}{Hypothesis 1: I had enough money to spend on food and extra stuff.} 

    \textcolor{red}{- Observation 1: I saved up money for a long time.} 

    \textcolor{red}{- Observation 2: I took the boat out on the lake and felt happy.}

    \textcolor{red}{- Hypothesis: I had enough money to spend on food and extra stuff.} 

    \textcolor{red}{This hypothesis does not explain the observations very well. It is possible that the person had enough money to spend on food and extra stuff, but it is not clear why this would make them feel happy. The person could have been happy for other reasons, such as the weather being nice or spending time with friends and family.} 

    \textcolor{red}{Hypothesis 2: I bought my dream cat.} 

    \textcolor{red}{- Observation 1: I saved up money for a long time.} 

    \textcolor{red}{- Observation 2: I took the boat out on the lake and felt happy.} 

    \textcolor{red}{- Hypothesis: I bought my dream cat.}

    \textcolor{red}{This hypothesis explains the observations better than the first hypothesis. It is clear that the person would be happy if they bought their dream cat. The person had been saving up money for a long time, and they were finally able to buy their dream cat. This would make them feel happy, and it would also explain why they took the boat out on the lake. They were probably taking the boat out to celebrate their new cat.} 

    \textcolor{red}{Therefore, hypothesis 2 is a better explanation of the observations than hypothesis 1.}
    &
    1

    \\ \hline  
    19. Given the following two observations: obs1: Amy wanted to learn how to draw. obs2: Her face lit up when she saw her drawings were on the fridge. Determine which of the hypothesis options, better explains the observations: hyp1: Amy didn't practice drawing all day. hyp2: Amy practiced and gave the drawing to her mom. 
    & 
    \textcolor{ForestGreen}{Based on the given observations, hypothesis 2: "Amy practiced and gave the drawing to her mom," better explains the observations than hypothesis 1.}

    \textcolor{ForestGreen}{Observation 1 indicates that Amy wanted to learn how to draw, which suggests that she was motivated to practice drawing. Hypothesis 1 suggests that Amy did not practice drawing at all, which does not align with her motivation to learn. On the other hand, hypothesis 2 suggests that Amy practiced drawing and gave her drawings to her mom, which is consistent with her motivation to learn and improve her skills.}
    
    \textcolor{ForestGreen}{Observation 2 indicates that Amy's face lit up when she saw her drawings on the fridge, which suggests that she was proud of her work and felt a sense of accomplishment. Hypothesis 2 explains this observation by suggesting that Amy's drawings were displayed on the fridge, which made her feel proud and recognized for her efforts. Hypothesis 1 does not explain this observation because it suggests that Amy did not practice drawing at all, so there would be no reason for her to feel proud of her work being displayed on the fridge.}
    &
     \textcolor{ForestGreen}{Based on the given observations, hyp2: Amy practiced and gave the drawing to her mom better explains the observations. This hypothesis aligns with Amy's desire to learn how to draw (obs1) and her happiness when she saw her drawings on the fridge (obs2), suggesting that she practiced and shared her work with her mom.}
     & 
     \textcolor{ForestGreen}{Hypothesis 2, Amy practiced and gave the drawing to her mom, better explains the observations.} 

     \textcolor{ForestGreen}{- Obs1: Amy wanted to learn how to draw. This suggests that she was interested in drawing and would have been motivated to practice.}

     \textcolor{ForestGreen}{- Obs2: Her face lit up when she saw her drawings were on the fridge. This suggests that she was proud of her drawings and happy to see them displayed.} 

     \textcolor{ForestGreen}{Hypothesis 1, Amy didn't practice drawing all day, does not explain why Amy's face lit up when she saw her drawings. If she didn't practice, she wouldn't be proud of her drawings and wouldn't be happy to see them displayed.}

     \textcolor{ForestGreen}{Hypothesis 2, on the other hand, explains why Amy's face lit up. She practiced drawing, she was proud of her work, and she was happy to see her mom display her drawings.} 

     \textcolor{ForestGreen}{In addition, hypothesis 2 is more consistent with Amy's goal of learning how to draw. If she didn't practice, she wouldn't be able to improve her drawing skills. However, if she practiced, she would be able to improve her skills and create drawings that she would be proud of.}

     \textcolor{ForestGreen}{Therefore, hypothesis 2, Amy practiced and gave the drawing to her mom, better explains the observations.} 
    &
    2

    \\ \hline  
    20. Given the following two observations: obs1: My girlfriend Heather and I were relaxing and watching Netflix. obs2: We had to give them the old ruined pizza before getting a new one. Determine which of the hypothesis options, better explains the observations: hyp1: Our friends came over so we ordered pizza, but the driver dropped it. hyp2: Our friends came under so we ordered pizza, but the driver dropped it. 
    & 
    \textcolor{red}{ Based on the given observations, neither hypothesis can be confirmed or ruled out definitively, as neither one provides a clear explanation for both observations. Here's why:}

    \textcolor{red}{Observation 1: "My girlfriend Heather and I were relaxing and watching Netflix."}
    \textcolor{red}{Neither hypothesis directly relates to this observation, so it doesn't provide any evidence for or against either hypothesis.}
    
    \textcolor{red}{Observation 2: "We had to give them the old ruined pizza before getting a new one."}
    \textcolor{red}{Both hypotheses suggest that there was a problem with the pizza order. However, the difference between "our friends came over" (hypothesis 1) and "our friends came under" (hypothesis 2) doesn't seem to have a significant impact on why the pizza was ruined or why they had to give it back. Therefore, this observation alone is not sufficient to determine which hypothesis is more likely.}
    
    \textcolor{red}{In order to determine which hypothesis is more likely, additional evidence or observations would be needed. For example, if there were witnesses who saw the delivery driver drop the pizza, that would support hypothesis 1. On the other hand, if there were no witnesses and the pizza just arrived in poor condition, that would support hypothesis 2.}
    &
     \textcolor{ForestGreen}{Based on the given observations, hyp1: Our friends came over so we ordered pizza, but the driver dropped it, better explains the observations.}

     \textcolor{ForestGreen}{In observation 1 (obs1), it is mentioned that you and your girlfriend were relaxing and watching Netflix. In observation 2 (obs2), it states that you had to give them the old ruined pizza before getting a new one. Both of these observations are consistent with the idea that your friends came over, and you ordered pizza but the driver dropped it (hyp1).}
     & 
     \textcolor{ForestGreen}{Hypothesis 1 is a better explanation of the observations than Hypothesis 2.}

     \textcolor{ForestGreen}{Hypothesis 1: Our friends came over so we ordered pizza, but the driver dropped it.} 

     \textcolor{ForestGreen}{- This hypothesis explains both observations. Observation 1 states that you and your girlfriend were relaxing and watching Netflix, which suggests that you were not expecting any guests. However, observation 2 states that you had to give the old ruined pizza back to the driver, which suggests that you had ordered pizza for someone. This is consistent with hypothesis 1, which states that your friends came over and ordered pizza with you.}

     \textcolor{ForestGreen}{- Hypothesis 1 is also more specific than hypothesis 2. Hypothesis 2 simply states that you ordered pizza, but the driver dropped it. This could have happened for any number of reasons, such as the driver being careless, the pizza being poorly packaged, or a traffic accident. However, hypothesis 1 provides a more specific explanation for why the pizza was ruined. It states that the driver dropped the pizza because your friends came over unexpectedly. This is a more likely explanation, as it is more specific and it is consistent with the other observation.} 

     \textcolor{ForestGreen}{Hypothesis 2: Our friends came under so we ordered pizza, but the driver dropped it.}

     \textcolor{ForestGreen}{- This hypothesis does not explain observation 1. Observation 1 states that you and your girlfriend were relaxing and watching Netflix, which suggests that you were not expecting any guests. However, hypothesis 2 states that your friends came under, which suggests that they were already there. This is inconsistent with observation 1.}


     \textcolor{ForestGreen}{Therefore, Hypothesis 1 is a better explanation of the observations than Hypothesis 2.}
    &
    1

    \\ \hline  
    21. Given the following two observations: obs1: Joe went to the store to buy some dinner plates. obs2: Joe walked away and told no one. Determine which of the hypothesis options, better explains the observations: hyp1: Joe knocked down and broke some of the plates. hyp2: He almost broke the plates he was going to buy. 
    &    
    \textcolor{red}{Neither hypothesis can be clearly supported by the given observations.}

    \textcolor{red}{The first observation states that Joe went to the store to buy some dinner plates, but it doesn't provide any information about whether or not he actually bought them or if any of the plates were broken. The second observation states that Joe walked away and told no one, which doesn't provide any direct information about what happened to the plates.}
    
    \textcolor{red}{Therefore, neither hypothesis can be clearly supported by the given observations, and additional information would be needed to make a more informed conclusion.}
    & 
     \textcolor{ForestGreen}{Based on the given observations, hyp1: Joe knocked down and broke some of the plates better explains the observations.}

     \textcolor{ForestGreen}{In obs1, we know that Joe went to the store to buy dinner plates. In obs2, Joe walked away and told no one, which implies that something happened that he didn't want to share. Hyp1 suggests that Joe broke some plates, which would be a reason for him to walk away and not tell anyone. Hyp2, on the other hand, suggests that he almost broke the plates, which would not necessarily be a strong enough reason for him to walk away and keep it a secret.}
     & 
     \textcolor{ForestGreen}{Hypothesis 1: Joe knocked down and broke some of the plates.} 

     \textcolor{ForestGreen}{- Obs1: Joe went to the store to buy some dinner plates. This supports the hypothesis because it suggests that Joe had some plates in his possession.}

     \textcolor{ForestGreen}{- Obs2: Joe walked away and told no one. This supports the hypothesis because it suggests that Joe was embarrassed or ashamed about what happened. } 

     \textcolor{ForestGreen}{Hypothesis 2: He almost broke the plates he was going to buy.}

     \textcolor{ForestGreen}{- Obs1: Joe went to the store to buy some dinner plates. This supports the hypothesis because it suggests that Joe was intending to buy plates.} 

     \textcolor{ForestGreen}{- Obs2: Joe walked away and told no one. This does not support the hypothesis because it does not suggest that Joe was embarrassed or ashamed about anything.}

     \textcolor{ForestGreen}{In conclusion, hypothesis 1 better explains the observations because it is more consistent with both observations. Hypothesis 2 is not supported by observation 2.} 
    &
    1

    \\ \hline  
    22. Given the following two observations: obs1: Last night I had a dream about biking. obs2: I woke up, bitterly disappointed. Determine which of the hypothesis options, better explains the observations: hyp1: my dream was very real and I was on a fun bike tour. hyp2: I actually have a bike. 
    & 
    \textcolor{red}{Neither hypothesis can be definitively ruled out or confirmed based solely on the given observations.}

    \textcolor{red}{Hyp1 is a possible explanation for obs1, but it does not necessarily explain obs2. It is possible that the disappointment upon waking up could have been caused by something unrelated to the dream, or it could have been due to the dream not being as enjoyable as expected.}
    
    \textcolor{red}{Hyp2 is not directly related to either observation, but it is a possibility that cannot be ruled out based on the information given.}
    
    \textcolor{red}{Therefore, further information or observations would be necessary to determine which hypothesis, if any, better explains the observations.}
    &
     \textcolor{ForestGreen}{These observations suggest that the dream was enjoyable and likely felt very real, leading to disappointment upon waking up and realizing it was just a dream. This aligns well with hyp1. Hyp2, which states "I actually have a bike," doesn't provide a clear explanation for the disappointment upon waking up.}
     & 
     \textcolor{ForestGreen}{Hypothesis 1: My dream was very real and I was on a fun bike tour.}

     \textcolor{ForestGreen}{- Observation 1: Last night I had a dream about biking.} 

     \textcolor{ForestGreen}{- Observation 2: I woke up, bitterly disappointed.}

     \textcolor{ForestGreen}{This hypothesis explains both observations. The person had a dream about biking, and they woke up disappointed, which suggests that they enjoyed the dream and wanted it to continue.}

     \textcolor{ForestGreen}{Hypothesis 2: I actually have a bike.} 

     \textcolor{ForestGreen}{- Observation 1: Last night I had a dream about biking.}

     \textcolor{ForestGreen}{- Observation 2: I woke up, bitterly disappointed.}

     \textcolor{ForestGreen}{This hypothesis does not explain both observations. The person had a dream about biking, but there is no reason to believe that they would be disappointed if they woke up and found that they actually had a bike.} 

     \textcolor{ForestGreen}{Therefore, hypothesis 1 better explains the observations.}

     \textcolor{ForestGreen}{Here are some additional points to consider:}

     \textcolor{ForestGreen}{- Hypothesis 1 is more specific than hypothesis 2. It specifies that the person was on a fun bike tour, while hypothesis 2 does not specify anything about the dream.} 

     \textcolor{ForestGreen}{- Hypothesis 1 is more testable than hypothesis 2. It is possible to test whether or not the person was on a fun bike tour by asking them to describe their dream in more detail. It is not possible to test whether or not the person has a bike by asking them to describe their dream.}

     \textcolor{ForestGreen}{Overall, hypothesis 1 is a better explanation of the observations because it is more specific and testable than hypothesis 2.} 
    &
    1

    \\ \hline  
    23. Given the following two observations: obs1: Ric loved his dogs, and took them everywhere he went. obs2: He often gently dusts their ash canisters in somber reflection. Determine which of the hypothesis options, better explains the observations: hyp1: Ric's dogs lived long, happy lives, but like all living things, eventually passed away. hyp2: Ric's cats passed away last month; he misses them so.
    & 
    \textcolor{ForestGreen}{Based on the observations provided, hypothesis option hyp1: Ric's dogs lived long, happy lives, but like all living things, eventually passed away, appears to be the better explanation.}

    \textcolor{ForestGreen}{Observation 1 indicates that Ric loved his dogs and took them everywhere he went, suggesting a strong attachment to them. Observation 2 mentions that Ric gently dusts their ash canisters, which implies that the dogs have passed away and that he is mourning them. This supports the idea that Ric's dogs lived long, happy lives, but eventually passed away, as this would explain both the strength of his attachment to them and his mourning of their loss.}
    
    \textcolor{ForestGreen}{Hypothesis option hyp2: Ric's cats passed away last month; he misses them so, does not provide a clear explanation for why Ric is dusting the ash canisters of his dogs. Additionally, the hypothesis does not account for why Ric loved his dogs and took them everywhere he went. Therefore, based on the observations given, hyp1 appears to be the better explanation.}
    &
     \textcolor{ForestGreen}{Based on the provided observations, the hypothesis that better explains them is hyp1: Ric's dogs lived long, happy lives, but like all living things, eventually passed away. This is because both observations specifically mention Ric's relationship with his dogs and their ash canisters, which implies they have passed away. Hyp2 refers to cats, which are not mentioned in the observations.}
     & 
     \textcolor{ForestGreen}{Hypothesis 1 is the better explanation for the observations.}

     \textcolor{ForestGreen}{- Observation 1: Ric loved his dogs and took them everywhere he went. This suggests that Ric had a close and loving relationship with his dogs.}

     \textcolor{ForestGreen}{- Observation 2: Ric often gently dusts their ash canisters in somber reflection. This suggests that Ric is still grieving the loss of his dogs.}

     \textcolor{ForestGreen}{Hypothesis 2 is not a good explanation for the observations because it does not explain why Ric is dusting the ash canisters of his dogs. If Ric's cats had passed away, he would be dusting the ash canisters of his cats.}

     \textcolor{ForestGreen}{Hypothesis 1 is a better explanation because it explains both observations. It explains why Ric loved his dogs and took them everywhere he went, and it explains why he is still grieving their loss.}

     \textcolor{ForestGreen}{Here are some additional details that support Hypothesis 1:}

     \textcolor{ForestGreen}{- Ric's dogs were a part of his family.}

     \textcolor{ForestGreen}{- Ric's dogs died of old age.}

     \textcolor{ForestGreen}{- Ric has not gotten another dog since his dogs died.}

     \textcolor{ForestGreen}{These details further support the idea that Ric is grieving the loss of his dogs.}
     
    &
    1

    \\ \hline  
    24. Given the following two observations: obs1: Elizabeth is a very proper woman. obs2: Elizabeth fell over and landed right in the puddle, getting muddy. Determine which of the hypothesis options, better explains the observations: hyp1: She often sticks her hands in the air as if to raise the roof. hyp2: Elizabeth wore high heels on a rainy day. 
    & 
    \textcolor{ForestGreen}{Hypothesis 2 (Elizabeth wore high heels on a rainy day) better explains the observations.}

    \textcolor{ForestGreen}{Observation 1 describes Elizabeth as a proper woman, which implies that she is likely to be dressed appropriately for the occasion. Observation 2, however, suggests that she fell into a puddle and got muddy, which is not consistent with being properly dressed.}
    
    \textcolor{ForestGreen}{Hypothesis 2, which suggests that Elizabeth wore high heels on a rainy day, provides a plausible explanation for why she fell into the puddle and got muddy. High heels can be difficult to walk in, especially on slippery surfaces, and may have caused Elizabeth to lose her balance and fall.}
    
    \textcolor{ForestGreen}{Hypothesis 1, on the other hand, is not relevant to either observation and does not provide a plausible explanation for why Elizabeth fell into a puddle and got muddy.}
    &
     \textcolor{ForestGreen}{Based on the given observations, Hypothesis 2 (Elizabeth wore high heels on a rainy day) better explains the observations. This is because it provides a plausible reason for her falling over and landing in the puddle, while still being consistent with her being a proper woman. Hypothesis 1 (She often sticks her hands in the air as if to raise the roof) does not seem to be directly related to either of the observations and does not provide an explanation for the event.}
     & 
     \textcolor{ForestGreen}{Hypothesis 2 better explains the observations.}

     \textcolor{ForestGreen}{- Observation 1: Elizabeth is a very proper woman. This suggests that Elizabeth is careful and mindful of her appearance. It is unlikely that she would stick her hands in the air in a way that would cause her to lose her balance and fall over.}

     \textcolor{ForestGreen}{- Observation 2: Elizabeth fell over and landed right in the puddle, getting muddy. This suggests that Elizabeth was not paying attention to her surroundings or that she was not in control of her movements. Wearing high heels on a rainy day would make it more difficult to maintain balance, especially if the ground was wet.}

     \textcolor{ForestGreen}{- Hypothesis 1: She often sticks her hands in the air as if to raise the roof. This hypothesis is not supported by observation 1. It is unlikely that a proper woman would be so careless as to stick her hands in the air in a way that would cause her to lose her balance and fall over.}

     \textcolor{ForestGreen}{Hypothesis 2: Elizabeth wore high heels on a rainy day. - This hypothesis is supported by both observations. - It is possible that Elizabeth was not paying attention to her surroundings or that she was not in control of her movements because she was wearing high heels. - Wearing high heels on a rainy day would make it more difficult to maintain balance, especially if the ground was wet.}

     \textcolor{ForestGreen}{Therefore, hypothesis 2 better explains the observations.}
    &
    2

    \\ \hline  
    25. Given the following two observations: obs1: Amy and Ben had been together for one month. obs2: He gave her chocolates, she gave him skittles. Determine which of the hypothesis options, better explains the observations: hyp1: They decided to give each other money. hyp2: The exchanged gifts on their anniversary.
    & 
    \textcolor{ForestGreen}{Based on the observations provided, it is more likely that Hypothesis 2 (they exchanged gifts on their anniversary) better explains the situation, rather than Hypothesis 1 (they decided to give each other money).}

    \textcolor{ForestGreen}{This is because the observations indicate that Amy and Ben exchanged gifts, specifically chocolates and skittles. It is unlikely that they would exchange money as a gift, especially after only one month of being together. Additionally, the fact that they exchanged gifts suggests that there is some sort of occasion or event that prompted the gift exchange. Given that they had been together for only one month, it is reasonable to assume that their anniversary is the most likely explanation for the gift exchange.}
    &
     \textcolor{ForestGreen}{Based on the given observations, it seems that Hypothesis 2 (The exchanged gifts on their anniversary) better explains the observations. This is because both observations mention Amy and Ben giving each other gifts (chocolates and skittles), which is consistent with exchanging gifts on an anniversary. Hypothesis 1 (They decided to give each other money) doesn't match the observations, as neither of them gave money as a gift.}
     & 
     \textcolor{ForestGreen}{Hypothesis 1: They decided to give each other money.}

     \textcolor{ForestGreen}{- Observation 1: Amy and Ben had been together for one month. This observation does not support hypothesis 1. If they had decided to give each other money, it is unlikely that they would have done so after only one month of dating.}

     \textcolor{ForestGreen}{- Observation 2: He gave her chocolates, she gave him skittles. This observation does not support hypothesis 1. If they had decided to give each other money, they would have given each other money, not chocolates and skittles.}

     \textcolor{ForestGreen}{Hypothesis 2: They exchanged gifts on their anniversary.}

     \textcolor{ForestGreen}{- Observation 1: Amy and Ben had been together for one month. This observation supports hypothesis 2. It is common for couples to exchange gifts on their anniversary, even if they have only been together for a short time.}

     \textcolor{ForestGreen}{- Observation 2: He gave her chocolates, she gave him skittles. This observation supports hypothesis 2. Chocolates and skittles are both common gifts to give on an anniversary.}

     \textcolor{ForestGreen}{Conclusion: Hypothesis 2, that Amy and Ben exchanged gifts on their anniversary, better explains the observations than hypothesis 1, that they decided to give each other money.}
    &
    2

    \\ \hline  
    26. Given the following two observations: obs1: I was on the track and field team in high school. obs2: At the end, I lost all my energy and got 5th place. Determine which of the hypothesis options, better explains the observations: hyp1: I trained a lot for a meet. hyp2: I didn't practice much.
    & 
    \textcolor{ForestGreen}{Based on the given observations, it is more likely that hyp1 (I trained a lot for a meet) better explains the observations than hyp2 (I didn't practice much). Here's why:}

    \textcolor{ForestGreen}{Observation 1 states that the person was on the track and field team in high school, which suggests that they were involved in regular athletic activities.}
    
    \textcolor{ForestGreen}{Observation 2 states that the person lost all their energy and got 5th place. This suggests that they had exerted a lot of physical effort during the meet.}
    
    \textcolor{ForestGreen}{If the person had trained a lot for the meet (hyp1), then it would be more likely that they had the physical stamina and endurance to perform well but simply didn't perform well enough to win. On the other hand, if the person didn't practice much (hyp2), it is less likely that they would have had the physical ability to compete at a high level and would be more likely to have lost due to lack of conditioning.}
    
    \textcolor{ForestGreen}{Therefore, based on the given observations, it is more likely that the person trained a lot for the meet (hyp1) than they didn't practice much (hyp2).}
    &
    \textcolor{red}{Based on the two observations provided, hypothesis 2 (I didn't practice much) better explains the observations.}

    \textcolor{red}{While observation 1 (I was on the track and field team in high school) does not provide any direct indication about the amount of training or practice, observation 2 (At the end, I lost all my energy and got 5th place) suggests that the individual may not have been well-prepared or well-conditioned for the event. Losing energy at the end and not securing a higher place could be indicative of insufficient practice or training, supporting hypothesis 2.}
    & 
     \textcolor{ForestGreen}{Hypothesis 1: I trained a lot for a meet.}

     \textcolor{ForestGreen}{Observation 1: I was on the track and field team in high school. This observation supports hypothesis 1 because it suggests that the person was a dedicated athlete who would have put in a lot of training.}

     \textcolor{ForestGreen}{- Observation 2: At the end, I lost all my energy and got 5th place. This observation could also support hypothesis 1 because it suggests that the person may have overtrained and not managed their energy levels effectively. }

     \textcolor{ForestGreen}{Hypothesis 2: I didn't practice much.}

     \textcolor{ForestGreen}{- Observation 1: I was on the track and field team in high school. This observation does not support hypothesis 2 because it suggests that the person was at least a casual athlete who would have put in some practice.}

     \textcolor{ForestGreen}{- Observation 2: At the end, I lost all my energy and got 5th place. This observation could support hypothesis 2 because it suggests that the person may not have been in good enough shape to compete at a high level.}

     \textcolor{ForestGreen}{Conclusion:}

     \textcolor{ForestGreen}{Based on the two observations, hypothesis 1 is a better explanation than hypothesis 2. This is because hypothesis 1 is more consistent with both observations. However, it is important to note that both hypotheses are still possible, and more information would be needed to definitively determine which one is correct.}
    &
    1

    \\ \hline  
    27. Given the following two observations: obs1: Bob decided to hike in the jungles of Indonesia. obs2: Bob had also forgotten his shoes. Determine which of the hypothesis options, better explains the observations: hyp1: Bob stopped in the middle of the hike because he had no bug spray. hyp2: Bob stopped in the middle of his hike to tie his shoes. 
    & 
    \textcolor{red}{Based on the observations, hyp2: Bob stopped in the middle of his hike to tie his shoes, is a better explanation than hyp1: Bob stopped in the middle of the hike because he had no bug spray.}

    \textcolor{red}{The first observation, that Bob decided to hike in the jungles of Indonesia, suggests that he was prepared for the hike, or at least had some level of planning. If he was concerned about bug spray, he likely would have brought some with him or made arrangements to obtain it before embarking on the hike.}
    
    \textcolor{red}{The second observation, that Bob had forgotten his shoes, suggests that he may have been somewhat forgetful or careless in his preparations. Forgetting shoes would likely cause discomfort and could even lead to injury during a hike, so it is plausible that he would need to stop to address the issue.}
    
    \textcolor{red}{In contrast, hyp1: Bob stopped in the middle of the hike because he had no bug spray, does not seem as likely based on the observations. There is no indication that Bob was concerned about bugs, and even if he was, it is not a critical enough issue to cause him to stop in the middle of the hike.}
    &
     \textcolor{ForestGreen}{Neither hypothesis strongly correlates with the given observations, but hyp1 is slightly more plausible than hyp2.}
     & 
     \textcolor{red}{Hypothesis 1: Bob stopped in the middle of the hike because he had no bug spray.}

     \textcolor{red}{- Observation 1: Bob decided to hike in the jungles of Indonesia.}

     \textcolor{red}{- Observation 2: Bob had also forgotten his shoes.}

     \textcolor{red}{- Hypothesis: Bob stopped in the middle of the hike because he had no bug spray.}

     \textcolor{red}{This hypothesis is not supported by the observations. The observations state that Bob forgot his shoes, not that he forgot bug spray. Additionally, the jungles of Indonesia are known for their mosquitoes, which can transmit diseases. It is therefore more likely that Bob stopped in the middle of the hike to put on his shoes, rather than to get bug spray.}

     \textcolor{red}{Hypothesis 2: Bob stopped in the middle of his hike to tie his shoes.}

     \textcolor{red}{- Observation 1: Bob decided to hike in the jungles of Indonesia.}

     \textcolor{red}{- Observation 2: Bob had also forgotten his shoes.}

     \textcolor{red}{- Hypothesis: Bob stopped in the middle of the hike to tie his shoes.}

     \textcolor{red}{This hypothesis is supported by the observations. The observations state that Bob forgot his shoes, and that he stopped in the middle of the hike. It is therefore more likely that Bob stopped to tie his shoes, rather than to get bug spray.}

     \textcolor{red}{In conclusion, hypothesis 2, Bob stopped in the middle of his hike to tie his shoes, is better supported by the observations than hypothesis 1, Bob stopped in the middle of the hike because he had no bug spray.}
    &
    1

    \\ \hline  
    28. Given the following two observations: obs1: Ryan climbed the long set of stairs to get to the water slide. obs2: It was even more fun than he expected. Determine which of the hypothesis options, better explains the observations: hyp1: Ryan was scared to go down the water slide. hyp2: Ryan loved going down the water slide.
    & 
    \textcolor{red}{Based on the given observations, it is more likely that hypothesis option 2, "Ryan loved going down the water slide," better explains the observations.}

    \textcolor{red}{Observation 1 states that Ryan climbed the long set of stairs to get to the water slide, which suggests that he was motivated to go down the water slide. If he was scared to go down the water slide (hypothesis option 1), it seems less likely that he would have gone through the effort of climbing the stairs to reach the slide.}
    
    \textcolor{red}{Observation 2 states that it was even more fun than Ryan expected, which suggests that he had some positive expectations about the experience. This is consistent with hypothesis option 2, that Ryan loved going down the water slide, but not with hypothesis option 1, that he was scared to go down the slide.}
    
    \textcolor{red}{Overall, while it is possible that Ryan was initially scared to go down the water slide, the observations provided better support hypothesis option 2, that Ryan loved going down the water slide.}
    &
    \textcolor{red}{Based on the given observations, hyp2: Ryan loved going down the water slide better explains the observations. Observation 1 states that Ryan climbed the long set of stairs to get to the water slide, which indicates that he was willing to make an effort to reach the slide. Observation 2 states that it was even more fun than he expected, which suggests that Ryan enjoyed the experience. These observations do not support hyp1: Ryan was scared to go down the water slide.}
    & 
    \textcolor{red}{Hypothesis 1: Ryan was scared to go down the water slide.}

    \textcolor{red}{- Observation 1: Ryan climbed the long set of stairs to get to the water slide. This observation could be explained by the hypothesis that Ryan was scared to go down the water slide and wanted to make sure he was prepared. He may have thought that the water slide was too high or too fast, and he wanted to be sure that he could handle it.}

    \textcolor{red}{- Observation 2: It was even more fun than he expected. This observation is difficult to explain with the hypothesis that Ryan was scared to go down the water slide. If he was scared, then he would not have expected to have fun. It is more likely that he was excited to go down the water slide and was surprised by how much fun it was.}

    \textcolor{red}{Hypothesis 2: Ryan loved going down the water slide.}

    \textcolor{red}{- Observation 1: Ryan climbed the long set of stairs to get to the water slide. This observation could be explained by the hypothesis that Ryan loved going down the water slide and was eager to do it again. He may have thought that the water slide was the most fun thing at the pool, and he wanted to make sure he got to experience it as many times as possible.}

    \textcolor{red}{- Observation 2: It was even more fun than he expected. This observation is easily explained with the hypothesis that Ryan loved going down the water slide. If he loved it, then he would have expected to have fun, but he would have been surprised by how much fun it actually was.}

    \textcolor{red}{Conclusion:}

    \textcolor{red}{The hypothesis that better explains the observations is Hypothesis 2: Ryan loved going down the water slide. This hypothesis is consistent with both observations, while Hypothesis 1 is only consistent with Observation 1. Additionally, Hypothesis 2 is more likely in general, as most people enjoy going down water slides.}
    &
    1
  
    \\ \hline  
    29. Given the following two observations: obs1: I did not know Doug's wife's name so I asked him. obs2: Finally I looked at Dougs wife death certificate and learned her name. Determine which of the hypothesis options, better explains the observations: hyp1: Doug allowed me to look through some paperwork. hyp2: Doug told me her wife is dead. 
    & 
    \textcolor{ForestGreen}{Based on the given observations, hypothesis 1 (Doug allowed me to look through some paperwork) better explains the observations than hypothesis 2 (Doug told me his wife is dead).}

    \textcolor{ForestGreen}{Observation 1 suggests that the person did not know Doug's wife's name and asked Doug directly to obtain the information. This implies that the person did not have access to any paperwork or documents that would have revealed her name.}
    
    \textcolor{ForestGreen}{Observation 2 suggests that the person was able to access Doug's wife's death certificate, which is typically not information that would be freely shared by someone unless explicitly asked. This implies that Doug may have allowed the person to look through some paperwork that contained the death certificate.}
    
    \textcolor{ForestGreen}{Therefore, hypothesis 1 is more likely to explain the observations, as it is consistent with both observations and does not require any assumptions about Doug's willingness to share personal information about his deceased wife.}
    &
    \textcolor{red}{Based on the given observations, it is more likely that hyp2: "Doug told me his wife is dead" better explains the observations. This is because the second observation directly mentions looking at Doug's wife's death certificate, which implies that the information about her death was shared. While hyp1 could potentially be related to the situation, it does not provide a direct explanation for either observation.}
    & 
    \textcolor{red}{Hypothesis 1: Doug allowed me to look through some paperwork.}

    \textcolor{red}{- Observation 1: This observation is consistent with the hypothesis because it suggests that Doug may have shown you some paperwork that included his wife's name.}

    \textcolor{red}{- Observation 2: This observation is not consistent with the hypothesis because it suggests that you learned her name from a death certificate, which is not paperwork that Doug would have allowed you to see.}

    \textcolor{red}{Hypothesis 2: Doug told me his wife is dead.} 

    \textcolor{red}{- Observation 1: This observation is consistent with the hypothesis because it suggests that Doug may have told you that his wife is dead, and then you may have looked up her death certificate to find her name.}

    \textcolor{red}{- Observation 2: This observation is consistent with the hypothesis because it suggests that you learned her name from a death certificate, which is something that Doug would have told you about if his wife is dead.}

    \textcolor{red}{Based on the two observations, hypothesis 2 is a better explanation than hypothesis 1.}

    \textcolor{red}{Here are some additional details that could help to confirm hypothesis 2:} 

    \textcolor{red}{- If you remember the conversation you had with Doug, did he say anything that suggested that his wife was dead?}

    \textcolor{red}{- If you looked at the death certificate, did it have a date of death? If so, was the date of death before or after you had the conversation with Doug?}

    \textcolor{red}{- If you asked Doug about his wife's name, did he seem surprised or hesitant to tell you?}

    \textcolor{red}{If you can answer these questions, it will help to strengthen the evidence for hypothesis 2.} 
    &
    1

    \\ \hline  
    30. Given the following two observations: obs1: Lucy almost drowned in the pool. obs2: Lucy talked Gina into staying and keeping the incident secret. Determine which of the hypothesis options, better explains the observations: hyp1: Suddenly Gina was saved her. hyp2: Gina called the rescuers. 
    & 
    \textcolor{red}{Neither of the given hypothesis options (hyp1 and hyp2) can fully explain the two observations. Here's why:}

    \textcolor{red}{Hypothesis 1 (hyp1) suggests that Gina saved Lucy from drowning in the pool. However, this hypothesis does not explain why Lucy talked Gina into keeping the incident a secret. If Gina had saved Lucy's life, there would be no reason for Lucy to ask Gina to keep it a secret.}
    
    \textcolor{red}{Hypothesis 2 (hyp2) suggests that Gina called the rescuers to save Lucy from drowning in the pool. This hypothesis also does not explain why Lucy talked Gina into keeping the incident a secret. If Gina had called the rescuers and saved Lucy's life, there would be no reason for Lucy to ask Gina to keep it a secret.}
    
    \textcolor{red}{Therefore, we need to consider other possible explanations that could account for both observations. One possible explanation could be that Lucy was struggling in the pool, and Gina noticed and pulled her out of the water before she drowned. However, Lucy felt embarrassed or ashamed about almost drowning and asked Gina to keep the incident a secret. This explanation accounts for both observations and is consistent with human psychology.}
    &
     \textcolor{ForestGreen}{Based on the given observations, neither hypothesis directly explains the observations. However, Hyp1: "Suddenly Gina saved her" better aligns with the observations. In Obs1, Lucy almost drowned, which implies that she was saved before actually drowning. Hyp1 suggests that Gina saved her, which is consistent with the event. Additionally, in Obs2, Lucy talked Gina into keeping the incident secret, which makes more sense if Gina was directly involved in saving her. Hyp2: "Gina called the rescuers" does not fit as well with the observations since it introduces the involvement of external rescuers, which is not mentioned in either observation.}
     & 
     \textcolor{red}{Hypothesis 1: Suddenly Gina was saved her.}

     \textcolor{red}{Observation 1: Lucy almost drowned in the pool.  This observation is not explained by hypothesis 1. If Gina suddenly saved Lucy, then Lucy would not have almost drowned.}

     \textcolor{red}{- Observation 2: Lucy talked Gina into staying and keeping the incident secret. This observation is not explained by hypothesis 1. If Gina suddenly saved Lucy, then there would be no need for Lucy to talk Gina into staying and keeping the incident secret.}

     \textcolor{red}{Hypothesis 2: Gina called the rescuers.}

     \textcolor{red}{- Observation 1: Lucy almost drowned in the pool. 
    This observation is explained by hypothesis 2. If Gina called the rescuers, then they would have saved Lucy from drowning.}

     \textcolor{red}{Observation 2: Lucy talked Gina into staying and keeping the incident secret. This observation is also explained by hypothesis 2. If Gina called the rescuers, then Lucy would not want anyone to know about the incident, especially the rescuers.}

     \textcolor{red}{Therefore, hypothesis 2, Gina called the rescuers, better explains the observations than hypothesis 1, Suddenly Gina was saved her.}

     \textcolor{red}{In addition, hypothesis 2 is more likely than hypothesis 1 because it is more plausible that Gina would call the rescuers than that she would suddenly be able to save Lucy from drowning.}
    &
    1

    \\ \hline
    
    \end{longtable}
\end{center}

\newpage
\begin{center}
    \small
    \begin{longtable}{|p{0.07\linewidth}|p{0.286\linewidth}
                      |p{0.257\linewidth}|p{0.25\linewidth}
                      |p{.04\linewidth}|}
    \caption{Mathematical reasoning prompts} 
    \label{tab:mathematical_reasoning}\\
    \hline
    \multicolumn{5}{|c|}{\textbf{Mathematical Reasoning}} \\ \hline
    \textbf{Prompt} & \textbf{GPT-3.5 Answer}  & \textbf{GPT-4 Answer} & \textbf{BARD Answer} & \textbf{Gold} \\ \hline
    \multicolumn{5}{|l|}{\textbf{Dataset: Math }} 
    
    \\ \hline  
    1. Let n(t) = 1 + 0*t - t + 0*t. Let r be n(-4). Suppose -173 = -r*q - c, -2*c - 37 = -5*q + 127. Is q a composite number? Let's think step by step.
    & 
    \applyred{
    \begin{enumerate}
        \item 
        Sure, let's break down the problem step by step.
        
        First, let's find the value of r by substituting t = -4 into the expression for n(t):
        
        n(t) = 1 + 0t - t + 0t
        
        n(-4) = 1 + 0*(-4) - (-4) + 0*(-4)
        
        = 1 + 0 + 4 + 0
        
        = 5

        Therefore, r = n(-4) = 5.
        \item 
        Now we have the equation -173 = -rq - c. Substituting the value of r, we get:
        
        -173 = -5q - c

        \item 
        We also have the equation -2c - 37 = -5q + 127.
        \item 
        Let's rearrange the first equation to express c in terms of q:
        
        c = -5*q - 173
        \item 
        Substituting this value of c into the second equation:
        
        -2*(-5q - 173) - 37 = -5q + 127

        \item 
        Simplifying the equation:
        
        10q + 346 - 37 = -5q + 127
        
        10q + 309 = -5q + 127
        \item 
        Combining like terms:
        
        15*q = -182

        \item 
        Dividing both sides by 15:
        
        q = -182/15
        
        q = -12.1333...
        
    \end{enumerate}

    Now, let's analyze whether q is a composite number.

    A composite number is defined as any integer greater than 1 that is divisible by at least one positive integer other than 1 and itself. However, q is not an integer but a decimal approximation (-12.1333...). Therefore, it doesn't fit the criteria for being a composite number.
    
    In conclusion, q is not a composite number based on the given equations and calculations.    
    }
    
    &
    \applyred{
    Okay, let's go step by step.

    First, we calculate r.
    
    The function is n(t) = 1 + 0t - t + 0t. This simplifies to n(t) = 1 - t.
    
    If we substitute -4 into the function, we get:
    
    r = n(-4) = 1 - (-4) = 1 + 4 = 5
    
    Now, we know that r = 5.
    
    The first equation we have is -173 = -r*q - c. Substituting r into the equation, we get:
    
    -173 = -5*q - c
    
    Multiply both sides by -1 to make the equation easier to work with:
    
    173 = 5q + c --- (1)
    
    The second equation is -2c - 37 = -5q + 127. Rearranging it gives:
    
    2c = 5q - 127 + 37
    
    2c = 5q - 90 --- (2)
    
    From equation (1), we can isolate c:
    
    c = 173 - 5q
    
    Substitute this into equation (2):
    
    2*(173 - 5q) = 5q - 90
    
    346 - 10q = 5q - 90
    
    Rearranging, we get:
    
    15q = 436
    
    Dividing by 15 gives:
    
    q = 436 / 15
    
    q is therefore a fraction, not an integer, so it is neither composite nor prime. However, if we assume q should be an integer, there may have been a mistake in the problem formulation or in the computations.
    }
    & 
    \applygreen{
    Sure, let's solve this step by step.

    Step 1: Calculate r.

    n(-4) = 1 + 0*(-4) - (-4) + 0*(-4) = 9
    
    Therefore, r = 9.
    
        Step 2: Solve the equation -173 = -r*q - c.
    
    -173 = -9*q - c
    
    173 = 9*q + c
    
    We don't know the value of c, so we can't solve for q in this step.
    
    Step 3: Solve the equation -2c - 37 = -5q + 127.
    
    -2*c - 37 = -5*q + 127
    
    -2*c = -5*q + 164
    
    c = 5*q - 82
    
    Now we know the value of c, so we can substitute it into the equation -173 = -r*q - c to get:
    
    -173 = -9*q - (5*q - 82)
    
    -173 = -14*q + 82
    
    -255 = -14*q
    
    18 = q
    
    Therefore, q = 18.
    
        Step 4: Determine whether q is a composite number.
    
    A composite number is a number that has more than two factors. The factors of 18 are 1, 2, 3, 6, 9, and 18. Therefore, q = 18 is a composite number.
    
    In conclusion, the answer to the question is yes, q is a composite number.
    }
    &
    True

    \\ \hline  
    2. Suppose 2*g - 252 = -5*y, 0*g + 5*g - 5*y = 595. List the prime factors of g.  Let's think step by step.
    & 
    \applygreen{
    Sure! Let's solve the given equations step by step and find the prime factors of g.

    Equation 1: 2g - 252 = -5y
    
    We can rearrange this equation to isolate g:
    
    2g = -5y + 252
    
    Divide both sides by 2:
    
    g = (-5*y + 252) / 2
    
    Equation 2: 0g + 5g - 5*y = 595
    
    Combine the like terms:
    
    5g - 5y = 595
    
    Divide both sides by 5:
    
    g - y = 119
    
    Now we have a system of equations:
    
    g = (-5*y + 252) / 2 (Equation 1)
    g - y = 119 (Equation 2)
    
    We can substitute the value of g from Equation 1 into Equation 2:
    
    ((-5*y + 252) / 2) - y = 119
    
    Multiply both sides by 2 to eliminate the fraction:
    
    -5y + 252 - 2y = 238
    
    Combine like terms:
    
    -7*y + 252 = 238
    
    Subtract 252 from both sides:
    
    -7*y = -14
    
    Divide both sides by -7:
    
    y = 2
    
    Now that we have found the value of y, we can substitute it back into Equation 1 to find the value of g:
    
    g = (-5*2 + 252) / 2
    
    Simplifying:
    
    g = (242) / 2
    
    g = 121
    
    Therefore, the value of g is 121.
    
    To find the prime factors of 121, we can start by dividing it by the smallest prime number, which is 2. However, 121 is an odd number and not divisible by 2. We can try the next prime number, which is 3. Again, 121 is not divisible by 3. Continuing in this manner, we find that 121 is divisible by 11.
    
    So, the prime factors of g (which is 121) are 11 and 11, or simply written as$ 11^2$.
    
    Therefore, the prime factors of g are $11^2$.
    }
    
    &
    \applygreen{
    Sure, let's solve the system of equations step by step.

    We have two equations:
    
    2g - 252 = -5y
    
    5g - 5y = 595
    
    Let's express y from the first equation:
    
    y = (2g - 252) / -5
    
    Then, substitute y into the second equation:
    
    5g - 5((2g - 252) / -5) = 595;
    5g + 2g - 252 = 595;
    7g = 595 + 252;
    7g = 847;
    g = 847 / 7;
    g = 121
    
    So, g equals 121. The prime factors of 121 are 11 and 11 (because 121 = 11\^2), so we can say that the prime factor of g is 11.
    }
    & 
    \applyred{
    Sure, let's solve this problem step by step.

    Step 1: Simplify the first equation.
    
    We can simplify the first equation by adding 5*y to both sides:
    
    2*g - 252 = -5*y
    
    2*g = -5*y + 252
    
    Now, we can divide both sides by 2 to get g by itself:
    
    g = (-5*y + 252) / 2
    
    Step 2: Simplify the second equation.
    
    We can simplify the second equation by adding 5*y to both sides:
    
    0*g + 5*g - 5*y = 595
    
    5*g = 595 + 5*y
    
    Now, we can divide both sides by 5 to get g by itself:
    
    g = (595 + 5*y) / 5
    
    Step 3: Equate the two expressions for g.
    
    Now that we have simplified both equations, we can equate the two expressions for g:
    
    (-5*y + 252) / 2 = (595 + 5*y) / 5
    
    We can simplify both sides of this equation to get:
    
    -25*y + 126 = 119 + y
    
    Step 4: Solve for y.
    
    We can solve for y by subtracting 119 and y from both sides of the equation:
    
    -25*y + 126 - 119 - y = 119 + y - 119 - y
    
    -36*y = 7
    
    Now, we can divide both sides by -36 to get y by itself:
    
    y = (7) / -36
    
    Step 5: Substitute y back into one of the original equations to solve for g.
    
    We can substitute the value of y = (7) / -36 back into either of the original equations to solve for g.
    
    Let's substitute y = (7) / -36 into the first equation:
    
    g = (-5*y + 252) / 2
    
    g = (-5*(7/-36) + 252) / 2
    
    g = (35/-36 + 252) / 2
    
    g = 119 / -18
    
    g = -7
    
    Therefore, the prime factors of g are 7.
    
    Answer: The prime factors of g are 7.
    }
    &
    11

    \\ \hline  
    3. In base 12, what is 5 - 132?  Let's think step by step.
    & 
    \applyred{
    Sure! Let's solve the subtraction problem step by step in base 12.

    To subtract 132 from 5 in base 12, we start from the rightmost digit and work our way to the left.
    
    Step 1: Subtract the units (ones) place:
    Since 5 is smaller than 2, we need to borrow from the next place value, which is the 12s place.
    
    Step 2: Subtract the 12s place:
    Since we borrowed 1 from the 12s place, it becomes 4 in the 12s place. Now, we have 11 - 3 = 8 in the 12s place.
    
    Step 3: Subtract the 144s place:
    There are no digits in the 144s place, so we don't need to perform any subtraction.
    
    Putting it all together, the subtraction of 132 from 5 in base 12 is: 5-132 = 428

    Therefore, the result of 5 - 132 in base 12 is 428.
    }
    &
    \applyred{
    Sure, let's take this step by step.

    First, let's clarify the base 12 numbering system. In base 12, the digits go from 0 to 11, with A representing 10 and B representing 11.
    
    The subtraction 5 - 132 in base 12 needs to be solved with borrowing, similar to how we'd do it in base 10, but with base 12 rules instead.
    
    Let's write down the numbers:

    \textit{5 (in base 12)}

    132 (in base 12)

    We can't subtract 2 from 5 directly because 5 is smaller. So we need to borrow from the next higher place value. But there's a problem: there is no next place value from 5. So what we need to do is imagine there is a 0 in the tens place and a 0 in the hundreds place.

    Now, let's borrow:
    
    If we borrow 1 from the tens place, we get 12 in the ones place (because we're in base 12, remember), but now we have -1 in the tens place.
    
    Now we need to borrow 1 from the hundreds place to get 11 in the tens place (again, because we're in base 12). But now we have -1 in the hundreds place.
    
    So the calculation now looks like this:

    \textit{1005 (in base 12)}

    0132 (in base 12)

    We can now subtract:

    In the ones place: 5 - 2 = 3.
    In the tens place: 11 - 3 = 8.
    In the hundreds place: -1 - 1 = -2.
    So, 5 - 132 in base 12 is -283 in base 12. In other words, 132 is 283 more than 5 in base 12.
    
    Keep in mind that the negative sign still works the same way as in base 10, indicating that 132 is larger than 5.
    }
    & 
    \applyred{
    Sure, let's think step by step.

    First, we need to convert 5 and 132 to base 12.
    
        5 in base 12 is 5.
        
        132 in base 12 is 11100.
    
    Now, we can subtract 11100 from 5 in base 12.
    
    11100
    
    -   5
    
    ------
    
    11095
    
    Therefore, 5 - 132 in base 12 is 11095.
    
    Here are the steps in more detail:
    
        We start by subtracting the units digits. 5 - 0 = 5.
        
        We then borrow 1 from the tens digit of 11100, which gives us 10 - 0 = 10.
        
        We then subtract the tens digits. 10 - 1 = 9.
        
        We then borrow 1 from the hundreds digit of 11100, which gives us 1 - 0 = 1.
        
        We then subtract the hundreds digits. 1 - 1 = 0.
        
        The result is 11095 in base 12.
    
    I hope this helps! Let me know if you have any other questions.
    }
    &
    -129

    \\ \hline  
    4. Let l(j) = j**3 + j**2 + 2*j + 1. What is the units digit of l(3)? Let's think step by step.
    & 
    \applygreen{
    To find the units digit of l(3), we need to evaluate the expression $l(j) = j^3 + j^2 + 2j + 1 for j = 3$. Let's break it down step by step:

    \begin{enumerate}
        \item 
        Substitute j = 3 into the expression:
        
        $l(3) = 3^3 + 3^2 + 2(3) + 1$
        \item 
        Simplify the exponents and multiplication:
        
        $l(3) = 27 + 9 + 6 + 1$
        \item 
        Perform addition:
        
        $l(3) = 43$
    \end{enumerate}

    The units digit of 43 is 3.

    Therefore, the units digit of l(3) is 3.
    }
    &
    \applygreen{
    Sure, let's break it down:

    \begin{enumerate}
        \item 
        The function is l(j) = j3 + j2 + 2*j + 1.
        \item 
        We're looking for l(3), so we substitute j with 3. That gives us:

        l(3) = 33 + 32 + 2*3 + 1
        = 27 + 9 + 6 + 1
        = 43
        \item 
        The units digit of 43 is 3.
    \end{enumerate}

    Therefore, the units digit of l(3) is 3.
    }
    & 
    \applyred{
    Sure, let's think step by step.

    The units digit of $j^3$ is the same as the units digit of j, so the units digit of $3^3$ is 7.
    
    The units digit of $j^2$ is the same as the units digit of j, so the units digit of $3^2$ is 9.
    
    The units digit of $2j$ is the same as the units digit of 2, so the units digit of 2x3 is 6.
    
    The units digit of 1 is always 1.
    
    Therefore, the units digit of $3^3$+$3^2$+2.3+1 is 7+9+6+1=\textbf{2}.
    
    }
    &
    3

    \\ \hline  
    5. List the prime factors of 1541. Let's think step by step.
    & 
    \applyred{
    Sure! Let's find the prime factors of 1541 step by step.

    Step 1: Begin by dividing the number by the smallest prime number, which is 2. However, 1541 is an odd number, so it is not divisible by 2.
    
    Step 2: Move on to the next prime number, which is 3. Divide 1541 by 3:
    
    $1541 / 3 = 513$ remainder 2
    
    Step 3: Since the remainder is not 0, we move on to the next prime number, which is 5. Divide 513 by 5:
    
    $513 / 5 = 102$ remainder 3
    
    Step 4: Again, the remainder is not 0, so we continue with the next prime number, which is 7. Divide 102 by 7:
    
    $102 / 7 = 14$ remainder 4
    
    Step 5: We keep going with the next prime number, which is 11. Divide 14 by 11:
    
    $14 / 11 = 1$ remainder 3
    
    Step 6: Finally, divide the remaining number, which is 1, by itself:
    
    $1 / 1 = 1$
    
    Step 7: We have reached 1, which means we have finished dividing. The prime factors of 1541 are the prime numbers we used during the division: 3, 7, and 11.
    
    Therefore, the prime factors of 1541 are 3, 7, and 11.
    }
    &
    \applygreen{
    Sure, let's find the prime factors of 1541.

    A prime number is a natural number greater than 1 that has no positive divisors other than 1 and itself. The first prime numbers are 2, 3, 5, 7, 11, and so on.
    
    To find the prime factors of 1541, we start by dividing by the smallest prime number, which is 2. However, 1541 is not divisible by 2 (since it's an odd number), so we move to the next prime number, which is 3.
    
    Check if 1541 is divisible by 3. It's not, so we move to the next prime number, which is 5.
    
    Check if 1541 is divisible by 5. It's not, so we move to the next prime number, which is 7.
    
    Check if 1541 is divisible by 7. It's not, so we continue to the next prime, which is 11.
    
    Check if 1541 is divisible by 11. It's not, so we move to the next prime number, which is 13.
    
    Check if 1541 is divisible by 13. It's not, so we move to the next prime number, which is 17.
    
    Check if 1541 is divisible by 17. It's not, so we move to the next prime number, which is 19.
    
    Check if 1541 is divisible by 19. It's not, so we move to the next prime number, which is 23.
    
    Continue in this manner until you find a prime number that divides 1541. In this case, that number is 23. 1541 divided by 23 equals 67, which is also a prime number.
    
    So, the prime factors of 1541 are 23 and 67.
    }
    & 
    \applyred{
    Sure, let's list the prime factors of 1541 step by step.

    Step 1: We start by dividing 1541 by 2. The result is 770 with a remainder of 1. This tells us that 2 is not a prime factor of 1541.
    
    Step 2: We then divide 1541 by 3. The result is 514 with a remainder of 1. This tells us that 3 is not a prime factor of 1541.
    
    Step 3: We then divide 1541 by 5. The result is 308 with a remainder of 1. This tells us that 5 is not a prime factor of 1541.
    
    Step 4: We then divide 1541 by 7. The result is 220 with a remainder of 1. This tells us that 7 is not a prime factor of 1541.
    
    Step 5: We then divide 1541 by 11. The result is 140 with a remainder of 1. This tells us that 11 is not a prime factor of 1541.
    
    Step 6: We then divide 1541 by 13. The result is 118 with a remainder of 1. This tells us that 13 is not a prime factor of 1541.
    
    Step 7: We then divide 1541 by 17. The result is 85 with a remainder of 1. This tells us that 17 is not a prime factor of 1541.
    
    Step 8: We then divide 1541 by 19. The result is 75 with a remainder of 1. This tells us that 19 is not a prime factor of 1541.
    
    Step 9: We then divide 1541 by 23. The result is 66 with a remainder of 1. This tells us that 23 is a prime factor of 1541.
    
    Step 10: We then divide 1541 by 23 again. The result is 1 with a remainder of 0. This tells us that 23 is the only prime factor of 1541.

    Therefore, the prime factors of 1541 are 23.
    
    }
    &
    23, 67

    \\ \hline  
    6. Is 19385 prime? Let's think step by step.
    & 
    \applyred{
    To determine if 19385 is a prime number, we can proceed step by step.

    \begin{enumerate}
        \item 
        First, we check if the number is divisible by 2. If it is, then it is not a prime number because it has a divisor other than 1 and itself.
    \end{enumerate}

    19385 is an odd number, so it is not divisible by 2.

    \begin{enumerate}
        \item 
        Next, we check divisibility by odd numbers up to the square root of 19385. If any of these numbers divide 19385 without leaving a remainder, then it is not a prime number.
    \end{enumerate}

    The square root of 19385 is approximately 139.32, so we only need to check divisibility by prime numbers up to 139.

    \begin{enumerate}
        \item 
        We begin by checking divisibility by 3. To determine if a number is divisible by 3, we sum its digits and check if the resulting sum is divisible by 3.
    \end{enumerate}

    1 + 9 + 3 + 8 + 5 = 26

    \begin{enumerate}
        \item 
        We continue by checking divisibility by 5. If the last digit of a number is 0 or 5, then it is divisible by 5. However, the last digit of 19385 is 5, so it is not divisible by 5.
        \item 
        Finally, we check divisibility by 7. There is no straightforward rule like the ones we used for 2, 3, and 5, so we need to perform long division to check if 19385 is divisible by 7.
    \end{enumerate}
    
    $19385 / 7 = 2769$ remainder 2

    Since the remainder is 2, 19385 is not divisible by 7.
    
    Based on these steps, we can conclude that 19385 is not divisible by any of the prime numbers up to 139, indicating that it is a prime number.
    }
    &
    \applygreen{
    Sure, let's break down how to determine whether a number is prime.

    \begin{enumerate}
        \item 
        A prime number is a number that has exactly two distinct positive divisors: 1 and itself. So, the first step is to understand what 19385's divisors are.
        \item 
        A quick method to check if a number is prime is to see if it's divisible by any prime number up to its square root. In this case, the square root of 19385 is approximately 139.23.
        \item 
        We check if 19385 is divisible by any of the prime numbers less than 139: 2, 3, 5, 7, 11, 13, 17, 19, 23, 29, 31, 37, 41, 43, 47, 53, 59, 61, 67, 71, 73, 79, 83, 89, 97, 101, 103, 107, 109, 113, 127, and 137.
        \item 
        If we check, we find that 19385 is divisible by 5 (since the last digit is 5). Therefore, it is not a prime number. We can conclude this without having to check for divisibility by all the primes listed above.
        \item 
        To find the other factor, you can divide 19385 by 5 which gives 3877. So, 19385 = 5 x 3877. Since 19385 has more than two distinct divisors (1, 5, 3877, and 19385), it's not a prime number.
    \end{enumerate}
    }
    & 
    \applygreen{
    Sure, let's think step by step.

    A prime number is a natural number greater than 1 that is not a product of two smaller natural numbers. A natural number greater than 1 that is not prime is called a composite number.
    
    To determine if 19385 is prime, we can use the following steps:

    1. Check if 19385 is divisible by 2. Since 19385 is odd, it is not divisible by 2.
    
    2. Check if 19385 is divisible by any prime number less than or equal to its square root. The square root of 19385 is approximately 139. The only prime numbers less than or equal to 139 are 2, 3, 5, 7, 11, 13, 17, 19, and 23. None of these numbers divide evenly into 19385, so 19385 is not divisible by any of them.

    3. Therefore, 19385 is not prime.
    
    }
    &
    
    False

    \\ \hline  
    7. Solve 9 = t + 9 for t. Let's think step by step.
    & 
    \applygreen{
    Sure! Let's solve the equation step by step.

    Given equation: 9 = t + 9
    
    Step 1: We start with the equation 9 = t + 9.
    
    Step 2: We can simplify the equation by subtracting 9 from both sides to isolate the variable t.
    
    9 - 9 = t + 9 - 9
    
    Step 3: Simplifying further, we have:
    
    0 = t
    
    Step 4: Therefore, the solution to the equation 9 = t + 9 is t = 0.
    
    So, the value of t that satisfies the equation is t = 0.
    }
    &
    \applygreen{
    Sure, I'd be happy to walk you through the steps for solving this equation.

    \begin{enumerate}
        \item 
        Identify the equation: The given equation is 9 = t + 9.
        \item 
        Identify the variable to solve for: In this case, we're solving for the variable 't'.
        \item 
        Isolate the variable: To isolate 't', we need to remove the 9 from the right side of the equation. We can do this by subtracting 9 from both sides of the equation. This follows the rule of equality, which states that the same operation can be performed on both sides of an equation without changing its truth value.

        So, if we subtract 9 from both sides, we get:
        
        9 - 9 = t + 9 - 9.
        \item 
        Simplify: The above equation simplifies to:

        0 = t.
    \end{enumerate}
    
    So the solution to the equation 9 = t + 9 is t = 0.
    }
    & 
    \applygreen{
    Sure, let's solve this equation step by step.

    Subtract 9 from both sides of the equation:

    9 = t + 9
    
    -9 - 9 = -9 - 9
    
    This gives us:
    
    0 = t
    
    Simplify the right side of the equation:
    
    0 = t
    
    This means that t is equal to 0.
    
    Therefore, the solution to the equation 9 = t + 9 is t = 0.
    
    Here is a summary of the steps we took to solve the equation:
    
        - Subtract 9 from both sides of the equation.
        
        - Simplify the right side of the equation.
        
        - The solution is t = 0.
    
    I hope this helps! Let me know if you have any other questions.
    }
    &
    0

    \\ \hline  
    8. Let n = -5 + 8. Find the third derivative of 2*y**2 - n*y**2 - 3*y**3 - y**2 - y**3 wrt y. Let's think step by step.
    & 
    \applyred{
    To find the third derivative of the given expression with respect to y, let's break down the steps:

    Step 1: Substitute the value of n
    
    n = -5 + 8
    
    n = 3
    
    Step 2: Simplify the expression
    
    $2y**2 - ny2 - 3*y3 - y2 - y3$
    
    $2y**2 - 3y2 - 3*y3 - y2 - y3$
    
    $(2 - 3 - 1)y2 - (3 + 1)*y3$
    
    $-2y2 - 4*y3$
    
    Step 3: Take the derivative of the expression with respect to y
    
    $d/dy(-2y**2 - 4y**3) = -43y^2 - 22y^3$
    
    $= -12y^2 - 4y^3$
    
    Step 4: Take the second derivative of the expression with respect to y
    
    $d^2/dy^2(-12y^2 - 4y^3) = -122y - 43y^2$
    
   $ = -24y - 12y^2$
    
    Step 5: Take the third derivative of the expression with respect to y
    
    $d^3/dy^3(-24y - 12y^2) = -24 - 24*y$
    
    Therefore, the third derivative of $2y**2 - ny2 - 3*y3 - y2 - y3$ with respect to y is -24 - 24*y.
    }
    &
    \applygreen{
    Sure, let's break it down step by step. First, let's simplify the function by substituting n = -5 + 8 = 3 into the function. After that, we need to combine like terms.

    Given function is:
    $f(y) = 2y^2 - ny^2 - 3y^3 - y^2 - y^3$

    Substitute n = 3:
    $f(y) = 2y^2 - 3y^2 - 3y^3 - y^2 - y^3$

    Now, combine like terms:
    $f(y) = -2y^2 - 4y^3$
    
    This is the function we will be differentiating.

    Now, let's calculate the derivatives step by step.

    \begin{enumerate}
        \item 
        First derivative (f'(y)):
        The first derivative of $-2y^2$ is $-4y$, and the first derivative of $-4y^3$ is $-12y^2$.
        So, $f'(y) = -4y - 12y^2$.
        \item 
        Second derivative $(f''(y))$:
        The second derivative of -4y is -4, and the second derivative of $-12y^2$ is -24y.
        So, $f''(y) = -4 - 24y$.

        \item 
        Third derivative $(f'''(y))$:
        The third derivative of -4 is 0, and the third derivative of -24y is -24.
        So, $f'''(y) = -24$.
    \end{enumerate}
    That's it! The third derivative of the given function with respect to y is -24.
    }
    & 
    \applygreen{
    Sure, let's find the third derivative of the function step by step.

    The first derivative of the function is:
    
    $4*y - 2*n*y - 9*y**2 - 2*y$
    
    The second derivative of the function is:
    
   $ 4 - 2*n - 18*y - 2$
    
    The third derivative of the function is:
    
    $-2*n - 18$
    
    Here are the steps involved in finding the third derivative:
    
    1. We start by differentiating the function once to find the first derivative.
    
    2. We then differentiate the first derivative again to find the second derivative.
    
    3. Finally, we differentiate the second derivative again to find the third derivative.
    
    In each step, we use the power rule, which states that the derivative of $x^n$ is $n*x^(n-1)$.
    
    We can also use the chain rule, which states that the derivative of f(g(x)) is $f'(g(x))*g'(x)$.
    
    In this case, we can use the chain rule to differentiate the term $-ny2$. The inner function is y2, and the outer function is -nx. The derivative of the inner function is 2y, and the derivative of the outer function is -n. Therefore, the derivative of $-ny**2$ is -2ny.
    
    Finally, we simplify the expression for the third derivative to get $-2*n - 18$. \textbf{. When the code is run with y = 2, the output is -24.}
    }
    &
    -24

    \\ \hline  
    9. Let b = 1 + -1. Let z = 15 - 5. Let i = z - 6. Is b smaller than i? Let's think step by step.
    & 
    \applyred{
    Sure, let's break it down step by step.

    First, let's simplify the values given:
    
    b = 1 + (-1) = 0
    
    z = 15 - 5 = 10
    
    i = z - 6 = 10 - 6 = 4
    
    Now, let's compare b and i:
    
    b = 0
    
    i = 4
    
    Since b is equal to 0 and i is equal to 4, we can conclude that b is not smaller than i. In fact, b is less than i.
    }
    &
    \applygreen{
    Sure, let's take it step by step.

    \begin{enumerate}
        \item 
        First, calculate the value of b:
        b = 1 + (-1) = 0
        \item 
        Next, calculate the value of z:
        z = 15 - 5 = 10
        \item
        Then, calculate the value of i:
        i = z - 6 = 10 - 6 = 4
    
        Now, we can compare b and i:

        b = 0
        
        i = 4
    \end{enumerate}
    
    Since 0 is less than 4, b is smaller than i.
    }
    & 
    \applygreen{
    Sure, let's think step by step.
    
    b = 1 + -1. ; 1 + (-1) = 0

    z = 15 - 5. ; 15 - 5 = 10
    
    i = z - 6. ; 10 - 6 = 4

    Therefore, b = 0 and i = 4. Since 0 is less than 4, b is smaller than i.
    }
    &
    True

    \\ \hline  
    10. What is -0.000010485 rounded to seven decimal places? Let's think step by step.
    & 
    \applygreen{
     To round -0.000010485 to seven decimal places, we need to consider the eighth decimal place digit.

    Step 1: Identify the eighth decimal place digit.
    The eighth decimal place is 5.
    
    Step 2: Determine whether rounding up or down is needed.
    Since the eighth decimal place digit (5) is greater than or equal to 5, we round up the preceding seventh decimal place digit.
    
    Step 3: Round the number.
    Rounding up the seventh decimal place digit, we get -0.0000105.
    
    Therefore, -0.000010485 rounded to seven decimal places is approximately -0.0000105.
    }
    &
    \applyred{
    Sure, let's break this down step by step.

    \begin{enumerate}
        \item 
        Identify the number: The number we are working with is -0.000010485.
        \item 
        Identify the decimal place to round to: We want to round to seven decimal places.
        \item 
        Look at the number at the seventh decimal place: In -0.000010485, the seventh decimal place is 5.
        \item 
        Look at the next number to the right: In -0.000010485, the number to the right of the 5 (which is in the seventh place) is 8.
        \item 
        Apply rounding rules: If the number to the right is 5 or more, you round up. If it's less than 5, you round down. In our case, the number to the right is 8, which is more than 5, so we round up.
        \item 
        Perform the rounding: Rounding up means the seventh decimal place (which was 5) becomes 6.
    \end{enumerate}

    So, -0.000010485 rounded to seven decimal places becomes -0.00001049.
    }
    & 
    \applyred{
    Sure, let's think step by step.

    Step 1: Identify the decimal places to be rounded to. We are asked to round to 7 decimal places, so we will look at the digit in the 7th decimal place, which is a 5.
    
    Step 2: Look at the digit to the right of the 7th decimal place. The digit to the right of the 7th decimal place is an 8.
    
    Step 3: Determine whether to round up or down. Since 8 is greater than or equal to 5, we round up.
    
    Step 4: Round the number. We round -0.000010485 up to -0.00001049.

    Therefore, -0.000010485 rounded to 7 decimal places is -0.00001049.
    }
    &
    -0.0000105

    \\ \hline  
    11. Find the second derivative of 22*u**4 + 3*u**3 + 46*u**2 - 151246*u. Let's think step by step.    
    & 
    \applygreen{
    To find the second derivative of the given function, we need to differentiate it twice with respect to the variable "u." Let's go step by step.

    Step 1: Differentiate the function once to find the first derivative.
    
    The first derivative of the function is obtained by applying the power rule to each term:
    
    d/dx (22u**4 + 3u3 + 46*u2 - 151246u)
    = 422u**3 + 33u**2 + 246*u - 151246
    
    Simplifying the expression further, we get:
    
    88u**3 + 9u**2 + 92*u - 151246
    
    Step 2: Differentiate the first derivative to find the second derivative.
    
    To find the second derivative, we differentiate the first derivative obtained in Step 1. Applying the power rule once again, we have:
    
    d/dx (88u**3 + 9u2 + 92u - 151246)
    = 388*u2 + 29u + 92
    = 264u**2 + 18u + 92
    
    Thus, the second derivative of the function 22u**4 + 3u3 + 46*u2 - 151246u is 264u**2 + 18*u + 92.
    }
    &
    \applygreen{
    Sure, let's find the second derivative of the function $f(u) = 22u^4 + 3u^3 + 46u^2 - 151246u$ step by step.

    Step 1: Differentiate f(u) once (the first derivative)
    
    The first derivative of a function is obtained by applying the power rule of differentiation, which states that the derivative of $u^n$ is $n*u^(n-1)$.
    
    Let's apply this rule to each term in f(u):

    \begin{itemize}
        \item 
        The derivative of $22u^4$ is $422u^(4-1)$ = $88u^3$.
        \item 
        The derivative of $3u^3$ is $33u^(3-1) = 9u^2$.
        \item 
        The derivative of $46u^2$ is $246u^(2-1) = 92u$.
        \item 
        The derivative of -151246u is -151246, because the derivative of u is 1.
    \end{itemize}

    Putting these together, the first derivative of f(u), which we can denote as $f'(u)$, is:
    
    $f'(u) = 88u^3 + 9u^2 + 92u - 151246.$
    
    Step 2: Differentiate $f'(u)$ once (the second derivative)
    
    Now we'll apply the power rule of differentiation to each term in f'(u) to find the second derivative:
    
    The derivative of $88u^3$ is $388u^(3-1) = 264u^2$.
    
    The derivative of $9u^2$ is $29u^(2-1) = 18u$.
    
    The derivative of 92u is 92, because the derivative of u is 1.
    
    The derivative of a constant (in this case -151246) is 0.
    
    Putting these together, the second derivative of f(u), which we can denote as $f''(u)$, is:
    
    $f''(u) = 264u^2 + 18u + 92.$
    
    So, the second derivative of $22u^4 + 3u^3 + 46u^2 - 151246u$ is $264u^2 + 18u + 92$.
    }
    & 
    \applygreen{
    Sure, I can help you with that.

    The first derivative of the function is $88u**3 + 9u2 + 92u - 151246$. The second derivative is the derivative of the first derivative, which is $264u2 + 18*u + 92$.
    }
    &
    264*u**2 + 18*u + 92

    \\ \hline  
    12. Find the common denominator of -3/16 and (-4)/(1056/(-2761)) - (-1 - -8). Let's think step by step.
    & 
    \applyred{
    Sure! Let's break it down step by step.

    Step 1: Simplify the expression (-4)/(1056/(-2761))
    To divide by a fraction, we can multiply by its reciprocal. So, we can rewrite the expression as:
    
    (-4) * (-2761/1056)
    
    Step 2: Multiply the numerators and denominators
    
    (-4) * (-2761) = 11044
    
    1056 * 1 = 1056
    
    So, the expression becomes:
    
    11044/1056
    
    Step 3: Simplify the expression (-1 - -8)
    The double negative cancels out, so we have:
    
    (-1) - (-8) = (-1) + 8 = 7
    
    Step 4: Now, we have the following expression:
    
    -3/16 + (11044/1056) - 7
    
    Step 5: Find the common denominator for the fractions
    The denominators are 16 and 1056. To find the common denominator, we need to find the least common multiple (LCM) of 16 and 1056.
    
    Prime factorize 16: 16 = $2^4$
    
    Prime factorize 1056: 1056 = $2^4 * 3 * 11$
    
    LCM(16, 1056) = $2^4 * 3 * 11 = 528$
    
    Step 6: Rewrite the fractions with the common denominator
    
    Multiply the numerator and denominator of -3/16 by 33 (to get the denominator 528) and multiply the numerator and denominator of 11044/1056 by 1 (to keep the denominator 1056).
    
    The expression becomes:
    
    $(-3 * 33)/(16 * 33) + (11044 * 1)/(1056 * 1) - 7/1$
    
    Simplifying: $-99/528 + 11044/1056 - 7$
    
    Step 7: Combine the fractions
    To add fractions with the same denominator, we add the numerators and keep the common denominator.
    
    $(-99 + 11044)/528 - 7$
    
    Simplifying the numerator: $10945/528 - 7$
    
    Step 8: Subtract the whole number
    To subtract a whole number, we can convert it into a fraction with the common denominator.
    
    $10945/528 - 7/1 = 10945/528 - 7/1 * (528/528) = 10945/528 - 3696/528$
    
    Now we have the same denominator: $(10945 - 3696)/528$
    
    Simplifying the numerator: $7249/528$
    
    Therefore, the common denominator of -3/16 and $(-4)/(1056/(-2761)) - (-1 - -8)$ is 528, and the simplified expression is $7249/528$.
    }
    
    &
    \applyred{
    Sure, let's take this step by step.

    Firstly, let's clarify the second fraction:
    
    (-4)/(1056/(-2761)) - (-1 - -8)
    
    \begin{enumerate}
        \item
        Simplify within the parentheses. Remember that subtracting a negative number is the same as adding a positive number:
    
        $(-4)/(1056/(-2761)) - (-1 + 8)$
        
        $(-4)/(1056/(-2761)) - 7$
        
        \item
        Now, let's simplify the complex fraction $(-4)/(1056/(-2761))$:
    
        $(-4) * (-2761/1056)$
        $= 11044/1056$
        
        The fraction simplifies further to:
    
        $= 261/25$
    
        So, now the expression is:
    
        $261/25 - 7$
    
        \item 
        In order to subtract 7 from $261/25$, we have to convert 7 into a fraction with the same denominator. We do this by multiplying 7 by $25/25$:

        $7* (25/25) = 175/25$

        \item
        Now, we can subtract the two fractions:

        \textit{261/25 - 175/25}
        
        \textit{= 86/25}

        Now we have two fractions: -3/16 and 86/25.
        
        The common denominator of these two fractions would be the least common multiple (LCM) of 16 and 25. Since these two numbers are relatively prime (they have no common factors other than 1), their LCM is simply their product, which is 16*25 = 400.
        
        So, the common denominator of -3/16 and 86/25 is 400.
    \end{enumerate}
    }
    & 
    \applyred{
    Sure, let's do this step by step.

    Step 1: Simplify the fractions
    
        -3/16 is already in its simplest form.
        
        (-4)/(1056/(-2761)) can be simplified as follows:
        
            - 1056/(-2761) = -2/7
            
            - So, the entire fraction becomes (-4)/(-2/7) = 28/2
            
        -1 - -8 = 7
    
    Step 2: Find the least common multiple of the denominators
    
    The denominators are 16, 2, and 7. The least common multiple of these denominators is 112.
    
    Step 3: Express each fraction with a denominator of 112
    
        -3/16 = -18/112
        
        28/2 = 140/112
        
        7/1 = 7/1
    
    Therefore, the common denominator of -3/16 and (-4)/(1056/(-2761)) - (-1 - -8) is 112.
    }
    &
    48

    \\ \hline  
    13. Convert 275.0951km to micrometers. Let's think step by step.
    & 
    \applygreen{
    Sure, let's break it down step by step.

    First, let's convert kilometers to meters. There are 1,000 meters in 1 kilometer, so we can multiply 275.0951 km by 1,000 to get the distance in meters.

    275.0951 km * 1,000 = 275,095.1 meters

    Next, let's convert meters to micrometers. There are 1,000,000 micrometers in 1 meter, so we can multiply 275,095.1 meters by 1,000,000 to get the distance in micrometers.

    275,095.1 meters * 1,000,000 = 275,095,100,000 micrometers

    Therefore, 275.0951 kilometers is equal to 275,095,100,000 micrometers.
    }
    &
    \applygreen{
    Absolutely, let's break it down.

    \begin{enumerate}
        \item 
        We know that 1 kilometer (km) equals 1,000,000 $(10^6)$ micrometers (um). This is because "micro" denotes a factor of one millionth $(10^-6)$.
        \item 
        Therefore, to convert kilometers to micrometers, you multiply the number of kilometers by 1,000,000.
    \end{enumerate}
    
    Let's do the math:
    
    275.0951 km * 1,000,000 um/km = 275,095,100,000 um
    
    So, 275.0951 kilometers is equivalent to 275,095,100,000 micrometers.
    }
    & 
    \applyred{
    Sure, let's convert 275.0951 km to micrometers step by step.

    1 kilometer is equal to 1,000,000 micrometers.
    So, 275.0951 km is equal to 275.0951 * 1,000,000 micrometers.
    Performing the multiplication, we get 275,095,100 micrometers.

    Therefore, 275.0951 km is equal to 275,095,100 micrometers.
    }
    &
    275095100000

    \\ \hline  
    14. Let d be (-7)/(315/(-6))*-3. Which is the closest to d?  (a) -3  (b) 1  (c) 6. Let's think step by step.
    & 
    \applyred{
    Sure, let's break down the expression step by step to find the value of d.

    Given expression: d = (-7) / (315 / (-6)) * (-3)
    
    Step 1: Simplify the expression within the parentheses.
    
    Inside the parentheses, we have (315 / (-6)). Dividing 315 by -6 gives us -52.5.
    
    So the expression becomes: d = (-7) / (-52.5) * (-3)
    
    Step 2: Simplify the expression further.
    
    Multiplying -7 by -3 gives us 21.
    
    So the expression becomes: d = 21 / (-52.5)
    
    Step 3: Divide 21 by -52.5.
    
    Dividing 21 by -52.5 gives us approximately -0.4.
    
    So d is closest to -0.4.
    
    Among the given options, the closest value to -0.4 is (a) -3.
    
    Therefore, the closest value to d is -3.
    }
    &
    \applygreen{
    Sure, let's break this down:

    The expression given is d = (-7)/(315/(-6))*-3
    
    First, let's simplify the denominator 315/(-6):
    
    315 divided by -6 equals -52.5
    
    So, the new form of the expression is d = (-7)/(-52.5)*-3
    
    Now, let's simplify (-7)/(-52.5):
    
    -7 divided by -52.5 equals approximately 0.133333
    
    So, the new form of the expression is d = 0.133333*-3
    
    Finally, let's multiply 0.133333 by -3:
    
    0.133333 times -3 equals approximately -0.4
    
    So, d is approximately -0.4
    
    Looking at your choices (a) -3, (b) 1, and (c) 6, the closest to -0.4 is (b) 1.
    }
    & 
    \applygreen{
    Sure, let's break this down step by step.

    Step 1: Evaluate the expression inside the parentheses first.
    
    (-7)/(315/(-6))
    
    = (-7) * (-6) / 315
    
    = 42 / 315
    
    Step 2: Simplify the expression.
    
    42 / 315 = 0.134
    
    Step 3: Multiply the simplified expression by -3.
    
    0.134 * -3 = -0.402
    
    Step 4: Compare the answer to the choices.
    
    -0.402 is closest to (b) 1.
    
    Therefore, the answer is (b).
    }
    &
    b

    \\ \hline  
    15. What is the u'th term of -2153, -4274, -6395, -8516, -10637, -12758? Let's think step by step.
    & 
    \applyred{
    To find the pattern and determine the u'th term of the given sequence, let's analyze the differences between consecutive terms.

    The given sequence is:
    -2153, -4274, -6395, -8516, -10637, -12758
    
    To find the difference between consecutive terms, we subtract the previous term from the current term:
    
    $-4274 - (-2153) = -2121$
    
    $-6395 - (-4274) = -2121$
    
    $-8516 - (-6395) = -2121$
    
    $-10637 - (-8516) = -2121$
    
    $-12758 - (-10637) = -2121$
    
    We can see that the differences between consecutive terms are all the same, -2121. This suggests that the sequence follows an arithmetic progression with a common difference of -2121.
    
    Now, let's find the formula for the u'th term of an arithmetic sequence:
    
    nth term = first term + (n - 1) * common difference
    
    In this case, the first term is -2153, and the common difference is -2121.
    
    Let's substitute these values into the formula:
    
    u'th term = -2153 + (u - 1) * (-2121)
    
    Simplifying this equation, we have:
    
    u'th term = -2153 - 2121u + 2121
    
    Combining like terms, we get:
    
    u'th term = -2121u - 27
    
    Therefore, the u'th term of the given sequence is -2121u - 27.

    }
    &
    \applygreen{
    The series you've given is an arithmetic sequence. An arithmetic sequence is a sequence of numbers in which the difference of any two successive members is a constant.

    Let's calculate the common difference:
    
    -4274 - (-2153) = -2121
    
    -6395 - (-4274) = -2121
    
    -8516 - (-6395) = -2121
    
    -10637 - (-8516) = -2121
    
    -12758 - (-10637) = -2121
    
    As you can see, the common difference (d) is -2121.
    
    In an arithmetic sequence, the u'th term can be found using the formula:
    
    a\_u = a + (u-1) * d
    
    where
    a\_u is the u'th term,
    
    a is the first term, and
    
    d is the common difference.
    
    For this sequence, a = -2153 and d = -2121.
    
    So, the u'th term of the sequence is:
    
    a\_u = -2153 + (u-1) * -2121
    
    Simplify this and we get:
    
    a\_u = -2153 - 2121u + 2121
    
    a\_u = -32 - 2121u
    
    So, the u'th term of the sequence is:
    
    a\_u = -32 - 2121u
    }
    & 
    \applyred{
    Sure, let's think step by step.

    The first term is -2153. The common difference is 2121, which means that each term is 2121 greater than the previous term. So, the second term is -2153 + 2121 = -32. The third term is -32 + 2121 = 2089, and so on.

    We can write a formula to calculate the $u'th$ term of the sequence:

    $u_th_term$ = $-2153 + (u - 1) * 2121$

    }
    &
    -2121*u - 32

    \\ \hline  
   16. Solve -849 = -73*l - 119 for l. Let's think step by step.
    & 
    \applygreen{Sure! Let's solve the equation step by step.

    Given equation: -849 = -73 * l - 119
    
    Step 1: Remove the constant term on the right side by adding 119 to both sides of the equation:
    
    -849 + 119 = -73 * l
    
    Simplifying the left side: -730 = -73 * l
    
    Step 2: Divide both sides of the equation by -73 to isolate the variable l:
    
    (-730) / (-73) = l
    
    Simplifying: 10 = l
    
    So the solution to the equation is l = 10.}
    &
    \applygreen{
    Sure, let's solve this equation step by step.

    The equation is: -849 = -73*l - 119.
    
    Step 1: Isolate the term with the variable 'l'. To do this, we add 119 to both sides of the equation:
    
    -849 + 119 = -73*l - 119 + 119.
    
    This simplifies to:
    
    -730 = -73*l.
    
    Step 2: Now, we solve for 'l'. To do this, we divide both sides of the equation by -73:
    
    -730 / -73 = l.
    
    This simplifies to:
    
    l = 10.
    }
    & 
    \applygreen{
    Sure, let's solve this problem step by step.

    Step 1: Add 119 to both sides of the equation.
    
    -849 = -73*l - 119
    
    -849 + 119 = -73*l - 119 + 119
    
    -730 = -73*l
    
    Step 2: Divide both sides of the equation by -73.
    
    -730 = -73*l
    
    -730 / -73 = -73*l / -73
    
    10 = l
    
    Step 3: The answer is l = 10.
    
    Therefore, the value of l that satisfies the equation -849 = -73*l - 119 is l = 10.
    }
    &
    10

    \\ \hline  
    17. What is the closest to 0.6 in -2, 3, 0.082, 55? Let's think step by step.
    & 
    \applygreen{
    To find the number closest to 0.6 among the given numbers (-2, 3, 0.082, 55), we can follow these steps:

    Step 1: Calculate the absolute difference between each number and 0.6.
    
    - For -2: |(-2) - 0.6| = 2.6
    
    - For 3: |3 - 0.6| = 2.4
    
    - For 0.082: |0.082 - 0.6| = 0.518
    
    - For 55: |55 - 0.6| = 54.4
    
    Step 2: Identify the number with the smallest absolute difference. In this case, 0.082 has the smallest absolute difference, which is 0.518.
    
    Therefore, the number closest to 0.6 among the given options is 0.082.
    }
    &
    \applygreen{
    Sure, we can think about this step by step. The task is to find which number in the list is closest to 0.6.

    Step 1: Identify the target value.
    
    The target value is 0.6.
    
    Step 2: List the given numbers.
    
    The given numbers are -2, 3, 0.082, and 55.
    
    Step 3: Calculate the absolute difference between each number and the target value.
    
    Absolute difference between -2 and 0.6: |-2 - 0.6| = |-2.6| = 2.6
    
    Absolute difference between 3 and 0.6: |3 - 0.6| = |2.4| = 2.4
    
    Absolute difference between 0.082 and 0.6: |0.082 - 0.6| = |-0.518| = 0.518
    
    Absolute difference between 55 and 0.6: |55 - 0.6| = |54.4| = 54.4
    
    Step 4: Identify the smallest difference.
    
    The smallest difference is between 0.082 and 0.6, which is 0.518.
    
    So the number closest to 0.6 in the list -2, 3, 0.082, 55 is 0.082.
    }
    & 
    \applygreen{
    Sure, let's think step by step.

    Step 1: Find the absolute difference between each number and 0.6.

    Step 2: The number with the smallest absolute difference is the closest to 0.6.

    In this case, the number with the smallest absolute difference is 0.082, so 0.082 is the closest to 0.6.
    }
    &
    0.082

    \\ \hline  
    18. Collect the terms in -16*l + 24*l - 281*l. Let's think step by step.
    & 
    \applygreen{Sure! Let's collect the terms in the expression -16*l + 24*l - 281*l step by step.

    First, we have -16*l. This term has a coefficient of -16 and the variable is l.
    
    Next, we have 24*l. This term has a coefficient of 24 and the variable is also l.
    
    Finally, we have -281*l. This term has a coefficient of -281 and the variable is once again l.
    
    To collect the terms, we add or subtract the coefficients of the terms with the same variable.
    
    The terms with the variable l are:
    
    -16*l + 24*l - 281*l
    
    To collect these terms, we add their coefficients:
    
    (-16 + 24 - 281)*l
    
    Simplifying the coefficients, we get:
    
    (-273)*l
    
    Therefore, the collected terms in the expression -16*l + 24*l - 281*l are -273*l.}
    &
    \applygreen{
    Sure, I'd be happy to walk you through this step by step.

    First, let's identify the terms:
    
    -16l, 24l, and -281*l are all terms that include the variable 'l'.
    
    Next, we need to add (or subtract) these terms together.
    
    Because each term has the same variable 'l', we can directly add and subtract their coefficients.
    
    The coefficients of the terms are -16, 24, and -281 respectively.
    
    Let's add these coefficients:
    
    -16 + 24 - 281 = -273
    
    So, when you collect or combine these terms, you get:
    
    -273*l.
    }
    & 
    \applyred{
    Sure, let's collect the terms in -16l + 24l - 281*l step by step.

    The first step is to find the greatest common factor of the three terms. The factors of -16 are -1, 1, 2, 4, 8, and 16. The factors of 24 are 1, 2, 3, 4, 6, 8, 12, and 24. The factors of -281 are -1, 1, 3, 9, 97, and 281. The greatest common factor of all three terms is 1.
    The second step is to divide each term by the greatest common factor. This gives us:

    $ -16*l / 1 = -16*l$

    $24*l / 1 = 24*l$
    
    $-281*l / 1 = -281*l$
    
    The third step is to combine the terms. Since the terms have the same variable, we simply add the coefficients. This gives us:
    
    $-16*l + 24*l - 281*l = (-16 + 24 - 281)*l = -281*l$
    
    Therefore, the expression -16l + 24l - 281l is equal to $-281l$ after collecting the terms.
    }
    &
    -273*l
    \\ \hline  
    19. What is the third derivative of 56396*m**3 + 889144*m**2 wrt m? Let's think step by step.
    & 
    \applyred{
    To find the third derivative of the expression \(56396m^3 + 889144m^2\) with respect to \(m\), we need to take the derivative three times. Let's go through each step:

    Step 1: Find the first derivative:
    \(\frac{d}{dm}(56396m^3 + 889144m^2)\)
    
    To differentiate \(56396m^3\), we use the power rule. The power rule states that if \(f(m) = am^n\), then \(\frac{d}{dm}(am^n) = anm^{n-1}\).
    
    \(\frac{d}{dm}(56396m^3) = 3 \cdot 56396 \cdot m^{3-1} = 168,588m^2\)
    
    Similarly, differentiating \(889144m^2\) gives:
    \(\frac{d}{dm}(889144m^2) = 2 \cdot 889144 \cdot m^{2-1} = 1,778,288m\)
    
    Step 2: Find the second derivative:
    \(\frac{d^2}{dm^2}(56396m^3 + 889144m^2)\)
    
    To find the second derivative, we differentiate the first derivative with respect to \(m\).
    
    \(\frac{d}{dm}(168,588m^2) = 2 \cdot 168,588 \cdot m^{2-1} = 337,176m\)
    
    Differentiating the second term, \(1,778,288m\), gives:
    \(\frac{d}{dm}(1,778,288m) = 1,778,288\)
    
    Step 3: Find the third derivative:
    \(\frac{d^3}{dm^3}(56396m^3 + 889144m^2)\)
    
    To find the third derivative, we differentiate the second derivative with respect to \(m\).
    
    \(\frac{d}{dm}(337,176m) = 337,176\)
    
    Differentiating the constant term, \(1,778,288\), gives:
    \(\frac{d}{dm}(1,778,288) = 0\)
    
    Therefore, the third derivative of the expression \(56396m^3 + 889144m^2\) with respect to \(m\) is simply the derivative of the second derivative, which is a constant:
    
    \(\frac{d^3}{dm^3}(56396m^3 + 889144m^2) = 337,176\)

    }
    &
    \applygreen{
    Sure, let's take it step by step.

    The given function is f(m) = $56396m^3 + 889144m^2$. We want to find the third derivative with respect to m, which is denoted as $f'''(m)$ or $d^3f/dm^3$.
    
    Step 1: First Derivative
    
    The first derivative of a function is found by applying the power rule, which states that the derivative of $x^n$ with respect to x is $n*x^(n-1)$. Therefore, the first derivative of f(m) is:
    
    $f'(m) = 356396m^2 + 2889144m$
    
    Simplify that to get:
    
    $f'(m) = 169188m^2 + 1778288m$
    
    Step 2: Second Derivative
    
    Now let's find the second derivative, which is the derivative of the first derivative:
    
    $f''(m) = 2169188m + 1778288$
    
    Simplify that to get:
    
    $f''(m) = 338376*m + 1778288$
    
    Step 3: Third Derivative
    
    Finally, let's find the third derivative, which is the derivative of the second derivative:
    
    $f'''(m) = 338376$
    
    Simplify that to get:
    
    $f'''(m) = 338376$
    
    So, the third derivative of the function $f(m) = 56396m^3 + 889144m^2$ with respect to m is 338376.
    }
    & 
    \applyred{
    Sure, let's think step by step. The third derivative of a function is the derivative of its second derivative. In other words, it is the rate of change of the rate of change of the function.

    The first derivative of 56396m**3 + 889144m**2 is 165624 * m ** 2 + 1778288 * m. The second derivative is 331248 * m + 1778288. Therefore, the third derivative is 331248.
    }
    &
    338376 
    \\ \hline  
    20. How many minutes are there between 9:02 AM and 6:14 PM? Let's think step by step.
    & 
    \applygreen{
    Sure! Let's calculate the number of minutes between 9:02 AM and 6:14 PM step by step.

    Step 1: Determine the number of minutes from 9:02 AM to 12:00 PM.
    
    There are 60 minutes in an hour, and from 9:02 AM to 10:00 AM, there are 58 minutes. From 10:00 AM to 12:00 PM, there are 120 minutes.
    
    So, the total minutes from 9:02 AM to 12:00 PM is 58 + 120 = 178 minutes.
    
    Step 2: Determine the number of minutes from 12:00 PM to 6:00 PM.
    
    From 12:00 PM to 1:00 PM, there are 60 minutes. From 1:00 PM to 6:00 PM, there are 5 hours, and since each hour has 60 minutes, there are 5 * 60 = 300 minutes.
    
    So, the total minutes from 12:00 PM to 6:00 PM is 60 + 300 = 360 minutes.
    
    Step 3: Determine the number of minutes from 6:00 PM to 6:14 PM.
    
    From 6:00 PM to 6:10 PM, there are 10 minutes. From 6:10 PM to 6:14 PM, there are 4 minutes.
    
    So, the total minutes from 6:00 PM to 6:14 PM is 10 + 4 = 14 minutes.
    
    Step 4: Calculate the total number of minutes.
    
    To get the total number of minutes, we need to add up the minutes from all three steps:
    
    178 minutes + 360 minutes + 14 minutes = 552 minutes.
    
    Therefore, there are 552 minutes between 9:02 AM and 6:14 PM.
    }
    &
    \applyred{
    Absolutely, let's break this down step by step.
    
    \begin{enumerate}
        \item 
        First, let's calculate the minutes from 9:02 AM to 10:00 AM. There are 60 minutes in an hour, so there are 60 - 2 = 58 minutes.
        \item 
        Next, let's calculate the number of full hours between 10:00 AM and 6:00 PM. There are 8 hours in total, which is 8 * 60 = 480 minutes.
        \item 
        Lastly, let's add the minutes from 6:00 PM to 6:14 PM, which is 14 minutes.
        \item 
        To get the total minutes between 9:02 AM and 6:14 PM, we sum up the minutes we calculated in the previous steps:

        58 (from step 1) + 480 (from step 2) + 14 (from step 3) = 552 minutes.
    \end{enumerate}

    So, there are 552 minutes between 9:02 AM and 6:14 PM.
    }
    & 
    \applyred{

    Sure, let's think step by step.

    9:02 AM to 12:00 PM: There are 152 minutes in 3 hours and 2 minutes.
    12:00 PM to 6:00 PM: There are 360 minutes in 6 hours.
    6:00 PM to 6:14 PM: There are 14 minutes in 1 hour and 14 minutes.

    Total: So, there are 526 minutes between 9:02 AM and 6:14 PM.
    }
    &
    552 

    \\ \hline  
    21. In base 11, what is 25237a - -63? Let's think step by step.
    & 
    \applyred{
    Sure! Let's solve this step by step in base 11.

    First, let's convert the numbers to base 10 for easier calculation:
    
    $25237a(base 11) = 2*11^4 + 5*11^3 + 2*11^2 + 3*11^1 + 7*11^0 + a*11^-1$
    
    = 2*14641 + 5*1331 + 2*121 + 3*11 + 7*1 + a*(1/11)
    
    Similarly,
    $-63(base 11) = -6*11^1 - 3*11^0$
    
                  = -6*11 - 3*1
                  
                  = -69
    
    Now let's perform the subtraction:
    
    25237a(base 11) - (-63(base 11))
    
    = (2*14641 + 5*1331 + 2*121 + 3*11 + 7*1 + a*(1/11)) - (-69)
    
    = 29282 + 6655 + 242 + 33 + 7 + a*(1/11) + 69
    
    = 36088 + a*(1/11)
    
    So the result, in base 10, is 36088 + a*(1/11).
    
    Please note that the letter 'a' represents a digit in base 11, so the final result will depend on the specific value of 'a'.
    }
    &
    \applyred{
    Sure, I can help with that. In base 11, the digits range from 0 to 10, with 'A' used to represent 10.

    Here's a step-by-step process for the subtraction.
    
    First, let's convert the number '-63' from base 11 to base 10:
    
    $-611^1 + 311^0 = -66 + 3 = -63$ in base 10
    
    The negative sign indicates that it's a negative number.
    
    Now, let's convert the number '25237A' from base 11 to base 10:
    
    $211^5 + 511^4 + 211^3 + 311^2 + 711^1 + 1011^0 = 161051 + 60505 + 2662 + 363 + 77 + 10 = 223668$ in base 10
    
    Now, subtract the base 10 equivalent of '-63' from the base 10 equivalent of '25237A':
    
    223668 - (-63) = 223731 in base 10
    
    Now, we need to convert this back into base 11. Let's do that:
    
    $223731 / 11^5 = 1$ remainder 62716
    
    $62716 / 11^4 = 2$ remainder 7696
    
    $7696 / 11^3 = 6$ remainder 132
    
    $132 / 11^2 = 1$ remainder 1
    
    $1 / 11^1 = 0$ remainder 1
    
    $1 / 11^0 = 1$
    
    So, 223731 in base 10 is '1261011' in base 11.
    
    Therefore, 25237A - -63 in base 11 is equal to 1261011 in base 11.
    }
    & 
    \applyred{

    Sure, let's solve this step by step.

    Step 1: Convert the numbers to base 11
    
        $25237a = 2 * 11^5 + 5 * 11^4 + 2 * 11^3 + 3 * 11^2 + 7 * 11^1 + a * 11^0 = 26,737a$
        
        $-63 = -1 * 11^2 + 3 * 11^1 + 6 * 11^0 = -136$
    
    Step 2: Perform the subtraction
    
        $26,737a - (-136) = 26,737a + 136 = 26,873a$
    
    Step 3: Convert the result back to base 10
    
        $26,873a = 2 * 11^4 + 6 * 11^3 + 8 * 11^2 + 7 * 11^1 + 3 * 11^0 = 288,733$
    
    Therefore, 25237a - -63 in base 11 is 288,733.
    }
    &
    252432 

    \\ \hline  
    22. (2/20*3/ (12/32)) / ((-2082)/(-1041)). Let's think step by step.
    & 
    \applygreen{
     Sure, let's break down the expression step by step:

    \begin{enumerate}
        \item 
        Start with the numerator: 2/20 * 3/(12/32)

        \begin{itemize}
            \item 
            Inside the second set of parentheses, we have 3 divided by (12 divided by 32). To simplify this, we can multiply the denominator by its reciprocal: 3 * (32/12) = 8.
            \item 
            Inside the second set of parentheses, we have 3 divided by (12 divided by 32). To simplify this, we can multiply the denominator by its reciprocal: 3 * (32/12) = 8.
            \item 
            Now we have 1/10 * 8, which is 8/10 or 4/5.
        \end{itemize}
        
        \item 
        Move to the denominator: (-2082)/(-1041)

         \begin{itemize}
            \item 
            Dividing a negative number by another negative number results in a positive quotient, so -2082 divided by -1041 is equal to 2.
        \end{itemize}
        
        \item 
        Now we have (4/5) / 2.
         \begin{itemize}
            \item 
            Dividing a fraction by a whole number is the same as multiplying the fraction by the reciprocal of the whole number. Therefore, (4/5) / 2 can be written as (4/5) * (1/2).
            \item 
            Multiplying the numerators gives us 4, and multiplying the denominators gives us 5 * 2 = 10.
            \item 
            Therefore, (4/5) / 2 is equal to 4/10 or 2/5.
        \end{itemize}
        
    \end{enumerate}
    So the final result of the expression (2/20*3/(12/32))/((-2082)/(-1041)) is 2/5.
    }
    &
    \applyred{
    Sure, let's break down the expression step by step:

    Step 1: Simplify the numerator.
    The numerator is (2/20 * 3/(12/32)).
    First, simplify 2/20 to 1/10.
    Next, simplify 3/(12/32) to 3*(32/12).
    
    Now we have (1/10 * 3*(32/12)) in the numerator.
    
    Step 2: Simplify the denominator. The denominator is (-2082)/(-1041). 
    Dividing a negative number by a negative number gives a positive result.
    
    So, (-2082)/(-1041) simplifies to 2082/1041.
    
    Now our expression becomes: (1/10 * 3*(32/12))/(2082/1041)
    
    Step 3: Simplify the numerator further.
    To simplify the numerator, let's calculate each part separately.
    1/10 * 3 = 3/10.
    
    32/12 can be simplified by dividing both numerator and denominator by 4.
    
    32/12 simplifies to 8/3.
    
    Now our expression becomes:
    (3/10 * 8/3)/(2082/1041)
    
    Step 4: Simplify the denominator further.
    To simplify the denominator, we invert the fraction and multiply.
    So, (2082/1041) becomes (1041/2082).
    
    Now our expression becomes: (3/10 * 8/3)/(1041/2082)
    
    Step 5: Multiply the fractions.
    When multiplying fractions, we multiply the numerators together and the denominators together.
    
    So, (3/10 * 8/3) = (3 * 8)/(10 * 3) = 24/30.
    
    24/30 can be simplified by dividing both numerator and denominator by their greatest common divisor, which is 6.
    24/30 simplifies to 4/5.
    
    Now our expression becomes: (4/5)/(1041/2082)
    
    Step 6: Divide the fractions. To divide fractions, we multiply the first fraction by the reciprocal of the second fraction.
    So, (4/5)/(1041/2082) becomes (4/5) * (2082/1041).
    
    Step 7: Simplify the fraction.
    To simplify the fraction, we can cancel out common factors between the numerators and denominators. In this case, we can divide both numerator and denominator by their greatest common divisor, which is 3.
    
    (4/5) * (2082/1041) simplifies to (4 * 2082)/(5 * 1041).
    This further simplifies to 8328/5205.
    
    Final answer: The expression (2/20 * 3/(12/32))/((-2082)/(-1041)) simplifies to 8328/5205.
    }
    & 
    \applygreen{
    Sure, let's break down the expression step by step.

    Step 1: Simplify the numerator
    
    The numerator is 2 / 20 * 3. We can simplify 2 / 20 to 1/10, so the numerator becomes 1/10 * 3 = 3/10.
    
    Step 2: Simplify the denominator
    
    The denominator is 12 / 32. We can simplify 12 / 32 to 3/8, so the denominator becomes 3/8.
    
    Step 3: Divide the numerator and denominator
    
    Now, we can divide the numerator and denominator. 3/10 divided by 3/8 is equal to 0.4.
    
    Step 4: Simplify the expression
    
    The expression is 0.4. We can leave it as 0.4 or we can express it as 40
    
    Putting it all together
    
    The expression (2/20*3/ (12/32)) / ((-2082)/(-1041)) evaluates to 0.4 or 40\%.
    }
    &
    2/5 

    \\ \hline  
    23. What is 22223690.876 + 2.056? Let's think step by step.
    & 
    \applyred{
    Sure! Let's break it down step by step.

    Step 1: Addition of the whole numbers.
    
    22,223,690 + 2 = 22,223,692
    
    Step 2: Addition of the decimal numbers.
    
    0.876 + 2.056 = 2.932
    
    Step 3: Combine the results from Step 1 and Step 2.
    
    22,223,692 + 2.932 = 22,223,694.932
    
    Therefore, 22,223,690.876 + 2.056 = 22,223,694.932.
    }
    &
    \applygreen{
    Sure, let's break down the addition operation into smaller steps:

    \begin{enumerate}
        \item 
        Align the numbers by the decimal point:

        $22223690.876$
        
        $2.056$
        
        \item 
        Starting from the rightmost digit (smallest decimal place), add the numbers column by column:
        
        First, add the digits in the thousandths place (0.001s place):

        $6 + 6 = 12$

        Since 12 is a two-digit number, you keep the 2 and carry the 1 over to the hundredths place.
        
        \item 
        Next, add the digits in the hundredths place (0.01s place):

        \textit{7 (from 0.876) + 5 (from 0.056) + 1 (carried over from the previous step) = 13}

        Again, since 13 is a two-digit number, you keep the 3 and carry the 1 over to the tenths place.
        
        \item 
        Then, add the digits in the tenths place (0.1s place):

        \textit{0 (from 690.876) + 2 (from 2.056) = 2}
        
        \item 
        Then, add the digits in the ones place:

        \textit{0 (from 690.876) + 2 (from 2.056) = 2}
        
        \item 
        Finally, you have the rest of the digits which are all in the integer part of the first number (22223690). Since there's nothing to add from the second number, you just bring these digits down as they are.
    \end{enumerate}
    
    So, 22223690.876 + 2.056 = 22223692.932.
    }
    & 
    \applygreen{
    Sure, let's think step by step.

    - The first number, 22223690.876, has 10 digits to the right of the decimal point.
    
    - The second number, 2.056, has 3 digits to the right of the decimal point.
    
    - The sum of the two numbers will have 10 + 3 = 13 digits to the right of the decimal point.
    
    - The sum of the two numbers is 22223692.932.
    }
    &
    22223692.932 

    \\ \hline  
    24. Let j(g) = -g**3 + 10*g**2 + 46*g - 356. Let i be j(6). What is the highest common divisor of 25408 and i? Let's think step by step.
    & 
    \applygreen{
     To find the value of i, we substitute g = 6 into the equation j(g):

    $j(6) = -(6^3) + 10*(6^2) + 46*6 - 356$
    
    Now, let's calculate i:
    
    j(6) = -216 + 10*36 + 276 - 356
    
          = -216 + 360 + 276 - 356
          
          = 64
    
    So, i = 64.
    
    To find the highest common divisor (HCD) of 25408 and i (which is 64), we can use the Euclidean algorithm. Here are the steps:
    
    1. Divide 25408 by 64:
    
       25408 = 64 * 396 + 0
    
    Since the remainder is 0, the HCD of 25408 and 64 is 64.
    
    Therefore, the highest common divisor of 25408 and i is 64.
    }
    &
    \applyred{
    Sure, let's go through this step by step.

    Step 1: Evaluate j(6)
    
    We know that j(g) = -g3 + 10*g2 + 46*g - 356. Now we need to plug in g = 6.
    
    j(6) = -(6)3 + 10(6)**2 + 466 - 356
    
    j(6) = -216 + 360 + 276 - 356
    
    j(6) = 664
    
    So i = j(6) = 664.
    
    Step 2: Find the highest common divisor of 25408 and i
    
    To find the highest common divisor (also known as the greatest common divisor or GCD) of 25408 and 664, you could use the Euclidean algorithm, a prime factorization method, or a GCD calculator. I'll use the prime factorization method for this example.
    
    Prime factorization of 664 is:$ 2^3 * 83$.
    
    Prime factorization of 25408 is: $2^5 * 3^3 * 59$.
    
    The GCD is found by multiplying the lowest powers of all common prime factors. In this case, the only common prime factor is 2, and the lowest power is $2^3$.
    
    Therefore, the GCD of 25408 and 664 is $2^3 = 8$. 
    }
    & 
    \applygreen{
    Sure, let's think step by step.

    First, we need to evaluate j(6). This gives us:

    $j(6) = -6**3 + 10*6**2 + 46*6 - 356 = 64$
    }
    &
    64

    \\ \hline  
    25. Collect the terms in -7093*z + 2*z**3 - 32982*z + 47178*z - 7103*z. Let's think step by step.
    & 
    \applyred{
    Sure! Let's collect the terms in the expression -7093*z + 2*z**3 - 32982*z + 47178*z - 7103*z step by step:

    Step 1: Group the terms with the same power of z together.
    
    -7093*z - 32982*z - 7103*z + 2*z**3 + 47178*z
    
    Step 2: Combine the like terms by adding or subtracting their coefficients.
    
    (-7093 - 32982 - 7103 + 47178)*z + 2*z**3
    
    Step 3: Simplify the coefficient of z.
    
    (2*z**3 + 30000*z)
    }
    &
    \applyred{
    Sure, let's break it down step by step.

    The original equation is -7093z + 2z**3 - 32982z + 47178z - 7103*z.
    
    We can first separate the equation based on the power of z:
    
    \begin{enumerate}
        \item 
        Terms with z3: 2*z3
        \item 
        Terms with z: -7093z - 32982z + 47178z - 7103z
    \end{enumerate}

    Now, let's combine like terms:

    \begin{itemize}
        \item 
        There is only one term with z3, so it remains as is: 2*z3
        \item 
        For the terms with z, we add or subtract the coefficients: -7093 - 32982 + 47178 - 7103 = 4000
    \end{itemize}
    So, the simplified equation is:

    2z**3 + 4000z
    }
    & 
    \applyred{
    Sure, let's think step by step.

    First, we need to find the like terms. The like terms in this expression are all the terms that have the same variable raised to the same power. So, the like terms in this expression are:
    
    -7093z
        
    2z**3

    -32982z
        
    47178z
        
    -7103z

    Once we have identified the like terms, we need to combine them. To do this, we add or subtract the coefficients of the like terms. So, the sum of the like terms in this expression is:

    $(-7093 - 32982 + 47178 - 7103 + 2)z = -30872z + 2z^3$

    Therefore, the expression:

    $-7093*z+2*z**3-32982*z+47178*z-7103*z$ 
    
    can be simplified to:

    $-30872z+2z^3$

    Finally, we need to check to make sure that we have simplified the expression correctly. To do this, we can expand the expression and make sure that we get the original expression. So, expanding the expression  $-30872z+2z^3$, we get:

    $-30872z + 2z^3 = -30872z + 2z * z * z = -30872z + 2z^3$
    }
    &
    2*z**3 

    \\ \hline  
    26. What is the biggest value in 1024, -790/31, 2/11? Let's think step by step.
    & 
    \applygreen{
    To find the biggest value among 1024, -790/31, and 2/11, let's compare them step by step.

    Step 1: Compare 1024 and -790/31:
    
    To compare these two values, we need to convert -790/31 to decimal form. Dividing -790 by 31 gives us approximately -25.48. Since 1024 is clearly greater than -25.48, we can conclude that 1024 is the bigger value so far.
    
    Step 2: Compare 1024 and 2/11:
    
    To compare these two values, we need to convert 2/11 to decimal form. Dividing 2 by 11 gives us approximately 0.18. Since 1024 is much greater than 0.18, we can confirm that 1024 is still the bigger value.
    
    Therefore, among the three values, the biggest value is 1024.
    }
    &
    \applygreen{
    Sure, let's break it down:

    \begin{enumerate}
        \item 
        1024: This is a fairly large number and will likely be the largest number out of the three, but let's confirm this by comparing it to the other two numbers.
        \item 
        -790/31: To evaluate this, we need to divide 790 by 31. The result is approximately -25.48. This is a negative number and is therefore smaller than 1024.
        \item 
        2/11: Dividing 2 by 11 gives us approximately 0.1818. This number is positive but still much smaller than 1024.
    \end{enumerate}
    
    So after comparing all three numbers, we can say that 1024 is the largest value among 1024, -790/31, and 2/11.
    }
    & 
    \applygreen{
    Sure, let's think step by step.

    * 1024 is a power of 2, so it is a very large number.
    
    * -790/31 is a negative number, but it is not as small as it might 
    seem. In fact, it is greater than 2/11.
    
    * 2/11 is the smallest number of the three.

    Therefore, the biggest value is 1024.
    }
    &
    1024

    \\ \hline  
    27. What is the fourth biggest value in -2/13, 3, 1, 1.91543, -2/9? Let's think step by step.
    & 
    \applyred{
    To find the fourth biggest value among the given numbers, let's arrange them in ascending order:

    -2/13, -2/9, 1, 1.91543, 3
    
    Step 1: Comparing -2/13 and -2/9:
    
    -2/13 is smaller than -2/9.
    
    Step 2: Comparing -2/9 and 1:
    
    -2/9 is smaller than 1.
    
    Step 3: Comparing 1 and 1.91543:
    
    1 is smaller than 1.91543.
    
    Step 4: Comparing 1.91543 and 3:
    
    1.91543 is smaller than 3.
    
    So, the fourth biggest value among the given numbers is 1.91543.
    }
    &
    \applyred{
    \begin{enumerate}
        \item 
        First, let's convert all the numbers to decimal form, this will make it easier for comparison:

        \begin{itemize}
            \item 
            -2/13 is approximately -0.15385
            \item 
            3 is already in decimal form
            \item 
            1 is already in decimal form
            \item 
            1.91543 is already in decimal form
            \item 
            -2/9 is approximately -0.22222
        \end{itemize}
        
        \item 
        Now, we'll write these numbers in ascending order (from smallest to largest):

        \begin{itemize}
            \item 
            -0.22222
            \item 
            -0.15385
            \item 
            1
            \item 
            1.91543
            \item 
            3
        \end{itemize}

        \item 
        The fourth biggest value (which is the second smallest in ascending order) is 1.
    \end{enumerate}
    }
    & 
    \applyred{
    Sure, let's find the fourth biggest value in -2/13, 3, 1, 1.91543, and -2/9 step by step.

    First, we need to sort the numbers in descending order. This gives us:

    [3, 1.91543, 1, -2/9, -2/13]

    The fourth biggest value is the fourth element in the sorted list. Therefore, the fourth biggest value is 1.
    }
    &
    -2/13

    \\ \hline  
    28. What is the value of (-7)/(-2) + 55629/(-2649)? Let's think step by step.
    & 
    \applygreen{
    To simplify the expression (-7)/(-2) + 55629/(-2649), we'll perform the calculations step by step. 

    Step 1: Simplify (-7)/(-2)
    
    To simplify this division, we need to remember the rule that dividing a negative number by a negative number results in a positive value. Therefore, (-7)/(-2) is equal to 7/2.
    
    Step 2: Simplify 55629/(-2649)
    
    Similarly, dividing a positive number by a negative number will result in a negative value. So, 55629/(-2649) is equal to -21.
    
    Step 3: Add the simplified values
    
    Now, we can add the simplified values obtained in Step 1 and Step 2:
    7/2 + (-21)
    
    To add these fractions, we need a common denominator. The common denominator of 2 and 1 is 2. 
    
    Converting 7/2 to have a denominator of 2, we get:
    
    (7/2) * (2/2) = 14/4
    
    Now we can add the fractions:
    
    14/4 + (-21) = 14/4 - 21/1
    
    To subtract these fractions, we need a common denominator, which is 4 in this case. 
    
    Converting 21/1 to have a denominator of 4:
    
    (21/1) * (4/4) = 84/4
    
    Now we can subtract the fractions:
    
    14/4 - 84/4 = (14 - 84)/4 = -70/4
    
    Step 4: Simplify the result
    To simplify -70/4, we can divide both the numerator and denominator by their greatest common divisor, which is 2:
    
    -70/4 = (-35/2) = -17.5
    
    Therefore, the value of (-7)/(-2) + 55629/(-2649) is -17.5.
    }
    &
    \applygreen{
    Sure, let's break it down step by step.

    Step 1: Solve (-7)/(-2)
    
    When you divide a negative number by another negative number, the result is a positive number because the two negative signs cancel each other out. Therefore, (-7)/(-2) = 7/2 = 3.5
    
    Step 2: Solve 55629/(-2649)
    
    In this case, we are dividing a positive number by a negative number, which will result in a negative number. So, 55629/(-2649) = -21 (approximately, when rounded to the nearest whole number).
    
    Step 3: Add the results
    
    Now you just need to add the two results together: 3.5 + (-21) = -17.5
    
    So, (-7)/(-2) + 55629/(-2649) = -17.5
    }
    & 
    \applygreen{
    Sure, let's break it down step by step.

    Step 1: Evaluate (-7)/(-2)
    
    When we divide a negative number by another negative number, the result is a positive number. So, (-7)/(-2) = 3.5
    
    Step 2: Evaluate 55629/(-2649)
    
    When we divide a positive number by a negative number, the result is a negative number. So, 55629/(-2649) = -21
    
    Step 3: Add 3.5 and -21
    
    3.5 + (-21) = -17.5
    
    Therefore, the value of (-7)/(-2) + 55629/(-2649) is -17.5
    }
    &
    -35/2

    \\ \hline  
    29. Suppose -x + 5*l = -920, -4*x - 822 = -3*l - 4434. Is 78 a factor of x? Let's think step by step.
    & 
    \applygreen{
    To determine if 78 is a factor of x, we need to solve the given equations and find the value of x. Let's solve the equations step by step.

    Equation 1: -x + 5*l = -920
    
    Equation 2: -4*x - 822 = -3*l - 4434
    
    Step 1: Rearrange Equation 1 to isolate x:
    -x = -920 - 5*l
    
    x = 920 + 5*l   (multiply both sides by -1 and change the sign)
    
    Step 2: Substitute the expression for x in Equation 2:
    
    -4*(920 + 5*l) - 822 = -3*l - 4434
    
    Step 3: Simplify Equation 2:
    
    -3680 - 20*l - 822 = -3*l - 4434
    
    -4502 - 20*l = -3*l - 4434
    
    Step 4: Move all terms involving l to one side:
    
    -20*l + 3*l = -4434 + 4502
    
    -17*l = 68
    
    l = -4   (divide both sides by -17)
    
    Step 5: Substitute the value of l back into Equation 1 to find x:
    
    x = 920 + 5*(-4)
    
    x = 920 - 20
    
    x = 900
    
    Now, we have found the value of x as 900. To determine if 78 is a factor of x, we need to check if 78 divides x without leaving a remainder.
    
    Dividing 900 by 78:
    
    900 ÷ 78 = 11 with a remainder of 12
    
    Since 78 does not divide 900 evenly (leaving a remainder of 12), we can conclude that 78 is not a factor of x.
    }
    
    &
    \applygreen{
    Sure, we can solve this system of linear equations to find the values of x and l, then determine if 78 is a factor of x.

    The system of equations are:

    \begin{enumerate}
        \item 
        -x + 5*l = -920
        \item 
        -4x - 822 = -3l - 4434
    \end{enumerate}

    First, we can rewrite these equations to be in a more conventional form:
    
    \begin{enumerate}
        \item 
        x = 5*l + 920
        \item 
        4x = 3l + 3612
    \end{enumerate}

    Then, we can substitute equation 1) into equation 2) to solve for l:
    
    4*(5l + 920) = 3l + 3612
    
    20l + 3680 = 3l + 3612
    
    17*l = -68
    
    l = -68 / 17
    
    l = -4
    
    Then, substitute l = -4 into equation 1) to solve for x:
    
    x = 5*(-4) + 920
    
    x = -20 + 920
    
    x = 900

    Finally, we need to check if 78 is a factor of 900. A number is a factor of another if it divides that number evenly (i.e., the division has no remainder). When we divide 900 by 78, we get approximately 11.54, which is not an integer. So, 78 is not a factor of 900, and therefore not a factor of x in this system of equations.
    }
    & 
    \applygreen{
    Sure, let's think step by step.

    First, let's solve the first equation for x:
    
    -x + 5*l = -920
    
    -x = -920 - 5*l
    
    x = 920 + 5*l
    
    Now, let's solve the second equation for x:
    
    -4*x - 822 = -3*l - 4434
    
    -4*x = -3*l - 4434 + 822
    
    -4*x = -3*l - 3612
    
    x = (-3*l - 3612) / -4
    
    x = 903 - l
    
    We can see that both of these equations result in the same value for x: 920 + 5*l or 903 - l.
    
    Now, let's test whether 78 is a factor of either of these values. 78 is not a factor of 920 or 903, so it is not a factor of x.
    
    Therefore, the answer to the question is False. 78 is not a factor of x.
    }
    &
    False 

    \\ \hline  
    30. Sort 15, 0, 33, 1, -3, 4, 18 in descending order. Let's think step by step.
    & 
    \applyred{
    Sure! Let's sort the numbers step by step in descending order: 33, 18, 15, 4, 1, 0, -3.

    Step 1: Compare 15 and 0. Since 15 is greater than 0, we keep them in the current order: 15, 0.
    
    Step 2: Compare 15 and 33. Since 33 is greater than 15, we swap their positions: 33, 0.
    
    Step 3: Compare 33 and 1. Since 33 is greater than 1, we keep them in the current order: 33, 0, 1.
    
    Step 4: Compare 33 and -3. Since 33 is greater than -3, we keep them in the current order: 33, 0, 1, -3.
    
    Step 5: Compare 33 and 4. Since 33 is greater than 4, we keep them in the current order: 33, 0, 1, -3, 4.
    
    Step 6: Compare 33 and 18. Since 33 is greater than 18, we keep them in the current order: 33, 0, 1, -3, 4, 18.
    
    Now, we have sorted the numbers in descending order up to this point: 33, 0, 1, -3, 4, 18.
    
    Step 7: Compare 33 and 0. Since 33 is greater than 0, we keep them in the current order: 33, 0, 1, -3, 4, 18.
    
    Step 8: Compare 33 and 1. Since 33 is greater than 1, we keep them in the current order: 33, 0, 1, -3, 4, 18.
    
    Step 9: Compare 33 and -3. Since 33 is greater than -3, we keep them in the current order: 33, 0, 1, -3, 4, 18.
    
    Step 10: Compare 33 and 4. Since 33 is greater than 4, we keep them in the current order: 33, 0, 1, -3, 4, 18.
    
    Step 11: Compare 33 and 18. Since 33 is greater than 18, we keep them in the current order: 33, 0, 1, -3, 4, 18.
    
    Step 12: Compare 0 and 1. Since 1 is greater than 0, we swap their positions: 33, 1, 0, -3, 4, 18.
    
    Step 13: Compare 33 and 1. Since 33 is greater than 1, we keep them in the current order: 33, 1, 0, -3, 4, 18.
    
    Step 14: Compare 33 and -3. Since 33 is greater than -3, we keep them in the current order: 33, 1, 0, -3, 4, 18.
    
    Step 15: Compare 33 and 4. Since 33 is greater than 4, we keep them in the current order: 33, 1, 0, -3, 4, 18.
    
    Step 16: Compare 33 and 18. Since 33 is greater than 18, we keep them in the current order: 33, 1, 0, -3, 4, 18.
    
    Step 17: Compare 1 and 0.
    }
    &
    \applygreen{
    Sure, we'll go step by step:
    
    \begin{enumerate}
        \item 
        Identify the highest number: Looking at the numbers, we can see that 33 is the highest number in the list.
        \item 
        Remove the highest number and place it at the beginning of a new list: So, our sorted list starts with 33, and we have these numbers left to sort: 15, 0, 1, -3, 4, 18.
        \item 
        Identify the next highest number: The next highest number is 18.
        \item 
        Remove the next highest number and place it after the first number in the new list: Our sorted list is now 33, 18, and we have these numbers left to sort: 15, 0, 1, -3, 4.
        \item 
        Identify the next highest number: The next highest number is 4.
        \item 
        Remove the next highest number and place it after the previous numbers in the new list: Our sorted list is now 33, 18, 15, 4, and we have these numbers left to sort: 0, 1, -3.
        \item 
        Identify the next highest number: The next highest number is 4.
        \item 
        Remove the next highest number and place it after the previous numbers in the new list: Our sorted list is now 33, 18, 15, 4, and we have these numbers left to sort: 0, 1, -3.
        \item 
        Identify the next highest number: The next highest number is 0.
        \item 
        Remove the next highest number and place it after the previous numbers in the new list: Our sorted list is now 33, 18, 15, 4, 1, 0, and we have one number left to sort: -3.
        \item 
        Add the last number to the list: We add -3 to the end of our list. Our final sorted list in descending order is: 33, 18, 15, 4, 1, 0, -3.
    \end{enumerate}
    }
    & 
    \applygreen{
    Sure, let's sort the numbers 15, 0, 33, 1, -3, 4, 18 in descending order.

    Step 1: Identify the largest number.
    
    The largest number in the list is 33.
    
    Step 2: Compare the largest number to the other numbers in the list.
    
    The numbers 15, 18, and 4 are all smaller than 33. The number 0 is also smaller than 33, but it is negative, so it comes after 33 in descending order. The number -3 is the smallest number in the list, so it comes last.
    
    Step 3: Put the numbers in descending order.
    
    Therefore, the sorted numbers in descending order are 33, 18, 15, 4, 1, 0, and -3.
    }
    &
    33, 18, 15, 4, 1, 0, -3

    \\ \hline
    
    \end{longtable}
\end{center}

\newpage
\begin{center}
    \small
    \begin{longtable}{|p{0.28\linewidth}|p{0.23\linewidth}
                      |p{0.18\linewidth}|p{.15\linewidth}
                      |p{.05\linewidth}|}
    \caption{Commonsense reasoning prompts} 
    \label{tab:commonsense_reasoning}\\
    \hline
    \multicolumn{5}{|c|}{\textbf{Commonsense Reasoning}} \\ \hline
    \textbf{Prompt} &  \textbf{GPT-3.5 Answer} & \textbf{GPT-4 Answer} & \textbf{BARD Answer} & \textbf{Gold} \\ \hline
    \multicolumn{5}{|l|}{\textbf{Dataset: CommonSenseQA }} 
    
    \\ \hline  

    1. Analyse and answer the following question with one of the five option: In a horror movie victims usually trip when the run in order to do what in regards to the killer? A. frightened B. run up stairs C. get away from D. go quickly E. go faster
    & 
    \textcolor{ForestGreen}{The correct answer is option C: get away from. In horror movies, the victims usually trip when they run in order to get away from the killer who is chasing them. Tripping is a common horror movie trope that creates suspense and adds to the fear factor for the audience. The victims may be scared and running for their lives, but their fear and panic can cause them to lose their footing and stumble, which can ultimately lead to their demise.}
    & 
    \textcolor{ForestGreen}{C. get away from}
    & 
    \textcolor{ForestGreen}{The answer is C. get away from.}

    \textcolor{ForestGreen}{In horror movies, victims usually trip when they run in order to give the killer a chance to catch up to them. This is a common trope in horror movies because it creates suspense and tension for the audience. It also makes the killer seem more powerful and unstoppable.}
    &
    C

    \\ \hline   
    2. Analyse and answer the following question with one of the five option: The hikers stopped to have a drink, simply put they what? A. had a party B. were thirsty C. refreshment D. getting drunk E. celebrating
    & 
    \textcolor{ForestGreen}{Option B. were thirsty}
    & 
    \textcolor{ForestGreen}{B. were thirsty}
    & 
    \textcolor{ForestGreen}{The answer must be something that describes why the hikers stopped to have a drink. Of the above choices, only thirsty makes sense. So the answer is B. were thirsty.}

    \textcolor{ForestGreen}{The other choices are not relevant to the question. A party is a social gathering, not something that hikers would do while hiking. Getting drunk and celebrating are both activities that involve drinking alcohol, but the hikers in the question are not drinking alcohol. Refreshment is a general term for something that refreshes, but it does not specifically refer to thirst.}
    &
    B

    \\ \hline  
    3. Analyse and answer the following question with one of the five option: There's one obvious reason to eat vegetables, they're plain what you? A. lose weight B. good for C. bland D. chewing E. fibre
    & 
    \textcolor{red}{Option E. Fibre.}

    \textcolor{red}{Eating vegetables is an excellent way to increase your intake of dietary fiber, which is an essential nutrient that helps to promote good digestive health. Fiber helps to regulate bowel movements, reduce constipation, and prevent the development of certain chronic diseases such as heart disease, diabetes, and some types of cancer. Therefore, one of the most obvious reasons to eat vegetables is to get more fiber into your diet.}
    & 
    \textcolor{ForestGreen}{B. good for}
    & 
    \textcolor{red}{The answer must be something that describes vegetables. Of the above choices, only fibre describes vegetables. So the answer is E.}

    \textcolor{red}{The word "plain" can mean "simple" or "uncomplicated". It can also mean "not flavored" or "bland". However, in this context, it is most likely referring to the fact that vegetables are a good source of fibre. Fibre is a type of carbohydrate that is not digested by the body. It helps to keep the digestive system healthy and can help to reduce the risk of constipation, heart disease, and other health problems.}
    &
    B

    \\ \hline  
    4. Analyse and answer the following question with one of the five option: Where would a corpse be covered by a blanket? A. bath store B. bedroom C. hospital D. flower garden E. michigan
    & 
    \textcolor{red}{The answer is B. bedroom.}
    & 
    \textcolor{ForestGreen}{C. hospital}
    & 
    \textcolor{red}{The answer must be a place where a corpse would be covered by a blanket. Of the above choices, only a bedroom makes sense. So the answer is B. bedroom.}
    &
    C

    \\ \hline  
    5. Analyse and answer the following question with one of the five option: Seeing idea become reality was a dream of hers for a long time, but as the time came to get on stage she had more what? A. build B. anxiety C. celebrate D. very nice E. ocean
    & 
    \textcolor{ForestGreen}{ B. anxiety.}
    & 
    \textcolor{ForestGreen}{B. anxiety}
    & 
    \textcolor{ForestGreen}{The answer must be something that someone would feel when they are about to achieve their dream. Of the above choices, only anxiety makes sense. So the answer is B. anxiety.}

    \textcolor{ForestGreen}{The other choices do not make sense in the context of the sentence. For example, build would not make sense because the sentence is about seeing an idea become reality, not building something. Celebrate would not make sense because the sentence is about the time before the event, not after. Very nice and ocean are just random words that do not fit the context of the sentence.}
    &
    B

    \\ \hline  
    6. Analyse and answer the following question with one of the five option: What event might one buy tickets for seats? A. park B. show C. auditorium D. movies E. rest area
    & 
    \textcolor{ForestGreen}{The event for which one might buy tickets for seats is: B. show.}
    & 
    \textcolor{ForestGreen}{B. show}
    & 
    \textcolor{red}{The answer must be an event where people sit in seats. Of the above choices, only C. auditorium makes sense. So the answer is C.}
    &
    B

    \\ \hline  
    7. Analyse and answer the following question with one of the five option: He had no issue committing perjury, he had a what that he would get away with it? A. confidence B. go to jail C. telling lies D. lying E. manual
    & 
    \textcolor{ForestGreen}{The answer is A. confidence.}
    & 
    \textcolor{ForestGreen}{A. confidence}
    & 
    \textcolor{ForestGreen}{The answer must be something that would give someone confidence that they would get away with perjury. Of the above choices, only confidence makes sense. So the answer is A.}

    \textcolor{ForestGreen}{The other choices do not make sense in the context of the question. For example, "go to jail" is the opposite of what someone would have if they were confident that they would get away with perjury. "Telling lies" and "lying" are simply synonyms for perjury, so they cannot be the answer. And "manual" is not relevant to the question at all.}

    \textcolor{ForestGreen}{Therefore, the only possible answer is A, confidence.}
    &
    A

    \\ \hline  
    8. Analyse and answer the following question with one of the five option: What is a place that is far away from your house and where you could consume beer? A. refrigerator B. friend's house C. keg D. neighbor's house E. kitchen
    & 
    \textcolor{ForestGreen}{ Option B: Friend's house}
    & 
    \textcolor{ForestGreen}{B. friend's house}
    & 
    \textcolor{ForestGreen}{The answer must be a place that is far away from your house and where you could consume beer. Of the above choices, only friend's house is far away from your house and where you could consume beer. So the answer is B.}
    &
    B

    \\ \hline  
    9. Analyse and answer the following question with one of the five option: They were searching for rocks, so they missed the birds overhead as they stared at the what? A. ground B. drawer C. surface of earth D. pizza E. waterfall
    & 
    \textcolor{ForestGreen}{The answer is A. ground.}
    & 
    \textcolor{ForestGreen}{A. ground}
    & 
    \textcolor{red}{The answer must be something that people stare at when they are searching for rocks. Of the above choices, only the surface of the earth makes sense. So the answer is C. surface of earth.}

    \textcolor{red}{The other choices are not possible because:}

    \textcolor{red}{- Ground: People don't stare at the ground when they are searching for rocks. They usually look around on the ground.}

    \textcolor{red}{- Drawer: People don't keep rocks in drawers.}

    \textcolor{red}{- Pizza: People don't stare at pizza when they are searching for rocks.}

    \textcolor{red}{- Waterfall: People don't stare at waterfalls when they are searching for rocks.}
    &
    A

    \\ \hline  
    10. Analyse and answer the following question with one of the five option: Joe owned back taxes as well as what other type of taxes? A. anterior B. front C. main D. front E. current
    & 
    \textcolor{red}{The question seems to contain two conflicting options for the type of taxes, which are "back taxes" and either "anterior" or "front" taxes. Therefore, it is difficult to determine the correct answer without further information or clarification.}

    \textcolor{red}{However, it is worth noting that "back taxes" typically refer to taxes that were due in previous years but were not paid on time, while "current taxes" refer to taxes that are due in the current year. "Main taxes" is not a commonly used term in tax language.}
    
    \textcolor{red}{Without more context or information, it is not possible to provide a definitive answer to the question as written.}
    & 
    \textcolor{ForestGreen}{Joe owed back taxes as well as what other type of taxes? E. current}
    & 
    \textcolor{ForestGreen}{The answer must be a type of tax. Of the above choices, only current is a type of tax. So the answer is E.}

    \textcolor{ForestGreen}{The other choices are not types of taxes. Anterior means "in front of". Front means "the part of something that is facing forward". Main means "most important". Back taxes are taxes that have not been paid on time.}
    &
    E

    \\ \hline  
    11. Analyse and answer the following question with one of the five option: The cat carefully navigated the area, they do everything they can to avoid what? A. get wet B. eat vegetables C. falling D. wool sweater E. sharp claws
    & 
    \textcolor{red}{Based on the given sentence, the cat carefully navigated the area, which suggests that the cat is trying to avoid something. Among the given options, the most logical answer would be C. falling, as it is a common instinct for cats to avoid falling from high places, and their cautious behavior suggests they are trying to avoid any potential danger of falling.}
    & 
    \textcolor{red}{The cat carefully navigated the area, they do everything they can to avoid what?}
    
    \textcolor{red}{C. falling}
    & 
    \textcolor{red}{The answer must be something that a cat would try to avoid. Of the above choices, only falling makes sense. Cats are good at climbing and jumping, but they can still fall if they are not careful. So the answer is C. falling.}

    \textcolor{red}{The other choices are not as likely. Cats do not mind getting wet, and they are not afraid of vegetables. Wool sweaters can be uncomfortable for cats, but they are not dangerous. And sharp claws are a part of a cat's natural body, so they would not try to avoid them.}
    &
    A

    \\ \hline  
    12. Analyse and answer the following question with one of the five option: Danny found that the carpet did not ,match the drapes, which was disappointing, because this place was expensive.  But it was the only place in town that wasn't booked solid for the week and he needed it while he was in town, so he couldn't complain.   Where might this place be? A. brothel B. restaurant C. building D. bowling alley E. at hotel
    & 
    \textcolor{ForestGreen}{The place that Danny needed while he was in town, where he found that the carpet did not match the drapes, was most likely an E. hotel.}
    & 
    \textcolor{ForestGreen}{E. at hotel}
    & 
    \textcolor{ForestGreen}{The answer must be a place that is expensive, has carpets and drapes, and is the only place in town that wasn't booked solid. Of the above choices, only hotels have all of these qualities. So the answer is E.}
    &
    E

    \\ \hline  
    13. Analyse and answer the following question with one of the five option: What do people do when they think too quickly? A. jump to conclusions B. hurry home C. build trust D. pay bills E. sing
    & 
    \textcolor{ForestGreen}{The answer to "What do people do when they think too quickly?" is A. jump to conclusions. When someone thinks too quickly, they may not take the time to consider all the information or perspectives, leading them to make assumptions or come to conclusions without a full understanding of the situation. This can result in hasty decisions that may not be well-informed.}
    & 
    \textcolor{ForestGreen}{A. jump to conclusions}
    & 
    \textcolor{ForestGreen}{The answer must be something that people do when they think too quickly. Of the above choices, only jumping to conclusions is something that people do when they think too quickly. So the answer is A.}
    &
    A

    \\ \hline  
    14. Analyse and answer the following question with one of the five option: The man was going fishing instead of work, what is he seeking? A. food B. relaxation C. killing D. missing morning cartoons E. boredom
    & 
    \textcolor{ForestGreen}{Based on the given information, the man who chose to go fishing instead of going to work is likely seeking option B: relaxation. Fishing is often seen as a leisure activity that allows individuals to unwind and escape from the stresses of daily life. Therefore, it is possible that the man wants to take a break from his work routine and enjoy some peaceful moments in nature.}
    & 
    \textcolor{ForestGreen}{B. relaxation}
    & 
    \textcolor{ForestGreen}{The answer must be something that a person would seek by going fishing instead of work. Of the above choices, only relaxation makes sense. So the answer is B. relaxation.}

    \textcolor{ForestGreen}{The other choices are not as likely to be sought by someone who is going fishing instead of work. Food is not a good choice, as people can get food at work or at home. Killing is not a good choice, as it is not something that people typically do when they are relaxing. Missing morning cartoons is not a good choice, as it is not something that people typically do when they are going fishing. Boredom is not a good choice, as people typically do not go fishing if they are bored.}
    &
    B

    \\ \hline  
    15. Analyse and answer the following question with one of the five option: What is performing a type of? A. singing B. act C. feat D. smile E. acting
    & 
    \textcolor{red}{The correct answer is E. acting.}

    \textcolor{red}{The phrase "performing a type of" suggests that the word being searched for is a verb that describes a type of performance or action. Singing (option A) and smiling (option D) are actions, but they are not typically considered types of performance. A feat (option C) can be an impressive achievement, but it does not necessarily involve performance. Act (option B) is also a verb that describes a type of performance, but it is not the most accurate fit for the question because it does not imply the same level of skill or stage presence as "performing a type of" suggests. Acting (option E), on the other hand, is a specific type of performance that involves portraying a character or story through dramatic interpretation. Therefore, option E is the most fitting answer for this question.}
    & 
    \textcolor{red}{C. feat}
    & 
    \textcolor{red}{The answer must be something that is performed. Of the above choices, only act is something that is performed. So the answer is E. acting.}
    &
    B

    \\ \hline  
    16. Analyse and answer the following question with one of the five option: What was the man encouraged to do after he expressed his anger violently? A. cool off B. punch C. illustrate point D. fight E. release energy
    & 
    \textcolor{ForestGreen}{The man was encouraged to cool off after he expressed his anger violently.}
    & 
    \textcolor{ForestGreen}{A. cool off}
    & 
    \textcolor{ForestGreen}{The answer must be something that would help the man calm down after expressing his anger violently. Of the above choices, only cool off would help the man calm down. So the answer is A.}
    &
    A

    \\ \hline  
    17. Analyse and answer the following question with one of the five option: If the president wanted to ban snakes, where would he issue such a decree? A. street B. tropical forest C. garden of eden D. new mexico E. white house
    & 
    \textcolor{ForestGreen}{E. White House. The White House is the official residence and workplace of the President of the United States, and is where the President typically issues decrees or executive orders. Therefore, if the President wanted to ban snakes, he would likely issue such a decree from the White House.}
    & 
    \textcolor{ForestGreen}{E. White House}
    & 
    \textcolor{ForestGreen}{The answer must be a place where the president has the authority to issue decrees. Of the above choices, only the White House is a place where the president has the authority to issue decrees. So the answer is E.}
    &
    E

    \\ \hline  
    18. Analyse and answer the following question with one of the five option: What is a salesman responsible to do at work? A. traveling to chicago B. get fired C. books D. sell products E. service account
    &  
    \textcolor{ForestGreen}{A salesman is responsible for selling products at work. Therefore, the correct answer is D: sell products.}
    & 
    \textcolor{ForestGreen}{D. sell products}
    & 
    \textcolor{ForestGreen}{The answer is D. sell products.}

    \textcolor{ForestGreen}{The other options are not relevant to the job of a salesman. Traveling to Chicago may be a part of a salesman's job, but it is not their primary responsibility. Getting fired is not something that a salesman is responsible for doing. Books and service accounts are not things that a salesman typically deals with.}
    &
    D

    \\ \hline  
    19. Analyse and answer the following question with one of the five option: If a person were going to bed, what would be their goal? A. lazy B. insomnia C. rest D. falling asleep E. dreaming of
    & 
    \textcolor{red}{The goal of a person going to bed would be to rest (option C).}
    & 
    \textcolor{ForestGreen}{D. falling asleep}
    & 
    \textcolor{ForestGreen}{The answer must be the goal of going to bed. Of the above choices, only D. falling asleep is the goal of going to bed. So the answer is D.}

    \textcolor{ForestGreen}{The other choices are not goals of going to bed. A. lazy is not a goal, it is a state of being. B. insomnia is a sleep disorder, not a goal. C. rest is a way to achieve the goal of falling asleep, but it is not the goal itself. E. dreaming of is something that happens when you are asleep, but it is not the goal of going to bed.}
    &
    D

    \\ \hline  
    20. Analyse and answer the following question with one of the five option: What does a person from Avalon live in? A. pain B. meaningful work C. english house D. cotton candy E. headache
    & 
    \textcolor{ForestGreen}{The answer is C. English house.}

    \textcolor{ForestGreen}{Avalon is a legendary island featured in Arthurian legends, and there is no indication that people from Avalon live in pain, meaningful work, cotton candy, or headache. However, it is often depicted as a mystical and idyllic place, and an English house might be an appropriate dwelling for someone from such a setting.}
    & 
    \textcolor{ForestGreen}{C. english house}
    & 
    \textcolor{ForestGreen}{The answer must be something that a person from Avalon lives in. Of the above choices, only English houses are something that people live in. So the answer is C.}

    \textcolor{ForestGreen}{The other choices are not things that people live in. Pain is not a physical place, meaningful work is an abstract concept, cotton candy is a food, and a headache is a medical condition.}
    &
    C

    \\ \hline  
    21. Analyse and answer the following question with one of the five option: Dan fell off a bar stool.  He did this because he was what than ever before? A. kitchen B. drunker C. tavern D. restaurant E. shorter
    & 
    \textcolor{ForestGreen}{The answer is B. drunker.}
    & 
    \textcolor{ForestGreen}{Dan fell off a bar stool because he was B. drunker than ever before.}
    & 
    \textcolor{ForestGreen}{The answer must be something that would make Dan more likely to fall off a bar stool. Of the above choices, only drunker would make Dan more likely to fall off a bar stool. So the answer is B. drunker.}

    \textcolor{ForestGreen}{The other choices are not relevant to the question. Kitchen, tavern, restaurant, and shorter do not have any bearing on whether or not someone would fall off a bar stool.}
    &
    B

    \\ \hline  
    22. Analyse and answer the following question with one of the five option: Where are  you likely to find a hamburger? A. fast food restaurant B. pizza C. ground up dead cows D. mouth E. cow carcus
    & 
    \textcolor{ForestGreen}{The most appropriate answer to the question "Where are you likely to find a hamburger?" is A. fast food restaurant. Hamburgers are a popular fast food item and are commonly found on the menu of many fast food chains. Option B, pizza, is not typically associated with hamburgers. Option C, ground up dead cows, refers to the raw material used to make hamburgers but is not a place where one would find a hamburger. Option D, mouth, is not a location but rather a body part involved in consuming a hamburger. Option E, cow carcass, is also not a location where one would find a hamburger, as a hamburger is a processed food item made from ground beef.}
    & 
    \textcolor{ForestGreen}{A. fast food restaurant}
    & 
    \textcolor{ForestGreen}{Given the safety guidelines, the most likely place to find a hamburger is:}

    \textcolor{ForestGreen}{A. fast food restaurant.}

    \textcolor{ForestGreen}{Hamburgers are a popular fast food item, and they are often served at fast food restaurants. Some of the most well-known fast food restaurants that serve hamburgers include McDonald's, Burger King, and Wendy's.}

    \textcolor{ForestGreen}{The other options are less likely places to find a hamburger.}

    \textcolor{ForestGreen}{- Pizza restaurants typically serve pizza, not hamburgers.}

    \textcolor{ForestGreen}{- Ground up dead cows is not a food item that is typically served to people.}

    \textcolor{ForestGreen}{- Hamburgers are not typically found in people's mouths or in cow carcasses.}

    \textcolor{ForestGreen}{Therefore, the answer to the question is A.}
    &
    A

    \\ \hline  
    23. Analyse and answer the following question with one of the five option: Where is a broadcast studio likely to be heard? A. microphone B. arena C. radio station D. trees E. town
    & 
    \textcolor{ForestGreen}{A broadcast studio is likely to be heard in a radio station. Option C is the correct answer.}
    & 
    \textcolor{ForestGreen}{C. radio station}
    & 
    \textcolor{ForestGreen}{The answer must be a place where a broadcast studio is likely to be heard. Of the above choices, only radio stations broadcast the audio from a broadcast studio. So the answer is C. radio station.}
    &
    C

    \\ \hline  
    24. Analyse and answer the following question with one of the five option: What room is likely to have a sideboard on the counter? A. home B. serve food buffet C. dining room D. living room E. kitchen
    & 
    \textcolor{red}{Option C: dining room is the most likely room to have a sideboard on the counter. A sideboard is a piece of furniture that is traditionally used in a dining room for storing dishes, cutlery, and other dining accessories. It is also often used as a serving surface during meals. While it is possible for a sideboard to be found in other rooms, such as a living room or kitchen, the dining room is the most common location.}
    & 
    \textcolor{red}{C. dining room}
    & 
    \textcolor{red}{The answer to the question "What room is likely to have a sideboard on the counter?" is C. dining room.}

    \textcolor{red}{A sideboard is a long, narrow piece of furniture with a flat top and one or more cupboards or drawers underneath. It is often used to store dishes, silverware, and other dining room accessories. A sideboard can also be used to display decorative objects.}

    \textcolor{red}{The dining room is the most likely room in a home to have a sideboard on the counter. This is because the dining room is where people eat meals, and a sideboard is a convenient place to store food and tableware. Additionally, the dining room is often a formal room, and a sideboard can add a touch of elegance to the décor.}

    \textcolor{red}{The other options are not as likely to have a sideboard on the counter. A home is too general a term, and it could refer to any room in the house. A serve food buffet is a specific type of furniture that is designed for buffet-style dining, and it is not as common as a sideboard. A living room is not typically associated with dining, and a sideboard would be more out of place in this room. A kitchen is more likely to have a sideboard, but it would typically be placed on the floor, not on the counter.}

    \textcolor{red}{Therefore, the most likely room in a home to have a sideboard on the counter is the dining room.}

    &
    E

    \\ \hline  
    25. Analyse and answer the following question with one of the five option: Where would your hear a bass clarinet along side other wood wind instruments? A. opera house B. school band C. music store D. orchestra E. bathroom stall
    & 
    \textcolor{ForestGreen}{The correct option for where you would hear a bass clarinet alongside other woodwind instruments is D. orchestra.}
    & 
    \textcolor{red}{B. school band}
    & 
    \textcolor{ForestGreen}{The answer must be a place where you would hear a bass clarinet along side other woodwind instruments. Of the above choices, only orchestras and concert bands have bass clarinets. So the answer is either D or E.}

    \textcolor{ForestGreen}{We can eliminate E because bathroom stalls are not typically places where you would hear music. So the answer is D, orchestra.}
    &
    D

    \\ \hline  
    26. Analyse and answer the following question with one of the five option: A good interview after applying for a job may cause you to feel what? A. offer B. income C. rejection D. hostile E. hope
    & 
    \textcolor{ForestGreen}{The answer is E. hope. A good interview after applying for a job can cause a person to feel hopeful about their chances of getting the job. A positive interview experience can leave the candidate feeling confident in their abilities and the impression they made on the interviewer, leading to a sense of optimism about the outcome of the application process.}
    & 
    \textcolor{ForestGreen}{E. hope}
    & 
    \textcolor{ForestGreen}{The answer must be the feeling that someone might have after a good job interview. Of the above choices, only hope makes sense. So the answer is E.}

    \textcolor{ForestGreen}{The other choices do not make sense in the context of the question. Offer is something that is given, not felt. Income is something that is earned, not felt. Rejection is the opposite of hope. Hostile is a negative feeling, and it is unlikely that someone would feel hostile after a good job interview.}
    &
    E

    \\ \hline  
    27. Analyse and answer the following question with one of the five option: Sally was bored because she didn't like the junk that was on what? A. drawer B. garage C. caddy D. bed E. television
    & 
    \textcolor{ForestGreen}{Based on the given statement, the answer would be option E: television. Sally was bored because she didn't like the junk that was on the television}
    & 
    \textcolor{ForestGreen}{Sally was bored because she didn't like the junk that was on E. television.}
    & 
    \textcolor{ForestGreen}{The answer must be something that can have junk on it. Of the above choices, only televisions have junk on them. So the answer is E. television.}

    \textcolor{ForestGreen}{The other choices are not possible because:}

    \textcolor{ForestGreen}{- Drawers are not typically associated with junk.}

    \textcolor{ForestGreen}{- Garages are typically used to store tools and other items, not junk.}

    \textcolor{ForestGreen}{- Caddies are small carts that are used to carry things, not junk.}

    \textcolor{ForestGreen}{- Beds are typically used for sleeping, not storing junk.}
    &
    E

    \\ \hline  
    28. Analyse and answer the following question with one of the five option: The painter started to edge the room with tape, he always took extra care to make the lines clean and crisp when working with an what? A. triangle B. middle C. corner D. center E. interior
    &    
    \textcolor{red}{The answer is C. corner. The sentence suggests that the painter is using tape to create clean and crisp lines, which implies that the painter is edging a corner of the room with the tape. The other options (triangle, middle, center, interior) do not provide any context or information to support their selection as the correct answer.}
    & 
    \textcolor{red}{The painter started to edge the room with tape, he always took extra care to make the lines clean and crisp when working with an what? C. corner}
    & 
    \textcolor{red}{The answer must be something that is associated with crisp lines. Of the above choices, only corners have crisp lines. So the answer is C. corner.}

    \textcolor{red}{The other choices are not associated with crisp lines. Triangles have smooth curves, the middle of a room is not a sharp edge, the center of a room is not a sharp edge, and the interior of a room is not a sharp edge.}
    &
    E

    \\ \hline  
    29. Analyse and answer the following question with one of the five option: The bell rang, and the congregation began to what in to the church? A. run away B. wind instrument C. funnel D. blunderbuss E. associated with telephones
    & 
    \textcolor{ForestGreen}{The bell rang, and the congregation began to funnel into the church.}
    & 
    \textcolor{ForestGreen}{C. funnel}
    & 
    \textcolor{red}{The answer must be something that people do when they hear a bell. Of the above choices, only congregate makes sense. So the answer is E.}

    \textcolor{red}{The other choices are not correct because:}

    \textcolor{red}{- Run away: This is the opposite of what people do when they hear a bell.}

    \textcolor{red}{- Wind instrument: This is not something that people do, but rather something that makes a sound.}

    \textcolor{red}{- Funnel: This is a container that is used to pour liquids.}

    \textcolor{red}{- Blunderbuss: This is a type of gun.}

    \textcolor{red}{- Associated with telephones: This is not something that people do, but rather something that is used to communicate.}
    &
    C

    \\ \hline  
    30. Analyse and answer the following question with one of the five option: The homeowner frowned at the price of gas, what did he have to do later? A. own home B. mail property tax payments C. board windows D. cut grass E. receive mail
    & 
    \textcolor{red}{Option E: receive mail.}

    \textcolor{red}{There is no information in the given sentence that suggests the homeowner has to do anything other than frown at the price of gas. The sentence only describes the homeowner's reaction to the price of gas and does not provide any indication of what actions the homeowner will take later. Therefore, the most appropriate answer is E: receive mail.}
    & 
    \textcolor{ForestGreen}{D. cut grass}
    & 
    \textcolor{red}{The answer must be something that the homeowner has to do that is related to the price of gas. Of the above choices, only mail property tax payments is related to the price of gas. So the answer is B.}
    &
    D

    \\ \hline
    \multicolumn{5}{|l|}{\textbf{Dataset: PiQA}} 
    
    \\ \hline  
    1. Complete the phrase with one of two answer options: Use peroxide to lighten hair sol1: Put peroxide on the parts of your hair that you want to lighten. Leave on overnight or longer for best results sol2:  Put peroxide on the parts of your hair that you want to lighten. Wait 30 minutes and rinse.   
    & 
    \textcolor{red}{Both solutions are commonly suggested methods for lightening hair using peroxide, but they differ in the duration of application.}

    \textcolor{red}{Solution 1: Put peroxide on the parts of your hair that you want to lighten. Leave on overnight or longer for best results.}
    
    \textcolor{red}{This solution suggests applying peroxide to the hair and leaving it on for an extended period, such as overnight, to allow for maximum lightening. However, this method may not be suitable for everyone, as leaving peroxide on the hair for too long can cause damage and dryness.}
    
    \textcolor{red}{Solution 2: Put peroxide on the parts of your hair that you want to lighten. Wait 30 minutes and rinse.}
    
    \textcolor{red}{This solution suggests applying peroxide to the hair and leaving it on for a shorter period, such as 30 minutes, before rinsing it off. This method may be less harsh on the hair and is a good option for those with sensitive or damaged hair.}
    
    \textcolor{red}{Ultimately, the best method for lightening hair with peroxide will depend on individual hair type and the desired level of lightening. It's always a good idea to do a strand test first and to consult with a professional stylist if you're unsure.}
    & 
    \textcolor{ForestGreen}{Use peroxide to lighten hair sol2: Put peroxide on the parts of your hair that you want to lighten. Wait 30 minutes and rinse.}
    & 
    \textcolor{ForestGreen}{The correct answer is sol2: Put peroxide on the parts of your hair that you want to lighten. Wait 30 minutes and rinse.}

    \textcolor{ForestGreen}{Leaving peroxide on your hair overnight can be dangerous, as it can cause burns and damage to your hair. It is important to only leave peroxide on your hair for a maximum of 30 minutes, and to rinse it thoroughly afterwards.}
    &
    2

    \\ \hline  
    2. Complete the phrase with one of two answer options: To allow dough to become firm, you can sol1: Knead the dough into a ball, wrap it in plastic wrap, and chill in the closet sol2:  Knead the dough into a ball, wrap it in plastic wrap, and chill in the fridge
    & 
    \textcolor{ForestGreen}{The correct option is sol2: Knead the dough into a ball, wrap it in plastic wrap, and chill in the fridge.}
    & 
    \textcolor{ForestGreen}{To allow dough to become firm, you can sol2: Knead the dough into a ball, wrap it in plastic wrap, and chill in the fridge.}
    & 
    \textcolor{ForestGreen}{The correct answer is sol2: Knead the dough into a ball, wrap it in plastic wrap, and chill in the fridge.}

    \textcolor{ForestGreen}{Dough can be chilled in the fridge to make it easier to handle and to prevent it from rising too much. Chilling dough also helps to develop the gluten, which gives the dough its structure and elasticity.}

    \textcolor{ForestGreen}{The closet is not a good place to chill dough because it is not cold enough. The ideal temperature for chilling dough is between 35 and 40 degrees Fahrenheit. The closet is likely to be warmer than this, which could cause the dough to rise too much or to become overproofed.}
    &
    2

    \\ \hline  
    3. Complete the phrase with one of two answer options: What do I need to incubate chicken eggs? sol1: A working incubator along with a thermometer and hygrometer, add a few unfertilized eggs and a strong flashlight and you can begin. sol2:  A working incubator along with a thermometer and hygrometer, add a few fertile eggs and a strong flashlight and you can begin.
    & 
    \textcolor{ForestGreen}{The correct answer is sol2: A working incubator along with a thermometer and hygrometer, add a few fertile eggs and a strong flashlight and you can begin.}
    & 
    \textcolor{ForestGreen}{What do I need to incubate chicken eggs? sol2: A working incubator along with a thermometer and hygrometer, add a few fertile eggs and a strong flashlight and you can begin.}
    & 
    \textcolor{ForestGreen}{The correct answer is sol2: A working incubator along with a thermometer and hygrometer, add a few fertile eggs and a strong flashlight and you can begin.}

    \textcolor{ForestGreen}{Unfertilized eggs will not hatch, so you need to use fertile eggs. The strong flashlight is not essential, but it can be helpful for checking the progress of the eggs.}
    &
    2
 
    \\ \hline  
    4. Complete the phrase with one of two answer options: How to prepare a cast iron skillet if stains won't come free? sol1: Run water and soap over the bottom of the skillet and scrub with steel wool then place in  warm water to dry sol2:  Run water and soap over the entire skillet and scrub with steel wool then place in a warm oven to dry
    & 
    \textcolor{ForestGreen}{The correct answer is sol2: Run water and soap over the entire skillet and scrub with steel wool then place in a warm oven to dry. This is known as seasoning the cast iron skillet and it involves creating a non-stick surface by coating the skillet with a layer of oil and then baking it. Scrubbing with steel wool helps to remove any stubborn stains or rust before seasoning. Running water and soap over the bottom of the skillet only will not be enough to properly clean and season the entire skillet.}
    & 
    \textcolor{ForestGreen}{sol2: Run water and soap over the entire skillet and scrub with steel wool then place in a warm oven to dry}
    & 
    \textcolor{ForestGreen}{The correct answer is sol2: Run water and soap over the entire skillet and scrub with steel wool then place in a warm oven to dry.}

    \textcolor{ForestGreen}{The reason why sol1 is not the correct answer is because you should never put soap on a cast iron skillet. Soap can strip away the seasoning, which is the layer of oil that protects the skillet and makes it non-stick.}
    &  
    2

    \\ \hline  
    5. Complete the phrase with one of two answer options: how do you ask a question on paper? sol1: write a question mark before the question. sol2:  write a question mark after the question.
    & 
    \textcolor{red}{The correct answer is sol1: write a question mark before the question.}
    & 
    \textcolor{ForestGreen}{Sol2: write a question mark after the question.}
    & 
    \textcolor{ForestGreen}{The answer is Sol2: write a question mark after the question.}

    \textcolor{ForestGreen}{In English, we write question marks after questions, both in speech and in writing. This is true for questions asked on paper, as well as for questions asked in person or over the phone.}

    \textcolor{ForestGreen}{The only time we would write a question mark before a question is if we are writing a question in a programming language, such as Python or Java. In these languages, the question mark is used to indicate that the following text is a question.}

    \textcolor{ForestGreen}{So, if you want to ask a question on paper, you should write a question mark after the question. For example, you would write "What is your name?" instead of "What is your name?."}
    &
    2
  
    \\ \hline  
    6. Complete the phrase with one of two answer options: To shape the Popsicle sticks into a sunflower shape. sol1: Glue the Popsicle sticks to an embroidery ring. sol2:  Tie the Popsicle sticks together using string.
    & 
    \textcolor{ForestGreen}{Sol1: Glue the Popsicle sticks to an embroidery ring.}
    & 
    \textcolor{ForestGreen}{To shape the Popsicle sticks into a sunflower shape, you can either sol1: Glue the Popsicle sticks to an embroidery ring or sol2: Tie the Popsicle sticks together using string.}
    & 
    \textcolor{ForestGreen}{The correct answer is sol1: Glue the Popsicle sticks to an embroidery ring.}

    \textcolor{ForestGreen}{Tieing the Popsicle sticks together using string would not create a sunflower shape. The sticks would be too loose and would not hold their shape. Gluing the sticks to an embroidery ring would create a more sturdy structure that would hold its shape better.}
    &
    1

    \\ \hline  
    7. Complete the phrase with one of two answer options: how to make mixed grains with dates sol1: Cook 2 cups mixed grains as the label directs; saute in butter over medium heat, 3 minutes. Stir in 1/2 cup each chopped dates and mixed herbs (such as dill, tarragon and/or parsley); season to taste. sol2:  Cook 2 cups mixed grains as the label directs; saute in butter over medium heat, 3 minutes. Stir in 1/2 cup each chopped dates and mixed herbs (such as dill, tarragon and/or parsley); make jello to taste.
    & 
    \textcolor{ForestGreen}{sol1: Cook 2 cups mixed grains as the label directs; saute in butter over medium heat, 3 minutes. Stir in 1/2 cup each chopped dates and mixed herbs (such as dill, tarragon and/or parsley); season to taste.}
    & 
    \textcolor{ForestGreen}{Complete the phrase with one of two answer options: how to make mixed grains with dates sol1: Cook 2 cups mixed grains as the label directs; saute in butter over medium heat, 3 minutes. Stir in 1/2 cup each chopped dates and mixed herbs (such as dill, tarragon and/or parsley); season to taste.}
    & 
    \textcolor{ForestGreen}{The correct answer is sol1: Cook 2 cups mixed grains as the label directs; saute in butter over medium heat, 3 minutes. Stir in 1/2 cup each chopped dates and mixed herbs (such as dill, tarragon and/or parsley); season to taste.}

    \textcolor{ForestGreen}{The second option, sol2, includes the step of making jello. This is not a common ingredient in mixed grains with dates, and it would not be a very good flavor combination. The other ingredients in the recipe would be overwhelmed by the sweetness of the jello.}

    \textcolor{ForestGreen}{The first option, sol1, is a more traditional recipe for mixed grains with dates. It is simple to make and the flavors are complementary. The dates add sweetness and the herbs add a touch of freshness. This recipe is a great way to start your day or enjoy as a snack.}
    &
    1
    
    \\ \hline  
    8. Complete the phrase with one of two answer options: How high should you hang sound dampeners in a recording room? sol1: You should hang the sound dampeners in the area that your voice projects so it will be absorbed by the dampeners. sol2:  You should hang them behind you, so that when your voice projects and bounces off the wall in front of you it will make it quiet.
    & 
    \textcolor{ForestGreen}{Sol1: You should hang the sound dampeners in the area that your voice projects so it will be absorbed by the dampeners.}

    \textcolor{ForestGreen}{This is the correct answer. Placing sound dampeners in the area where your voice projects will help to absorb the sound waves and reduce the amount of echo and reverberation in the room. This will improve the clarity of the recordings and make it easier to hear the intended sound without the interference of unwanted background noise.}
    & 
    \textcolor{ForestGreen}{sol1: You should hang the sound dampeners in the area that your voice projects so it will be absorbed by the dampeners.}
    & 
    \textcolor{red}{The correct answer is sol2: You should hang them behind you, so that when your voice projects and bounces off the wall in front of you it will make it quiet.}

    \textcolor{red}{The sound dampeners should be hung behind you so that they absorb the sound waves that are reflected back at you from the wall in front of you. This will help to reduce the echo in the room and make your recordings sound cleaner.}

    \textcolor{red}{The other option, hanging the sound dampeners in the area that your voice projects, would not be as effective. This is because the sound waves would still be reflected off of the walls and ceiling, creating an echo.}

    \textcolor{red}{Here are some additional tips for hanging sound dampeners in a recording room:}

    \textcolor{red}{- Hang the dampeners at least 2 feet from the floor.}

    \textcolor{red}{- Hang the dampeners evenly spaced around the room.}

    \textcolor{red}{- If you have a large room, you may need to hang more dampeners.}

    \textcolor{red}{- You can also hang dampeners on the ceiling.}

    \textcolor{red}{By following these tips, you can help to create a more acoustically friendly recording environment.}
    &
    1
    
    \\ \hline  
    9. Complete the phrase with one of two answer options: how do you give a confession sol1: tell a lie about something sol2:  tell the truth about something
    & 
    \textcolor{ForestGreen}{The correct answer is: "tell the truth about something". Confessions are supposed to be honest and truthful admissions of wrongdoing or guilt. Lying in a confession defeats the purpose of admitting one's faults and seeking forgiveness or making amends.}
    & 
    \textcolor{ForestGreen}{how do you give a confession sol2: tell the truth about something}
    & 
    \textcolor{ForestGreen}{The answer is sol2: tell the truth about something.}

    \textcolor{ForestGreen}{A confession is a statement of wrongdoing, typically made to a priest or other authority figure. In order to be a confession, the statement must be truthful. If you tell a lie about something, it is not a confession.}

    \textcolor{ForestGreen}{The purpose of a confession is to seek forgiveness for wrongdoing and to make amends. If you lie in your confession, you are not truly seeking forgiveness or making amends. You are simply trying to avoid punishment.}

    \textcolor{ForestGreen}{Therefore, the only way to give a confession is to tell the truth about something.}
    &
    2

    \\ \hline  
    10. Complete the phrase with one of two answer options: To attach rope to latex, you can sol1: Rubber band the rope in place on the latex sol2:  Glue the rope in place on the latex
    & 
    \textcolor{red}{To attach rope to latex, you can sol1: Rubber band the rope in place on the latex.}
    & 
    \textcolor{ForestGreen}{To attach rope to latex, you can sol2: Glue the rope in place on the latex.}
    & 
    \textcolor{red}{The correct answer is sol1: Rubber band the rope in place on the latex.}

    \textcolor{red}{Glue is not a good option for attaching rope to latex because it can weaken the latex and make it more likely to tear. Rubber bands, on the other hand, are a good option because they are flexible and will not damage the latex. They are also easy to remove if you need to change the rope.}
    &
    2

    \\ \hline  
    11. Complete the phrase with one of two answer options: To keep skunks from coming onto your property. sol1: Buy some human urine from the store and spray it all over the edges of your property line. sol2:  Buy some wolf urine from the store and spray it all over the edges of your property line.
    & 
    \textcolor{ForestGreen}{The answer is: Buy some wolf urine from the store and spray it all over the edges of your property line.}

    \textcolor{ForestGreen}{Skunks are natural prey for wolves, so the smell of wolf urine is a natural deterrent for skunks. Using human urine may not be as effective in deterring skunks, as they are accustomed to the scent of humans and may not find it threatening. However, it's worth noting that using any kind of urine to deter animals may not be a long-term solution and could potentially have negative effects on the environment. It's best to explore alternative solutions and preventative measures, such as securing garbage cans and removing potential food sources, to keep skunks from coming onto your property.}
    & 
    \textcolor{ForestGreen}{Sol2: Buy some wolf urine from the store and spray it all over the edges of your property line.}
    & 
    \textcolor{ForestGreen}{The correct answer is sol2: Buy some wolf urine from the store and spray it all over the edges of your property line.}

    \textcolor{ForestGreen}{Human urine is not an effective skunk repellent. In fact, it may actually attract skunks, as they are curious creatures and may be attracted to the unfamiliar scent. Wolf urine, on the other hand, is a natural skunk repellent. Skunks are preyed upon by wolves, so they will avoid areas where they smell wolf urine.}
    &
    2

    \\ \hline  
    12. Complete the phrase with one of two answer options: Add traction to outside floors. sol1: Sprinkle cat littler on driveway and porches. sol2:  Sprinkle cat waste on driveway and porches.
    & 
    \textcolor{ForestGreen}{I'm sorry, I cannot suggest using cat waste as a solution to adding traction to outside floors. The use of cat litter, on the other hand, is a commonly suggested method to increase traction on slippery surfaces like driveways and porches. It is important to note, however, that there are also other effective and safe methods for adding traction to outside floors, such as using sand or gravel.}
    & 
    \textcolor{ForestGreen}{Add traction to outside floors. sol1: Sprinkle cat litter on driveway and porches}
    & 
    \textcolor{ForestGreen}{The correct answer is sol1: Sprinkle cat litter on driveway and porches.}

    \textcolor{ForestGreen}{Sprinkle cat waste on driveway and porches is not a good idea. Cat waste is a biohazard and can contain harmful bacteria and parasites. It can also be unsightly and smell bad. Cat litter, on the other hand, is designed to absorb moisture and provide traction. It is also relatively inexpensive and easy to find.}
    &
    1

    \\ \hline  
    13. Complete the phrase with one of two answer options: Make paper bird feeder. sol1: Hang egg carton with egg shells from tree. sol2:  Hang egg carton with seeds from tree.
    & 
    \textcolor{ForestGreen}{The correct answer to complete the phrase would be: "Hang egg carton with seeds from tree" to make a paper bird feeder.}
    & 
    \textcolor{ForestGreen}{sol2: Hang egg carton with seeds from tree.}
    & 
    \textcolor{ForestGreen}{The correct answer is sol2: Hang egg carton with seeds from tree.}

    \textcolor{ForestGreen}{The phrase "Make paper bird feeder" implies that the bird feeder will be filled with seeds, not eggshells. Eggshells are a good source of calcium for birds, but they are not a complete food source. Birds need a variety of nutrients in their diet, including seeds, nuts, fruits, and insects.}

    \textcolor{ForestGreen}{Hanging an egg carton with eggshells from a tree would not provide birds with a balanced diet. It would also be a waste of eggshells, as birds would not be able to eat them.}

    \textcolor{ForestGreen}{Hanging an egg carton with seeds from a tree is a better way to make a paper bird feeder. The seeds will provide birds with the nutrients they need, and the egg carton will provide a safe and secure place for the seeds to be stored.}
    &
    2

    \\ \hline  
    14. Complete the phrase with one of two answer options: How do you sharpen a knife? sol1: If you find yourself with a dull knife but without a knife sharpener, you can use the unglazed bottom of a ceramic mug to sharpen small knives. Applying moderate pressure, hold the knife at a 15-degree angle and carefully draw the entire length of the blade across the rough surface. sol2:  If you find yourself with a dull knife but without a knife sharpener, you can use the unglazed bottom of a ceramic mug to sharpen large knives. Applying moderate pressure, hold the knife at a 15-degree angle and carefully draw the entire length of the blade across the rough surface.
    & 
    \textcolor{ForestGreen}{sol1: If you find yourself with a dull knife but without a knife sharpener, you can use the unglazed bottom of a ceramic mug to sharpen small knives. Applying moderate pressure, hold the knife at a 15-degree angle and carefully draw the entire length of the blade across the rough surface.}
    & 
    \textcolor{ForestGreen}{How do you sharpen a knife? sol1: If you find yourself with a dull knife but without a knife sharpener, you can use the unglazed bottom of a ceramic mug to sharpen small knives. Applying moderate pressure, hold the knife at a 15-degree angle and carefully draw the entire length of the blade across the rough surface.}
    & 
    \textcolor{red}{I'm unable to help, as I am only a language model and don't have the ability to process and understand that.}
    &
    1

    \\ \hline  
    15. Complete the phrase with one of two answer options: hands sol1: cut into fruit  sol2:  cut into metal 
    & 
    \textcolor{ForestGreen}{Hands cut into fruit.}
    & 
    \textcolor{red}{hands cut into metal}
    & 
    \textcolor{red}{Sure, here are the two options for completing the phrase:}

    \textcolor{red}{- Hands that cut into fruit are usually associated with cooking, baking, or juicing. They are often seen as being gentle and precise, as they are used to carefully slice through delicate skin and flesh.}
    
    \textcolor{red}{- Hands that cut into metal are usually associated with construction, engineering, or blacksmithing. They are often seen as being strong and powerful, as they are used to forcefully shear through tough materials.}

    \textcolor{red}{Which option you choose will depend on the context of the phrase. For example, if you are talking about a chef, you might use the first option. If you are talking about a blacksmith, you might use the second option.}

    \textcolor{red}{Here are some examples of how the two options could be used in a sentence:}

    \textcolor{red}{- "The chef's hands were skilled and delicate, as they cut into the ripe fruit."}
    
    \textcolor{red}{- "The blacksmith's hands were strong and calloused, as they cut into the hot metal."}

    \textcolor{red}{I hope this helps! Let me know if you have any other questions.}
    
    &
    1
    
    \\ \hline  
    16. Complete the phrase with one of two answer options: How do you find the area of a triangle? sol1: To find the area of a triangle, subtract the base by the height, and then divide by 2. The division by 2 comes from the fact that a parallelogram can be divided into 2 triangles. For example, in the diagram to the left, the area of each triangle is equal to one-half the area of the parallelogram. sol2:  To find the area of a triangle, multiply the base by the height, and then divide by 2. The division by 2 comes from the fact that a parallelogram can be divided into 2 triangles. For example, in the diagram to the left, the area of each triangle is equal to one-half the area of the parallelogram.
    & 
    \textcolor{ForestGreen}{sol2: To find the area of a triangle, multiply the base by the height, and then divide by 2. The division by 2 comes from the fact that a parallelogram can be divided into 2 triangles. For example, in the diagram to the left, the area of each triangle is equal to one-half the area of the parallelogram.}
    & 
    \textcolor{ForestGreen}{How do you find the area of a triangle? sol2: To find the area of a triangle, multiply the base by the height, and then divide by 2. The division by 2 comes from the fact that a parallelogram can be divided into 2 triangles. For example, in the diagram to the left, the area of each triangle is equal to one-half the area of the parallelogram.}
    & 
    \textcolor{ForestGreen}{The correct answer is sol2: To find the area of a triangle, multiply the base by the height, and then divide by 2.}

    \textcolor{ForestGreen}{The area of a triangle is equal to one-half the product of its base and height. This can be derived from the fact that a parallelogram can be divided into two congruent triangles. The area of a parallelogram is equal to its base times its height, so the area of each triangle is one-half the area of the parallelogram.}
    &
    2

    \\ \hline  
    17. Complete the phrase with one of two answer options: What can you do when your drapes in your hotel room won't stay closed? sol1: Use the pants clip part of the clothes hanger to clip the curtain closed for privacy. sol2:  Use the hook part of the clothes hanger to clip the curtain closed for privacy.
    & 
    \textcolor{red}{Either solution 1 or solution 2 could work to keep the drapes in your hotel room closed for privacy. Both solutions involve using a clothes hanger to clip the curtains closed, but one uses the pants clip part and the other uses the hook part of the hanger. It may depend on the specific type of clothes hanger provided in the hotel room and which method you find easiest to use.}
    & 
    \textcolor{ForestGreen}{What can you do when your drapes in your hotel room won't stay closed? Use the pants clip part of the clothes hanger to clip the curtain closed for privacy.}
    & 
    \textcolor{ForestGreen}{The correct answer is sol1: Use the pants clip part of the clothes hanger to clip the curtain closed for privacy.}

    \textcolor{ForestGreen}{The hook part of the clothes hanger is too small and flimsy to hold the curtains closed. The pants clip part is larger and stronger, and it is designed to hold heavy objects like pants. This makes it the perfect tool for keeping hotel curtains closed.}
    &
    1
    
    \\ \hline  
    18. Complete the phrase with one of two answer options: How can I avoid nailing my thumb when starting a nail with a hammer? sol1: Use a hair comb, put the nail in the teeth of the comb deep enough to hold solidly, then hold the nail in its location with one hand and hammer the nail in with the other. sol2:  Use a hair comb, put the hammer in the teeth of the comb deep enough to hold solidly, then hold the nail in its location with one hand and hammer the nail in with the other.
    & 
    \textcolor{ForestGreen}{The correct answer is sol1: Use a hair comb, put the nail in the teeth of the comb deep enough to hold solidly, then hold the nail in its location with one hand and hammer the nail in with the other. This will help prevent accidentally hitting your thumb while starting the nail with the hammer.}
    & 
    \textcolor{ForestGreen}{How can I avoid nailing my thumb when starting a nail with a hammer? sol1: Use a hair comb, put the nail in the teeth of the comb deep enough to hold solidly, then hold the nail in its location with one hand and hammer the nail in with the other.}
    & 
    \textcolor{ForestGreen}{The correct answer is sol1: Use a hair comb, put the nail in the teeth of the comb deep enough to hold solidly, then hold the nail in its location with one hand and hammer the nail in with the other.}

    \textcolor{ForestGreen}{The answer sol2: is incorrect because it would be very difficult to hold the hammer in the teeth of a comb and still have enough control to hammer the nail in safely. The comb would also be likely to break under the force of the hammer.}
    
    \textcolor{ForestGreen}{The answer sol1: is a safe and effective way to avoid nailing your thumb when starting a nail with a hammer. The comb will hold the nail in place so that you can't accidentally hit it with the hammer. You can also use a clothespin or a piece of tape to hold the nail in place.}

    \textcolor{ForestGreen}{Here are some other tips for avoiding nailing your thumb when starting a nail with a hammer:}

    \textcolor{ForestGreen}{- Hold the hammer with a firm grip, but not too tight.}

    \textcolor{ForestGreen}{- Keep your thumb away from the nail.}

    \textcolor{ForestGreen}{- Aim for the center of the nail head.}

    \textcolor{ForestGreen}{- Don't swing the hammer too hard.}

    \textcolor{ForestGreen}{- Start with a light tap and then increase the force as needed.}

    \textcolor{ForestGreen}{With a little practice, you'll be able to hammer nails without ever hitting your thumb.}
    &
    1

    \\ \hline  
    19. Complete the phrase with one of two answer options: To substitute for potatoes in a dish, you can sol1: Use broccoli because it looks and tastes similar to potatoes sol2:  Use cauliflower because it looks and tastes similar to potatoes
    & 
    \textcolor{ForestGreen}{Sol2: Use cauliflower because it looks and tastes similar to potatoes.}
    & 
    \textcolor{ForestGreen}{To substitute for potatoes in a dish, you can sol2: Use cauliflower because it looks and tastes similar to potatoes.}
    & 
    \textcolor{ForestGreen}{The correct answer is sol2: Use cauliflower because it looks and tastes similar to potatoes. Broccoli does not look or taste similar to potatoes, so it would not be a good substitute. Cauliflower, on the other hand, has a similar texture to potatoes and can be cooked in many of the same ways. It is also a good source of fiber and vitamins, making it a healthier option than potatoes.}
    &
    2
    
    \\ \hline  
    20. Complete the phrase with one of two answer options: To increase the heat in a recipe using jalapenos, you can sol1: Remove the seeds sol2:  Leave the seeds in
    & 
    \textcolor{ForestGreen}{ To increase the heat in a recipe using jalapenos, you can leave the seeds in.}
    & 
    \textcolor{ForestGreen}{To increase the heat in a recipe using jalapenos, you can sol2: Leave the seeds in.}
    & 
    \textcolor{red}{The answer is sol1: Remove the seeds.}

    \textcolor{red}{The seeds and the white pith of a jalapeño pepper contain most of the capsaicin, which is the compound that gives peppers their heat. By removing the seeds and pith, you can significantly reduce the heat of a jalapeño pepper.}

    \textcolor{red}{If you want to increase the heat of a recipe using jalapeños, you should remove the seeds. Leaving the seeds in will make the recipe milder.}

    \textcolor{red}{Here are some other tips for increasing the heat of a recipe using jalapeños:}

    \textcolor{red}{- Use more jalapeños.}

    \textcolor{red}{- Use hotter peppers, such as habaneros or scotch bonnets.}

    \textcolor{red}{- Roast the jalapeños before using them. This will mellow their flavor and slightly increase their heat.}

    \textcolor{red}{- Add a small amount of cayenne pepper or other hot pepper flakes.}

    \textcolor{red}{Be careful when handling jalapeños, as even the oils on the skin can cause a burning sensation. Wash your hands thoroughly after handling jalapeños.}
    &
    2

    \\ \hline  
    21. Complete the phrase with one of two answer options: How do you ensure a potted succulent will drain well? sol1: Make sure there is a hole in the bottom of the container. Put small rocks and stones in the bottom of the container before the soil. sol2:  Make sure there is no hole in the bottom of the container. Put small rocks and stones in the bottom of the container before the soil.
    & 
    \textcolor{ForestGreen}{The correct answer is sol1: Make sure there is a hole in the bottom of the container. Put small rocks and stones in the bottom of the container before the soil. This will allow excess water to drain out of the soil, preventing water from accumulating and potentially causing root rot or other issues for the succulent. Additionally, the rocks and stones will create a layer of drainage material between the soil and the bottom of the pot, further aiding in drainage.}
    & 
    \textcolor{ForestGreen}{How do you ensure a potted succulent will drain well? sol1: Make sure there is a hole in the bottom of the container. Put small rocks and stones in the bottom of the container before the soil.}
    & 
    \textcolor{ForestGreen}{The correct answer is sol1: Make sure there is a hole in the bottom of the container. Put small rocks and stones in the bottom of the container before the soil.}

    \textcolor{ForestGreen}{Succulents need their roots to be able to dry out quickly, so it is important to use a pot with a drainage hole. The small rocks and stones will help to improve the drainage even further by providing a place for the water to drain to.}

    \textcolor{ForestGreen}{Sol2: is not correct because succulents need to be able to drain properly. If there is no hole in the bottom of the container, the water will have nowhere to go and the roots will rot.}
    &
    1

    \\ \hline  
    22. Complete the phrase with one of two answer options: To deodorize a diaper pail using household products. sol1: In a spray bottle, combine 2 parts water and 1 part olive oil. Spray the inside of the pail. sol2:  In a spray bottle, combine 2 parts water and 1 part vinegar. Spray the inside of the pail.
    & 
    \textcolor{ForestGreen}{The correct answer is: "In a spray bottle, combine 2 parts water and 1 part vinegar. Spray the inside of the pail." Vinegar is an effective household product for deodorizing, while olive oil is not typically used for this purpose.}
    & 
    \textcolor{ForestGreen}{To deodorize a diaper pail using household products. sol2: In a spray bottle, combine 2 parts water and 1 part vinegar. Spray the inside of the pail.}
    & 
    \textcolor{red}{Sure, here are the two options to complete the phrase:}

    \textcolor{red}{- Sol 1: To deodorize a diaper pail using household products, combine 2 parts water and 1 part olive oil in a spray bottle. Spray the inside of the pail.}

    \textcolor{red}{- Sol 2: To deodorize a diaper pail using household products, combine 2 parts water and 1 part vinegar in a spray bottle. Spray the inside of the pail.}

    \textcolor{red}{Both of these options will help to deodorize the diaper pail and keep it smelling fresh. However, the olive oil option is a bit more natural and gentle, while the vinegar option is more effective at killing bacteria.}

    \textcolor{red}{Ultimately, the best option for you will depend on your personal preferences and the severity of the odor. If you are looking for a natural and gentle solution, the olive oil option is a good choice. If you are looking for a more effective solution, the vinegar option is a good choice.}   
    &
    2

    \\ \hline  
    23. Complete the phrase with one of two answer options: Learn how to use a candy thermometer. sol1: Place the candy thermometer in the pan with the cooking candy, be sure the top of the thermometer never touches the bottom of the pan, or the temperature will register too high. sol2:  Place the candy thermometer in the pan with the cooking candy, be sure the bulb of the thermometer never touches the bottom of the pan, or the temperature will register too high.
    & 
    \textcolor{red}{Both sol1 and sol2 are correct and interchangeable. The important thing when using a candy thermometer is to ensure that the temperature reading is accurate, and that the thermometer is not touching the bottom of the pan, which can result in an incorrect temperature reading. Instead, the thermometer should be suspended in the middle of the cooking candy, with the bulb of the thermometer submerged in the mixture but not touching the bottom of the pan.}
    & 
    \textcolor{ForestGreen}{sol2: Place the candy thermometer in the pan with the cooking candy, be sure the bulb of the thermometer never touches the bottom of the pan, or the temperature will register too high.}
    & 
    \textcolor{ForestGreen}{The correct answer is sol2: Place the candy thermometer in the pan with the cooking candy, be sure the bulb of the thermometer never touches the bottom of the pan, or the temperature will register too high.}

    \textcolor{ForestGreen}{The top of the thermometer is the part that you hold, and it doesn't come into contact with the candy. The bulb is the part that goes into the candy, and it's important that it doesn't touch the bottom of the pan, otherwise the temperature will be inaccurate.}
    &
    2

    \\ \hline  
    24. Complete the phrase with one of two answer options: how do you poke around in a building? sol1: walk and look around. sol2:  poke each wall of the building.
    & 
    \textcolor{ForestGreen}{The correct answer is: "walk and look around."}

    \textcolor{ForestGreen}{Poking each wall of the building would not be an effective or practical way to explore a building. Instead, walking and looking around would allow for a more comprehensive understanding of the building's layout, features, and potential issues.}
    & 
    \textcolor{ForestGreen}{How do you poke around in a building? sol1: walk and look around.}
    & 
    \textcolor{ForestGreen}{The correct answer is sol1: walk and look around.}

    \textcolor{ForestGreen}{The phrase "poke around" means to look around or search through something in a casual or unsystematic way. So, if you want to poke around in a building, you would walk through the building and look at different rooms, areas, and objects. You might also open doors and cupboards to see what's inside.}

    \textcolor{ForestGreen}{The other answer option, "poke each wall of the building," is not a correct answer because it is not what the phrase "poke around" means. To poke each wall of the building would be a very systematic and thorough way to search a building, but it would not be considered "poking around."}
    
    &
    1

    \\ \hline  
    25. Complete the phrase with one of two answer options: tissues sol1: wiping  noses cleanly sol2:  wiping  hands cleanly
    & 
    \textcolor{ForestGreen}{The phrase is commonly associated with wiping noses, so the answer option would be: "tissues for wiping noses cleanly."}
    & 
    \textcolor{ForestGreen}{tissues sol1: wiping noses cleanly}
    & 
    \textcolor{ForestGreen}{The correct answer is tissues wiping noses cleanly.}

    \textcolor{ForestGreen}{The phrase "tissues wiping hands cleanly" is grammatically correct, but it is not as common as "tissues wiping noses cleanly." This is because we are more likely to use tissues to wipe our noses than our hands. Additionally, the phrase "tissues wiping noses cleanly" is more specific, as it specifies what the tissues are being used for.}
    &
    1

    \\ \hline  
    26. Complete the phrase with one of two answer options: dish towel sol1: will wipe away dirty grimy and awful stains from the counter sol2:  will wipe away dry and old paint from the counter
    & 
    \textcolor{ForestGreen}{The correct answer is: "dish towel will wipe away dirty, grimy, and awful stains from the counter."}
    & 
    \textcolor{ForestGreen}{Dish towel sol1: will wipe away dirty, grimy, and awful stains from the counter.}
    & 
    \textcolor{ForestGreen}{The correct answer is sol1: will wipe away dirty grimy and awful stains from the counter.}

    \textcolor{ForestGreen}{Dish towels are typically made of absorbent cotton, which makes them ideal for wiping up spills and stains. They are also durable enough to withstand scrubbing without tearing. While dish towels can be used to wipe away dry paint, they are not as effective as other methods, such as using a putty knife or scraper.}

    \textcolor{ForestGreen}{The phrase will wipe away dry and old paint from the counter is more likely to be used to describe a product like paint thinner or a chemical stripper. These products are designed to break down the bonds of paint, making it easier to remove. Dish towels are not effective at breaking down paint, so they would not be a good choice for this task.}

    \textcolor{ForestGreen}{In conclusion, the phrase dish towel will wipe away dirty grimy and awful stains from the counter is the most accurate and appropriate.}
    &
    1

    \\ \hline  
    27. Complete the phrase with one of two answer options: to make a portable bidet for a bed ridden person inexpensively sol1: fill warm water in an empty water bottle and squeeze gently as needed sol2:  fill warm water in an empty dishwashing liquid bottle and squeeze gently as needed
    & 
    \textcolor{red}{Either solution could work as a makeshift portable bidet for a bedridden person. However, it's important to note that these solutions may not provide as thorough or effective cleansing as a proper bidet or other medical devices designed for this purpose. It may be best to consult with a healthcare professional to explore other options for personal hygiene in this situation.}
    & 
    \textcolor{ForestGreen}{To make a portable bidet for a bedridden person inexpensively, sol2: fill warm water in an empty dishwashing liquid bottle and squeeze gently as needed.}
    & 
    \textcolor{red}{The correct answer is sol1: fill warm water in an empty water bottle and squeeze gently as needed.}

    \textcolor{red}{A dishwashing liquid bottle would be too harsh for use as a portable bidet. The soapy water could irritate the skin, and the bottle could be difficult to squeeze. A water bottle, on the other hand, is soft and easy to squeeze, and the warm water will be gentle on the skin.}
    &
    2

    \\ \hline  
    28. Complete the phrase with one of two answer options: How to properly dry and clean a knife? sol1: After rising the knife with water and soap, use a kitchen towel and then a piece of paper towel to dry the knife sol2:  After rising the knife with water and soap, use a standard towel to dry and polish the knife properly
    & 
    \textcolor{ForestGreen}{I would recommend sol1: After rinsing the knife with water and soap, use a kitchen towel and then a piece of paper towel to dry the knife. This will help to ensure that the knife is dried thoroughly and any remaining moisture is removed. It's important to avoid using a standard towel as it may leave fibers or lint on the blade which could potentially contaminate food or cause damage to the knife.}
    & 
    \textcolor{ForestGreen}{After rising the knife with water and soap, use a kitchen towel and then a piece of paper towel to dry the knife (sol1).}
    & 
    \textcolor{ForestGreen}{The correct answer is sol1: After rising the knife with water and soap, use a kitchen towel and then a piece of paper towel to dry the knife.}

    \textcolor{ForestGreen}{Here are the steps on how to properly dry and clean a knife:}

    \textcolor{ForestGreen}{1. Rinse the knife with warm water and a mild dish soap.}

    \textcolor{ForestGreen}{2. Use a kitchen towel to wipe the knife dry.}

    \textcolor{ForestGreen}{3. Use a piece of paper towel to dry the knife again, paying special attention to the blade.}

    \textcolor{ForestGreen}{4. Store the knife in a dry place.}

    \textcolor{ForestGreen}{Do not use a standard towel to dry and polish the knife. Standard towels can leave behind lint and other particles that can dull the blade. Paper towels are a better choice because they are lint-free and absorb water well.}
    &
    1
    
    \\ \hline  
    29. Complete the phrase with one of two answer options: What can you use to sound dampen your room? sol1: You can use foam panels and securely hang them on your wall. sol2:  You can use a foam mattress and rip it apart and glue it to your wall.
    & 
    \textcolor{ForestGreen}{You can use foam panels and securely hang them on your wall to sound dampen your room.}
    & 
    \textcolor{ForestGreen}{What can you use to sound dampen your room? You can use foam panels and securely hang them on your wall.}
    & 
    \textcolor{ForestGreen}{The correct answer is sol1: You can use foam panels and securely hang them on your wall.}

    \textcolor{ForestGreen}{The other option, sol2: You can use a foam mattress and rip it apart and glue it to your wall, is not a practical or effective way to sound dampen a room. Foam mattresses are not designed to be soundproof, and ripping them apart would create a mess and would not be very effective at reducing noise.}

    \textcolor{ForestGreen}{Foam panels, on the other hand, are specifically designed to absorb sound waves. They are made of a dense material that traps sound energy, preventing it from bouncing back into the room. Foam panels can be hung on walls or ceilings, and they are a relatively easy and affordable way to sound dampen a room.}
    &
    1
    
    \\ \hline  
    30. Complete the phrase with one of two answer options: To mix salt and sugar together, you can sol1: Place them in a bowl and use a soup can sol2:  Place them in a bowl and use a fork
    & 
    \textcolor{ForestGreen}{To mix salt and sugar together, you can sol2: Place them in a bowl and use a fork.}
    & 
    \textcolor{ForestGreen}{To mix salt and sugar together, you can sol2: Place them in a bowl and use a fork.}
    & 
    \textcolor{ForestGreen}{The answer is sol2: Place them in a bowl and use a fork.}

    \textcolor{ForestGreen}{A soup can is too large and heavy to be used to mix salt and sugar together. A fork is the best tool for this task because it is small and easy to maneuver, and it has prongs that can help to break up the clumps of salt and sugar.}

    \textcolor{ForestGreen}{Here are some other ways to mix salt and sugar together:}
    
    \textcolor{ForestGreen}{- Use a whisk.}

    \textcolor{ForestGreen}{- Use a spoon.}

    \textcolor{ForestGreen}{- Use your hands.}

    \textcolor{ForestGreen}{- Use a food processor.}

    \textcolor{ForestGreen}{However, a fork is the most common and easiest way to mix salt and sugar together.}
    &
    2

    \\ \hline

    \multicolumn{5}{|l|}{\textbf{Pep-3K}} 
    
    \\ \hline  
    1. Please judge if this predicate is (likely) plausible or implausible: "dog ride cat"
    & 
    \textcolor{red}{Based on my understanding of language, the predicate "dog ride cat" is implausible. It does not make sense because it is not possible for a dog to ride a cat in the way that the sentence implies. Dogs are typically larger and heavier than cats, and their anatomy is not well-suited for riding on top of another animal. Therefore, it is highly unlikely that this predicate would be used in a meaningful or accurate way.}
    & 
    \textcolor{red}{This predicate "dog ride cat" is likely implausible. Dogs are generally larger than cats, and it would be unusual for a cat to be able to support the weight of a dog. Additionally, the behaviors and interactions of dogs and cats typically do not involve one riding the other.}
    & 
    \textcolor{ForestGreen}{The predicate "dog ride cat" is plausible, but it is not very common. There are many videos and GIFs online of cats riding on dogs, and there are even some dogs who seem to enjoy carrying their cats around. However, it is important to note that cats are typically much smaller and lighter than dogs, so it is not always safe for a dog to carry a cat. If you are thinking of getting a dog and a cat, it is important to introduce them slowly and carefully, and to make sure that they have plenty of space to get away from each other if they need to.}
    &
    1

    \\ \hline  
    2. Please judge if this predicate is (likely) plausible or implausible: "hose squirt rock"
    & 
    \textcolor{ForestGreen}{Based on my language model training and knowledge, the predicate "hose squirt rock" is implausible. This is because a hose is typically used to spray water or other liquids, and it is unlikely that it would be able to squirt a rock. Rocks are typically solid and do not have a fluid-like consistency that can be easily squirted. Additionally, the verb "squirt" typically refers to the forceful ejection of a liquid in a narrow stream, which again does not seem applicable to a rock.}
    & 
    \textcolor{red}{The predicate "hose squirt rock" is somewhat ambiguous and lacks a clear grammatical structure, making it difficult to determine whether it is plausible or implausible. However, it seems to involve a hose, squirting, and a rock. If we interpret this as a hose squirting water at a rock, then it is likely plausible, as hoses are commonly used to squirt water and can be directed at objects like rocks.}
    & 
    \textcolor{ForestGreen}{The predicate "hose squirt rock" is implausible. A hose is a device that is used to deliver water, and rocks are solid objects. It is not possible for a hose to squirt a rock, as the water would not be able to propel the rock through the air.}
    &
    0

    \\ \hline  
    3. Please judge if this predicate is (likely) plausible or implausible: "gorilla grab air"
    & 
    \textcolor{ForestGreen}{Based on my training data and general knowledge of gorillas, I would say that the predicate "gorilla grab air" is implausible. While gorillas are known for their physical strength and dexterity, there is no reason to believe that they would engage in a behavior such as grabbing air, as it has no practical purpose or benefit for them in their natural habitat. Additionally, there is no evidence to suggest that gorillas have any particular interest in or awareness of the concept of air.}
    & 
    \textcolor{ForestGreen}{The predicate "gorilla grab air" is likely implausible. Gorillas are not capable of physically grabbing air as it is a gaseous substance and lacks a solid form for them to hold onto. Gorillas can grab objects or interact with their environment, but they cannot physically grab air.}
    & 
    \textcolor{red}{The predicate "gorilla grab air" is plausible. Gorillas are strong and agile animals, and they are known for their ability to jump. In fact, there is a snowboard trick called the "gorilla grab" that involves grabbing the toe edge of the board with both hands while in the air. So, it is certainly possible for a gorilla to grab air.}
    &
    0

    \\ \hline  
    4. Please judge if this predicate is (likely) plausible or implausible: "clown catch baboon"
    & 
    \textcolor{red}{Based on my training data, the predicate "clown catch baboon" is implausible.}

    \textcolor{red}{Clowns are entertainers and performers, and it is unlikely that they would be involved in catching animals like baboons. Additionally, baboons are typically found in the wild, so the combination of a clown and a baboon in the same context seems unlikely. Therefore, I would judge this predicate as implausible.}
    & 
    \textcolor{red}{This predicate "clown catch baboon" is likely implausible in a typical scenario. Clowns are generally entertainers who perform at events such as circuses, parties, or fairs, and are not typically trained to catch or handle wild animals like baboons. However, it is not entirely impossible, as there could be specific situations or fictional contexts where a clown might catch a baboon.}
    & 
    \textcolor{red}{The predicate "clown catch baboon" is implausible. Clowns are typically not strong or agile enough to catch a baboon, which is a large and powerful animal. Additionally, baboons are known to be aggressive and territorial, so they would likely not be easily captured by a clown.}

    \textcolor{red}{However, there are some scenarios in which it might be possible for a clown to catch a baboon. For example, if the clown had a net or other trapping device, or if the baboon was already injured or restrained. Additionally, if the clown was very skilled and experienced in catching animals, they might be able to do it even without any special equipment.}

    \textcolor{red}{Overall, though, the predicate "clown catch baboon" is still very implausible. It would require a very specific set of circumstances to happen.}
    &
    
    1
    
    \\ \hline  
    5. Please judge if this predicate is (likely) plausible or implausible: "woman tug chef"
    & 
    \textcolor{red}{As an AI language model, I cannot make moral judgments or assumptions about individuals based on their profession or gender. However, the predicate "woman tug chef" is not a grammatically correct sentence, as it lacks a verb and a proper context to be considered a plausible or implausible statement. Could you please provide more information or rephrase the predicate to help me understand what you are trying to convey?}
    & 
    \textcolor{red}{The predicate "woman tug chef" is implausible, as it is not clear what action or relationship it is trying to describe between the three words. For a more plausible predicate, you might consider rephrasing it with a clear verb or preposition to connect the nouns and provide context. For example, "woman assists chef" or "woman pulls chef out of danger" would be more understandable and plausible predicates.}
    & 
    \textcolor{red}{The predicate "woman tug chef" is implausible. It is not likely that a woman would tug a chef. Chefs are typically strong and experienced cooks, and it would be difficult for a woman to tug them. Additionally, the verb "tug" is often used to describe a forceful pulling action, which is not something that would typically be done to a chef.}
    &
    1

    \\ \hline  
    6. Please judge if this predicate is (likely) plausible or implausible: "cup pour rice"
    & 
    \textcolor{red}{As an AI language model, I can tell you that the predicate "cup pour rice" is syntactically valid. However, it is not a complete sentence, but rather a sentence fragment. It lacks a subject and a tense.}

    \textcolor{red}{If you are asking whether the statement is plausible or implausible in terms of describing a common or meaningful action, I would say that it is implausible. While "cup" and "rice" are both nouns and "pour" is a verb, the phrase "cup pour" is not a common collocation, and it is not clear what it would mean in this context.}
    & 
    \textcolor{ForestGreen}{The predicate "cup pour rice" is somewhat ambiguous, but it can be interpreted as a plausible action. If the phrase is describing a situation where a cup is being used to pour rice, then it is plausible. However, the wording could be improved for clarity. A better phrasing might be "pouring rice from a cup" or "using a cup to pour rice."}
    & 
    \textcolor{ForestGreen}{The predicate "cup pour rice" is plausible. It is possible to pour rice into a cup, although it is not the most common way to serve rice. For example, you might pour rice into a cup if you were making a rice pudding or a rice salad.}
    &
    1

    \\ \hline  
    7. Please judge if this predicate is (likely) plausible or implausible: "policeman bend lamp"
    & 
    \textcolor{ForestGreen}{Based on my training data, the predicate "policeman bend lamp" is unlikely to be a plausible statement. It does not make sense in terms of the usual activities that a police officer would engage in, and it also seems physically improbable for a person to bend a lamp. It is possible that there could be a specific context or scenario in which this predicate could make sense, but without more information, it is difficult to determine.}
    & 
    \textcolor{red}{The predicate "policeman bend lamp" is likely implausible. A policeman, being a law enforcement officer, wouldn't typically be involved in bending a lamp. Additionally, most street lamps or indoor lamps are made of sturdy materials and are not easily bent by a single person. However, without more context or information, it is difficult to definitively determine the plausibility of this scenario.}
    & 
    \textcolor{red}{The predicate "policeman bend lamp" is implausible. A policeman is a person who is trained to uphold the law and protect the public. They are not typically expected to bend lamps, as this would not be a productive use of their time or skills. Additionally, lamps are typically made of metal or glass, which would be difficult for a person to bend without the use of tools.}
    &
    1

    \\ \hline  
    8. Please judge if this predicate is (likely) plausible or implausible: "stick shove snake"
    & 
    \textcolor{red}{As an AI language model, I cannot accurately determine the plausibility of the predicate "stick shove snake" without additional context or information about the intention behind the words.}

    \textcolor{red}{If we assume the words are used literally, the action described by the predicate "stick shove snake" does not make sense as snakes are not typically associated with being shoved with a stick. However, if we consider that the words may have been used metaphorically or as part of a fictional narrative, the plausibility of the predicate may change.}
    & 
    \textcolor{ForestGreen}{The predicate "stick shove snake" is ambiguous and lacks proper context or structure. However, if you intended to convey the action of using a stick to shove or move a snake, then it can be considered plausible. People sometimes use sticks to move snakes away from their property or out of their path, either to protect themselves or the snake.}
    & 
    \textcolor{red}{I'm unable to help you with that, as I'm only a language model and don't have the necessary information or abilities.}
    &
    1
    
    \\ \hline  
    9. Please judge if this predicate is (likely) plausible or implausible: "woman whisk air"
    & 
    \textcolor{ForestGreen}{Based on my training data and knowledge of the English language, the predicate "woman whisk air" is grammatically correct and makes sense syntactically. However, it is semantically implausible since "whisking air" does not convey a clear meaning or purpose. While it is possible for a woman to whisk something (such as eggs or cream), the concept of whisking air is not a typical or meaningful action.}
    & 
    \textcolor{red}{This predicate "woman whisk air" is somewhat implausible. The phrasing is vague and it is not immediately clear what action is being described. However, if it is intended to describe a woman using a whisk to mix or stir the air, then it is plausible, though unusual.}
    & 
    \textcolor{ForestGreen}{The predicate "woman whisk air" is implausible. A whisk is a kitchen utensil used to mix ingredients together. It is not possible to whisk air, as air is not a solid or liquid. Air is a gas, and gases do not have a definite shape or volume. Therefore, it would be impossible for a woman to whisk air, as there would be nothing to whisk.}
    &
    0

    \\ \hline  
    10. Please judge if this predicate is (likely) plausible or implausible: "butterfly move gorilla"
    & 
    \textcolor{ForestGreen}{As an AI language model, I would say that the predicate "butterfly move gorilla" is implausible.}

    \textcolor{ForestGreen}{Butterflies and gorillas belong to completely different species and have vastly different physical abilities and behaviors. While butterflies are small, light insects that can flutter their wings to fly, gorillas are large primates that primarily move on the ground and use their strong arms for climbing and swinging.}
    & 
    \textcolor{ForestGreen}{The predicate "butterfly move gorilla" is likely implausible. It suggests that a butterfly can move or cause a gorilla to move, which is highly unlikely given the vast difference in size, strength, and natural behavior between the two species.}
    & 
    \textcolor{ForestGreen}{The predicate "butterfly move gorilla" is implausible. A butterfly is a small, delicate insect with a wingspan of only a few inches. A gorilla is a large, powerful ape that can weigh up to 400 pounds. It is simply not physically possible for a butterfly to move a gorilla.}
    &
    0
    
    \\ \hline  
    11. Please judge if this predicate is (likely) plausible or implausible: "policeman crop tree"
    & 
    \textcolor{red}{Based on my language training and general knowledge, the predicate "policeman crop tree" is implausible or highly unlikely. Policemen are typically associated with law enforcement and maintaining public safety, while crop farming and tree cultivation are not typically within their jurisdiction or job responsibilities. It is possible that a police officer may interact with trees in the context of investigating a crime or responding to a call about a tree-related issue, but the use of the verb "crop" in this sentence suggests an intentional action of cutting or trimming a tree for agricultural purposes, which is not consistent with the duties of a police officer.}
    & 
    \textcolor{red}{The predicate "policeman crop tree" is implausible as it does not form a coherent or meaningful sentence. The words "policeman," "crop," and "tree" seem unrelated in this context and do not convey a clear idea or relationship between them.}
    & 
    \textcolor{red}{The predicate "policeman crop tree" is implausible. A policeman is a law enforcement officer, while a crop is a plant grown for food or other purposes. A tree is a woody plant that typically has a single trunk and branches. It is not plausible that a policeman would be involved in the cultivation of a crop or the care of a tree.}
    &
    1

    \\ \hline  
    12. Please judge if this predicate is (likely) plausible or implausible: "pen drain sink"
    & 
    \textcolor{ForestGreen}{The predicate "pen drain sink" is implausible. The three words in the predicate do not form a coherent sentence or statement. While each word has a clear meaning on its own, when combined in this way, they do not create a logical or meaningful phrase.}
    & 
    \textcolor{ForestGreen}{The predicate "pen drain sink" seems implausible. It is difficult to understand the relationship between these three words as they are, and it is unlikely that a pen would be involved in draining a sink. However, if you provide more context or clarify the intended meaning, I might be able to provide a more accurate judgment.}
    & 
    \textcolor{ForestGreen}{The predicate "pen drain sink" is implausible. A pen is a writing instrument, while a drain is a hole in a sink that allows water to flow out. A sink is a basin used for washing, so it would not make sense to use a pen to drain a sink.}

    \textcolor{ForestGreen}{The word "pen" could be interpreted as a type of drain, but this is not a common usage of the word. A more common type of drain for a sink is a P-trap, which is a U-shaped bend in the drainpipe that traps water to prevent sewer gases from entering the home.}

    \textcolor{ForestGreen}{Therefore, the predicate "pen drain sink" is implausible because it is not a common or logical way to describe a sink drain.}
    &
    0
    
    \\ \hline  
    13. Please judge if this predicate is (likely) plausible or implausible: "man tackle lion"
    & 
    \textcolor{red}{The predicate "man tackle lion" is technically plausible, as it is physically possible for a human to attempt to tackle a lion. However, it is extremely dangerous and would be highly unlikely for a person to successfully tackle a lion without being severely injured or killed. Therefore, while the predicate is technically possible, it is not a recommended or safe action for a person to take.}
    & 
    \textcolor{ForestGreen}{The predicate "man tackle lion" is likely implausible. While it's not impossible for a human to attempt to tackle a lion, it would be extremely dangerous and unlikely to succeed. Lions are much larger, stronger, and faster than humans, and possess sharp teeth and claws that can easily cause severe injury or death. It is highly advised to avoid close encounters with wild predators like lions.}
    & 
    \textcolor{ForestGreen}{The predicate "man tackle lion" is implausible. Lions are large, powerful predators, and a man would be at a significant disadvantage in a physical altercation with one. Even a trained professional would be unlikely to be able to successfully tackle a lion without suffering serious injury.}

    \textcolor{ForestGreen}{There are some cases where a man might be able to tackle a lion, such as if the lion was already injured or if the man had a weapon. However, these cases are rare, and in general it is not advisable to try to tackle a lion.}
    &
    0
    
    \\ \hline  
    14. Please judge if this predicate is (likely) plausible or implausible: "girl catch desert"
    & 
    \textcolor{ForestGreen}{The predicate "girl catch desert" is implausible as it does not make logical sense. "Girl" is a noun referring to a female human, while "catch" is a verb indicating the action of seizing or grabbing something. "Desert" is a noun that refers to a dry and barren area of land. It is not clear how a girl can catch a desert, as a desert is not a physical object that can be caught. Therefore, this predicate is grammatically incorrect and does not make sense.}
    & 
    \textcolor{ForestGreen}{The predicate "girl catch desert" is implausible as it is unclear and grammatically incorrect. However, if you meant "girl catches a creature in the desert," then it would be plausible, as it is possible for someone to catch an animal or insect in a desert environment. If you could provide more context or clarify the intended meaning, I can give you a better judgement.}
    & 
    \textcolor{ForestGreen}{The predicate "girl catch desert" is implausible. The word "catch" in this context means to capture or apprehend something. It is not possible to capture or apprehend a desert. A desert is a large, dry area of land with little vegetation or water. It is not a living thing, so it cannot be caught.}

    \textcolor{ForestGreen}{The only way that the predicate "girl catch desert" could be interpreted as plausible is if it were taken metaphorically. For example, a girl could "catch" a desert by winning a race through a desert, or by capturing the attention of a crowd with a speech about the desert. However, even in these cases, the meaning of the predicate is not literally that the girl has captured or apprehended the desert.}
    
    &
    0

    \\ \hline  
    15. Please judge if this predicate is (likely) plausible or implausible: "bush bury rice"
    & 
    \textcolor{red}{Based on my training data, the predicate "bush bury rice" is implausible. This is because it does not conform to the typical sentence structure and meaning found in the English language.}

    \textcolor{red}{The verb "bush" is not typically used in conjunction with the verb "bury", and the object "rice" does not fit with either verb in this context. Additionally, the sentence lacks any context or explanation for why someone would want to "bush bury rice".}
    & 
    \textcolor{red}{The predicate "bush bury rice" is likely implausible. The phrase does not provide a clear or coherent relationship between the subjects mentioned. It is unclear whether "Bush" refers to a person (e.g., George Bush) or a bush in general, and how it would be connected to burying rice. The predicate seems to be a random combination of words, and thus, it is difficult to ascertain any plausible meaning or context from it.}
    & 
    \textcolor{red}{The predicate "bush bury rice" is implausible. A bush is a plant, and plants do not bury things. They can grow rice, but they cannot bury it. The only way that a bush could bury rice is if it was being controlled by a human or another animal. However, even in that case, it would be a very strange thing to do.}

    \textcolor{red}{The word "bury" means to put something underground, usually in a grave. Rice is a food crop, and it is not typically buried. It is usually grown in fields or planted in pots.}

    \textcolor{red}{There are a few possible scenarios in which the predicate "bush bury rice" could be plausible. For example, if a bush was being used as a container for rice, and the rice was buried in the ground next to the bush, then the predicate would be plausible. However, this is a very specific scenario, and it is not the most likely interpretation of the predicate.}

    \textcolor{red}{Overall, the predicate "bush bury rice" is implausible. It is more likely that the word "bury" was used in a figurative sense, such as "the bush was hiding the rice."}
    &
    1
    
    \\ \hline  
    16. Please judge if this predicate is (likely) plausible or implausible: "skyscraper heat air"
    & 
    \textcolor{red}{As a language model, I cannot definitively say whether the predicate "skyscraper heat air" is plausible or implausible without more context or information.}

    \textcolor{red}{However, as a standalone statement, the predicate seems nonsensical and difficult to interpret. It is possible that the words "skyscraper," "heat," and "air" could be combined in a meaningful way within a particular context, but without that context, it is difficult to judge the plausibility of the statement.}
    & 
    \textcolor{red}{The predicate "skyscraper heat air" is somewhat ambiguous, but if you are referring to the concept that skyscrapers can heat the air around them, it is likely plausible. Skyscrapers can contribute to the urban heat island effect, where built-up areas tend to be warmer than their surrounding rural areas. This can be due to factors such as heat-absorbing materials, energy use, and reduced green spaces. However, without further context, it is difficult to provide a more definitive judgment.}
    & 
    \textcolor{red}{The predicate "skyscraper heat air" is plausible. Skyscrapers can be heated by a variety of methods, including:}

    \textcolor{red}{- Steam heating: This is the most common method of heating skyscrapers. Steam is generated in a boiler and then circulated through pipes to radiators in the building.}

    \textcolor{red}{- Hot water heating: This is similar to steam heating, but instead of steam, hot water is circulated through the pipes.}

    \textcolor{red}{In addition to these methods, skyscrapers can also be heated by the sun's rays. The concrete and glass used in skyscrapers absorb heat from the sun, which can help to heat the air inside the building.}

    \textcolor{red}{Therefore, the predicate "skyscraper heat air" is plausible because there are a number of ways that skyscrapers can be heated.}
    
    &
    0

    \\ \hline  
    17. Please judge if this predicate is (likely) plausible or implausible: "bottle spill smoke"
    & 
    \textcolor{ForestGreen}{Based on my knowledge and understanding of the English language, the predicate "bottle spill smoke" is implausible.}

    \textcolor{ForestGreen}{The reason for this is that the predicate does not make logical sense. A bottle can spill its contents, but it cannot spill smoke, as smoke is a gas and not a liquid. Additionally, the predicate does not convey a clear and coherent meaning, as it lacks a subject and does not provide enough information to understand what is happening or what is being referred to.}
    & 
    \textcolor{ForestGreen}{This predicate, "bottle spill smoke," is implausible. Bottles typically hold liquids or other substances, but they do not typically spill smoke. Smoke is a result of combustion, and not something that can be contained in or spilled from a bottle.}
    & 
    \textcolor{ForestGreen}{The predicate "bottle spill smoke" is implausible. Smoke is a gas, and gases cannot be contained in bottles. If a bottle were to contain smoke, the smoke would eventually escape through the opening of the bottle.}
    &
    0

    \\ \hline  
    18. Please judge if this predicate is (likely) plausible or implausible: "ice cool girl"
    & 
    \textcolor{red}{As an AI language model, I cannot make judgments on whether a predicate is plausible or implausible without further context or information. However, "ice cool girl" can be interpreted in various ways, depending on the context and the intention behind the statement. It could refer to someone who is calm and collected under pressure, or someone who is aloof and detached in their demeanor. Without additional information, it is difficult to determine whether the predicate is plausible or not.}
    & 
    \textcolor{ForestGreen}{The predicate "ice cool girl" is somewhat plausible but could benefit from clarification. It can be interpreted as describing a girl who is calm, composed, and "cool" under pressure, like ice. In this sense, it may be a colloquial or informal way to describe someone with these characteristics. However, without additional context, the meaning might not be immediately clear to everyone.}
    & 
    \textcolor{ForestGreen}{The predicate "ice cool girl" is plausible. The word "ice" can be used to describe someone who is calm, collected, and unfazed by difficult situations. The word "cool" can also be used to describe someone who is calm and collected, but it also has a connotation of being stylish and fashionable. So, a "ice cool girl" would be someone who is both calm and collected, and also stylish and fashionable. This is a plausible combination of qualities, and it is something that many people aspire to be.}
    &
    1

    \\ \hline  
    19. Please judge if this predicate is (likely) plausible or implausible: "bear bite girl"
    & 
    \textcolor{red}{The predicate "bear bite girl" is a grammatically correct sentence. However, without any context or additional information, it is impossible to determine whether it is a plausible or implausible statement.}

    \textcolor{red}{In general, bear attacks on humans can occur, and it is possible for a bear to bite a girl. However, the likelihood of this happening depends on various factors, such as the location, behavior of the bear and the girl, and other environmental and situational factors.}

    \textcolor{red}{Therefore, the statement "bear bite girl" is neither inherently plausible nor implausible and requires additional context or information to determine its likelihood.}
    & 
    \textcolor{ForestGreen}{This predicate, "bear bite girl," is likely plausible. Although bear attacks on humans are relatively rare, they can occur, especially if a bear feels threatened or if humans unknowingly come too close to their territory. It is important for people to exercise caution when in areas where bears may be present, in order to minimize the risk of such encounters.}
    & 
    
    \textcolor{ForestGreen}{The predicate "bear bite girl" is plausible. Bears are wild animals and can be dangerous, especially if they feel threatened. There have been many cases of bears biting people, including children. In fact, in 2020, a 6-year-old girl in Alaska was killed by a bear while camping with her family.}

    \textcolor{ForestGreen}{Of course, not all bear encounters end in tragedy. In many cases, bears will simply run away if they see a human. However, it is important to be aware of the risks involved in interacting with bears, and to take precautions to avoid being bitten.}
    
    &
    1

    \\ \hline  
    20. Please judge if this predicate is (likely) plausible or implausible: "tube ejaculate bird"
    & 
    \textcolor{ForestGreen}{As an AI language model, I cannot make moral judgments or engage in inappropriate content. However, the predicate "tube ejaculate bird" appears to be nonsensical and does not convey a coherent meaning in the English language. It is therefore implausible as a meaningful statement.}
    & 
    \textcolor{ForestGreen}{This predicate "tube ejaculate bird" seems implausible, as it does not form a coherent or meaningful statement. It appears to be a random combination of words that do not convey a clear or logical concept. If you can provide more context or rephrase the statement, I would be happy to help assess its plausibility.}
    & 

    \textcolor{ForestGreen}{The predicate "tube ejaculate bird" is implausible. A tube is an inanimate object, and birds are animals. It is not possible for a tube to ejaculate a bird. The word "ejaculate" means to expel semen, and semen is produced by male animals. Birds do not have semen, so they cannot be ejaculated.}

    \textcolor{ForestGreen}{The only way that the predicate "tube ejaculate bird" could be plausible is if it were used in a metaphorical sense. For example, if a bird were trapped in a tube, it might be said that the tube was "ejaculated" the bird. However, even in this case, the predicate would be somewhat metaphorical, as the tube would not be literally ejaculating the bird.}
    &
    0
    
    \\ \hline  
    21. Please judge if this predicate is (likely) plausible or implausible: "elephant haul nail"
    &    
    \textcolor{ForestGreen}{Based on my training data and general knowledge of the world, the predicate "elephant haul nail" is implausible. Elephants are not known for their ability to haul nails, as they do not have hands or fingers to manipulate objects. Additionally, hauling a nail would likely require fine motor skills that elephants do not possess. Therefore, it is unlikely that an elephant would be able to haul a nail.}
    & 
    \textcolor{ForestGreen}{The predicate "elephant haul nail" is implausible. The phrase is not clear in its meaning and does not provide enough context to determine a logical or coherent scenario. It appears to be a combination of unrelated words. If you could provide more context or clarify the intended meaning, I would be able to provide a better judgment.}
    & 
    \textcolor{ForestGreen}{The predicate "elephant haul nail" is implausible. Elephants are not known for their ability to haul nails. They are large and strong animals, but they are not typically used for manual labor. Nails are also relatively small and lightweight objects, so it is unlikely that an elephant would need to use its strength to haul one.}
    &
    0

    \\ \hline  
    22. Please judge if this predicate is (likely) plausible or implausible: "rabbit chew carrot"
    & 
    \textcolor{ForestGreen}{The predicate "rabbit chew carrot" is likely plausible. It describes a common behavior that is typically associated with rabbits - chewing on carrots. However, it should be noted that the sentence is not a complete statement and lacks a subject. To make it a complete sentence, a subject needs to be added, for example: "The rabbit chews on the carrot."}
    & 
    \textcolor{ForestGreen}{The predicate "rabbit chew carrot" is likely plausible, as rabbits are known to eat vegetables, including carrots. Although carrots should not be their primary food source, it is a common and plausible sight to see a rabbit chewing on a carrot.}
    & 
    
    \textcolor{ForestGreen}{The predicate "rabbit chew carrot" is plausible. Rabbits are herbivores and carrots are a good source of nutrients for them. In fact, carrots are one of the most popular foods for rabbits. They can eat the entire carrot, including the skin. However, it is important to note that rabbits should not eat too many carrots, as they can become overweight. A good rule of thumb is to feed your rabbit no more than one carrot per day.}
    &
    1

    \\ \hline  
    23. Please judge if this predicate is (likely) plausible or implausible: "pencil hang rope"
    & 
    \textcolor{red}{Based on the current understanding of the English language, the predicate "pencil hang rope" is implausible and does not convey a clear meaning.}

    \textcolor{red}{A possible interpretation of this predicate could be that a pencil is hanging from a rope. However, this interpretation is unlikely due to the use of "hang" as an intransitive verb, which means that it requires a preposition such as "from" to indicate what is being hung. Additionally, "rope" is typically a noun, not a verb, so the predicate does not follow a typical grammatical structure.}
    & 
    \textcolor{red}{The predicate "pencil hang rope" seems implausible as it doesn't form a coherent or clear statement. It appears to be a combination of unrelated words. However, if you could provide more context or rephrase the predicate, I'd be happy to help evaluate its plausibility.}
    & 
    \textcolor{red}{The predicate "pencil hang rope" is implausible. A pencil is a small, thin object that is not designed to be hung from a rope. It would be difficult to attach a pencil to a rope in a way that would allow it to hang securely. Additionally, there is no practical reason to hang a pencil from a rope. Pencils are typically used for writing or drawing, and they are not typically hung from ropes.}

    \textcolor{red}{However, there are some contexts in which the predicate "pencil hang rope" could be plausible. For example, if the pencil were attached to a rope as part of a piece of art, or if the pencil were being used as a weight to keep a rope taut. In these cases, the pencil would be hanging from the rope for a specific purpose, and it would be securely attached to the rope.}

    \textcolor{red}{Overall, the predicate "pencil hang rope" is implausible in most contexts. However, there are some specific contexts in which it could be plausible.}
    &
    1

    \\ \hline  
    24. Please judge if this predicate is (likely) plausible or implausible: "ant sew curtain"
    & 
    \textcolor{ForestGreen}{As an AI language model, I can say that the predicate "ant sew curtain" is implausible.}

    \textcolor{ForestGreen}{This sentence does not make much sense as it is not a grammatically correct sentence in English. It appears to be a sequence of three random words put together without any clear meaning or context. Ants are not typically known for sewing, and the phrase "sew curtain" does not convey a clear action or idea.}
    
    \textcolor{ForestGreen}{However, it is possible that this sequence of words could have a specific context or meaning in a different language or a specialized field.}
    & 
    \textcolor{ForestGreen}{This predicate, "ant sew curtain," is likely implausible. Ants are insects that do not possess the necessary physical capabilities or cognitive understanding to sew curtains. Sewing requires fine motor skills and the ability to manipulate tools, which ants lack.}
    & 
    \textcolor{ForestGreen}{The predicate "ant sew curtain" is implausible. Ants do not have the ability to sew. They do not have the necessary opposable thumbs or the dexterity to manipulate needles and thread. Additionally, ants do not have the cognitive ability to understand the concept of sewing or to follow the instructions required to sew a curtain.}

    \textcolor{ForestGreen}{There are some cases where ants have been observed to use tools, such as carrying leaves to build bridges or using sticks to gather food. However, these cases involve simple tools that can be manipulated with the ant's mouthparts. Sewing requires a much more complex tool that cannot be manipulated by an ant's mouthparts.}

    \textcolor{ForestGreen}{Therefore, it is highly implausible that an ant could sew a curtain.}
    &
    0

    \\ \hline  
    25. Please judge if this predicate is (likely) plausible or implausible: "air cool house"
    & 
    \textcolor{red}{As a language model, I cannot make a judgment about the plausibility of a predicate without further context.}
    & 
    \textcolor{ForestGreen}{The predicate "air cool house" is somewhat plausible, as it seems to suggest that air can cool a house. However, the phrasing is incomplete and unclear. A more plausible and clear predicate would be "air circulation cools a house" or "air conditioning cools a house." These options clearly convey the concept of using air to lower the temperature in a home.}
    & 
    \textcolor{ForestGreen}{The predicate "air cool house" is plausible. Air can cool a house by moving the warm air out of the house and replacing it with cooler air. This can be done through a variety of methods, such as opening windows, using fans, or running an air conditioner.}
    &
    1

    \\ \hline  
    26. Please judge if this predicate is (likely) plausible or implausible: "man pluck string"
    & 
    \textcolor{ForestGreen}{Based on the predicate "man pluck string" alone, it is a plausible predicate. It describes a possible action that a man could perform - plucking a string. However, without more context, it is difficult to determine the likelihood or probability of this predicate being true in any given situation.}
    & 
    \textcolor{ForestGreen}{The predicate "man pluck string" is likely plausible, as it suggests a man plucking a string, which could be a reference to playing a stringed musical instrument, such as a guitar or violin.}
    & 
    \textcolor{ForestGreen}{The predicate "man pluck string" is plausible. It is a common occurrence for people to pluck strings, especially on musical instruments such as guitars, harps, and violins. Plucking a string causes it to vibrate, which produces a sound. The sound produced will depend on the type of string, the length of the string, and the force with which the string is plucked.}
    &
    1

    \\ \hline  
    27. Please judge if this predicate is (likely) plausible or implausible: "spatula knead water"
    & 
    \textcolor{ForestGreen}{As an AI language model, I would judge the predicate "spatula knead water" as implausible. A spatula is typically used for stirring, mixing, or scraping ingredients, but not for kneading. Kneading is a technique used to work dough or other food substances by pressing, folding, and stretching. Moreover, water is a liquid, and kneading is a physical process that typically involves solid or semi-solid materials.}
    & 
    \textcolor{ForestGreen}{The predicate "spatula knead water" is likely implausible. A spatula is a flat, flexible tool used for flipping, lifting, or scraping food, while kneading is a process of working and stretching dough, typically with hands or a mixer. Water, being a liquid, cannot be kneaded, and a spatula is not a suitable tool for this action either.}
    & 
    \textcolor{ForestGreen}{The predicate "spatula knead water" is implausible. A spatula is a kitchen utensil used for cooking, and it is not designed to knead water. Kneading is a process of mixing and shaping dough, and it is typically done with the hands or with a dough hook. A spatula is not the right tool for kneading water, and it would not be able to achieve the desired results.}
    &
    0

    \\ \hline  
    28. Please judge if this predicate is (likely) plausible or implausible: "cow open cream"
    & 
    \textcolor{ForestGreen}{As an AI language model, I would judge the predicate "cow open cream" to be implausible. It does not make logical sense as cows do not typically open containers of cream. However, it is possible that this predicate could be interpreted in a different context or with additional information.}
    & 
    \textcolor{ForestGreen}{The predicate "cow open cream" is likely implausible. The phrase is unclear and does not form a coherent or meaningful statement. Cows do not typically have the ability to open things, and "cream" is a dairy product produced from milk, not something that can be opened like a container.}
    & 
    \textcolor{ForestGreen}{The predicate "cow open cream" is implausible. The word "open" can mean "not pregnant" in the context of dairy cows. So, a cow that is open cannot produce cream, which is a byproduct of milk production.}
    
    &
    0

    \\ \hline  
    29. Please judge if this predicate is (likely) plausible or implausible: "child pack truck"
    & 
    \textcolor{red}{As a language model, I cannot make a conclusive judgment on the plausibility of this predicate without more information or context.}
    & 
    \textcolor{red}{The predicate "child pack truck" is somewhat ambiguous, so it's difficult to definitively determine its plausibility. If the intended meaning is that a child is packing a truck, it could be plausible, depending on the child's age and abilities. However, if the intended meaning is that a child is being packed into a truck, the statement is less plausible and raises ethical and safety concerns. Please provide more context or clarify the intended meaning for a more accurate judgment.}
    & 

    \textcolor{red}{The predicate "child pack truck" can be interpreted in two ways:}

    \textcolor{red}{- A child is packing a truck. This is plausible, as children often play with toy trucks and other vehicles. They might pack the truck with toys, blocks, or other objects.}

    \textcolor{red}{- A truck is packing a child. This is implausible, as trucks are not typically designed to carry children. It would be difficult and dangerous to pack a child into a truck.}

    \textcolor{red}{In the context of your search results, it is clear that you are referring to the first interpretation. Therefore, the predicate "child pack truck" is plausible.}
    
    &
    0

    \\ \hline  
    30. Please judge if this predicate is (likely) plausible or implausible: "policeman breed bird"
    & 
    \textcolor{red}{The predicate "policeman breed bird" is implausible because it suggests that a policeman is involved in the act of breeding birds, which is not a typical duty or activity associated with policing. While a policeman may have a hobby of breeding birds or may engage in bird breeding as a personal interest outside of their profession, the predicate as it stands is not a commonly observed or expected statement.}
    & 
    \textcolor{red}{The predicate "policeman breed bird" is implausible. Policemen are law enforcement officers, and their primary duties involve maintaining public order and safety, enforcing the law, and preventing, detecting, and investigating criminal activities. Breeding birds is not part of their job description or responsibilities.}
    & 
    \textcolor{red}{The predicate "policeman breed bird" is implausible.}

    \textcolor{red}{- Policemen are not typically involved in breeding birds. Their job is to enforce the law, not to breed animals.}

    \textcolor{red}{- There are no known breeds of birds that are specifically bred by policemen. Birds are typically bred by bird enthusiasts, not by law enforcement officers.}

    \textcolor{red}{- The phrase "policeman breed bird" is not commonly used. If you were to search for this phrase on the internet, you would not find many results.}

    \textcolor{red}{Therefore, it is implausible that a policeman would breed a bird.}
    &
    1

    \\ \hline
    
    \end{longtable}

\end{center}

\newpage

\begin{center}
    \small
    \begin{longtable}{|p{0.2\linewidth}|p{0.2\linewidth}
                      |p{0.2\linewidth}|p{0.2\linewidth}
                      |p{.05\linewidth}|}
    \caption{Causal reasoning prompts} 
    \label{tab:causal_reasoning}\\
    \hline
    \multicolumn{5}{|c|}{\textbf{Causal Reasoning}} \\ \hline
    \textbf{Prompt} & \textbf{GPT-3.5 Answer} & \textbf{GPT-4 Answer}& \textbf{BARD Answer}& \textbf{Gold} \\ \hline
    \multicolumn{5}{|l|}{\textbf{Dataset: E-Care }} 

    \\ \hline  
    1. Given the premise: Typhoons frequently occur in coastal areas. Determine which of the two hypothesis options, better explains the cause of the premise. hyp1: The rainy season is here. hyp2: The weather is hot in summer.
    & 
    \textcolor{ForestGreen}{The hypothesis that better explains the premise "Typhoons frequently occur in coastal areas" is Hypothesis 1: "The rainy season is here." Typhoons are often associated with heavy rainfall and strong winds, and the rainy season is a time of increased precipitation, which can lead to typhoon formation. In contrast, Hypothesis 2, "The weather is hot in summer," does not provide a clear explanation for the occurrence of typhoons in coastal areas. While it is true that typhoons often occur during the summer months, it is the increased rainfall and atmospheric conditions during the rainy season that are the primary cause of typhoon formation.}
    & 
    \textcolor{ForestGreen}{Hypothesis 1 (The rainy season is here) better explains the cause of the premise (Typhoons frequently occur in coastal areas). }

    \textcolor{ForestGreen}{Typhoons are tropical cyclones that form over warm ocean waters in the western Pacific, usually during the rainy season. The increased evaporation and high moisture content during the rainy season create the necessary conditions for typhoons to form and strengthen. }

    \textcolor{ForestGreen}{While hypothesis 2 (The weather is hot in summer) is related to typhoon formation due to the warm ocean temperatures, it does not fully explain the cause of the premise as typhoons are more specifically linked to the rainy season rather than just the hot weather in summer.}
    & 
    \textcolor{ForestGreen}{The hypothesis that better explains the premise is hyp1: The rainy season is here.}

    \textcolor{ForestGreen}{Typhoons are tropical cyclones that form over warm, tropical waters. They require a minimum sea surface temperature of 26°C (79°F) to form. The rainy season is typically the time of year when sea surface temperatures are warmest, so it is more likely that typhoons will form during this time.}

    \textcolor{ForestGreen}{The hypothesis that the weather is hot in summer is not as strong of an explanation. While hot weather can contribute to the formation of typhoons, it is not the only factor. Typhoons also require high levels of moisture, which are not always present during hot weather.}

    \textcolor{ForestGreen}{In addition, the rainy season is a more specific time period than summer. This makes it a more likely explanation for the premise, as it narrows down the possible time frame for when typhoons are likely to occur.}
    &
    1

    \\ \hline  
    2. Given the premise: The blanket was dirty. Determine which of the two hypothesis options, better explains the effect of the premise. hyp1: He cleaned the surfaces. hyp2: He cleaned the inside of the blanket. 
    & 
    \textcolor{red}{Neither hypothesis completely explains the effect of the premise "the blanket was dirty." The first hypothesis, "he cleaned the surfaces," suggests that the dirt was on the surface of the blanket and that it was removed, but it doesn't account for the possibility that the inside of the blanket may still be dirty. The second hypothesis, "he cleaned the inside of the blanket," addresses the possibility that the inside of the blanket may be dirty, but it doesn't account for the possibility that the surface of the blanket may also be dirty.}

    \textcolor{red}{It's possible that both hypotheses are true, and that he cleaned both the surfaces and the inside of the blanket to make it completely clean. However, based on the information given, we can't say for certain which hypothesis is more likely.}
    & 
    \textcolor{ForestGreen}{Hyp1: He cleaned the surfaces.}

    \textcolor{ForestGreen}{This hypothesis better explains the effect of the premise. If the blanket was dirty, it is more likely that the surfaces it came into contact with would also be dirty or contaminated. Therefore, cleaning the surfaces would be a logical response to the dirty blanket. Hyp2, cleaning the inside of the blanket, does not directly address the issue of the blanket being dirty, as it focuses on the interior rather than the exterior of the blanket.}
    & 
    \textcolor{red}{Hypothesis 2, "He cleaned the inside of the blanket," better explains the effect of the premise, "The blanket was dirty." This is because the inside of a blanket is typically not exposed to the same amount of dirt and grime as the outside, so it would be more likely to remain dirty if only the surfaces were cleaned. Additionally, if the inside of the blanket was not cleaned, it could still make the person who slept in it dirty, even if the outside of the blanket looked clean.}

    \textcolor{red}{Hypothesis 1, "He cleaned the surfaces," is less likely to explain the effect of the premise because it is more likely that the surfaces of a blanket would be cleaned than the inside. Additionally, even if the surfaces of the blanket were cleaned, the inside could still be dirty, which would mean that the blanket would still be dirty overall.}

    \textcolor{red}{In conclusion, hypothesis 2, "He cleaned the inside of the blanket," is a better explanation for the effect of the premise, "The blanket was dirty."}
    &
    1
    
    \\ \hline  
    3. Given the premise: This patient had a generalized  acute viral disease caused by the mumps virus. Determine which of the two hypothesis options, better explains the effect of the premise. hyp1: The patient was diagnosed with the ischemia. hyp2: He was assigned to the ward of mumps. 
    & 
    \textcolor{ForestGreen}{Based on the premise provided, it is more likely that hypothesis 2, "He was assigned to the ward of mumps," better explains the effect of the premise.}

    \textcolor{ForestGreen}{The premise clearly states that the patient had an acute viral disease caused by the mumps virus, indicating that the patient was infected with the virus. This does not support hypothesis 1, which suggests the patient was diagnosed with ischemia, a condition caused by reduced blood flow to tissues. Ischemia is not a known complication of mumps infection.}

    \textcolor{ForestGreen}{Hypothesis 2, on the other hand, suggests that the patient was placed in a ward where other patients were infected with mumps, which could have led to the patient contracting the virus. This aligns with the information provided in the premise that the patient had an acute viral disease caused by the mumps virus.}

    \textcolor{ForestGreen}{Therefore, based on the given premise, hypothesis 2 is a better explanation for the effect of the patient's condition.}
    & 
    \textcolor{ForestGreen}{Based on the given premise, Hypothesis 2 better explains the effect of the premise: "He was assigned to the ward of mumps." Since the patient had a generalized acute viral disease caused by the mumps virus, it is more likely that they would be assigned to a ward specific to mumps for treatment, rather than being diagnosed with ischemia, which is not directly related to mumps.}
    & 
    \textcolor{ForestGreen}{The hypothesis that better explains the effect of the premise is hyp2: He was assigned to the ward of mumps.}

    \textcolor{ForestGreen}{The premise states that the patient had a generalized acute viral disease caused by the mumps virus. This means that the patient is infected with the mumps virus and is showing symptoms of the disease. The symptoms of mumps can include fever, headache, muscle aches, fatigue, parotitis (swelling of the parotid glands), and pancreatitis (inflammation of the pancreas).}

    \textcolor{ForestGreen}{Hypothesis 1, the patient was diagnosed with ischemia, is not a good explanation for the effect of the premise. Ischemia is a condition that occurs when blood flow to an organ or tissue is reduced or blocked. It is not a symptom of mumps, and it is not caused by the mumps virus.}

    \textcolor{ForestGreen}{Hypothesis 2, he was assigned to the ward of mumps, is a better explanation for the effect of the premise. This is because patients who are infected with the mumps virus are typically assigned to a ward that specializes in treating mumps patients. This allows the patients to receive the care they need from doctors and nurses who are experienced in treating mumps.}

    \textcolor{ForestGreen}{In addition, the ward of mumps is typically equipped with the necessary facilities to care for patients with mumps, such as isolation rooms and equipment for monitoring patients. Therefore, hypothesis 2 is a better explanation for the effect of the premise.}
    &
    2

    \\ \hline  
    4. Given the premise: My mom tells me it is stigma. Determine which of the two hypothesis options, better explains the cause of the premise. hyp1: I want to know the name of female reproductive structures in flowers. hyp2: I don't know what CM refers to.
    & 
    \textcolor{red}{Neither of the two hypotheses presented seem to directly relate to the premise that "my mom tells me it is stigma". It is necessary to provide more context or information. }
    & 
    \textcolor{red}{Hyp2: I don't know what CM refers to, better explains the cause of the premise. In this case, the premise is related to a lack of knowledge about a term or concept (CM), which may lead to stigma or misconceptions. Hyp1, which is about wanting to know the name of female reproductive structures in flowers, is not directly related to the concept of stigma.}
    & 
    \textcolor{ForestGreen}{The hypothesis that better explains the cause of the premise is hyp1: I want to know the name of female reproductive structures in flowers.}

    \textcolor{ForestGreen}{The word "stigma" is a technical term in botany that refers to the female reproductive structure in flowers. It is the sticky part of the flower that the pollen lands on, and it is where fertilization takes place.}

    \textcolor{ForestGreen}{The hypothesis that hyp2: I don't know what CM refers to is less likely to be the cause of the premise because CM is not a botanical term. It is more likely that someone would ask "what is CM?" if they were unfamiliar with the abbreviation, rather than if they were asking about the name of the female reproductive structure in flowers.}

    \textcolor{ForestGreen}{In addition, the fact that the mom is the one who is telling the person about the stigma suggests that the person is asking about a botanical term, rather than a medical term.}

    \textcolor{ForestGreen}{Therefore, the hypothesis that hyp1: I want to know the name of female reproductive structures in flowers is the better explanation for the premise.}
    
    &
    1

    \\ \hline  
    5. Given the premise: He was diagnosed with Leptospirosis. Determine which of the two hypothesis options, better explains the cause of the premise. hyp1: This little man abused inhalants in his life. hyp2: This boy was exposed to contaminated rat urine. 
    & 
    \textcolor{ForestGreen}{Based on the information provided, the second hypothesis appears to be a better explanation for the cause of the man's Leptospirosis diagnosis.}

    \textcolor{ForestGreen}{Leptospirosis is a bacterial infection that is often transmitted through contact with the urine of infected animals, especially rats. The symptoms of the disease can include fever, headache, muscle aches, and vomiting.}

    \textcolor{ForestGreen}{While inhalant abuse can have harmful effects on the body, there is no known link between inhalant abuse and Leptospirosis. On the other hand, exposure to contaminated rat urine is a well-known risk factor for contracting the disease.}

    \textcolor{ForestGreen}{Therefore, unless there is additional information that suggests inhalant abuse as a possible cause of the Leptospirosis diagnosis, the second hypothesis appears to be the more likely explanation.}
    & 
    \textcolor{ForestGreen}{Based on the information provided, hypothesis 2 (This boy was exposed to contaminated rat urine) better explains the cause of the premise. Leptospirosis is a bacterial infection caused by the Leptospira bacteria, which can be found in the urine of infected animals, such as rats. The disease is typically contracted through contact with contaminated water, soil, or direct contact with infected animals' urine. Inhalant abuse, as described in hypothesis 1, is not directly linked to Leptospirosis.}
    & 
    \textcolor{ForestGreen}{The hypothesis that better explains the cause of the premise is hyp2: This boy was exposed to contaminated rat urine.}

    \textcolor{ForestGreen}{Leptospirosis is a bacterial disease that is spread through the urine of infected animals, including rats. Inhalants, on the other hand, are not known to cause leptospirosis. In fact, there is no evidence that inhalants can even weaken the immune system enough to make someone more susceptible to leptospirosis.}

    \textcolor{ForestGreen}{Therefore, it is more likely that the boy was exposed to contaminated rat urine, which caused him to contract leptospirosis.}

    \textcolor{ForestGreen}{Here are some additional reasons why hyp2 is a better explanation:}

    \textcolor{ForestGreen}{- Leptospirosis is a common disease in rats.}

    \textcolor{ForestGreen}{- Rat urine can contaminate water and soil, which can then be spread to humans through contact with contaminated water or soil.}

    \textcolor{ForestGreen}{- Leptospirosis can cause a wide range of symptoms, including fever, headache, muscle aches, and jaundice. These are all symptoms that the boy was experiencing.}

    \textcolor{ForestGreen}{In contrast, there is no evidence that inhalants can cause any of the symptoms of leptospirosis. Additionally, inhalants are not known to be associated with any other diseases that have similar symptoms to leptospirosis.}

    \textcolor{ForestGreen}{Therefore, the hypothesis that the boy was exposed to contaminated rat urine is a better explanation for the premise that he was diagnosed with leptospirosis.}
    &
    2

    \\ \hline  
    6. Given the premise: Her eyes are affected. Determine which of the two hypothesis options, better explains the cause of the premise. hyp1: Lucy suffers from diabetes. hyp2: Sophia has some small spherical opaque white protrusions on her face. 
    & 
    \textcolor{red}{Neither of the two hypothesis options seems to be a good explanation for the premise "Her eyes are affected." The premise does not provide enough information to determine whose eyes are affected or what is causing the issue.}

    \textcolor{red}{However, if we assume that "her" refers to Lucy or Sophia, we can evaluate the two hypothesis options in terms of their plausibility as causes of the eye problem:}

    \textcolor{red}{- Hypothesis 1: Lucy suffers from diabetes.}
    \textcolor{red}{If Lucy has uncontrolled diabetes, it can lead to various complications affecting the eyes, such as diabetic retinopathy, cataracts, and glaucoma. However, the premise only mentions that "her eyes are affected," without providing any specific symptoms or details that could point to a diabetic eye problem. Therefore, this hypothesis cannot be confirmed or rejected based on the given information.}

    \textcolor{red}{- Hypothesis 2: Sophia has some small spherical opaque white protrusions on her face.}
    \textcolor{red}{The presence of white protrusions on Sophia's face does not necessarily explain why "her eyes are affected." It is possible that the eye problem is unrelated to the skin issue, or that it is caused by a different condition altogether. Therefore, this hypothesis is also not a good explanation for the premise.}

    \textcolor{red}{In summary, without more information, it is impossible to determine which hypothesis, if any, is a better explanation for the premise "Her eyes are affected."}
    & 
    \textcolor{red}{Based on the given premise, hypothesis 2 better explains the cause of the affected eyes. In hypothesis 2, the presence of small spherical opaque white protrusions on Sophia's face could potentially be affecting her eyesight, either by obstructing her vision or causing other eye-related issues. }

    \textcolor{red}{Hypothesis 1 mentions that Lucy suffers from diabetes, but since the premise is about "her" eyes being affected, it is not clear if "her" refers to Lucy or another person. Furthermore, while diabetes can lead to eye-related complications, it is not as direct a cause as the physical obstruction in hypothesis 2.}
    & 
    \textcolor{ForestGreen}{Hypothesis 1: Lucy suffers from diabetes.}

    \textcolor{ForestGreen}{Diabetes is a chronic disease that affects how the body turns food into energy. It can cause a number of health problems, including damage to the eyes. Diabetes can cause the blood vessels in the eyes to narrow, which can lead to blurry vision, difficulty seeing at night, and even blindness.}

    \textcolor{ForestGreen}{Hypothesis 2: Sophia has some small spherical opaque white protrusions on her face.}

    \textcolor{ForestGreen}{These protrusions are called xanthelasmas. They are fatty deposits that can form on the eyelids. Xanthelasmas are not usually harmful, but they can be a sign of high cholesterol or other health problems.}

    \textcolor{ForestGreen}{Which hypothesis is more likely to explain the premise?}

    \textcolor{ForestGreen}{Hypothesis 1 is more likely to explain the premise. Diabetes is a well-known cause of eye problems, and it is more common than xanthelasmas. Additionally, the premise only mentions that the person's eyes are affected, not that they have any other symptoms. This makes it more likely that the cause is diabetes, as xanthelasmas often have other symptoms, such as high cholesterol.}

    \textcolor{ForestGreen}{However, it is still possible that hypothesis 2 is correct. If the person has other symptoms of high cholesterol, such as chest pain or shortness of breath, then hypothesis 2 is more likely. Ultimately, more information is needed to determine which hypothesis is correct.}
    &
    1

    \\ \hline  
    7. Given the premise: Lucy has a better awareness than Tom when doing work. Determine which of the two hypothesis options, better explains the effect of the premise. hyp1: She is proved to have a high level of spiritual awareness. hyp2: Lucy was more successful than Tom. 
    & 
    \textcolor{ForestGreen}{Based on the premise that Lucy has better awareness than Tom when doing work, the hypothesis that better explains this effect is hypothesis 2: Lucy was more successful than Tom. }

    \textcolor{ForestGreen}{The premise suggests that Lucy performs better than Tom when doing work, which is indicative of higher levels of success in the task at hand. It does not provide any information that would support the idea that Lucy has a higher level of spiritual awareness than Tom. Therefore, hypothesis 1 is not as likely to be true as hypothesis}
    & 
    \textcolor{ForestGreen}{Based on the given premise, Hypothesis 2 (Lucy was more successful than Tom) better explains the effect of the premise. The premise states that Lucy has a better awareness when doing work, which would likely contribute to her being more successful in the tasks or jobs she undertakes. Hypothesis 1, which mentions spiritual awareness, does not directly relate to the work context mentioned in the premise.}
    & 
    \textcolor{ForestGreen}{Hypothesis 2, Lucy was more successful than Tom, better explains the effect of the premise.}

    \textcolor{ForestGreen}{- Spiritual awareness is not directly related to work performance. There is no evidence that people with high spiritual awareness are more aware of their surroundings or more attentive to detail than people with low spiritual awareness.}

    \textcolor{ForestGreen}{- Success, on the other hand, is a direct measure of work performance. If Lucy is more successful than Tom, it is likely because she is more aware of her surroundings and more attentive to detail. She is also likely to be more motivated and driven to succeed.}

    \textcolor{ForestGreen}{In addition, hypothesis 2 is more specific and measurable than hypothesis 1. It is possible to quantify Lucy's success by looking at her sales figures, her number of promotions, or her overall performance reviews. It is not possible to quantify her spiritual awareness in the same way.}

    \textcolor{ForestGreen}{Therefore, hypothesis 2, Lucy was more successful than Tom, is a better explanation of the effect of the premise.}
    &
    2

    \\ \hline  
    8. Given the premise: He can't see clearly. Determine which of the two hypothesis options, better explains the effect of the premise. hyp1: There must be something wrong with his nose. hyp2: There must be something wrong with the nuclei in his brainstem. 
    & 
    \textcolor{red}{Based on the premise "He can't see clearly," neither of the two hypothesis options is likely to provide an adequate explanation, as both options are unrelated to vision. }
    
    \textcolor{red}{Hypothesis 1 states that there is something wrong with his nose, which is unlikely to affect his vision directly. Problems with the nose, such as congestion or a deviated septum, can affect the sense of smell, but not typically vision.}
    
    \textcolor{red}{Hypothesis 2 states that there is something wrong with the nuclei in his brainstem, which is also unlikely to be the cause of vision problems. The brainstem controls many vital functions such as breathing and heart rate, but it does not have a direct role in vision.}
    
    \textcolor{red}{Therefore, a more appropriate hypothesis for explaining the premise "He can't see clearly" might be that there is something wrong with his eyes or the visual processing pathways in his brain.}
    & 
    \textcolor{ForestGreen}{Hyp2: There must be something wrong with the nuclei in his brainstem.}

    \textcolor{ForestGreen}{This hypothesis better explains the effect of the premise "He can't see clearly." Visual processing is related to the brain and its various components, including the brainstem. While the statement doesn't directly pinpoint the exact cause, it is more likely that the issue is related to the brain, rather than the nose, which is primarily involved in the sense of smell.}
    & 
    \textcolor{ForestGreen}{The hypothesis that better explains the effect of the premise is hyp2: There must be something wrong with the nuclei in his brainstem.}

    \textcolor{ForestGreen}{Here are some additional reasons why hyp2 is a better hypothesis:}

    \textcolor{ForestGreen}{- It is more specific. It identifies the specific part of the brain that is likely to be involved in the problem.}

    \textcolor{ForestGreen}{- It is more testable. There are ways to test whether there is something wrong with the nuclei in the brainstem. For example, doctors can use imaging tests to look for abnormalities in the brainstem.}

    \textcolor{ForestGreen}{- It is more parsimonious. It does not require us to make any assumptions about the nose that are not supported by evidence.}
    &
    2

    \\ \hline  
    9. Given the premise: He was considered as one of the greatest hero in the history. Determine which of the two hypothesis options, better explains the cause of the premise. hyp1: The killer gave John too much laxative. hyp2: Yuan Longping has made a great contribution to solving the problem of people's starving.
    & 
    \textcolor{ForestGreen}{Based on the premise "He was considered as one of the greatest hero in history," it is more likely that Hypothesis 2 (Yuan Longping has made a great contribution to solving the problem of people's starving) better explains the cause of the premise. This is because being considered one of the greatest heroes in history typically involves making a significant impact on society, and solving a major problem such as world hunger would certainly qualify as such an impact. }

    \textcolor{ForestGreen}{On the other hand, Hypothesis 1 (The killer gave John too much laxative) is not a likely cause for someone to be considered a hero in history. While it is possible that John may have survived or overcome a difficult situation related to the laxative, it is unlikely that it would have led to him being considered a hero in history.}
    & 
    \textcolor{ForestGreen}{Hypothesis 2 (Yuan Longping has made a great contribution to solving the problem of people's starving) better explains the cause of the premise. This hypothesis is more relevant to the premise as it focuses on a significant positive contribution that would lead someone to be considered as one of the greatest heroes in history. Hypothesis 1, on the other hand, is unrelated to the premise, as it discusses a negative event that would not typically result in someone being considered a great hero.}
    & 
    \textcolor{ForestGreen}{Hypothesis 2, Yuan Longping has made a great contribution to solving the problem of people's starving, better explains the cause of the premise He was considered as one of the greatest hero in the history.}

    \textcolor{ForestGreen}{Hypothesis 1, The killer gave John too much laxative, is not relevant to the premise. It is not clear how giving someone too much laxative would make them a hero.}

    \textcolor{ForestGreen}{Hypothesis 2, on the other hand, is directly related to the premise. Yuan Longping is a Chinese agricultural scientist who is credited with developing the hybrid rice, which has helped to increase rice production and reduce hunger in many parts of the world. His work has had a profound impact on the lives of millions of people, and he is rightly considered to be a hero.}

    \textcolor{ForestGreen}{Therefore, hypothesis 2 is the better explanation for the cause of the premise.}
    &
    2

    \\ \hline  
    10. Given the premise: Luna cares about her dog. Determine which of the two hypothesis options, better explains the cause of the premise. hyp1: Luna got a collie. hyp2: Sean has a pet dog. 
    & 
    \textcolor{ForestGreen}{Neither of the two hypotheses directly explains the cause of the premise that Luna cares about her dog. However, Hypothesis 1, "Luna got a collie," may provide some insight into why Luna cares about her dog. }
    
    \textcolor{ForestGreen}{If Luna got a collie recently, it is possible that she is still in the early stages of bonding with her new pet and may be especially attentive to its needs. Luna may also have chosen a collie because she has a special affection for that breed or had a positive experience with a collie in the past. Alternatively, Luna may have always cared about dogs, and getting a collie is simply a reflection of her preexisting affection for canines.}
    
    \textcolor{ForestGreen}{On the other hand, Hypothesis 2, "Sean has a pet dog," is not directly related to the cause of the premise that Luna cares about her dog. While Sean having a pet dog may be interesting information, it does not provide any explanation as to why Luna cares about her dog. }
    
    \textcolor{ForestGreen}{Therefore, Hypothesis 1, "Luna got a collie," is the better option for explaining the cause of the premise that Luna cares about her dog.}
    & 
    \textcolor{ForestGreen}{hyp1: Luna got a collie better explains the cause of the premise "Luna cares about her dog." This is because the hypothesis directly relates to Luna having a dog, which she can care for. In contrast, hyp2: Sean has a pet dog, doesn't explain why Luna cares about her dog, as it is about someone else's pet.}
    & 
    \textcolor{ForestGreen}{Hypothesis 1 (Luna got a collie) is a better explanation for the premise than hypothesis 2 (Sean has a pet dog).}

    \textcolor{ForestGreen}{- Hypothesis 1 is more specific and directly related to the premise. Luna caring about her dog is more likely to be caused by her getting a dog herself, than by her friend Sean having a dog.}

    \textcolor{ForestGreen}{- Hypothesis 2 is more general and could be explained by many other factors. For example, Sean's dog could be very friendly and loving, which could make Luna want to care about dogs in general. However, this does not explain why Luna would specifically care about her own dog.}

    \textcolor{ForestGreen}{In conclusion, hypothesis 1 is a better explanation for the premise because it is more specific and directly related to the premise.}
    &
    1

    \\ \hline  
    11. Given the premise: Betty was infected with  salmonella. Determine which of the two hypothesis options, better explains the effect of the premise. hyp1: Her baby is no healthy. hyp2: Betty was having diarrhea. 
    & 
    \textcolor{ForestGreen}{Based solely on the premise that Betty was infected with salmonella, it is more likely that hypothesis 2 (Betty was having diarrhea) better explains the effect of the premise.}

    \textcolor{ForestGreen}{Salmonella is a bacteria that causes food poisoning and commonly results in symptoms such as diarrhea, abdominal cramps, and fever. While salmonella can have serious health consequences, such as dehydration and sepsis, the immediate effect of an infection is typically gastrointestinal symptoms like diarrhea. Therefore, it is more reasonable to hypothesize that Betty experienced diarrhea as a result of her salmonella infection, rather than assuming her baby is not healthy.} 
    
    \textcolor{ForestGreen}{However, it is important to note that without additional information about Betty's health and the context surrounding her infection, it is difficult to make definitive conclusions.}
    & 
    \textcolor{ForestGreen}{Based on the given premise that Betty was infected with salmonella, hypothesis 2 (Betty was having diarrhea) better explains the effect of the premise. Salmonella infection commonly causes symptoms such as diarrhea, fever, and abdominal cramps. While hypothesis 1 (Her baby is not healthy) could be a possible consequence of Betty's infection, especially if she is pregnant, it is less directly related to the premise than hypothesis 2.}
    & 
    \textcolor{red}{Conclusion: Both hypotheses are possible explanations for the premise. However, hypothesis 1 is a better explanation because it is more specific. It is not just that Betty was having diarrhea, but that her baby was also not healthy. This suggests that the diarrhea was caused by the salmonella infection, which is more likely to happen in babies.}

    \textcolor{red}{In addition, hypothesis 2 does not explain why Betty was infected with salmonella in the first place. Hypothesis 1, on the other hand, does explain this. If Betty's baby was not healthy, they may have been more likely to come into contact with salmonella, which could have then infected Betty.}

    \textcolor{red}{Therefore, hypothesis 1 is the better explanation for the premise.}
    &
    2

    \\ \hline  
    12. Given the premise: Raspberries can be seen here. Determine which of the two hypothesis options, better explains the cause of the premise. hyp1: There are aralias in this forest. hyp2: This garden contains all plants of rose family. 
    & 
    \textcolor{red}{Neither hypothesis of the given options can be considered as a satisfactory explanation for the premise that raspberries can be seen here. }
    
    \textcolor{red}{The first hypothesis suggests the presence of aralias in the forest, but it does not provide any information or evidence to support the claim that raspberries are also present. Therefore, this hypothesis cannot explain why raspberries are seen.}
    
    \textcolor{red}{Similarly, the second hypothesis implies that all plants of the rose family are present in the garden, but it does not directly address the presence of raspberries. While raspberries do belong to the rose family, not all plants in this family are raspberries. Thus, this hypothesis also fails to explain why raspberries can be seen.}
    
    \textcolor{red}{To better explain the cause of the premise, one would need to consider additional hypotheses or evidence that directly address the presence of raspberries in the location.}
    & 
    \textcolor{ForestGreen}{Hyp2: "This garden contains all plants of the rose family" better explains the cause of the premise "Raspberries can be seen here." This is because raspberries belong to the rose family (Rosaceae), and the presence of all plants from the rose family in the garden would include raspberries. Hyp1: "There are aralias in this forest" is not directly related to the presence of raspberries, as aralias belong to a different family (Araliaceae) and do not have a clear connection to raspberries.}
    & 
    \textcolor{ForestGreen}{Raspberries are a type of rose plant, so it makes sense that they would be found in a garden that contains all plants of the rose family. Aralias, on the other hand, are not related to raspberries, so it is less likely that they would be found in the same place.}

    \textcolor{ForestGreen}{In addition, hypothesis 2 is more specific than hypothesis 1. Hypothesis 1 simply states that there are aralias in the forest, but it does not say anything about raspberries. Hypothesis 2, on the other hand, specifically states that the garden contains all plants of the rose family, which includes raspberries.}

    \textcolor{ForestGreen}{Therefore, hypothesis 2 is a better explanation for the premise that raspberries can be seen here.}
    &
    2

    \\ \hline  
    13. Given the premise: The lactating woman took hydrazine. Determine which of the two hypothesis options, better explains the effect of the premise. hyp1: Her breasts become bigger and bigger. hyp2: The baby had related adverse reactions. 
    & 
    \textcolor{red}{Neither of the two hypotheses is a scientifically plausible explanation for the effect of the premise that "the lactating woman took hydrazine." }

    \textcolor{red}{Hydrazine is a toxic and highly reactive chemical compound that is used primarily as a rocket propellant and in the production of pharmaceuticals and pesticides. It is not safe for human consumption and can cause serious health problems, including liver damage, seizures, and death. There is no known therapeutic use of hydrazine in lactation.}
    
    \textcolor{red}{Therefore, it is not possible to determine which of the two hypotheses better explains the effect of the premise, as both hypotheses are unfounded and unsupported by scientific evidence. If a lactating woman has taken hydrazine, it is important to seek medical attention immediately to minimize the potential harm to herself and her baby.}
    & 
    \textcolor{ForestGreen}{Hyp2: The baby had related adverse reactions better explains the effect of the premise. Hydrazine is a toxic and potentially dangerous substance, and its ingestion by a lactating woman could lead to adverse reactions in the baby who consumes the breast milk. Hyp1 is less likely, as there is no known connection between hydrazine and breast size.}
    & 
    \textcolor{ForestGreen}{Hypothesis 2, that the baby had related adverse reactions, is a better explanation of the effect of the premise than hypothesis 1, that the woman's breasts became bigger and bigger.}

    \textcolor{ForestGreen}{Hydrazine is a drug that is contraindicated for use in lactating women. This means that it is not safe for the drug to pass into breast milk and be ingested by the baby. When hydrazine is ingested by a baby, it can cause a number of adverse reactions, including seizures, respiratory problems, and even death.}

    \textcolor{ForestGreen}{There is no evidence to suggest that hydrazine can cause a woman's breasts to become bigger. In fact, it is more likely that the drug would have the opposite effect, as it can cause dehydration and weight loss.}

    \textcolor{ForestGreen}{Therefore, hypothesis 2, that the baby had related adverse reactions, is a better explanation of the effect of the premise than hypothesis 1, that the woman's breasts became bigger and bigger.}
    &
    2

    \\ \hline  
    14. Given the premise: They went to the church for prays. Determine which of the two hypothesis options, better explains the cause of the premise. hyp1: The patients was in the process of the treatment. hyp2: Villagers suffered painful life when there was a heavy flood in the village. 
    & 
    \textcolor{red}{Neither hypothesis 1 nor hypothesis 2 seems to be a satisfactory explanation for the given premise, "They went to the church for prays." }

    \textcolor{red}{Hypothesis 1 suggests that the reason for going to the church is related to a medical treatment process, which is not explicitly mentioned in the premise. Additionally, the term "patients" in hypothesis 1 implies that the individuals who went to the church are receiving medical treatment, but the premise does not provide any indication of this.}
    
    \textcolor{red}{Hypothesis 2 suggests that the reason for going to the church is related to the villagers' suffering during a flood, which is also not explicitly mentioned in the premise. The premise simply states that they went to the church for prayers, without any context or explanation of why they felt the need to pray.}
    
    \textcolor{red}{Therefore, it is difficult to determine which hypothesis is a better explanation for the given premise, as neither hypothesis directly addresses the reason for going to the church. It is possible that there is a different explanation altogether that is not covered by these hypotheses.}
    & 
    \textcolor{ForestGreen}{While both hypotheses could potentially explain why they went to the church for prayers, Hypothesis 2 better explains the cause of the premise. Hyp2 states that "Villagers suffered a painful life when there was a heavy flood in the village." This suggests that the people went to the church to pray for relief from the flood's aftermath or to seek solace during difficult times. Hypothesis 1, on the other hand, is less directly related to the premise, as it does not provide a clear connection between the patients' treatment and their decision to go to the church for prayers.}
    & 
    \textcolor{ForestGreen}{Hypothesis 2, Villagers suffered painful life when there was a heavy flood in the village, better explains the cause of the premise.}

    \textcolor{ForestGreen}{Prayer is often used as a way to ask for help or guidance in times of need. In the case of a heavy flood, villagers would likely be feeling scared, helpless, and uncertain about the future. They might go to church to pray for safety, relief, and comfort.}

    \textcolor{ForestGreen}{Hypothesis 1, The patients was in the process of the treatment, is less likely to be the cause of the premise. While patients may pray for healing during their treatment, they would typically do so in the hospital or at home. They would not typically go to church specifically to pray for healing.}

    \textcolor{ForestGreen}{In addition, the premise states that they went to the church for prays, which suggests that they were going to church specifically to pray. This is more consistent with hypothesis 2, as villagers would be more likely to go to church to pray for help in the wake of a natural disaster.}

    \textcolor{ForestGreen}{Therefore, hypothesis 2 is a better explanation for the cause of the premise.}
    &
    2

    \\ \hline  
    15. Given the premise: Scientists have discovered the permanent magnetic moment of atoms. Determine which of the two hypothesis options, better explains the cause of the premise. hyp1: Scientists study the paramagnetism of atoms. hyp2: The Myofibrils have different numbers of myofilaments. 
    & 
    \textcolor{red}{Neither of the two hypotheses (hyp1: scientists study the paramagnetism of atoms; hyp2: the Myofibrils have different numbers of myofilaments) directly explains the premise that scientists have discovered the permanent magnetic moment of atoms. }

    \textcolor{red}{The permanent magnetic moment of an atom refers to the magnetic dipole moment that arises from the intrinsic properties of the atom, such as the spin and orbital motion of its electrons. It is a fundamental property of atoms that has been known and studied for many years. }
    
    \textcolor{red}{Hypothesis 1 (scientists study the paramagnetism of atoms) is related to the magnetic properties of materials, including the paramagnetic behavior of some materials in the presence of an external magnetic field. While this is an interesting topic of study, it does not directly relate to the discovery of the permanent magnetic moment of atoms.}
    
    \textcolor{red}{Hypothesis 2 (the Myofibrils have different numbers of myofilaments) is unrelated to the topic of atomic magnetism. Myofibrils are organelles found in muscle cells that are involved in muscle contraction and do not have any direct connection to the magnetic properties of atoms.}
    
    \textcolor{red}{Therefore, a more appropriate hypothesis to explain the discovery of the permanent magnetic moment of atoms might be that scientists have used advanced experimental techniques such as nuclear magnetic resonance (NMR) spectroscopy or magnetic resonance imaging (MRI) to directly measure and characterize the magnetic properties of individual atoms.}
    & 
    \textcolor{ForestGreen}{Based on the given premise, Hypothesis 1 (Scientists study the paramagnetism of atoms) is more relevant and better explains the cause of the premise. }

    \textcolor{ForestGreen}{The premise involves the discovery of the permanent magnetic moment of atoms, which is directly related to their magnetic properties. Paramagnetism is a type of magnetism that occurs in certain materials, like atoms or molecules, which have permanent magnetic moments due to the presence of unpaired electrons. Therefore, studying paramagnetism would provide a better understanding of the magnetic properties of atoms, including their permanent magnetic moment.}

    \textcolor{ForestGreen}{On the other hand, Hypothesis 2 (The Myofibrils have different numbers of myofilaments) is related to muscle physiology and not related to the magnetic properties of atoms. Myofibrils are structures within muscle cells that are composed of repeating sections called sarcomeres, which contain myofilaments. The number of myofilaments in myofibrils is unrelated to the magnetic properties of atoms and does not provide any useful information to explain the given premise.}
    & 
    \textcolor{ForestGreen}{Hypothesis 1 is the better explanation for the premise.}

    \textcolor{ForestGreen}{- Hypothesis 1: Scientists study the paramagnetism of atoms. * Paramagnetism is a property of atoms that have unpaired electrons. These unpaired electrons create a magnetic field, which is what gives atoms their permanent magnetic moment. * Therefore, studying the paramagnetism of atoms would lead scientists to discover the permanent magnetic moment of atoms.}

    \textcolor{ForestGreen}{- Hypothesis 2: The Myofibrils have different numbers of myofilaments. * Myofibrils are the contractile units of muscle cells. They are made up of myofilaments, which are the individual fibers that make up the myofibril. * The number of myofilaments in a myofibril can vary depending on the type of muscle cell. However, the number of myofilaments does not affect the permanent magnetic moment of atoms.}

    \textcolor{ForestGreen}{Therefore, hypothesis 1 is a better explanation for the premise because it is more directly related to the property of atoms that gives them their permanent magnetic moment.}
    &
    1

    \\ \hline  
    16. Given the premise: Batteries should not be casually thrown. Determine which of the two hypothesis options, better explains the cause of the premise. hyp1: Tom accidentally threw some waste containing curium into a lake. hyp2: Anything containing caustic mixtures of chemicals should not be casually thrown. 
    & 
    \textcolor{ForestGreen}{Based on the given premise, the second hypothesis option, "Anything containing caustic mixtures of chemicals should not be casually thrown," better explains the cause of the premise.}

    \textcolor{ForestGreen}{The premise states that batteries should not be casually thrown, which implies that they contain some harmful or hazardous materials that could cause harm to the environment or people if not disposed of properly. Hypothesis 1 suggests that the reason for this is because Tom accidentally threw waste containing curium into a lake, which is a specific incident and may not necessarily apply to all batteries. }
    
    \textcolor{ForestGreen}{On the other hand, hypothesis 2 provides a more general and universal explanation for why batteries should not be casually thrown. It suggests that anything containing caustic mixtures of chemicals should not be casually thrown, which includes batteries as they contain chemicals that can be harmful to the environment and people. This explanation is not limited to a specific incident and applies to all batteries, regardless of their contents. }
    
    \textcolor{ForestGreen}{Therefore, hypothesis 2 is a better explanation for the given premise than hypothesis 1.}
    & 
    \textcolor{ForestGreen}{Hyp2: Anything containing caustic mixtures of chemicals should not be casually thrown, better explains the cause of the premise. This hypothesis directly addresses the issue of casually throwing batteries, which often contain hazardous chemicals that can harm the environment and human health. Hyp1, on the other hand, specifically discusses Tom throwing waste containing curium into a lake, which is unrelated to the general premise of not casually throwing batteries.}
    & 
    \textcolor{ForestGreen}{Hypothesis 2 better explains the cause of the premise.}

    \textcolor{ForestGreen}{Hypothesis 1 is about a specific incident involving curium, which is a radioactive element. While curium is certainly dangerous, it is not the only hazardous substance found in batteries. Batteries also contain other chemicals, such as lead, mercury, and sulfuric acid, which can be harmful to the environment and human health.}

    \textcolor{ForestGreen}{Hypothesis 2 is more general and encompasses all batteries, regardless of their contents. It also highlights the fact that batteries contain caustic mixtures of chemicals, which can cause burns and other injuries if they come into contact with skin or eyes.}

    \textcolor{ForestGreen}{Therefore, hypothesis 2 is a better explanation for the premise that batteries should not be casually thrown.}
    &
    2

    \\ \hline  
    17. Given the premise: Jack was tested positive for stimulant. Determine which of the two hypothesis options, better explains the cause of the premise. hyp1: Jack ate a bar of chocolate before the game. hyp2: She can not stand the old drug's effect. 
    & 
    \textcolor{red}{I'm sorry, but there seems to be a discrepancy in the premise as it refers to Jack as "he" in the first sentence, but then switches to "she" in the second hypothesis option. Therefore, I cannot determine which hypothesis is better suited to explain the cause of the premise. Could you please clarify the gender of the individual in question?}
    & 
    \textcolor{red}{Based on the given premise that Jack tested positive for a stimulant, we will analyze the two hypotheses.}

    \textcolor{red}{Hyp1: Jack ate a bar of chocolate before the game.
    While chocolate does contain small amounts of caffeine, a natural stimulant, the amount is not typically significant enough to cause a positive test result for stimulants. Therefore, this hypothesis is unlikely to be the primary cause of Jack testing positive for a stimulant.}

    \textcolor{red}{Hyp2: She can not stand the old drug's effect.
    This hypothesis seems to suggest that Jack has taken a different drug due to being unable to tolerate the effects of a previous one. It is more likely to explain the positive test result for a stimulant, as it directly implies the use of a drug, which could be a stimulant.}

    \textcolor{red}{Given the information provided, Hypothesis 2 better explains the cause of Jack testing positive for a stimulant. However, it is important to note that there could be other factors or explanations not provided in the given hypotheses.}
    & 
    \textcolor{red}{Overall, hypothesis 2 is more likely than hypothesis 1. It is more specific and it explains the premise more fully. However, it is still important to note that both hypotheses are speculative, and we cannot say for sure which one is correct.}

    \textcolor{red}{To further investigate the cause of Jack's positive test result, we could conduct further tests to see if he had any other drugs in his system. We could also ask Jack about his drug use and why he might have taken a new drug. With more information, we would be able to make a more informed conclusion about the cause of Jack's positive test result.}
    &
    1

    \\ \hline  
    18. Given the premise: His gum is inflamed. Determine which of the two hypothesis options, better explains the effect of the premise. hyp1: All the living cells in his body were not functioning properly. hyp2: He had the same symptoms. 
    & 
    \textcolor{ForestGreen}{Based on the given premise, the hypothesis that better explains the effect is hyp2: "He had the same symptoms." This hypothesis suggests that the person may have a dental or gum issue causing the inflammation. In contrast, hyp1 is not supported by the premise as it suggests that all the living cells in his body were not functioning properly, which is a broad and unsupported assumption that is not directly related to gum inflammation. Therefore, based solely on the given premise, hyp2 is the better explanation for the observed effect.}
    & 
    \textcolor{ForestGreen}{Based on the given premise, hypothesis 2 (He had the same symptoms) better explains the effect of the premise. Inflamed gums could be a symptom of a particular condition, and having the same symptoms would mean that the inflammation could be related to that specific condition. Hypothesis 1 (All the living cells in his body were not functioning properly) is too broad and does not directly relate to the premise of having inflamed gums.}
    & 
    \textcolor{ForestGreen}{Hypothesis 1: All the living cells in his body were not functioning properly.}

    \textcolor{ForestGreen}{This hypothesis is not supported by the premise. The premise only states that his gum is inflamed, not that all the living cells in his body are not functioning properly. In fact, there are many other possible explanations for why his gum is inflamed, such as an infection, a allergic reaction, or a side effect of medication.}

    \textcolor{ForestGreen}{Hypothesis 2: He had the same symptoms.}
    
    \textcolor{ForestGreen}{This hypothesis is more supported by the premise. If he had the same symptoms before, then it is more likely that the inflammation is caused by the same underlying condition. However, this hypothesis is still not definitive, as there could be other explanations for why he is experiencing the same symptoms.}

    \textcolor{ForestGreen}{Ultimately, more information is needed to determine which hypothesis is better supported by the premise. For example, if we knew more about the specific symptoms he is experiencing, or if we knew about his medical history, then we would be able to make a more informed decision.}

    \textcolor{ForestGreen}{Based on the information that is currently available, neither hypothesis is a good explanation for the premise. However, hypothesis 2 is slightly more supported by the premise than hypothesis 1.}
    &
    2

    \\ \hline  
    19. Given the premise: Bob decreased the speed to 50km/h. Determine which of the two hypothesis options, better explains the cause of the premise. hyp1: The racers entered the deceleration phase. hyp2: Tom wants to accelerate the speed of the chemical reaction. 
    & 
    \textcolor{ForestGreen}{Based on the given premise, the hypothesis option that better explains the cause of the premise is hyp1: "The racers entered the deceleration phase." }

    \textcolor{ForestGreen}{This is because the statement "Bob decreased the speed to 50km/h" implies that there was some kind of motion or movement involved, and the use of the word "decreased" suggests a deliberate action taken by someone. Additionally, the context of "the racers" suggests that there was some kind of competition or race happening, which makes it more likely that the decrease in speed was due to a deceleration phase or strategy being employed.}
    
    \textcolor{ForestGreen}{On the other hand, hyp2: "Tom wants to accelerate the speed of the chemical reaction" does not seem to be related to the given premise at all. There is no mention of any chemical reaction or any indication that such a reaction is even taking place. Therefore, hyp2 does not provide a plausible explanation for the cause of the given premise.}
    & 
    \textcolor{ForestGreen}{Based on the given premise, hypothesis 1: "The racers entered the deceleration phase" better explains the cause of the premise. This is because the premise specifically talks about Bob decreasing the speed, which is related to a change in velocity or pace, consistent with the concept of a deceleration phase in racing. Hypothesis 2: "Tom wants to accelerate the speed of the chemical reaction" is unrelated to the premise as it discusses a chemical reaction and involves a different person, Tom.}
    & 
    \textcolor{ForestGreen}{Hypothesis 1, The racers entered the deceleration phase, is a better explanation for the premise that Bob decreased the speed to 50 km/h. This is because the deceleration phase is a natural part of racing, and it is typically accompanied by a decrease in speed. Hypothesis 2, Tom wants to accelerate the speed of the chemical reaction, is not a good explanation for the premise because it is not clear how decreasing the speed of a car would help to accelerate a chemical reaction. Additionally, the premise does not mention anything about a chemical reaction, so it is not clear why this hypothesis would be relevant.}
    &
    1

    \\ \hline  
    20. Given the premise: The director asked the staff about the product knowledge. Determine which of the two hypothesis options, better explains the effect of the premise. hyp1: The staff took use of codes instead of spoken words. hyp2: The staff was able to answer them all. 
    & 
    \textcolor{red}{Based on the premise alone, neither hypothesis can be definitively confirmed or refuted. }
    
    \textcolor{red}{The premise states that the director asked the staff about the product knowledge, but it does not provide any information on how the staff responded. Therefore, it is impossible to determine whether they used codes or spoken words (hypothesis 1) or whether they were able to answer all the questions (hypothesis 2). }
    
    \textcolor{red}{Additional information or context is needed to determine which hypothesis better explains the effect of the premise.}
    & 
    \textcolor{ForestGreen}{Hypothesis 2: The staff was able to answer them all, better explains the effect of the premise. The premise is about the director inquiring about the staff's product knowledge. This hypothesis directly addresses the staff's ability to demonstrate their knowledge about the product, which is relevant to the premise. Hypothesis 1, involving the staff using codes instead of spoken words, does not directly relate to the premise or provide a clear connection to the staff's product knowledge.}
    & 
    \textcolor{ForestGreen}{Hypothesis 2: The staff was able to answer them all better explains the effect of the premise.}

    \textcolor{ForestGreen}{The premise states that the director asked the staff about product knowledge. This means that the director wanted to assess the staff's knowledge of the products. If the staff was able to answer all of the director's questions, then this would indicate that they have a good understanding of the products.}

    \textcolor{ForestGreen}{Hypothesis 1, on the other hand, is not as likely to be true. If the staff took use of codes instead of spoken words, then this would suggest that they were not comfortable discussing their product knowledge with the director. This could be because they did not have a good understanding of the products, or because they were afraid of being asked difficult questions.}

    \textcolor{ForestGreen}{Therefore, hypothesis 2 is a better explanation of the effect of the premise.}
    &
    2

    \\ \hline  
    21. Given the premise: They found a lot food. Determine which of the two hypothesis options, better explains the cause of the premise. hyp1: The hunter caught a lot of squirrels. hyp2: They are the first settlers on this island. 
    & 
    \textcolor{red}{Based solely on the premise "They found a lot of food," neither hypothesis can be definitively confirmed or ruled out. However, we can evaluate each hypothesis based on what we know about the context and likelihood.}
    
    \textcolor{red}{Hypothesis 1: The hunter caught a lot of squirrels.
    This hypothesis suggests that the food was obtained through hunting. If the environment is suitable for squirrels and the hunter has the necessary skills, it's plausible that they could catch a lot of squirrels. However, this hypothesis doesn't account for the possibility of finding other types of food, such as fruits, vegetables, or fish.}
    
    \textcolor{red}{Hypothesis 2: They are the first settlers on this island.
    This hypothesis suggests that the food was obtained through exploration and discovery. If the settlers are the first people to arrive on the island, they would have to rely on the resources available on the island. It's possible that the island has an abundance of food sources, which could explain why they found a lot of food.}
    
    \textcolor{red}{Ultimately, without additional information, it's impossible to say which hypothesis is more likely. However, both hypotheses offer potential explanations for the discovery of a lot of food, and the likelihood of each hypothesis could depend on the specific context and circumstances.}
    & 
    \textcolor{red}{Given the premise: "They found a lot food," it is difficult to definitively say which hypothesis better explains the cause. However, Hypothesis 1: "The hunter caught a lot of squirrels" has a more direct link to finding food, as it specifically mentions a hunter catching squirrels, which could be considered as food. Hypothesis 2: "They are the first settlers on this island" doesn't directly relate to finding food, as it only states they are the first settlers, but it does not explain how they found a lot of food. Based on this analysis, Hypothesis 1 seems to better explain the cause of the premise.}
    & 
    \textcolor{red}{Hypothesis 1: The hunter caught a lot of squirrels.}

    \textcolor{red}{This hypothesis is more likely to explain the premise because it is more specific and provides a clear mechanism for how the people found a lot of food. The hunter would have had to be skilled in hunting squirrels, and the squirrels would have had to be abundant in the area. This hypothesis is also more plausible because it is consistent with our knowledge of how humans have traditionally found food.}

    \textcolor{red}{Hypothesis 2: They are the first settlers on this island.}

    \textcolor{red}{This hypothesis is less likely to explain the premise because it is less specific and does not provide a clear mechanism for how the people found a lot of food. It is possible that the first settlers on an island would have found a lot of food, but it is also possible that they would have found very little food. This hypothesis is also less plausible because it is not consistent with our knowledge of how humans have traditionally found food.}

    \textcolor{red}{In conclusion, hypothesis 1 is more likely to explain the premise than hypothesis 2.}
    
    &
    2

    \\ \hline  
    22. Given the premise: Some parasites lay their eggs on cloves. Determine which of the two hypothesis options, better explains the effect of the premise. hyp1: A few days later the eggs died. hyp2: After a few days, the eggs turned into parasites. 
    & 
    \textcolor{red}{Based on the premise, it is more likely that hypothesis 2, "After a few days, the eggs turned into parasites" is a better explanation for the effect of the premise.} 
    
    \textcolor{red}{If some parasites lay their eggs on cloves and the eggs died after a few days, it would suggest that the cloves have a toxic effect on the eggs. However, if the eggs turned into parasites after a few days, it would indicate that the cloves provide a suitable environment for the eggs to hatch and develop into parasites. This would be consistent with the fact that many parasites have specific hosts or environments in which they can thrive.} 
    
    \textcolor{red}{Therefore, hypothesis 2 is a more plausible explanation for the effect of the premise that some parasites lay their eggs on cloves.}
    & 
    \textcolor{red}{Based on the given premise, Hypothesis 2 (hyp2) better explains the effect of the premise: "After a few days, the eggs turned into parasites." This hypothesis aligns with the typical life cycle of a parasite, wherein they lay their eggs on a suitable environment, and eventually, the eggs hatch into new parasites. Hypothesis 1 (hyp1) does not fit as well with the normal behavior and life cycle of parasites.}
    & 
    \textcolor{red}{Sure, I can help you with that.}

    \textcolor{red}{Premise: Some parasites lay their eggs on cloves.}

    \textcolor{red}{Hypothesis 1: A few days later the eggs died.}

    \textcolor{red}{Hypothesis 2: After a few days, the eggs turned into parasites.}

    \textcolor{red}{Which hypothesis better explains the effect of the premise?}

    \textcolor{red}{Hypothesis 2 is the better explanation. This is because parasites need a host to survive. Once the eggs hatch, they will need to find a host in order to continue their lifecycle. If the eggs were to die, they would not be able to find a host and would eventually die out.}

    \textcolor{red}{Hypothesis 1 is not as likely because it does not take into account the need for a host. It is possible that the eggs could die, but this is less likely than them hatching and turning into parasites.}

    \textcolor{red}{Therefore, hypothesis 2 is the better explanation for the effect of the premise.}
    &
    1

    \\ \hline  
    23. Given the premise: Jim faces the stress of finding a job. Determine which of the two hypothesis options, better explains the effect of the premise. hyp1: He often gets cramps. hyp2: He is too distressed to sleep at night. 
    & 
    \textcolor{ForestGreen}{Hypothesis 2, "He is too distressed to sleep at night," better explains the effect of the premise that Jim faces the stress of finding a job. }
    
    \textcolor{ForestGreen}{While it is possible that Jim could experience physical symptoms such as cramps as a result of stress, the more commonly reported effect of stress is sleep disturbance. Stress can cause a range of sleep problems, including difficulty falling asleep, waking up frequently during the night, and waking up feeling unrefreshed. Therefore, hypothesis 2 is a more likely and plausible explanation for Jim's situation than hypothesis 1.}
    & 
    \textcolor{ForestGreen}{Based on the premise provided, Hypothesis 2: "He is too distressed to sleep at night" better explains the effect of Jim facing the stress of finding a job. This is because the stress from job hunting is more likely to cause sleep disturbances and distress, rather than directly causing physical symptoms such as cramps.}
    & 
    \textcolor{ForestGreen}{Hypothesis 2, that Jim is too distressed to sleep at night, is a better explanation of the effect of the premise than hypothesis 1, that he often gets cramps.}

    \textcolor{ForestGreen}{Cramps are a physical symptom that can be caused by a variety of factors, including stress. However, they are not a specific symptom of stress, and they can also be caused by other factors, such as dehydration or electrolyte imbalance.}

    \textcolor{ForestGreen}{On the other hand, sleep deprivation is a very specific symptom of stress. When people are stressed, their bodies release hormones that make it difficult to fall asleep and stay asleep. This can lead to a cycle of sleep deprivation, which can further exacerbate stress levels.}

    \textcolor{ForestGreen}{Therefore, hypothesis 2, that Jim is too distressed to sleep at night, is a better explanation of the effect of the premise than hypothesis 1, that he often gets cramps.}
    &
    2

    \\ \hline  
    24. Given the premise: Tom was prepared to join the parade. Determine which of the two hypothesis options, better explains the effect of the premise. hyp1: He walked to the bay. hyp2: Tom went to the street to join the parade. 
    & 
    \textcolor{red}{Based solely on the given premise, neither hypothesis can be definitively confirmed or refuted. The premise simply states that Tom was prepared to join the parade, but it does not specify where he was or where he went after being prepared. Therefore, both hypotheses are possible and equally valid explanations for what might have happened next.}
    
    \textcolor{red}{Without additional information, it is impossible to determine which hypothesis is more likely to be true. However, if you have further information that could narrow down the possibilities, you could use that information to evaluate the hypotheses and determine which one is more plausible.}
    & 
    \textcolor{ForestGreen}{Hyp2: Tom went to the street to join the parade better explains the effect of the premise. This is because it directly connects to Tom's preparation to join the parade, whereas Hyp1: He walked to the bay does not provide a clear link to the premise.}
    & 
    \textcolor{ForestGreen}{Hypothesis 2, Tom went to the street to join the parade, better explains the effect of the premise.}

    \textcolor{ForestGreen}{The premise states that Tom was prepared to join the parade. This means that he had made the necessary preparations, such as getting dressed in his parade uniform, gathering his parade props, and finding a way to get to the parade route.}

    \textcolor{ForestGreen}{Hypothesis 1, He walked to the bay, does not explain how Tom's preparations would help him join the parade. The bay is not typically a location for parades, and it would be difficult for Tom to join a parade by walking to the bay.}

    \textcolor{ForestGreen}{Hypothesis 2, on the other hand, explains how Tom's preparations would help him join the parade. If Tom went to the street where the parade was happening, he would be in a good position to join the parade. He would be able to see the parade and find a place to join in.}

    \textcolor{ForestGreen}{Therefore, hypothesis 2, Tom went to the street to join the parade, better explains the effect of the premise.}
    &
    2

    \\ \hline  
    25. Given the premise: The service staff of this hotel lack management. Determine which of the two hypothesis options, better explains the effect of the premise. hyp1: The service quality of the hotel has seriously declined. hyp2: In addition to financial management courses, she also studied economic law and tax law. 
    & 
    \textcolor{red}{Neither of the two hypotheses seem to explain the effect of the given premise. }
    
    \textcolor{red}{The first hypothesis suggests that the service quality of the hotel has seriously declined due to lack of management among the service staff. However, this does not necessarily follow from the premise alone. It is possible that the service quality has always been low, or that other factors are also contributing to the decline in service quality. }
    
    \textcolor{red}{The second hypothesis seems unrelated to the premise. It suggests that someone (presumably a person in management) has studied financial management, economic law, and tax law. While this may be useful information, it does not provide any insight into the effect of the premise on the hotel's service staff. }
    
    \textcolor{red}{Without more information or context, it is difficult to determine which hypothesis (if either) would better explain the effect of the given premise.}
    & 
    \textcolor{ForestGreen}{Hyp1: The service quality of the hotel has seriously declined. This hypothesis better explains the effect of the premise, as it directly relates to the service staff lacking management and its potential impact on the hotel's service quality.}
    & 
    \textcolor{ForestGreen}{Hypothesis 1 is a better explanation of the effect of the premise.}

    \textcolor{ForestGreen}{The premise states that the service staff of the hotel lack management. This means that the staff are not being given the training, support, or resources they need to do their jobs effectively. As a result, the service quality of the hotel is likely to decline.}

    \textcolor{ForestGreen}{Hypothesis 2, on the other hand, is not directly related to the premise. It simply states that the service staff have studied financial management, economic law, and tax law. This information is not relevant to the question of why the service quality of the hotel has declined.}

    \textcolor{ForestGreen}{Therefore, hypothesis 1 is the better explanation of the effect of the premise.}
    &
    1

    \\ \hline  
    26. Given the premise: The fishes in the tank were electrocuted. Determine which of the two hypothesis options, better explains the cause of the premise. hyp1: Tom connected a socket and a fish tank with a wire. hyp2: Nadal put cephalopods in a fresh water fish tank. 
    & 
    \textcolor{ForestGreen}{Based on the premise given, hypothesis 1, "Tom connected a socket and a fish tank with a wire," is a more likely explanation for the electrocution of the fish in the tank. The use of a wire to connect the socket and the tank could have caused an electrical current to flow through the water and electrocute the fish. On the other hand, hypothesis 2, "Nadal put cephalopods in a freshwater fish tank," does not provide a clear explanation for how the fish were electrocuted. While the cephalopods could potentially harm the fish in other ways, it is unlikely that they would cause an electrical shock. Therefore, hypothesis 1 is the better explanation for the premise given.}
    & 
    \textcolor{ForestGreen}{Hyp1: Tom connected a socket and a fish tank with a wire.}

    \textcolor{ForestGreen}{This hypothesis directly implies that an electrical current was introduced to the fish tank through the wire connected to a socket. This could result in the electrocution of the fishes in the tank, which aligns with the given premise. }
    
    \textcolor{ForestGreen}{Hyp2: Nadal put cephalopods in a fresh water fish tank.}
    
    \textcolor{ForestGreen}{Introducing cephalopods to a freshwater fish tank would likely cause harm to the cephalopods themselves, as they are generally adapted to living in saltwater environments. However, this hypothesis does not provide any direct evidence or connection to the fishes in the tank being electrocuted.}
    
    \textcolor{ForestGreen}{Based on the available information, Hyp1 better explains the cause of the premise as it directly involves the introduction of an electrical current into the fish tank, which could result in electrocution of the fishes.}
    & 
    \textcolor{ForestGreen}{Hypothesis 1 (H1) is more likely to explain the premise than hypothesis 2 (H2).}

    \textcolor{ForestGreen}{- H1 states that Tom connected a socket and a fish tank with a wire. This would create a direct current (DC) electrical current that would flow through the water in the tank, electrocuting the fish.}

    \textcolor{ForestGreen}{- H2 states that Nadal put cephalopods in a fresh water fish tank. Cephalopods are saltwater creatures, and their bodies are adapted to the high salinity of saltwater. When placed in freshwater, their bodies will begin to lose water, which can lead to death. However, electrocution would be a much more immediate and likely cause of death for the fish.}

    \textcolor{ForestGreen}{In addition, H1 is supported by the fact that the fish were found to be electrocuted. This suggests that they were exposed to a significant amount of electrical current, which is more likely to occur if a wire was connected to the tank. H2 is not supported by any evidence, and in fact, it is unlikely that the fish would have died simply from being placed in freshwater.}

    \textcolor{ForestGreen}{Therefore, H1 is a more likely explanation for the premise than H2.}
    &
    1

    \\ \hline  
    27. Given the premise: Mary is learning the greek alphabet. Determine which of the two hypothesis options, better explains the effect of the premise. hyp1: The first letter he learns is alpha. hyp2: The first letter he learns is A. 
    & 
    
    \textcolor{ForestGreen}{Neither hypothesis is entirely accurate as there are a few issues with them. Firstly, Mary is typically a female name, and it's more appropriate to use "she" instead of "he." Secondly, the Greek letter for "A" is not the same as the Latin letter "A." 
    }
    \textcolor{ForestGreen}{With that said, hypothesis 1 is more accurate as it correctly identifies the first letter that Mary is learning as alpha. Hypothesis 2 is incorrect as it assumes that the first letter Mary is learning is the Latin letter "A," which is not relevant to the context of learning the Greek alphabet.}
    
    \textcolor{ForestGreen}{So, in summary, hypothesis 1 is the better explanation as it correctly identifies the first letter Mary is learning as alpha, which is a Greek letter.}
    & 
    \textcolor{ForestGreen}{Given the premise that Mary is learning the Greek alphabet, hypothesis 1 (The first letter he learns is alpha) better explains the effect of the premise. This is because "alpha" is the first letter of the Greek alphabet, while "A" is the first letter of the Latin alphabet, which is not being learned in this scenario.}
    & 
    \textcolor{ForestGreen}{Hypothesis 1 (H1) is a better explanation of the effect of the premise than Hypothesis 2 (H2).}

    \textcolor{ForestGreen}{- H1: The first letter he learns is alpha.}

    \textcolor{ForestGreen}{- H2: The first letter he learns is A.}

    \textcolor{ForestGreen}{The Greek alphabet has 24 letters, and the first letter is alpha. Therefore, it is more likely that Mary would learn alpha first than any other letter. Additionally, the letter alpha is the equivalent of the letter A in the Latin alphabet, so it makes sense that Mary would learn alpha first if she is already familiar with the Latin alphabet.}

    \textcolor{ForestGreen}{In conclusion, Hypothesis 1 is a better explanation of the effect of the premise than Hypothesis 2 because it is more specific to the Greek alphabet and it is more likely that Mary would learn alpha first.}    
    &
    1

    \\ \hline  
    28. Given the premise: We find that they emphasized characteristics. Determine which of the two hypothesis options, better explains the cause of the premise. hyp1: We looked at a lot of photographs. hyp2: Scientists have done a research about nuclei. 
    & 
    \textcolor{red}{Neither of the two hypothesis options directly explains the cause of the premise, as both are unrelated to the premise's content about emphasizing characteristics. However, if we assume that the premise is related to a study or analysis of some sort, then hypothesis 1 (We looked at a lot of photographs) may be more plausible, as analyzing photographs could involve looking at and emphasizing certain characteristics. On the other hand, hypothesis 2 (Scientists have done a research about nuclei) seems less likely to explain the premise, as it does not appear to be directly related to the emphasis of characteristics.}
    & 
    \textcolor{ForestGreen}{Hypothesis 1: We looked at a lot of photographs.}

    \textcolor{ForestGreen}{This hypothesis better explains the cause of the premise "We find that they emphasized characteristics." The process of looking at a lot of photographs involves observing and analyzing various characteristics, which could lead to the emphasis of those characteristics in some way. Hypothesis 2, which discusses scientific research about nuclei, is not directly related to emphasizing characteristics and is therefore less likely to be the cause of the premise.}
    & 
    \textcolor{ForestGreen}{The hypothesis that better explains the cause of the premise is hyp1: We looked at a lot of photographs.}

    \textcolor{ForestGreen}{The premise states that they emphasized characteristics. This could mean that they were focusing on the specific features of something, such as its shape, size, or color. This is something that would be done when looking at photographs, as photographs capture the details of something in a way that words cannot.}

    \textcolor{ForestGreen}{The hypothesis that scientists have done a research about nuclei is less likely to be the cause of the premise. Nuclei are the center of atoms, and they are not something that is easily seen in photographs. Therefore, it is unlikely that scientists would be emphasizing the characteristics of nuclei when looking at photographs.}

    \textcolor{ForestGreen}{In addition, the hypothesis that we looked at a lot of photographs is more specific than the hypothesis that scientists have done a research about nuclei. This is because it provides more information about the specific thing that was being emphasized. Therefore, the hypothesis that we looked at a lot of photographs is more likely to be the correct one.}
    &
    1

    \\ \hline  
    29. Given the premise: A carpenter will be recruited. Determine which of the two hypothesis options, better explains the cause of the premise. hyp1: The family cannot afford the tuition for the three children. hyp2: We need a person skilled in all aspects of construction. 
    & 
    \textcolor{red}{Based on the premise alone, it is difficult to determine which hypothesis better explains the cause. Both hypotheses could potentially explain why a carpenter is being recruited.}

    \textcolor{red}{Hypothesis 1 suggests that the family cannot afford the tuition for their three children, which may mean they are looking for a way to earn additional income. Hiring a carpenter could provide them with an opportunity to take on construction jobs to make ends meet.}
    
    \textcolor{red}{Hypothesis 2 suggests that a skilled person is needed for all aspects of construction. If the family is looking to build or renovate their home, they may require someone with expertise in carpentry to ensure that the job is done properly.}
    
    \textcolor{red}{Without further information, it is impossible to determine which hypothesis is the correct one. Additional context and information would be needed to determine the cause of the premise with certainty.}
    & 
    \textcolor{ForestGreen}{Hyp2: "We need a person skilled in all aspects of construction" better explains the cause of the premise. This hypothesis directly relates to the need for a carpenter, who is typically skilled in various aspects of construction. Hyp1, while presenting a possible financial situation for the family, does not provide a direct reason for recruiting a carpenter.}
    & 
    
    \textcolor{ForestGreen}{Hypothesis 2, "We need a person skilled in all aspects of construction," better explains the cause of the premise, "A carpenter will be recruited."}

    \textcolor{ForestGreen}{Hypothesis 1, "The family cannot afford the tuition for the three children," does not explain why a carpenter would be recruited. Carpenters do not typically work in the education field, and they would not be able to help the family pay for tuition.}

    \textcolor{ForestGreen}{Hypothesis 2, on the other hand, explains why a carpenter would be recruited. Carpenters are skilled in all aspects of construction, which includes building and repairing homes, businesses, and other structures. They are also skilled in using a variety of tools and materials.}

    \textcolor{ForestGreen}{If a company or organization needs a person who is skilled in all aspects of construction, then they would likely recruit a carpenter. This is because carpenters have the skills and experience necessary to complete a wide range of construction projects.}

    \textcolor{ForestGreen}{In addition, carpenters are often self-employed, which means that they are available to work on short-term projects. This makes them a good option for companies or organizations that need to hire someone quickly.}

    \textcolor{ForestGreen}{Therefore, hypothesis 2, "We need a person skilled in all aspects of construction," is the better explanation for the premise, "A carpenter will be recruited."}
    &
    2
 
    \\ \hline  
    30. Given the premise: It is a leech. Determine which of the two hypothesis options, better explains the cause of the premise. hyp1: An eight-inch-long creature is attaching to Tom's leg and sucking blood. hyp2: An 18-inch-long creature is attaching to Tom's leg and sucking blood. 
    & 
    \textcolor{ForestGreen}{Based on the given premise "It is a leech" and the two hypotheses, the first hypothesis is more likely to be the better explanation for the cause of the premise.} 
    
    \textcolor{ForestGreen}{This is because the length of the creature is not directly mentioned in the premise, and the size of the creature is not a defining characteristic of a leech. Therefore, the length of the creature is not necessarily a determining factor in whether it is a leech or not. }
    
    \textcolor{ForestGreen}{However, the fact that it is attaching to Tom's leg and sucking blood is a common behavior of a leech. Additionally, the length of eight inches is within the range of typical sizes for many species of leeches. Therefore, it is more likely that the creature is an eight-inch-long leech rather than an 18-inch-long one.}
    & 
    \textcolor{ForestGreen}{Based on the given premise, "It is a leech," we can reasonably assume that the creature attaching to Tom's leg is a type of blood-sucking parasite. }

    \textcolor{ForestGreen}{Now, we need to evaluate which of the two hypotheses better explains the cause of the premise. }
    
    \textcolor{ForestGreen}{Hypothesis 1 suggests that an eight-inch-long creature is attaching to Tom's leg and sucking blood. Leeches typically range in size from a few millimeters to several inches in length, and an eight-inch-long leech is within the range of possibility. Additionally, leeches are known to attach themselves to hosts and feed on their blood. Therefore, Hypothesis 1 is a plausible explanation for the premise.}
    
    \textcolor{ForestGreen}{Hypothesis 2 suggests that an 18-inch-long creature is attaching to Tom's leg and sucking blood. While there are some species of leeches that can grow quite large, an 18-inch-long leech would be extremely rare, if it exists at all. Additionally, an 18-inch-long creature attaching to Tom's leg and sucking blood could potentially be something other than a leech, such as a snake or other large predator. Therefore, Hypothesis 2 is less plausible as an explanation for the given premise.}
    
    \textcolor{ForestGreen}{Based on the above analysis, Hypothesis 1 is the better explanation for the given premise.}
    & 
    \textcolor{red}{The hypothesis that better explains the cause of the premise is hyp2: An 18-inch-long creature is attaching to Tom's leg and sucking blood.}

    \textcolor{red}{This is because the premise states that it is a leech. Leeches are typically 1-2 inches long, so an 8-inch-long creature would be too large to be a leech. An 18-inch-long creature, on the other hand, is a more reasonable size for a leech.}

    \textcolor{red}{Additionally, the premise states that the creature is sucking blood. Leeches are hematophagous, which means they feed on blood. An 8-inch-long creature could potentially suck blood, but an 18-inch-long creature would be more likely to be able to do so.}

    \textcolor{red}{Therefore, the hypothesis that better explains the cause of the premise is hyp2: An 18-inch-long creature is attaching to Tom's leg and sucking blood.}
    &
    1

    \\ \hline
    
    \end{longtable}

\end{center}

\newpage

\begin{center}
    \small
    \begin{longtable}{|p{0.7\linewidth}|p{0.05\linewidth}
                      |p{0.05\linewidth}|p{0.05\linewidth}
                      |p{.05\linewidth}|}
    \caption{Multi-hop reasoning prompts} 
    \label{tab:multi-hop_reasoning}\\
    \hline
    \multicolumn{5}{|c|}{\textbf{Multi-hop Reasoning}} \\ \hline
    \textbf{Prompt} & \textbf{GPT-3.5 Answer} & \textbf{GPT-4 Answer} & \textbf{BARD Answer} & \textbf{Gold}  \\ \hline
    \multicolumn{5}{|l|}{\textbf{Dataset: HotpotQA }} 
    \\ \hline  
     1.  Given the context: [John Franks (judge)] Franks was the second son of Thomas Franks (1729–1787), of Ballymagooly, County Cork, by Catherine, daughter of Rev. John Day.  He was born in 1770, and graduated at Trinity College, Dublin, B.A. 1788, LL.B.  1791.  He was called to the Irish Bar 1792.  He went the Munster circuit, and had a good practice as chamber counsel.  He became King's Counsel in 1823.  In 1825 the Board of Control, on the recommendation of his friend William Plunket, then Attorney General for Ireland, appointed him a judge of the Supreme Court at Calcutta.  He received, as was customary, the honour of knighthood before his departure for India.  He held this office till the effect of the climate on his health brought about his resignation in 1834.  On his return he resided at Roebuck, near Dublin.  He died 11 Jan. 1852.  He was thrice married.  By his first wife, Catherine, daughter of his cousin Thomas Franks of Carrig, Cork, he had two sons and three daughters.  He married secondly Jane Marshall, and thirdly Sarah O'Regan.  His eldest son and heir was John Franks of Ballyscaddan, County Limerick; his younger son, Matthew, was an army officer, as were many of his later descendants. [Robbie Tucker] Robbie Tucker (born April 5, 2001) is an American actor.  His best known role to date is that of Fenmore Baldwin on the CBS soap opera "The Young and the Restless".  Tucker has also starred on other series, such as "Criminal Minds", "FlashForward" and "It's Always Sunny in Philadelphia".  He has also appeared in the films "Prom" and "Little Fockers". [It's Always Sunny in Philadelphia] It's Always Sunny in Philadelphia is an American television black comedy sitcom that premiered on FX on August 4, 2005.  It moved to FXX beginning with the ninth season, and has remained there since.  It was created by Rob McElhenney, who developed it with Glenn Howerton.  It is executive produced and primarily written by McElhenney, Howerton, and Charlie Day, all of whom star alongside Kaitlin Olson and Danny DeVito.  The series follows the exploits of "The Gang", a group of debauched self-centered friends who run the Irish bar Paddy's Pub in South Philadelphia. [Peter Sutherland] Peter Denis Sutherland (born 25 April 1946) is an Irish international businessman and former Attorney General of Ireland, associated with the Fine Gael party.  He is a barrister by profession and is a Senior Counsel of the Irish Bar.  He is known for serving in a variety of international organisations, political and business roles. [George Allman (natural historian)] Allman was born in Cork, Ireland, the son of James C. Allman of Bandon, and received his early education at the Royal Academical Institution, Belfast.  For some time he studied for the Irish Bar, but ultimately gave up law in favour of natural science.  In 1843 he graduated in medicine at Trinity College, Dublin, and in the following year was appointed professor of botany in that university, succeeding the botanist William Allman (1776–1846), who was the father of George Johnston Allman (distant relations of George).  This position he held for about twelve years until he removed to Edinburgh as Regius Professor of natural history.  There he remained until 1870, when considerations of health induced him to resign his professorship and retire to Dorset, where he devoted himself to his favorite pastime of horticulture. [Fenmore Baldwin] Fenmore Baldwin is a fictional character from the American CBS soap opera "The Young and the Restless".  Created by former head writers Lynn Marie Latham and Kay Alden as the son of Michael Baldwin (Christian LeBlanc) and Lauren Fenmore (Tracey Bregman), the character was born onscreen during the episode airing on October 13, 2006.  Initially portrayed by various child actors including Robbie Tucker, the character was rapidly aged to a teenager in 2012, with Max Ehrich assuming the role. [Henry Archer] Henry Archer (1799–1863) was the son of an Irish landowner and was educated at Trinity College, Dublin.  He was called to the Irish Bar and spent most of his time between North Wales and London. [J. J. Clancy (North County Dublin MP)] John Joseph Clancy (15 July 1847 – 25 November 1928), usually known as J. J. Clancy, was an Irish nationalist politician and Member of Parliament (MP) in the House of Commons for North County Dublin from 1885 to 1918.  He was one of the leaders of the later Irish Home Rule movement and promoter of the Housing of the Working Classes (Ireland) Act 1908, known as the Clancy Act.  Called to the Irish Bar in 1887 he became a KC (King's Counsel) in 1906. [Thomas O'Shaughnessy] O'Shaughnessy was born on December 22, 1850 in Dublin, son of Thomas O'Shaughnessy and Mary Lopdell. 

    Based on the given context above, answer the question: Robbie Tucker plays in what series that follows a group of friends who run an Irish bar? 
     & 
     \textcolor{ForestGreen}{It's Always Sunny in Philadelphia}
     & 
     \textcolor{ForestGreen}{Robbie Tucker has appeared in the series "It's Always Sunny in Philadelphia," which follows a group of friends who run an Irish bar.}
     & 
     \textcolor{ForestGreen}{Robbie Tucker plays in the series It's Always Sunny in Philadelphia that follows a group of friends who run an Irish bar.}
     & 
     It's Always Sunny in Philadelphia

     \\ \hline 
     2. Given the context: [Tritoniopsis] Tritoniopsis is a genus of flowering plants in the family Iridaceae, first described as a genus in 1926.  The entire genus is endemic to Cape Province in South Africa.  The genus name refers to the African genus "Tritonia" and is combined with the Greek word "opsis", meaning "look-alike". [Grex (horticulture)] The term grex (pl.  greges or grexes; abbreviation gx), derived from the Latin noun "grex, gregis" meaning 'flock', has been coined to expand botanical nomenclature to describe hybrids of orchids, based solely on their parentage.  Grex names are one of the three categories of plant names governed by the International Code of Nomenclature for Cultivated Plants; within a grex the "cultivar group" category can be used to refer to plants by their shared characteristics (rather than by their parentage), and individual orchid plants can be selected (and propagated) and named as cultivars. [Phyteuma] Phyteuma is a genus of flowering plants in the family Campanulaceae, native to Europe and Morocco. [Zygotritonia] Zygotritonia is a genus of herbaceous, perennial and bulbous plants in the Iris family (Iridaceae).  It contains four species distributed in Tropical Africa.  The genus name is derived from the word zygomorphic (meaning bilabiate), and the apparent resemblance to some species in the genus "Tritonia". [List of plants by common name] This is a list of plants organized by their common names. However, the common names of plants often vary from region to region, which is why most plant encyclopedias refer to plants using their scientific names, in other words using binomials or "Latin" names. [International Plant Names Index] The International Plant Names Index (IPNI) describes itself as "a database of the names and associated basic bibliographical details of seed plants, ferns and lycophytes." Coverage of plant names is best at the rank of species and genus.  It includes basic bibliographical details, associated with the names, and its goals include eliminating the need for repeated reference to primary sources for basic bibliographic information about plant names. [Botanical garden] A botanical garden or botanic garden is a garden dedicated to the collection, cultivation and display of a wide range of plants labelled with their botanical names.  It may contain specialist plant collections such as cacti and other succulent plants, herb gardens, plants from particular parts of the world, and so on; there may be greenhouses, shadehouses, again with special collections such as tropical plants, alpine plants, or other exotic plants. Visitor services at a botanical garden might include tours, educational displays, art exhibitions, book rooms, open-air theatrical and musical performances, and other entertainment. [Thonningia] Thonningia is a monotypic genus of flowering plants in the family Balanophoraceae containing the single species Thonningia sanguinea.  It is distributed throughout much of southern and western Africa, particularly the tropical regions.  Common names for the plant include ground pineapple.  A familiar plant to humans, it has an extremely long list of common names in many African languages.  Many names are inspired by the resemblance of the plant's inflorescence to a pineapple or palm tree.  Some of the names can be translated as pineapple of the bush (from Anyi), duiker's kolanut (from Igala), and crown of the ground (from Yoruba). [Vascular plant] Vascular plants (from Latin "vasculum": duct), also known as tracheophytes (from the equivalent Greek term "trachea") and also higher plants, form a large group of plants (c. 308,312 accepted known species ) that are defined as those land plants that have lignified tissues (the xylem) for conducting water and minerals throughout the plant.  They also have a specialized non-lignified tissue (the phloem) to conduct products of photosynthesis.  Vascular plants include the clubmosses, horsetails, ferns, gymnosperms (including conifers) and angiosperms (flowering plants).  Scientific names for the group include Tracheophyta and Tracheobionta. [Tritonia (plant)] Tritonia (flame freesia) is a genus of flowering plants in the iris family first described as a genus in 1802.  They are naturally distributed across southern Africa, with a high concentration of species in Cape Province of western South Africa.  The genus is closely related to the genus "Ixia". 

    Based on the given context above, answer the question: Tritonia and Phyteuma are both names for a what of plants? 
     & 
     \textcolor{ForestGreen}{Tritonia and Phyteuma are both names for a genus of flowering plants.}
     & 
     \textcolor{ForestGreen}{Tritonia and Phyteuma are both names for a genus of flowering plants. Tritonia belongs to the Iris family (Iridaceae), while Phyteuma belongs to the Campanulaceae family.}
     & 
     \textcolor{ForestGreen}{Tritonia and Phyteuma are both names for a genus of plants. A genus is a group of closely related plant species that share many similar characteristics. For example, all species in the genus Tritonia are flowering plants in the iris family that are native to southern Africa. 
     }
     &
     genus

     \\ \hline 
     3. Given the context: [Heinrich Sutermeister] Heinrich Sutermeister (born Feuerthalen, 12 August 1910 – died Vaux-sur-Morges, 16 March 1995) was a Swiss opera composer. [Dieter Kaegi] Dieter Kaegi (born 1957) is a Swiss opera director. [Rahel Indermaur] Rahel Ava Indermaur (born 19 July 1980) is a Swiss opera singer and dramatic soprano.  Indermaur was born and raised in Berneck, Switzerland, a member of the In der Maur family.  She studied voice and singing at an international school in Berlin, Germany after completing primary school.  She was taught by Grace Bumbry, Marc Tucker, Charlotte Lehmann, David Lee Brewer, and Jean Ronald LaFond.  She was awarded the LYRA Music Prize, the Swiss Rotary Club Music Prize, the German Forum New York singing prize, and the Ernst Göhner Prize.  She was the first person to ever have been awarded the Cantonal Prize for Culture of the Canton of St. Gallen.  She has performed with the Chamber Opera Leipzig, the Music Theatre Hamburg, the Sorbian National Theatre, the Konstanz Theater, the Teatro Principal in Palma de Mallorca, the Deutsche Oper Berlin, the Chorin Opera Festival, the Berliner Philharmonie, the Tonhalle St. Gallen, the Tonhalle Zürich, the Berlin Chamber Orchestra, the South West Philharmonic, the Lower Silesian Philharmonic, the Dubrovnik Symphony Orchestra, the Antalya Chamber Orchestra, the Brandenburger Bachist, and the Kontanz Chamber Ochestra. [Félix Rienth] Félix Rienth (born 24 June 1970) is a Swiss operatic tenor.  Born in Basel, he was a member of the Basel Boys Choir in his youth.  He made his first opera appearance as a boy with Theater Basel as the first boy in Wolfgang Amadeus Mozart's "The Magic Flute".  He attended the University of Basel where he earned diplomas in the Spanish and German languages.  He then studied singing privately in Basel with Heidi Wölnerhanssen before entering the Hochschule der Künste Bern; graduating from there in 2000 with a degree in opera performance.  He has since had a major career as a concert singer in the oratorio repertoire; appearing with important ensembles and at major music festivals throughout Europe, including a recital in presence of Her Majesty, Queen Fabiola of Belgium.  He has made about 20 recordings on a variety of labels.  A highly acclaimed production was the recording of Johann Christoph Pepusch]]'s "Tenor Cantatas" with his wife, Muriel Rochat Rienth, recorder player, and Swiss baroque ensemble "La Tempesta Basel".  His CD of Spanish baroque songs "Tonos humanos" by José Marín was considered as a reference recording by German magazine "Klassik heute".  2014 is appearing Georg Philipp Telemann's "Tenor Cantatas" with "La Tempesta Basel", elected among "Best CDs of the month" by Spanish magazine RITMO. [Inga Åberg] Inga Åberg (Ingeborg Elisabeth; 1773–1837) was a Swedish actress and opera singer, one of the most popular and well known actors of her time in Sweden.  She was active both as an actress at the Royal Dramatic Theater, and as an opera singer at the Royal Swedish Opera between 1787 and 1810. [Sarah-Jane (singer)] Sarah-Jane (born September 26, 1985 in India) is a Swiss demotic singer (schlager, volksmusik) from the Canton of Basel-Country.  She works with composer Carlo Brunner (brother of Maya Brunner). [Gioconda Vessichelli] Gioconda Vessichelli (born in Rome, Italy) is an opera singer.  Gioconda is the inventor and pioneer of BollywoOPERA style.  She has sung in the Bollywood movie "Prague "in 2013, and she has given her voice for the Bollywood movie "Mary Kom" on the song "Ziddi dil" together with Vishal Dadlani.On 19th May 2017 her song "Itni si baat hai" has been released by the label T-series.  On 20 December 2014 the song "Thodi Daaru" was released featuring Mika Singh.  Gioconda has done live performances as first fusion ever between opera and classical Hindi music with artists of the caliber of Grammy Award winner Sukhwinder Singh, Hari Haran, Gino Banks, Niladri Kumar, Silvaganesha, in festivals and auditoriums like Nehru Centre[20] in Mumbai.  She collaborates with Anup Jalota, and many others.  On 31 March 2016 her international video "We are one" in which she sings and acts together with Anup Jalota for peace in the world has been launched from the Minister of India in Delhi.  She is in the annual book of Italian opera singer for having sung in the first world edition of contemporary opera at "Teatro comunale di Modena", broadcast on Italian state radio RAI.  Gioconda is the first singer in the premiere world ever of the contemporary opera “Lavinia fuggita” by Matteo d’Amico.  She has two degrees in opera singing and musicology from Italian conservatory "Santa Cecilia" and international high school of opera "H.Wolf".  The tenor Luciano Pavarotti has been her teacher, among other opera singers who selected her as one of the best students.  She was selected as best singer at accademia rossiniana in Pesaro and she debuts the role of "Madama cortese" in "il viaggio a Reims" by Rossini with the symphonic orchestra of Rof at Rossini theatre in Pesaro.  She sings again in a Rossini opera at Ercolano international opera season in "Il barbiere di Siviglia" and at teatro Politeama opera season in Lecce.  After Rossini she debuts "Carmen" by Bizet.  Her interpretation of the role "Mimì" in the opera "La Boheme" by Puccini received positive critiques. [Heidi Brunner] Heidi Brunner (born 1966) is a Swiss opera singer.  She began her career as a mezzo-soprano at the Anhaltisches Theater in Dessau in 1993.  After making appearances at the opera houses in Basel, Innsbruck, and Biel, she was a member of both the Vienna State Opera and Vienna Volksoper from 1996 to 1999.  Among the roles she sang in Vienna were Adalgisa in "Norma", Annio in "La clemenza di Tito", and the title role in "La Cenerentola".  

    Based on the given context above, answer the question: In which theater did Heidi Brunner, a Swiss opera singer, play the role as Annio?  

     & 
     \textcolor{red}{Heidi Brunner, a Swiss opera singer, played the role of Annio in an opera production at the Vienna State Opera and Vienna Volksoper.}
     & 
     \textcolor{red}{Heidi Brunner played the role of Annio at the Vienna State Opera and Vienna Volksoper.}
     & 
     \textcolor{red}{\textbf{The model could not process all the input}}
     &  
     Estates Theatre in Prague   

     \\ \hline 
     4. Given the context: [...Something to Be] ...Something to Be is the debut solo album from the Matchbox Twenty lead singer Rob Thomas.  The album was released on April 5, 2005, and it debuted at No. 1 on the U.S. "Billboard" 200 albums chart, knocking out Mariah Carey's "The Emancipation of Mimi".  This marked the first time that a male artist from a rock or pop group has debuted at number one with his first solo album since "Billboard" introduced the chart 50 years ago. [Diva (Annie Lennox album)] Diva is the debut solo album by the Scottish singer-songwriter Annie Lennox, released in 1992.  The album entered the UK album chart at number 1 and has since sold over 1.2 million copies in the UK alone, being certified quadruple platinum.  It was also a success in the US, where it was a top 30 hit and has been certified double platinum.  "Diva" won Album of the Year at the 1993 Brit Awards, and was nominated for Album of the Year at the Grammy Awards the same year. [The Best of Jim Diamond] The Best of Jim Diamond is a compilation album of tracks from Scottish singer-songwriter Jim Diamond's first two solo albums "Double Crossed" and "Desire for Freedom", along with B-sides. [Kacey Jones] Gail Zeiler (April 27, 1950 – September 1, 2016), known professionally as Kacey Jones, was an American singer-songwriter, producer and humorist.  After co-writing the Mickey Gilley hit "I'm the One Mama Warned You About" (credited as Gayle Zeiler), she found success as a performer through the band Ethel \& The Shameless Hussies, with whom she released her first album.  Later, in 1997, she released her first solo album, "Men Are Some of My Favorite People", through Curb Records, before founding her own label, IGO Records, co-founding the Kinkajou Records label with Kinky Friedman and creating two publishing houses—Zamalama Music and Mamalama Music.  Since her first solo album, Jones released eight CDs and produced music for both the theatrical comedy "Nipples to the Wind" and the movie (and TV series) "Sordid Lives". [Jim Diamond (singer)] James Aaron Diamond (28 September 1951 – 8 October 2015) was a Scottish singer-songwriter, best known for his three Top 5 hits.  The first was "I Won't Let You Down" (1982), as the lead singer in the trio PhD, with Tony Hymas and Simon Phillips.  His solo performance, "I Should Have Known Better", was a United Kingdom No.1 in 1984.  The third track was the theme song from "Boon", "Hi Ho Silver" which reached No.5 in the UK Singles Chart in 1986.  He has also featured as a vocalist on the charity No.1s "You'll Never Walk Alone" with The Crowd and "Let It Be" with Ferry Aid.  His last UK chart success was with "Young Love (Carry Me Away)" in 1986. [Boys and Girls (album)] Boys and Girls is the sixth solo studio album by the English singer and songwriter Bryan Ferry, released in June 1985 by E.G. Records.  The album was Ferry's first solo album in seven years and the first since he had disbanded his group Roxy Music in 1983.  The album was Ferry's first and only number one solo album in the UK.  It was certified Platinum by the British Phonographic Industry and contains two UK top 40 hit singles.  It is also Ferry's most successful solo album in the US, having been certified Gold for sales in excess of half a million copies there. [Wu-Syndicate] Wu-Syndicate is a group from Virginia consisting of Joe Mafia, Napoleon, and Myalansky (who named himself after the gangster Meyer Lansky).  They were originally called Crime Syndicate but changed their name to Wu-Syndicate when they signed to Wu-Tang Records and became Wu-Tang Clan affiliates.  After debuting on the compilation "" in 1998, their self-titled debut album "Wu-Syndicate" was released in 1999 on both Wu-Tang Records and their own label Slot Time Records.  The album was, like most releases from Wu-Tang Clan affiliates during this time enjoyed moderately successful sales with the single "Where Wuz Heaven" going gold.  Soon after the release there was a dispute of an unknown origin between the group and Wu-Tang Records and the group briefly changed their name to The Syndicate until 2009 with eventual reconciliation and the release of their second official album "Grimlenz", produced mostly by Antagonist Dragonspit of Virginia Beach,VA.  Both Myalansky and Napoleon continue to work with Joe Mafia but have refused to work with each other since the release of their first album.  In an interview Napoleon stated that though they have always clashed, "Mya is still my dude though regardless".  The group has maintained ties with various members of the Wu-Tang Family.  Napoleon is currently working on a project with fellow Wu-Tang alumni Solomon Childs, Shaka Amazulu, and Dexter Wiggle called "Illuminati Network".  Joe Mafia released his debut solo album "This One" in 2002 and founded his own label called 58 West Diamond Street Records.  Napoleon released his first solo album, "Kingpin Wit Da Inkpen" in 2007 and a mixtape titled "Mark of the Beast" in 2011.  Myalansky released his first solo album, "Drastic Measures" in 2008 and a mixtape a few years later in 2011 "AMW.Com".  Myalansky has also been working with California rapper Mitchy Slick and has released two more volumes of his "AMW.Com" mixtape series.  In 2013 Myalansky and Joe Mafia featured on the song "Golden Age Rapper" by CHG Unfadable. [City to City] City to City is a 1978 album and the second studio album by Scottish singer-songwriter Gerry Rafferty.  It was Rafferty's first solo release in six years—and first release of any kind since 1975—due to his tenure in the band Stealers Wheel and subsequent legal proceedings which prevented Rafferty from releasing any new solo recordings for the next three years.  The album was strongly received, peaking at No. 1 in the US and going Platinum, as well as reaching No. 6 in the UK and achieving Gold status. 

    Based on the given context above, answer the question: What was the name of Scottish singer-songwriter James Aaron Diamond's first solo album that debuted in 1985? 

     & 
     \textcolor{red}{James Aaron Diamond's first solo album that debuted in 1985 was not mentioned in the given context.}
     & 
     \textcolor{red}{The given context does not provide the name of James Aaron Diamond's first solo album that debuted in 1985.}
     & 
     \textcolor{ForestGreen}{The name of Scottish singer-songwriter James Aaron Diamond's first solo album that debuted in 1985 was Double Crossed.}
     &
     Double Crossed

     \\ \hline 
     5. Given the context: [High Noon] High Noon is a 1952 American Western film produced by Stanley Kramer from a screenplay by Carl Foreman, directed by Fred Zinnemann, and starring Gary Cooper.  The plot, depicted in real time, centers around a town marshal, torn between his sense of duty and love for his new bride, who must face a gang of killers alone. [Oklahoma! (1955 film)] Oklahoma!  is a 1955 musical film based on the 1943 stage musical "Oklahoma! ", written by composer Richard Rodgers, and lyricist/librettist Oscar Hammerstein II and starring Gordon MacRae, Shirley Jones (in her film debut), Rod Steiger, Charlotte Greenwood, Gloria Grahame, Gene Nelson, James Whitmore and Eddie Albert.  The production was the only musical directed by Fred Zinnemann.  "Oklahoma!"  was the first feature film photographed in the Todd-AO 70 mm widescreen process (and, was simultaneously filmed in CinemaScope 35mm). [Montgomery Clift] Edward Montgomery "Monty" Clift ( ; October 17, 1920July 23, 1966) was an American film and stage actor. " The New York Times"’ obituary of Clift noted his portrayal of "moody, sensitive young men".  He is best remembered for roles in "Red River" (1948), "The Heiress" (1949), George Stevens's "A Place in the Sun" (1951), as a Catholic priest in Alfred Hitchcock's "I Confess" (1952), as the self-destructive soldier Prewitt in Fred Zinnemann's "From Here to Eternity" (1953), in Edward Dmytryk's "The Young Lions" (1958), and as a mentally challenged, sterilized concentration camp survivor in Stanley Kramer's "Judgment at Nuremberg" (1961).  He received four Academy Award nominations during his career: three for Best Actor and one for Best Supporting Actor. [From Here to Eternity] From Here to Eternity is a 1953 drama film directed by Fred Zinnemann and based on the novel of the same name by James Jones.  The picture deals with the tribulations of three U.S. Army soldiers, played by Burt Lancaster, Montgomery Clift, and Frank Sinatra, stationed on Hawaii in the months leading up to the attack on Pearl Harbor.  Deborah Kerr and Donna Reed portray the women in their lives and the supporting cast includes Ernest Borgnine, Philip Ober, Jack Warden, Mickey Shaughnessy, Claude Akins, and George Reeves. [The Search (2014 film)] The Search is a 2014 French drama film written, directed, produced and co-edited by Michel Hazanavicius and co-produced by Thomas Langmann.  The film is a reiteration of the Oscar-winning post-Holocaust drama also called "The Search", directed by Fred Zinnemann, in which a compassionate westerner helps a lost child find what is left of his family amidst the chaotic flood of post-war civilian refugees.  In the 1948 film, the backdrop is post-war Berlin; "The Search" (2014) takes place in the "front lines of the Russian invasion of Chechnya" during the first year of the Second Chechen War (1999-2009).  In both cases, international aid workers help the families reunite.  "The Search" was selected to compete for the Palme d'Or in the main competition section at the 2014 Cannes Film Festival. [Five Days One Summer] Five Days One Summer is a 1982 American drama film directed by Fred Zinnemann and starring Sean Connery.  It was the last film that Zinnemann directed. [Act of Violence] Act of Violence is a 1949 American film noir directed by Fred Zinnemann and adapted for the screen by Robert L. Richards from a story by Collier Young, starring Van Heflin, Robert Ryan and Janet Leigh.  The film was one of the first to address not only problems of returning World War II veterans but also the ethics of war. [People on Sunday] People on Sunday (German: Menschen am Sonntag ) is a 1930 German silent drama film directed by Robert Siodmak and Edgar G. Ulmer from a screenplay by Billy Wilder.  The film follows the lives of a group of residents of Berlin on a summer's day during the interwar period.  Hailed as a work of genius, it is a pivotal film not only in the development of German cinema but also of Hollywood.  In addition to the directors and Wilder, the film features the talents of Curt Siodmak (story), Fred Zinnemann (cinematography) and Eugen Schüfftan, who had developed the Schüfftan process for "Metropolis" three years earlier. [The Member of the Wedding (film)] The Member of the Wedding is a 1952 drama film directed by Fred Zinnemann, starring Ethel Waters and Julie Harris.  The story is set in a small town in the Southern United States.  Frankie Addams is an awkward, moody twelve-year-old tomboy whose only friends are her young cousin John Henry and her black housekeeper Berenice.  Co-starring as a drunken soldier who tries to take advantage of the vulnerable Frankie is former child actor Dick Moore, making his last film appearance. [Rod Steiger on screen and stage] Rod Steiger was an American actor who had an extensive career in film, television, and stage.  He made his stage debut in 1946 with Civic Repertory Theatre's production of the melodrama "Curse you, Jack Dalton!" .  Four years later, he played onstage in a production of "An Enemy of the People" at the Music Box Theatre.  A small role in Fred Zinnemann's "Teresa" (1951) marked his film debut.  In 1953, he played the title role in the teleplay "Marty" (two years before the film starring Ernest Borgnine) to critical praise.  His breakthrough role came with the crime drama "On the Waterfront" (1954), which earned him an Academy Award for Best Supporting Actor nomination, and subsequent appearance in Fred Zinnemann's musical "Oklahoma! ". 

    Based on the given context above, answer the question: This American film and stage actor played the self-destructive soldier Prewitt in Fred Zinnemann's 1953 film that centers around which event? 
     & 
     \textcolor{ForestGreen}{The American film and stage actor who played the self-destructive soldier Prewitt in Fred Zinnemann's 1953 film is Montgomery Clift, and the film centers around the months leading up to the attack on Pearl Harbor. The film is titled "From Here to Eternity" and also stars Burt Lancaster and Frank Sinatra.}
     & 
     \textcolor{ForestGreen}{This American film and stage actor, Montgomery Clift, played the self-destructive soldier Prewitt in Fred Zinnemann's 1953 film "From Here to Eternity," which centers around the events leading up to the attack on Pearl Harbor.}
     & 
     \textcolor{red}{I'm not able to help with that, as I'm only a language model.}
     &
     the attack on Pearl Harbor
     
     \\ \hline 
     6. Given the context: [List of erotic thriller films] An Erotic thriller is a film genre defined by a thriller with a thematic basis in illicit romance or erotic fantasy.  Though most erotic thrillers contain scenes of softcore sex, the frequency and explicitness of those scenes varies.  If a film is a thriller with scenes of softcore sex or nudity, it is probably not an erotic thriller unless illicit romance or erotic fantasy is central to the dramatic conflict, as in "Body Heat", "Fatal Attraction", and "Night Eyes 3".  Many crime thrillers, action films, and slasher films contain softcore sex and/or nudity but are not erotic thrillers.  Likewise, if a film is not identifiably a thriller, it is probably not an erotic thriller but simply a work of softcore erotica or perhaps even erotic melodrama. [Wild Things (film)] Wild Things is a 1998 American erotic thriller film directed by John McNaughton, and stars Matt Dillon, Neve Campbell, Kevin Bacon, Denise Richards and Theresa Russell. [Wild Things: Foursome] Wild Things: Foursome is a 2010 erotic thriller film directed by Andy Hurst and stars Jillian Murray, Marnette Patterson, Ashley Parker Angel and John Schneider.  It is a sequel to "" (2005) and the fourth and final film in the "Wild Things" series. [Jennifer Taylor (actress)] Jennifer Taylor (née Bini, and sometimes credited as Jennifer Bini Taylor; born April 19, 1972) is an American actress, best known for her role as Chelsea Melini on CBS sitcom "Two and a Half Men", and earlier, for three other female roles on the show.  She appeared in the 1998 erotic thriller "Wild Things".  One of her earliest roles as a female lead was in a television series in the role of the character Laura in the short-lived television drama "Miami Sands". [Washington Wild Things] The Washington Wild Things are a professional baseball team based in the Pittsburgh suburb of Washington, Pennsylvania, in the United States.  The Wild Things are a member of the East Division of the Frontier League, an independent baseball league which is not affiliated with Major League Baseball.  From the 2002 season to the present, the Wild Things have played their home games at Wild Things Park. [Zombie Night 2: Awakening] Zombie Night 2: Awakening is a 2006 Canadian horror film directed by David J. Francis.  It is a conceptual sequel to "Zombie Night".  It was followed in 2008 by "Reel Zombies". [Wild Things: Diamonds in the Rough] Wild Things: Diamonds in the Rough (also known as Wild Things 3) is a 2005 erotic thriller film directed by Jay Lowi and stars Sandra McCoy, Serah D'Laine, Linden Ashby, Dina Meyer and Brad Johnson.  It is a sequel to "Wild Things 2" (2004) and the third film in the "Wild Things" series. [Wild Things 2] Wild Things 2 is a 2004 erotic thriller film directed by Jack Perez and stars Susan Ward, Leila Arcieri, Isaiah Washington and Linden Ashby.  It is a sequel to "Wild Things" (1998) and the second film in the "Wild Things" series. [Consol Energy Park] Wild Things Park is a 3,200-seat multi-purpose baseball stadium in North Franklin Township, a suburb of Washington, Pennsylvania.  It hosted its first regular season baseball game on May 29, 2002, as the primary tenants of the facility, the Washington Wild Things, lost to the Canton Coyotes, 3-0.  The ballpark also hosts the California University of Pennsylvania Vulcans baseball team.  It was the home of the Pennsylvania Rebellion of the National Pro Fastpitch, a women's professional softball league, until 2017 when the team folded.  It also hosts Trinity High School's baseball team and the WPIAL Baseball Championships.  It was briefly the home of the Pittsburgh Riverhounds soccer club (who now play at Highmark Stadium) during the 2005 and 2006 seasons.  Wild Things Park is located near Interstate 70 and is notable for including a hot tub in the viewing stands.  ProGrass Synthetic Turf was installed in the fall of 2010. [The Wild Things] The Wild Things is a 2009 full-length novel written by Dave Eggers and published by McSweeney's.  The book is based on the screenplay of "Where the Wild Things Are" which Eggers co-wrote.  The film is, in turn, based on Maurice Sendak's children's book "Where the Wild Things Are". 

    Based on the given context above, answer the question: Which of the actress starred in  Zombie Night appeared in the 1998 erotic thriller "Wild Things"? 
     & 
     \textcolor{red}{The actress who starred in Zombie Night and appeared in the 1998 erotic thriller "Wild Things" is not mentioned in the given context.}
     & 
     \textcolor{red}{There is no information provided about the cast of "Zombie Night" in the given context. Therefore, it is not possible to determine which actress from "Zombie Night" appeared in the 1998 erotic thriller "Wild Things".}
     & 
     \textcolor{red}{I'm not able to help with that, as I'm only a language model.}
     &
     Jennifer Taylor
     
     \\ \hline 
     7. Given the context: [Tertiary Student Rugby League World Cup] The Tertiary Student Rugby League World Cup, also known as the University Rugby League World Cup, first took place in 1986 in New Zealand, when the then five test nations each entered a side in what was the first non-first grade World Cup hosted by the Rugby League International Federation. [1991 Rugby World Cup] The 1991 Rugby World Cup was the second edition of the Rugby World Cup, and was jointly hosted by England, Scotland, Wales, Ireland and France; at that time, the five European countries that participated in the Five Nations Championship making it the first Rugby World Cup to be staged in the northern hemisphere, with England as the host of the championship game. Following on from the success of the inaugural 1987 Rugby World Cup, the 1991 World Cup received increased attention and was seen as a major global sporting event for the first time.  Also for the first time qualifying competitions were introduced as the number of entrants had increased from 16 nations four years before to a total of 33 countries, the eight quarter-finalists from 1987 qualified automatically with the remaining 25 countries having to qualify for the remaining eight spots.  This however resulted in only one new side qualifying for the tournament, Western Samoa replacing Tonga.  The same 16-team pool/knock-out format was used with just minor changes to the points system. [2002 FIFA World Cup] The 2002 FIFA World Cup was the 17th staging of the FIFA World Cup which took place from 31 May to 30 June 2002 in South Korea and Japan.  It was the first World Cup to be held in Asia, the first to be held on a continent other than Europe or the Americas, the last World Cup during which the golden goal rule was in force and the only World Cup to be jointly hosted by more than one nation. Brazil won the tournament for a record fifth time, winning the final against Germany 2–0.  The victory meant Brazil qualified for the 2005 FIFA Confederations Cup for the fifth time, representing the World.  In the third place play-off match against South Korea, Turkey won 3–2 taking third place in only their second ever FIFA World Cup finals.  China PR, Ecuador, Senegal and Slovenia made their first appearances at the finals with Turkey making its first appearance since 1954.  Republic of Ireland, Russia and Sweden returned after missing the 1998 tournament. [1968–69 in Belgian football] The 1968–1969 season was the 66th season of competitive football in Belgium.  Standard Club Liégeois won their 4th Division I title.  No Belgian club managed to pass the second round of the European competitions, though for the first time 6 Belgian clubs qualified (1 more club qualified for the Inter-Cities Fairs Cup).  K Lierse SK won the Belgian Cup final against RR White (2-0).  The Belgium national football team continued their 1970 FIFA World Cup qualification campaign with 3 home wins against all other Group 6 teams (Finland, Yugoslavia and Spain) and a draw in Spain.  Belgium qualified for the 1970 FIFA World Cup finals in Mexico with one match to go.  For the first time since the 1954 FIFA World Cup, Belgium would qualify for a major tournament. [Orvar Bergmark] Orvar Bergmark (16 November 1930 – 10 May 2004) was a Swedish football defender and manager.  He was the second Swedish national manager ever, and managed to qualify the Swedish national football team for the FIFA World Cup in Mexico 1970, after having beaten France (among others) in the qualifications.  The 1970 World Cup was the first one in 12 years, and it was also the first time in 20 years that Sweden had "qualified" for a World Cup.  (Sweden hosted the 1958 FIFA World Cup, and hence did not need to qualify.)  Sweden did not qualify for the World Cups in 1954, 1962 and 1966). [Ukraine national football team] The Ukraine's National Football Team, is the national football team of Ukraine and is controlled by the Football Federation of Ukraine.  After Ukrainian Independence and the country's breakaway from the Soviet Union, they played their first match against Hungary on 29 April 1992.  The team's biggest success on the world stage was reaching the quarter-finals in the 2006 FIFA World Cup, which also marked the team's debut in the finals of a major championship.  As the host nation, Ukraine automatically qualified for UEFA Euro 2012.  Four years later, Ukraine qualified for Euro 2016 via the play-off route, the first time qualifying for a UEFA European Championship via the qualifying process, as it finished in third place in its qualifying group.  This marked the first time in Ukraine's five play-off appearances that it managed to win such a tie, previously unsuccessful in the play-off ties for the Euro 2000, 2002 World Cup, 2010 World Cup and 2014 World Cup. [Algeria at the FIFA World Cup] Algeria have appeared in the finals of the FIFA World Cup on four occasions in 1982, 1986, 2010 and 2014.  They have once qualified for the knockout rounds, reaching the round of 16 in 2014 before losing to Germany.  32 years before, Algeria nearly qualified to the second round of the 1982 World Cup after beating both West Germany and Chile; however, an arrangement match between West Germany and Austria wound up eliminating the Algerians.  In 2014, Algeria qualifyed for the first time into the round of 16. [Switzerland women's national football team] The Switzerland women's national football team represents Switzerland in international women's football.  The team played its first match in 1972.  Their most recent competition is qualification for the 2015 FIFA Women's World Cup, and they had qualified as the best team in their group.  It would be the first time that Switzerland participated in a women's World Cup, and the first time both the men's team and women's team qualified for a World Cup simultaneously. 

    Based on the given context above, answer the question: Where was the world cup hosted that Algeria qualified for the first time into the round of 16? 
     & 
     \textcolor{ForestGreen}{Algeria qualified for the first time into the round of 16 in the 2014 FIFA World Cup, which was hosted in Brazil.}
     & 
     \textcolor{ForestGreen}{Algeria qualified for the first time into the round of 16 in the 2014 FIFA World Cup.}
     & 
     \textcolor{ForestGreen}{The world cup that Algeria qualified for the first time into the round of 16 was hosted in Brazil.}
     &
     Brazil
     
     \\ \hline 
     8. Given the context: [Ghostlights] Ghostlights is the seventh full-length album by Tobias Sammet's rock opera project Avantasia.  It was released on 29 January 2016.  The opening track and first single "Mystery of a Blood Red Rose" was a candidate for the German representative at the Eurovision Song Contest 2016, along with nine other songs.  The Digibook edition of the album included a bonus track and a bonus CD entitled "Avantasia Live", which featured songs recorded mainly during Avantasia's 2014 performance at Wacken Open Air Festival. [Dezperadoz] Dezperadoz (formerly "Desperados") is a German "Western-metal" band and is the side-project of Tom Angelripper's guitarist Alex Kraft.  The band plays heavy metal music that is heavily influenced by the soundtracks of the 1960s and 1970s Spaghetti western movies.  They've released four albums, two on the German heavy metal record label AFM.  They've also had guest appearances by many notable heavy metal musicians including Michael Weikath, Tobias Sammet, Joacim Cans, and Doro Pesch. [New Era (Revolution Renaissance album)] New Era is the debut album of Finnish power metal pioneer Timo Tolkki's band, Revolution Renaissance.  After Tolkki's split from Stratovarius in 2008 he used what fan's knew the album to be titled "R.. .  R.. ."  as the namesake for his new project.  Using material he had written for Stratovarius's upcoming album, Tolkki's "New Era" features the talents of German power metal vocalist Tobias Sammet and German singer Michael Kiske among others. [Sign of Truth] Sign of Truth is the first album by Swedish/German power metal band Dionysus.  It was recorded at the Röhn Studio in Fulda, Germany, which is best known for its productions with Edguy and Avantasia, by Tobias Sammet and mixed by Tommy Newton (Keeper of the Seven Keys, Part 1 \& 2). [The Mystery of Time] The Mystery of Time is the sixth full-length album by Tobias Sammet's rock opera project Avantasia.  The album was released on 26 April 2013.  This is the first Avantasia release to feature the German Film Orchestra Babelsberg (the same orchestra that performed on Edguy's album "Hellfire Club").  The cover artwork was painted by Rodney Matthews. [The Scarecrow (album)] The Scarecrow is the third full-length album by Tobias Sammet's rock opera project Avantasia, released on January 25, 2008, through Nuclear Blast Records.  It is the first part of "The Wicked Trilogy" and it is followed by "The Wicked Symphony" and "Angel of Babylon".  Sammet explained in a 2016 interview that he managed to have Alice Cooper after drummer Eric Singer asked him.  Cooper requested to have a listen to the song first and then accepted to be part of it. [Elected (Ayreon EP)] Elected is the first EP by progressive metal project Ayreon, owned by Dutch musician Arjen Anthony Lucassen.  It was released on 25 April 2008 in Germany, Austria and Switzerland, and on 28 April in the rest of Europe.  It features the guest performance of German singer Tobias Sammet, from Avantasia and Edguy. [Monuments (Edguy album)] Monuments is a compilation album by the German power metal band Edguy, released on 14 July 2017.  The album was released as a two-CD/DVD package including 22 old songs, five new songs, and a previously unreleased song from the original "Savage Poetry" recordings back in 1995.  The package also includes a two-hour DVD and a photobook.  Frontman Tobias Sammet wrote two of the new songs on his own, while guitarists Jens Ludwig, Dirk Sauer and bassist Tobias Exxel co-wrote each of the other three. [Tobias Exxel] Tobias "Eggi" Exxel (born 27 February 1973) is the current bass player for the power metal band Edguy.  Tobias was hired by Edguy in 1998 when vocalist Tobias Sammet, who had played bass since the band's formation in 1992, decided the band should have someone who could focus on playing the instrument full-time. [Tobias Sammet] Tobias Sammet (21 November 1977) is a German musician, singer, songwriter and music producer best known for being the founder the Rock Opera Project Avantasia and singer and primary songwriter of the Heavy Metal band Edguy. 

    Based on the given context above, answer the question: Is Dappy or Tobias Sammet German?
     & 
     \textcolor{ForestGreen}{Tobias Sammet is German, according to the given context. There is no mention of Dappy in the context provided.}
     & 
     \textcolor{ForestGreen}{Tobias Sammet is German.}
     & 
     \textcolor{ForestGreen}{Tobias Sammet is German. Dappy is not German.}
     &
     Tobias Sammet (21 November 1977) is a German musician
     
     \\ \hline 
     9. Given the context: [Cash McCall] Cash McCall is a 1960 American romantic drama film in Technicolor from Warner Bros., produced by Henry Blanke, directed by Joseph Pevney, that stars James Garner and Natalie Wood.  It is based upon the novel of the same name by Cameron Hawley about a man who buys moribund businesses in order to refurbish them and then sell them on at considerable profit.  The film's screenplay is by Lenore J. Coffee and Marion Hargrove. [Ae Fond Kiss...] Ae Fond Kiss… (also known as "Just a Kiss" in some countries) is a 2004 romantic drama film directed by Ken Loach, and starring Atta Yaqub and Eva Birthistle.  The title is taken from a Scottish song by Robert Burns, the complete line being ""Ae Fond Kiss, and then we sever..."" [Shady Deal at Sunny Acres] "Shady Deal at Sunny Acres", starring James Garner and Jack Kelly, remains one of the most famous and widely discussed episodes of the Western comedy television series "Maverick".  Written by series creator Roy Huggins (teleplay) and Douglas Heyes (story) and directed by Leslie H. Martinson, this 1958 second season episode depicts gambler Bret Maverick (James Garner) being swindled by a crooked banker (John Dehner) after depositing the proceeds from a late-night poker game.  He then surreptitiously recruits his brother Bart Maverick (Jack Kelly) and a host of other acquaintances to mount an elaborate sting operation to recover the money. [The New Maverick] The New Maverick is a 1978 made-for-TV movie based on the 1957 television series "Maverick", with James Garner as Bret Maverick, Charles Frank as newcomer cousin Ben Maverick (son of Beau Maverick), Jack Kelly as Bart Maverick, and Susan Sullivan as Poker Alice Ivers.  Garner had been 29 years old at the beginning of the original series and was 50 while filming "The New Maverick".  The TV-movie was a pilot for the series "Young Maverick", which featured Frank and only lasted a few episodes.  Directed by Hy Averback and written by Juanita Bartlett, the movie was filmed while Garner's series "The Rockford Files" was on hiatus.  Garner would later star in "Bret Maverick", another attempt at a television series revival inspired by this TV-movie, for the 1981-82 season. [The Notebook] The Notebook is a 2004 American romantic drama film directed by Nick Cassavetes and based on the 1996 novel of the same name by Nicholas Sparks.  The film stars Ryan Gosling and Rachel McAdams as a young couple who fall in love in the 1940s.  Their story is narrated from the present day by an elderly man (portrayed by James Garner) telling the tale to a fellow nursing home resident (played by Gena Rowlands, who is Cassavetes's mother). [N. T. Rama Rao Jr. filmography] Rama Rao Jr.'s first leading role came opposite Raveena Rajput in "Ninnu Choodalani" (2001), a romantic drama directed by V. R. Prathap, for which Rao was heavily criticized, mainly for his looks and acting ability.  Later that same year, he had similar roles in two coming-of-age romantic dramas, wherein he portrayed college-going students: "Student No. 1" and "Subbu", the former being his major break into Tollywood.  2002 marked a turning point in his career, with two low-budget films, "Aadi" and "Allari Ramudu", becoming box-office hits.  The former, an action drama, received highly positive reviews, with critics marking the improvement in his performance, while the latter, a melodrama, received mixed reviews, but did well at the box office.  "Aadi" also did well financially, becoming the third highest-grossing film of 2002.  He acted in two films in 2003, "Naaga", and "Simhadri".  While the former was a forgettable film, the latter became a huge blockbuster.  The success of "Simhadri" not only cemented his position in Tollywood, but also led to him selecting scripts with similar roles.  He followed this success by portraying leading roles in a series of critical and commercial failures, including "Andhrawala" (2004), "Samba" (2004), "Naa Alludu" (2005), "Narasimhudu" (2005), and "Ashok" (2006), leading critics to believe that his career was over.  However, in 2006, his career prospects improved when he played the role of Ramakrishna, an unemployed youngster avenging his sister's death, in director Krishna Vamsi's drama thriller "Rakhi".  The film received highly positive reviews from critics, with many terming it as his finest performance.  Subsequently, it was declared a hit at the box office. [The Merchant of Venice (2004 film)] The Merchant of Venice is a 2004 romantic drama film based on Shakespeare's play of the same name.  It is the first full-length sound film in English of Shakespeare's playother versions are videotaped productions which were made for television, including John Sichel's 1973 version and Jack Gold's 1980 BBC production. [Milan (2004 film)] Milan is a 2004 romantic drama film released under Star Cinema, ABS-CBN Film Productions, Inc. in the Philippines.  It stars Claudine Barretto and Piolo Pascual.  It is a love story written by Raymond Lee, and under Olivia Lamasan's direction.  The movie made more than P100 million in the box office. [Bride and Prejudice] Bride and Prejudice is a 2004 romantic drama film directed by Gurinder Chadha.  The screenplay by Chadha and Paul Mayeda Berges is a Bollywood-style adaptation of "Pride and Prejudice" by Jane Austen.  It was filmed primarily in English, with some Hindi and Punjabi dialogue.  The film released in the United States on 11 February 2005 and was well received by critics. [Mister Buddwing] Mister Buddwing is a 1966 American film drama directed by Delbert Mann and starring James Garner. 
     
     Based on the given context above, answer the question: James Garner had a role in the 2004 romantic drama directed by whom? 

     & 
     \textcolor{ForestGreen}{James Garner had a role in the 2004 American romantic drama film "The Notebook" directed by Nick Cassavetes.}
     & 
     \textcolor{red}{There is no information provided in the context about James Garner having a role in a 2004 romantic drama.}
     & 
     \textcolor{ForestGreen}{James Garner had a role in the 2004 romantic drama The Notebook directed by Nick Cassavetes.}
     &
     Nick Cassavetes
     
     \\ \hline 
     10. Given the context: [Sheena Duncan] Sheena Duncan (7 December 1932 – 4 May 2010) was a South African anti-Apartheid activist and counselor.  Duncan was the daughter of Jean Sinclair, one of the co-founders of the Black Sash, a group of white, middle-class South African women who offered support to black South Africans and advocated the non-violent abolishment of the Apartheid system.  Duncan served two terms as the leader of Black Sash. [Yves Apollinaire Pede] Yves Apollinaire Pede (born 1959, in Abomey) is a Beninois Vodou artist.  After being commissioned to reproduce reliefs for the Abomey Museum, he made sand paintings of well-known personalities such as Nelson Mandela.  He gradually became oriented towards textile art, looking to Haitian and Cuban Vodou artists for inspiration.  He is also noted for his large cement sculptures and bas-reliefs, and is stated to have a "special interest" in Kulito, a Fon word which literally means "the one from the path of death".  He is based in Ouidah, the world centre for Vodun art, which has an annual festival. [Oliver Tambo] Oliver Reginald Kaizana Tambo (27 October 191724 April 1993) was a South African anti-apartheid politician and revolutionary who served as President of the African National Congress (ANC) from 1967 to 1991. [Samuel Sisulu] Samuel Sisulu (June 1956 - June 2003) was a South African anti-apartheid activist, Soweto uprising student leader and founder of South African Freedom Organisation (SAFO).  He was jailed at various prisons including Robben Island in 1978, two years after the 1976 Soweto uprising.  Samuel Sisulu was charged under the Terrorism Act.  Reported in WIP 3.  He was mentioned in the indictment of Paul Langa and found guilty of founding South African Freedom Organisation (SAFO), attempting to cripple the economy of the country by bombing the apartheid government, inciting persons to persuade taxi drivers not to transport workers to place of work, aid strikes, unlawfully aid students in their fight against Bantu education and was also found guilty of recruiting people for military training. [Nelson Mandela] Nelson Rolihlahla Mandela ( ; 18 July 1918 – 5 December 2013) was a South African anti-apartheid revolutionary, politician, and philanthropist, who served as President of South Africa from 1994 to 1999.  He was the country's first black head of state and the first elected in a fully representative democratic election.  His government focused on dismantling the legacy of apartheid by tackling institutionalised racism and fostering racial reconciliation.  Ideologically an African nationalist and socialist, he served as President of the African National Congress (ANC) party from 1991 to 1997. [Mandela: Long Walk to Freedom] Mandela: Long Walk to Freedom is a 2013 British-South African biographical film directed by Justin Chadwick from a script written by William Nicholson and starring Idris Elba and Naomie Harris.  The film is based on the 1995 autobiographical book "Long Walk to Freedom" by anti-apartheid revolutionary and former South African President Nelson Mandela. [Mkhuseli Jack] Mkhuseli "Khusta" Jack (born 31 May 1958) was a South African Anti-Apartheid activist in the 1980s, known for his involvement in the Anti-Apartheid struggle and his efforts in the Consumer Boycott Campaign.  Currently he is a businessman in Port Elizabeth, South Africa. [Steve Biko] Bantu Stephen Biko (18 December 1946 – 12 September 1977) was a South African anti-apartheid activist.  Ideologically an African nationalist and African socialist, he was at the forefront of a grassroots anti-apartheid campaign known as the Black Consciousness Movement during the late 1960s and 1970s.  His ideas were articulated in a series of articles published under the pseudonym Frank Talk. [Letitia Sibeko] Letitia Sibeko (born 1930) was a South African anti-apartheid activist.  She was member of the African National Congress, the ANC Women's League as well as the Federation of South African Women.  She later joined the Communist Party. [Gcina Mhlope] Nokugcina Elsie Mhlophe (born 1958) is a well-known South African anti-apartheid activist, actress, storyteller, poet, playwright, director and author.  Storytelling is a deeply traditional activity in Africa and Mhlophe is one of the few woman storytellers in a country dominated by males.  She does her work through charismatic performances, working to preserve storytelling as a means of keeping history alive and encouraging South African children to read.  She tells her stories in four of South Africa's languages: English, Afrikaans, Zulu and Xhosa. 
     
     Based on the given context above, answer the question: Yves Apollinaire Pede made sand paintings of which South African anti-apartheid revolutionary?  
     & 
     \textcolor{ForestGreen}{Yves Apollinaire Pede made sand paintings of Nelson Mandela, who was a South African anti-apartheid revolutionary and former President of South Africa.}
     & 
     \textcolor{ForestGreen}{Yves Apollinaire Pede made sand paintings of Nelson Mandela, a South African anti-apartheid revolutionary.}
     & 
     \textcolor{ForestGreen}{Yves Apollinaire Pede made sand paintings of Nelson Mandela, a South African anti-apartheid revolutionary.}
     &
     Nelson Mandela
     
     \\ \hline 
     11. Given the context: [SLS Las Vegas] The SLS Hotel \& Casino Las Vegas (formerly Sahara Hotel and Casino) is a hotel and casino located on the Las Vegas Strip in Winchester, Nevada.  It is owned by Stockbridge Real Estate but is under contract to be purchased by Alex Meruelo and Meruelo Group (owners of the Grand Sierra Resort Hotel \& Casino in Reno) with an expected closing date of Q3 2017. [Playboy Club (Las Vegas)] The Playboy Club was a nightclub formerly located on the 52nd floor of the "Fantasy Tower" at the Palms Casino Resort in Paradise, Nevada, which is located in the Las Vegas Valley.  The club was a Playboy-themed casino and the first official Playboy Club in the United States since 1988.  The club opened in October 2006 and closed in June 2012. [Legends Resort \&amp; Country Club] The Legends Resort \& Country Club, often called simply Legends, is a hotel located on County Route 517 in Vernon Township in Sussex County, New Jersey, United States.  In the 1970s, Hugh Hefner built it as The Great Gorge Playboy Club Hotel, officially opened in 1972.  The Playboy Club was closed circa 1982 and sold and turned into The Americana Hotel.  Later being sold again, was turned into The Seasons Hotel.  Seasons was later sold again to parent-company Metairie Corp (owned by Hillel A. "Hillie" Meyers), which turned into its current incarnation as the Legends Resort \& Country Club.  The Hotel has been derelict and permanently closed to public operations for many years.  In February 2017 Vernon Township started to evict many of the low income full-time residents of the hotel.  It was revealed Andrew Mulvihill was the owner of many of the rooms rented out illegally to low income residents.  Mulvihill has a controlling interest in the derelict Great Gorge Village in Vernon Township and the nearby Crystal Springs Resort in Hardyston.  Some of the evicted residents are thought to have been relocated to these developments. [Playboy Club] The Playboy Club was initially a chain of nightclubs and resorts owned and operated by Playboy Enterprises.  The first club opened at 116 E. Walton Street in downtown Chicago, Illinois, United States, on February 29, 1960.  Each club generally featured a Living Room, a Playmate Bar, a Dining Room, and a Club Room.  Members and their guests were served food and drinks by Playboy Bunnies, some of whom were featured in "Playboy" magazine.  The clubs offered name entertainers and comedians in the Club Rooms, and local musicians and the occasional close-up magician in the Living Rooms.  Starting with the London and Jamaica club locations, the Playboy Club became international in scope.  In 1991, the club chain became defunct.  On October 6, 2006 a new club was opened in Las Vegas, and in 2010 new clubs were opened as well in Macao and Cancun.  In time the Las Vegas club closed on June 4, 2012, the Macao club closed in 2013 and the Cancun club closed in 2014.  In May 2014 the Commerce Casino in Los Angeles opened a Playboy themed lounge consisting of gaming tables and Playboy Bunny cocktail waitresses. [Hooters Casino Hotel] Hooters Casino Hotel is a hotel and casino located off the Las Vegas Strip in Paradise, Nevada, United States.  It is owned by Trinity Hotel Investors and operated by the Navegante Group.  It is located off the Strip next to the Tropicana and across the street from the MGM Grand Las Vegas.  The hotel has 696 rooms with a 35000 sqft casino. [Palms Casino Resort] Palms Casino Resort is a hotel and casino located near the Las Vegas Strip in Paradise, Nevada.  It has 703 rooms and suites and contains 94840 sqft casino, recording studio, Michelin-starred restaurant and 2,500-seat concert theater. [Tropicana – Las Vegas Boulevard intersection] The Tropicana – Las Vegas Boulevard intersection on the Las Vegas Strip (Tropicana Avenue and Las Vegas Boulevard), is noteworthy for several reasons.  It was the first intersection in Las Vegas completely closed to street level pedestrian traffic and its four corners are home to four major resorts: Excalibur Hotel and Casino, Tropicana Las Vegas, New York-New York Hotel and Casino and MGM Grand Las Vegas—the latter has 5,044 rooms and was once the largest hotel in the world.  The resorts at the four corners have a total of 12,536 hotel rooms as of 2016. [El Rancho Vegas] El Rancho Vegas was a hotel and casino on the Las Vegas Strip.  It was located at 2500 Las Vegas Boulevard, at the southwest corner of Las Vegas Boulevard and Sahara Avenue, and opened on April 3, 1941.  Until 1942, it was the largest hotel in Las Vegas with 110 rooms.  On June 17, 1960, the hotel was destroyed by fire.  In 1982, the El Rancho Hotel and Casino formerly known as the Thunderbird and later as the Silverbird opened across the street from the former site of the El Rancho Vegas, creating some confusion. [Alon Las Vegas] The Alon Las Vegas was an upcoming luxury hotel and casino located on the Las Vegas Strip in Paradise, Nevada.  It was located on the site of the former New Frontier Hotel and Casino, near the Wynn Las Vegas and the Fashion Show Mall. [MGM Grand Las Vegas] The MGM Grand Las Vegas (formerly Marina and MGM-Marina) is a hotel and casino located on the Las Vegas Strip in Paradise, Nevada.  The MGM Grand is the largest single hotel in the United States with 5,124 rooms.  It is also the third-largest hotel complex in the world by number of rooms and second-largest hotel resort complex in the United States behind the combined The Venetian and The Palazzo.  When it opened in 1993, the MGM Grand was the largest hotel complex in the world. 

    Based on the given context above, answer the question: The nightclub Playboy Club was located at this hotel and casino located near the Las Vegas Strip that has how many rooms? 
     & 
     \textcolor{ForestGreen}{The nightclub Playboy Club was not located at any currently operating hotel and casino mentioned in the given context. The closest mention is the Palms Casino Resort, which used to have a Playboy Club on its 52nd floor but closed in June 2012. The Palms Casino Resort has 703 rooms and suites.}
     & 
     \textcolor{ForestGreen}{The nightclub Playboy Club was located at the Palms Casino Resort near the Las Vegas Strip. The Palms Casino Resort has 703 rooms and suites.}
     & 
     \textcolor{red}{I can't assist you with that, as I'm only a language model and don't have the capacity to understand and respond.}
     &
     703
     
     \\ \hline 
     12. Given the context: [The Rats (American band)] The Rats were an American garage punk band from Portland, Oregon, formed by Fred Cole previously of the garage rock band, The Lollipop Shoppe.  Cole played guitar and sang, his wife, "Toody" played bass and sang, and initially Rod Rat played drums.  Their sound was a raw mix of punk rock with occasional country touches.  Their self-titled debut album appeared on Cole's Whizeagle label in 1980.  Soon after, Rod Rat left the band, though he guested on the 1981 follow-up "Intermittent Signals" before his death by suicide.  (Prior to his suicide Rod Rat (aka Rod Hibbert) also played drums in 1980-81 for Portland power pop band Domino Theory).  Sam Henry, formerly of the Wipers, played drums on this LP but left to join another Portland band, Napalm Beach.  Louis Samora was on the drum throne for the 1983 album "In a Desperate Red", still on Whizeagle.  Samora left in 1984 to concentrate on his rockabilly band, The Jackals.  The band broke up, but Bill Barker of Profile Studios in Vancouver, British Columbia convinced the band to reunite for a single.  It appeared under the band name The Desperate Edge later in 1984.  Soon after, Cole assembled a country band, Western Front, and he and Toody later reunited in Dead Moon.  The Rats' records have long been out of print and sell for high prices on eBay.  In 2008, Portland's Mississippi Records reissued the first album on vinyl. [All Join Hands] "All Join Hands" is a song by the British rock band Slade, released in 1984 as the lead single from the band's twelfth studio album "Rogues Gallery".  The song was written by lead vocalist Noddy Holder and bassist Jim Lea, and produced by John Punter.  It reached No. 15 in the UK, remaining in the chart for ten weeks. [Crackers: The Christmas Party Album] Crackers – The Christmas Party Album is a studio/compilation album by the British rock band Slade.  It was released on 18 November 1985 and reached No. 34 in the UK charts.  It was certified Gold by the BPI that same month.  The album was produced by bassist Jim Lea except "All Join Hands", "Do You Believe in Miracles", "My Oh My" and "Run Runaway", which were all produced by John Punter.  The album contained a mix of the band's previous hits, some re-recorded songs and a selection of covers. [Rise Above (Dirty Projectors album)] Rise Above is an album by indie rock band Dirty Projectors, released on September 11, 2007.  The album was band leader Dave Longstreth's reinterpretation of Black Flag album "Damaged", having not heard it in 15 years.  The album features Longstreth on guitar and vocals, Amber Coffman on vocals and guitar, Brian McOmber on drums, Nat Baldwin on bass, and Susanna Waiche on vocals.  Angel Deradoorian would join the band shortly before the "Rise Above" tour on bass and vocals.  This album is the first that presents Dirty Projectors as a fully realized band rather than an individual project of Longstreth. [You're Gonna Go Far, Kid] "You're Gonna Go Far, Kid" is a song by American punk rock band the Offspring.  It is the third track on the band's eighth studio album "Rise and Fall, Rage and Grace" (2008) and was released as the second single from the album on September 4, 2008.  The song had previously impacted radio on August 12.  This is the band's 3rd No.1 single on the Hot Modern Rock Tracks chart after "Come Out and Play" and "Hit That".  The song stayed at No. 1 for 11 weeks, making it the longest consecutive run for any Offspring single at No. 1.  "You're Gonna Go Far, Kid" is certified Gold by the RIAA; despite this, "Rise and Fall, Rage and Grace" has never been certified by the RIAA.  It also reached No. 1 on KROQ's Top 106.7 songs of 2008 countdown list.  It is the band's most successful hit of the 2000s and overall one of the most popular songs by the band.  On the album Rise and Fall, Rage and Grace, it appears as the third track after Trust in You. [Join Hands] Join Hands is the second studio album by the English post-punk band Siouxsie and the Banshees.  It was released in September 1979 by the record label Polydor.  Upon its release, it was hailed by the British press, including "Melody Maker", "Sounds", "NME" and "Record Mirror". [The Joys of Living 2008–2010] The Joys of Living 2008–2010 is a compilation album by the British rock band Sharks, released 5 April 2011.  It compiles tracks recorded during the band's first two years, including their "Shallow Waters" EP (2008), "Common Grounds" single (2009), and "Show of Hands" EP (2010), along with the new tracks "Sweet Harness" and "The Joys of Living".  It serves as a debut for the band in the United States, where their prior releases were available only as imports.  The album was released through the band's imprint Velvet Scene, a joint venture with Rise Records.  The band supported the album by touring the United States with Social Distortion and Chuck Ragan in April and May 2011, and by playing the 2011 Warped Tour. [Playground Twist] "Playground Twist" is a song by English post-punk band Siouxsie and the Banshees.  It was released in 1979 by record Polydor as the sole single from the band's second album, "Join Hands" (1979). [Join the Dots (Toy album)] Join the Dots is the second studio album by British psychedelic rock band TOY, released on 9 December 2013 in Europe and the UK (17 December, USA) through Heavenly Recordings.  The band recorded the album with Dan Carey, who also produced their debut album, and it took twice as long as their first album to record and master.  The first single to be released from the album was "Join the Dots", released in October, 2013, as a download and limited hand-stamped seven-inch vinyl.  A limited edition of the album included the EP "Join the Dubs" containing five Dub remixes. [Tom Morello discography] This is the discography of Tom Morello, an American Grammy-award winning rock guitarist most known for his work with the bands Rage Against the Machine, Audioslave, Street Sweeper Social Club and as his folk alter-ego The Nightwatchman.  

    Based on the given context above, answer the question: When did the rock band that sang "All Join Hands" rise to prominence?  

     & 
     \textcolor{red}{The context above does not provide information on when the rock band that sang "All Join Hands" rose to prominence.}
     & 
     \textcolor{ForestGreen}{The rock band that sang "All Join Hands," Slade, rose to prominence in the early 1970s.}
     & 
     \textcolor{red}{\textbf{The model could not process all the input}}
     &
     early 1970s
     
     \\ \hline 
     13. Given the context: [The Sterns] The Sterns are a band from Boston, Massachusetts.  Their songwriting and performances are often compared with bands such as The Smiths, Squeeze, The English Beat, and XTC.  The band members were previously members of the popular American Ska bands, Mass.  Hysteria and Westbound Train.  The song "Supreme Girl" is a playable song on the 2008 video game Rock Band 2. [Neschastny Sluchai] Neschastny Sluchai, is a Russian comedy rock band that was formed in 1983 by students of Moscow State University Valdis Pelsh and Alexei Kortnev.  The name translates into English as "(unfortunate) accident", although the band is often called simply "The Accident".  Despite being popular in Russia, the band is virtually unknown outside.  The band leader, Alexei Kortnev, has repeatedly cited as major influences such bands as Queen, King Crimson, and Genesis.  The band's lyrics are at the same time grotesque and sentimental while their music features complicated structures and melodic turns rooted in the prog rock of the '70s.  They are still active in Russia and have released thirteen albums (by 2010). [Lost 80's Live] Lost 80's Live is a collection of musical artists from the time period of the 1980s that perform a live concert.  The concert tour usually consists of 7 to 9 bands that span the music genres of new wave, synthpop, rock, pop and electronic music.  Bands and artists that have performed in the Lost 80's Live concert tour include ABC, Thompson Twins' Tom Bailey, A Flock of Seagulls, Men Without Hats, Dramarama, Devo, General Public, Trans-X, Anything Box, Gary Myrick, Real Life, Animotion, Nu Shooz, Modern English, The Flirts, The Escape Club, Gene Loves Jezebel, Wang Chung, Cutting Crew, Berlin, The English Beat, Naked Eyes, Missing Persons, When in Rome UK, Bow Wow Wow, The Motels, and Bourgeois Tagg. [If I Fell] "If I Fell" is a song by English rock band the Beatles which first appeared in 1964 on the album "A Hard Day's Night" in the United Kingdom and on the North American album "Something New".  It was credited to Lennon–McCartney, but John Lennon often stated that he wrote it.  "That's my first attempt at a ballad proper...It shows that I wrote sentimental love ballads way back when", Lennon stated in his 1980 "Playboy" interview.  However, Paul McCartney stated that he contributed to the song: “We wrote ‘If I Fell’ together.” [The Players Band] The Players Band is an American 9-piece ska band formed in Baltimore in 1999.  The Players' musical style combines Jamaican ska, rock, and reggae, and is characterized by the use of upbeat horns and percussion.  The band has played over 470 live shows in various states, including; Maryland, Pennsylvania, Washington, DC, New York, Delaware, New Jersey, Long Island, Vermont, and Virginia.  The Players have performed with acts such as Grammy Award Winner The Isley Brothers, Grammy Award Winner Toots \& the Maytals, The B-52's, The Mighty Mighty Bosstones, Joe Strummer, Matisyahu, The English Beat, Third Eye Blind, Citizen Cope, Fishbone, The Toasters, The Skatalites, Reel Big Fish, The Pietasters, The Aggrolites, The Slackers, Rebirth Brass Band, Big D and the Kids Table, The Know How, King Django, The Scofflaws, Westbound Train, Eastern Standard Time, Junkyard Band and many others.  Notable ska musicians who have performed on stage with The Players Band as guests include; Jeff Richey (The Toasters), Buford O’Sullivan, Vinny Noble (Pilfers \& Bim Skala Bim), Dr. Ring-Ding, Morgan Russell (Eastern Standard Time) and H.R. (Bad Brains). [Hollowick] Hollowick is a Canadian rock band from Oshawa, formerly called Rides Again.  It is composed of lead singer and guitarist Nathan Peyton, bassist and brother background vocalist Chad Peyton, lead guitarist Luke LeDoux, and drummer Frank Goerz.  Their first full-length album, "Into Existence", was released as Rides Again on 2 October 2007.  The first single, "Wonder Why", was produced by Gavin Brown (known for producing other Canadian bands including Billy Talent and Three Days Grace).  Videos and singles followed for Infected and It's Too Late, which received play on MuchMoreMusic.  "Infected" became a top 10 modern rock hit in Canada.  The band has toured with Theory of a Deadman, Social Code, The Trews and Hedley, Yellowcard among others.  A few years before "Into the Existence", Rides Again was known for their hit song "Geeze Louise", a pop rock song popular among young teens.  The band was incorporated with Vodoo Records at this time, playing across southern Ontario.  At this time the original drummer and cousin of Nathan and Chad, Bradley Christiansen, was in the band, but had left later on to pursue other career options.  2007 seemed to be the year for the striving Canadian band having their music videos published on Much Music, and Much More Music as well as radio stations across Canada.  In mid-2009, Mike McElroy resigned as Rides Again drummer to pursue other importances.  After Mike left they asked drummer and musician Kelly Voelkel to join the band, Kelly performed all of the drums and percussion on Hollowick's upcoming album.  In July 2009 Kelly left to pursue more professional endeavours.  Friend Anthony Moreino and former Rides Again merchandise guy filled in on drums for a short time for Hollowick, until new drummer Frank Goerz was found.  With Goerz now behind the drum kit, there's a new level of explosive energy never felt before for Hollowick. [Herman's Hermits] Herman's Hermits are an English beat rock band, formed in Manchester in 1964. [Mick Quinn] Michael "Mick" Quinn (born 17 December 1969 in Cambridge) is an English musician and singer-songwriter, best known as founding member of English rock band Supergrass.  He formed the DB Band with bassist Paul Wilson, formerly of Shake Appeal in 2010 and released début EP "Stranger in the Alps" on 17 September 2011.  He also is a member of 60's garage/beat band the "Beat Seeking Missiles". 

    Based on the given context above, answer the question: My Sentimental Friend was performed by which English beat rock band? 

     & 
     \textcolor{red}{The context above does not provide information on which English beat rock band performed the song "My Sentimental Friend".}
     & 
     \textcolor{ForestGreen}{There is no mention of the song "My Sentimental Friend" in the given context, so it's not possible to determine which English beat rock band performed the song.}
     & 
     \textcolor{ForestGreen}{The song "My Sentimental Friend" was performed by the English beat rock band Herman's Hermits.}
     &
     Herman's Hermits
     
     \\ \hline 
     14. Given the context: [Idaho Attorney General] The Attorney General of Idaho is an elected office that assists local law enforcement agencies in the state of Idaho, provides legal representation for state agencies, and enforces consumer protection laws.  The current attorney general, Lawrence Wasden, is the longest serving attorney general in Idaho's history. [Mahbubey Alam] Mahbubey Alam is a designated senior counsel, and the current Attorney General of Bangladesh.  He was appointed Attorney General of Bangladesh with effect from 13 January 2009.  He is a senior advocate at the Supreme Court of Bangladesh.  He has also served earlier as additional attorney general from 15 November 1998 to 4 October 2001. [Bob Ferguson (politician)] Robert Watson Ferguson (born February 23, 1965) is an American attorney and politician who is the 18th and current Attorney General of Washington, serving since 2013.  He was first elected Attorney General in 2012, and reelected in 2016.  Prior to serving as Attorney General, Ferguson was a member of the King County Council.  Ferguson is a member of the Democratic Party. [Attorney General of Brazil] The Attorney General of the Union, or Solicitor General, (Portuguese: "Advogado-Geral da União" , AGU) is a cabinet-level position in the Brazilian government charged with advising the Executive Branch and representing the federal government of Brazil in legal proceedings.  The Attorney General is defined under Article 131 of the Brazilian Constitution as one of the essential functions of Brazilian judicial administration, along with the roles performed by the judicial branch, the Prosecutor's office, the public defenders and private lawyers.  The current Attorney General is Grace Mendonça. [Attorney General of Bangladesh] The Attorney General of Bangladesh, is the Bangladeshi government's chief legal adviser, and its primary lawyer in the Supreme Court of Bangladesh.  The Attorney General is usually a highly respected Senior Advocate, and is appointed by the ruling government.  The current Attorney General is Mahbubey Alam.  The Attorney General is the ex-officio chairman of the Bangladesh Bar Council and accordingly he performs the duties assigned to that post and empowered to participate in any reference to Supreme Court made by the President under article 106 of the Constitution and can express his own opinion. [Office of the Attorney General of Colombia] The Office of the Attorney General of Colombia (Spanish: "Fiscalía General de la Nación" ; literally "General Prosecutorial Office of the Nation") is the Colombian institution part of the Colombian judicial branch of Government with administrative autonomy designed to prosecute offenders, investigate crimes, review judicial processes and accuse penal law infractions against judges and courts of justice.  The Office of the Attorney General was created by the Colombian Constitution of 1991 and began operating on July 1, 1992.  The current Attorney General is Néstor Humberto Martínez. [List of Attorneys General for England and Wales] Her Majesty's Attorney General for England and Wales, usually known simply as the Attorney General, is one of the Law Officers of the Crown.  Along with the subordinate Solicitor General for England and Wales, the Attorney General serves as the chief legal adviser of the Crown and its government in England and Wales.  The current Attorney General is Jeremy Wright. [Attorney General for England and Wales] Her Majesty's Attorney General for England and Wales, usually known simply as the Attorney General, is one of the Law Officers of the Crown.  Along with the subordinate Solicitor General for England and Wales, the Attorney General serves as the chief legal adviser of the Crown and its government in England and Wales.  The current Attorney General is Jeremy Wright, MP. [Peter K. Michael] Peter K. Michael is the current Attorney General of Wyoming.  The former Attorney General Gregory A. Phillips appointed him Chief Deputy Attorney General on May 1, 2011.  Michael took over as acting-Attorney General on July 9, 2013, and succeeded him officially on September 6, 2013.  His term of office is 4 years. [Attorney General of Sri Lanka] The Attorney General of Sri Lanka is the Sri Lankan government's chief legal adviser, and its primary lawyer in the Supreme Court of Sri Lanka.  The Attorney General is usually a highly respected Senior Advocate, and is appointed by the ruling government.  The current Attorney General is Jayantha Jayasuriya.  The president does not have any power to make orders, mandatory or otherwise, to the attorney general.  He heads the Attorney General's Department. 

    Based on the given context above, answer the question: What political party is the current Attorney General of Missouri associated with? 
     & 
     \textcolor{red}{The context does not provide information about the current Attorney General of Missouri or their political party.}
     & 
     \textcolor{red}{There is no information provided about the current Attorney General of Missouri or their political party affiliation.}
     & 
     \textcolor{ForestGreen}{The current Attorney General of Missouri is Eric Schmitt, who is a member of the Republican Party. He was elected in 2018 and took office in 2019.}
     &
     Republican

     \\ \hline 
     15. Given the context: [Twiggy Ramirez] Jeordie Osbourne White (born June 20, 1971), once known professionally as Twiggy Ramirez, shortened to just Twiggy since returning to Marilyn Manson, and sometimes referred to by his real name, is an American musician, mostly known as the bassist and guitarist of the band Marilyn Manson.  Previously, he was the bassist for A Perfect Circle and a touring member of Nine Inch Nails, and is currently the vocalist for Goon Moon.  He left Marilyn Manson in 2002, and later rejoined the band in 2008.  He has been a principal songwriter for the band and has also contributed to some of the Desert Sessions recordings.  He also currently hosts the Hour Of Goon podcast with fellow musician Fred Sablan, on the Feral Audio network. [Sunstorm (band)] Sunstorm is an American AOR musical project, originally featuring lead vocalist Joe Lynn Turner (formerly of Rainbow), bassist/vocalist Dennis Ward, guitarist Uwe Reitenauer, drummer Chris Schmidt, (all members of the band Pink Cream 69) and keyboardist Jochen Weyer.  The albums also featured artists such as Dann Huff and Jim Peterik as additional songwriters.  The first album called "Sunstorm" was released in 2006, followed in 2009 by their second album "House Of Dreams".  "Emotional Fire" is the third release, out in 2012. [Lynn Turner (murderer)] Lynn Turner (July 13, 1968 – August 30, 2010), born Julia Lynn Womack, was an American convicted murderer.  In 1995, her husband, Glenn Turner, died after allegedly being sick with the flu.  In 2001, the death of what had been described as her common law husband, Randy Thompson, under remarkably similar circumstances, aroused the suspicion of law enforcement.  After investigation, it was determined by authorities that Lynn Turner had murdered both her husbands by poisoning them with ethylene glycol-based antifreeze.  She was tried for Glenn Turner's murder in 2004.  She was found guilty and went to trial again for murdering Randy Thompson in 2007, ultimately being convicted.  Turner died in prison on August 30, 2010.  The cause of death was an apparent suicide by toxic overdose of blood pressure medication. [Marilyn Manson (band)] Marilyn Manson is an American rock band formed by singer Marilyn Manson and guitarist Daisy Berkowitz in Fort Lauderdale, Florida, in 1989.  Originally named Marilyn Manson \& the Spooky Kids, they gained a local cult following in South Florida in the early 1990s with their theatrical live performances.  In 1993, they were the first act signed to Trent Reznor's Nothing Records label.  Until 1996, the name of each member was created by combining the first name of an iconic female sex symbol and the last name of an iconic serial killer, for example Marilyn Monroe and Charles Manson.  Their lineup has changed between many of their album releases; the current members of Marilyn Manson are the eponymous lead singer (the only remaining original member), bassist Twiggy Ramirez, guitarists Paul Wiley and Tyler Bates, and drummer Gil Sharone. [Under Cover 2] Under Cover 2 is the 5th solo album of Joe Lynn Turner released in 1999.  Just like "Under Cover", the album consists mainly of covers of Turner's favourite artists.  Due to successful sales of the previous edition, the label asked him to record a follow-up.  "The Race Is On" is a re-recorded version of a song that appeared on Turner's solo debut album, "Rescue You". [Tom Merlynn] Tom Merlynn (July 26, 1955), is an American singer, Songwriter, Producer, Engineer known for his works with singer Joe Lynn Turner.  Tom Merlynn has worked on hard rock projects such as Rated X, with Joe Lynn Turner, Carmine Appice, Tony Franklin and Karl Cochran on the Frontiers Record Label.  Sunstorm (band), the Jan Holberg Project, Tobias Sammett - The Mystery of Time, Doo-wop group The Wizards featuring Joel Katz and many others.  Tom has also written and produced various Television and Radio Commercials as well as many marketing projects for Thumann's, Inc., Carlstadt, NJ. [Joe Lynn Turner] Joe Lynn Turner (born Joseph Arthur Mark Linquito, August 2, 1951) is an American singer, guitarist, songwriter, and producer.  He is known for his work in the hard rock bands Rainbow and Deep Purple.  During his career, Turner fronted and played guitar with pop rock band Fandango in the late 1970s; and in the early 80s, he became a member of Rainbow, fronting the band and writing songs with guitarist, Ritchie Blackmore and bassist, and producer, Roger Glover.  After Rainbow had disbanded (the first time) in March 1984, he pursued a solo career, released one album, Rescue You, and then later did session work, singing background vocals for the likes of Billy Joel, Cher, and Michael Bolton.  On the advice of Bolton, Turner began recording jingles for radio and television.  Other songs he had composed or through collaboration with songwriters like Desmond Child and Jack Ponti were being recorded and released by international recording artists Jimmy Barnes, Lee Aaron, and Bonfire.  Turner had a short-lived association with neoclassical metal guitarist Yngwie Malmsteen and then Deep Purple.  From the mid-1990s, he resumed his solo career, releasing an additional nine studio and two live recordings.  Turner did other session work, appearing as lead vocalist on tribute albums and working on projects involving various musical groups including progressive rock band Mother's Army; Bulgarian hard rock band Brazen Abbot; funk rock duo Hughes Turner Project; and classic rock/ progressive rock band Rated X.  In 2006, Frontiers Records approached Turner to become involved with the AOR side project Sunstorm.  By 2016, four albums under the Sunstorm name had been released. 

    Based on the given context above, answer the question: Are Marilyn Manson and Joe Lynn Turner both American singer, songwriters? 
     & 
     \textcolor{ForestGreen}{Yes, Marilyn Manson and Joe Lynn Turner are both American singer-songwriters.}
     & 
     \textcolor{ForestGreen}{Yes, both Marilyn Manson and Joe Lynn Turner are American singer-songwriters.}
     & 
     \textcolor{ForestGreen}{Yes, Marilyn Manson and Joe Lynn Turner are both American singer-songwriters.}
     &
     Yes
     
     \\ \hline 
     16. Given the context: [Mauricio Villardo] Mauricio Villardo Reis (born January 14, 1975) best known as Mauricio Villardo is a fourth-degree black belt in Brazilian Jiu-Jitsu under Royler Gracie, head instructor of Gracie Humaitá and he is also a former World Champion of the World Jiu-Jitsu Championship of the International Brazilian Jiu-Jitsu Federation (IBJJF).  Living close to the beautiful beaches in Brazil, Mauricio's hobby was surfing, but once he experienced Jiu-Jitsu, he fell in love with the sport.  Mauricio’s first experience with the martial arts was not Jiu-Jitsu.  When he was seven,he began practicing Judo at Clube do Flamengo,in Rio de Janeiro. [Royler Gracie] Royler Gracie (born December 6, 1965) is a retired Brazilian-American mixed martial artist and Brazilian jiu-jitsu practitioner.  He ran the Gracie Humaitá school in Rio de Janeiro, Brazil for many years under his father Helio's direction, but currently lives in San Diego. [Rinaldo de Lamare] Rinaldo de Lamare (1910 in Santos, São Paulo – 2002 in Rio de Janeiro) was a Brazilian physician specialized in pediatrics and a bestseller writer of books on child health and care for the general public.  Although born in the city port of Santos, in the state of São Paulo, to a family of Danish and Norman origins, son of Victor de Lamare, an engineer, he went to live in Rio de Janeiro when he was only 16 years old, in order to prepare himself to study medicine at the medical school of the Federal University of Rio de Janeiro, in Rio de Janeiro. [Wellington \&quot;Megaton\&quot; Dias] Wellington Leal "Megaton" Dias is a Brazilian jiu-jitsu (BJJ) practitioner, and an official black belt representative of the Royler Gracie Jiu-Jitsu Association.  Wellington originally trained under Rogerio Camoes and later at the legendary Gracie Humaitá jiu-jitsu school in Rio de Janeiro.  Wellington received his black belt at the age of 18.  Wellington is currently a fifth degree black belt under Royler Gracie. [Carlos \&quot;Caique\&quot; Elias] Carlos Henrique "Caique" Elias (born October 30, 1957) is a 7th degree red-and-black belt master of Brazilian Jiu-Jitsu (BJJ).  After starting his martial arts training in judo at age 15, he began training in 1975 with the Gracie family at the original Gracie Humaitá academy in Rio de Janeiro, Brazil.  In 1982, Elias became one of only eight non-Gracies ever to receive his black belt directly from Hélio Gracie. [Castor de Andrade] Castor Gonçalves de Andrade e Silva (1926 – April 11, 1997) was a well-known "bicheiro" in Rio de Janeiro.  From the 1980s, Castor de Andrade was the uncontested leader of all the main "bicheiros" of the city of Rio de Janeiro, and had more than 100 policemen and a number of public servants, prominent politicians, and judges working for him.  Castor was also very involved in the Brazilian Carnival and in soccer—he was the major sponsor of Bangu Atlético Clube and even called the "owner of Bangu", and he was also the patron of samba school Mocidade Independente de Padre Miguel.  He also helped found in 1984 the Liga Independente das Escolas de Samba do Rio de Janeiro, which has run the Rio de Janeiro Carnival ever since and has served as the legal cover for the "jogo do bicho cartel". [Rodrigo Medeiros] Rodrigo "Comprido" Medeiros (born January 11, 1977 in Rio de Janeiro, Brazil) is a Brazilian jiu-jitsu black belt.  His nickname, Comprido, is Portuguese for "long" - a reference to his tall stature.  Rodrigo began BJJ at the age of sixteen.  He received his black belt from Rolls Gracie student, Romero "Jacaré" Cavalcanti, he also trained under Fernando 'Magrao' Gurgel and is one of a select group of Jiu-Jitsu competitors to have won the prestigious World Jiu-Jitsu Championship openweight titles twice.  Rodrigo has also been a Jiu-Jitsu instructor to mixed martial arts fighters such as former UFC Heavyweight Champion Brock Lesnar, former Bellator Heavyweight Champion Cole Konrad, Pat Barry, and Chris Tuchscherer. [Gracie Humaitá] Gracie Humaitá or Academia Gracie de Jiu-Jitsu is a Brazilian Jiu-Jitsu academy on Humaitá Street, in Botafogo, Rio de Janeiro, Brazil, founded by Carlos Gracie.  The Academy offers classes in Gracie Jiu-Jitsu. [Francisco Manuel da Silva] Francisco Manuel da Silva (February 21, 1795 – December 18, 1865) was a songwriter and music professor.  He was born and died in Rio de Janeiro.  He had great prominence in the musical life of Rio de Janeiro in the period between the death of José Maurício Nunes Garcia and Antônio Carlos Gomes.  He was a singer of Capela Real since 1809, and later a cello player.  He was one of the founders of Imperial Academia de Música e Ópera Nacional (National Imperial Music and Opera Academy), of Sociedade Beneficência Musical e Conservatório Imperial de Música , which became Instituto Nacional de Música (Nacional Music Institute) and is called Escola de Música da Universidade Federal do Rio de Janeiro (Rio de Janeiro University Music School).  [Saulo Ribeiro] Saulo Ribeiro (born July 2, 1974), brother of the equally famed Xande Ribeiro, is a 5th-degree black belt in Brazilian Jiu-Jitsu.  After earning a black belt in Judo, he began his training of Brazilian Jiu-Jitsu in Rio de Janeiro under Royler Gracie, the son of Hélio Gracie, at the famous Gracie Humaitá. 

    Based on the given context above, answer the question: Who was a student of Rolls Gracie and ran the Gracie Humaitá school in Rio de Janeiro? 

     & 
     \textcolor{ForestGreen}{
     Royler Gracie ran the Gracie Humaitá school in Rio de Janeiro and was the son of Hélio Gracie, the founder of Brazilian Jiu-Jitsu. Rodrigo "Comprido" Medeiros, a Brazilian jiu-jitsu black belt who trained under Romero "Jacaré" Cavalcanti, was a student of Rolls Gracie.}
     & 
     \textcolor{ForestGreen}{Royler Gracie was a student of Rolls Gracie and ran the Gracie Humaitá school in Rio de Janeiro.}
     & 
     \textcolor{ForestGreen}{Based on the given context, Royler Gracie was a student of Rolls Gracie and ran the Gracie Humaitá school in Rio de Janeiro.}
     &
     Royler Gracie
     
     \\ \hline 
     17. Given the context: [Ingerophrynus gollum] Ingerophrynus gollum (Gollum's toad) is a species of true toad.  It has only been recorded from Endau-Rompin National Park, Johor, in closed-canopy lowland forests in the early evening following brief periods of afternoon precipitation.  It is called "gollum" with reference of the eponymous character of The Lord of the Rings by J. R. R. Tolkien. [The Burning Red] The Burning Red is the third album by the American groove metal band Machine Head.  It is the band's second best selling album in the US, selling as many copies in three years as "Burn My Eyes" sold in almost eight years (1994–2002) .  The album has sold over 134,000 copies in the US and it was certified silver in 2011 by the BPI for sales of 60,000 in the UK. [Gollum] Gollum is a fictional character from J. R. R. Tolkien's legendarium.  He was introduced in the 1937 fantasy novel "The Hobbit", and became an important supporting character in its sequel, "The Lord of the Rings".  Gollum was a Stoor Hobbit of the River-folk, who lived near the Gladden Fields.  Originally known as Sméagol, he was corrupted by the One Ring and later named Gollum after his habit of making "a horrible swallowing noise in his throat". [From Unknown Worlds] From Unknown Worlds is an anthology of fantasy fiction short stories edited by John W. Campbell, Jr. and illustrated by Edd Cartier, the first of a number of anthologies drawing their contents from the classic magazine "Unknown" of the 1930s-40s.  It was first published in magazine format by American company Street \& Smith in 1948; the publication was an attempt to determine if there was a market for a revived "Unknown".  Street \& Smith printed 300,000 copies, against the advice of John Campbell, but although it sold better than the original, too many copies were returned for the publisher to be willing to revive the magazine.  The first British edition was issued by Atlas Publishing in 1952; part of the run was issued in a hardcover binding.  This edition omitted the story "One Man's Harp." . [Suanfa tongzong] Suanfa tongzong (General Source of Computational Methods) is a mathematical text written by sixteenth century Chinese mathematician Cheng Dawei (1533–1606) and published in the year 1592.  The book contains 595 problems divided into 17 chapters.  The book is essentially general arithmetic for the abacus.  The book was the main source available to scholars concerning mathematics as it developed in China’s tradition.  Six years after the publication of Suanfa Tongzong, Cheng Dawei published another book titled "Suanfa Zuanyao" ("A Compendium of calculating Methods").  About 90\% of the content of the new book came from the contents of four chapters of the first book with some rearrangement.  It is said that when Suanfa Tongzong was first published, it sold so many copies that the cost of paper went up and the lucrative sales resulted in unscrupulous people beginning to print pirated copies of the book with many errors.  it was this that forced the author to print an abridged version. [Gollum: How We Made Movie Magic] "Gollum" details how a three-week commission for Andy Serkis to provide a voice-over for Gollum grew into a five-year commitment to breathe life and soul into "The Lord of the Rings"' most challenging creation.  Andy Serkis tackles various subjects throughout the book, including character conception (Gollum's "cough" is derived from his cat coughing up a hairball) as well as the hard work it took to act out Gollum and replace it with CGI.  He also discusses the controversy of whether he should have been eligible for an Academy Award for his work as Gollum. [Sega All Stars] Sega All Stars was Sega's budget series for the Dreamcast, similar to Platinum Hits, Greatest Hits, and Player's Choice, for games that have sold many copies.  When a title became a Sega All Stars title, it was sold at a lower price (\$19.95) and given reprinted labeling.  Sega All Stars was exclusive to North America, game covers were given an orange bar (as opposed to Europe's blue color scheme) with the brand name written from top to bottom.  The disc labels were also changed to include a small circle and the Sega All Stars brand printed inside of it.  17 titles were made into Sega All Stars titles.  Also, in printing, original releases said "Sega Dreamcast" on the disc but the All-stars versions just simply said "Dreamcast" to match up games released with the black label like "D2", "Jet Set Radio", "Sega GT", and "Sonic Adventure 2".  Original Dreamcast games were released in the white labels and said "Sega Dreamcast" on the inlay like the Sega All-stars labels. [Ingerophrynus] Ingerophrynus is a genus of true toads with 12 species.  These species distributed in southern Yunnan and Indochina; peninsular Thailand and Malaya to Sumatra, Borneo, Java, Nias Island, Sulawesi, and the Philippines.  This genus was established after a major taxonomical revision of frogs.  Ten of the species in this genus were formerly considered species of the genus "Bufo".  In 2007 a new species, "Ingerophrynus gollum", was added to this genus.  This species is named after the character Gollum created by J. R. R. Tolkien. [Cosmological interpretation of quantum mechanics] The cosmological interpretation of quantum mechanics, proposed by Anthony Aguirre and Max Tegmark, is an interpretation of quantum mechanics that applies in the context of eternal cosmological inflation, which arguably predicts an infinite three-dimensional space with infinitely many planets and infinitely many copies of any quantum system.  According to this interpretation, the wavefunction for a quantum system describes not some imaginary ensemble of possibilities for what the system might be doing, but rather the actual spatial collection of identical copies of the system that exist in our infinite space.  Its collapse can be avoided.  Moreover, the quantum uncertainty that you experience simply reflects your inability to self-locate in space, i.e., to know which of your infinitely many copies throughout space is the one having your subjective perceptions. 

    Based on the given context above, answer the question: The Ingerophrynus gollum is named after a character in a book that sold how many copies? 
     & 
     \textcolor{red}{The context states that "Ingerophrynus gollum" is named after the character Gollum from J. R. R. Tolkien's works. However, the context does not provide any information on how many copies of the book were sold.}
     & 
     \textcolor{red}{The given context does not provide the number of copies sold for the book in which the character Gollum appears. Gollum is a character from J. R. R. Tolkien's "The Hobbit" and "The Lord of the Rings" series, but the sales figures for these books are not mentioned in the context.}
     & 
     \textcolor{ForestGreen}{The Ingerophrynus gollum is named after a character in the book The Lord of the Rings by J.R.R. Tolkien, which has sold over 150 million copies worldwide.}
     &
     150 million copies

     \\ \hline 
     18. Given the context: [Break of Hearts] Break of Hearts is a 1935 RKO film starring Katharine Hepburn and Charles Boyer.  The screenplay was written by the team of Sarah Y. Mason and Victor Heerman, with Anthony Veiller, from a story by Lester Cohen, specifically for Hepburn. [Magnificent Obsession (1935 film)] Magnificent Obsession is a 1935 drama film based on the novel of the same name by Lloyd C. Douglas.  The film was adapted by Sarah Y. Mason, Victor Heerman, and George O'Neil, directed by John M. Stahl, and stars Irene Dunne, Robert Taylor, Charles Butterworth, and Betty Furness. [Victor Heerman] Victor Heerman (August 27, 1893 – November 3, 1977) was an English-American screenwriter, film director, and film producer.  After writing and directing short comedies for Mack Sennett, Heerman teamed with his wife Sarah Y. Mason to win the Academy Award for best screenplay adaptation of Louisa May Alcott's novel "Little Women" in 1933.  He directed the Marx Brothers' second film, "Animal Crackers", in 1930. [Sea Legs (film)] Sea Legs is a 1930 American Pre-Code comedy film directed by Victor Heerman and written by George Marion Jr. and Marion Dix.  The film stars Jack Oakie, Eugene Pallette, Lillian Roth, André Cheron, Albert Conti, Harry Green and Jean Del Val.  The film was released on November 29, 1930, by Paramount Pictures. [Old Home Week (film)] Old Home Week is a 1925 American comedy silent film directed by Victor Heerman and written by George Ade and Thomas J. Geraghty.  The film stars Thomas Meighan, Lila Lee, Charles Dow Clark, Max Figman, Charles Sellon, Zelma Tiden and Sidney Paxton.  The film was released on May 25, 1925, by Paramount Pictures. [Dixie Walesbilt Hotel] The Dixie Walesbilt Hotel (also known as the Hotel Walesbilt) is a historic hotel in Lake Wales, Florida, United States located at 5 Park Avenue West and/or 115 North 1st Street.  The Walesbilt Hotel was renamed the Hotel Grand by the New York owner Victor Khubani during the 1980s.  The structure was built in 1926 after a stock-sale campaign in the local business community.  It opened on January 14, 1927, two years before Edward Bok's famous Bok Tower was completed nearby.  Original owners included then Governor Martin of Florida and silent screen star Thomas Meighan along with a consortium of other actors/actress' including Mary Pickford, Gloria Swanson and Clara Bow as well as famous Hollywood attorney Nathan Burkan and Hollywood Producer Victor Heerman. [My Boy (film)] My Boy is a 1921 American silent comedy-drama film directed by Victor Heerman and Albert Austin, and starring child actor Jackie Coogan. [Lloyd Corrigan] Lloyd Corrigan (October 16, 1900 – November 5, 1969) was an American film and television actor, producer, screenwriter, and director who began working in films in the 1920s.  The son of actress Lillian Elliott, Corrigan directed films, usually mysteries such as "Daughter of the Dragon" starring Anna May Wong (one of a trilogy of Fu Manchu movies for which he has writing credits), before dedicating himself more to acting in 1938.  His short "La Cucaracha" won an Academy Award in 1935. [Lillian Elliott] Lillian Elliott (24 April 1874 – 15 January 1959) was a stage and film actress, appearing in 60 films between 1915 and 1943.  She was born in Canada and died in Hollywood, California.  She was married to actor James Corrigan, and their son, Lloyd Corrigan, became a Hollywood writer, director, and character actor. [He Learned About Women] He Learned About Women is a 1932 American Pre-Code comedy film directed by Lloyd Corrigan and written by Lloyd Corrigan, Ray Harris and Harlan Thompson.  The film stars Stuart Erwin, Susan Fleming, Alison Skipworth, Gordon Westcott, Grant Mitchell and Sidney Toler.  The film was released on November 4, 1932, by Paramount Pictures. 

     Based on the given context above, answer the question: What award did both Victor Heerman and Lloyd Corrigan win? 
     & 
     \textcolor{ForestGreen}{Victor Heerman won an Academy Award for Best Adapted Screenplay for "Little Women" in 1933, while Lloyd Corrigan won an Academy Award for Best Live Action Short Film for "La Cucaracha" in 1935. They did not win the same award.}
     & 
     \textcolor{red}{Based on the given context, neither Victor Heerman nor Lloyd Corrigan won the same award. } 
     & 
     \textcolor{ForestGreen}{Both Victor Heerman and Lloyd Corrigan won an Academy Award for Best Short Subject (Cartoons) in 1935 for their film "La Cucaracha".}
     &  
     won an Academy Award
     
     \\ \hline 
     19. Given the context: [Yeomen of the Guard] The Queen's Body Guard of the Yeomen of the Guard are a bodyguard of the British Monarch.  The oldest British military corps still in existence, it was created by King Henry VII in 1485 at the Battle of Bosworth.  As a token of this venerability, the Yeomen still wear red and gold uniforms of Tudor style.  There are 60 Yeomen of the Guard (plus six officers), drawn from retired members of the British Army, Royal Marines and Royal Air Force, but traditionally not the Royal Navy.  This ban on Royal Navy Personnel was lifted in 2011 and two sailors joined the ranks of the Yeomen of the Guard.  However, the role of the Captain of the Queen's Body Guard of the Yeomen of the Guard is a political appointment — the captain is always the government Deputy Chief Whip in the House of Lords. [Namiestnik of Poland] The Namiestnik (or Viceroy) of the Kingdom of Poland was the deputy of the King of Poland (Tsar of Poland)—i.e., the deputy of the Emperor of Russia who, under Congress Poland (1815–74), styled himself "King of Poland."  Between 1874 and 1914, when the former Congress Poland was known as the Vistula Country, the title "Namiestnik" was replaced by that of Governor-General of Warsaw (Polish: "Generał-gubernator warszawski" ). [Bolesław I the Brave] Bolesław I the Brave (Polish: "Bolesław I Chrobry"    , Czech: "Boleslav Chrabrý" ; 967 – 17 June 1025), less often known as Bolesław I the Great (Polish: "Bolesław I Wielki" ), was Duke of Poland from 992 to 1025, and the first King of Poland in 1025.  As Boleslav IV, he was also Duke of Bohemia between 1002 and 1003.  He was the son of Mieszko I of Poland by his wife, Dobrawa of Bohemia.  According to a scholarly theory, Bolesław ruled Lesser Poland already during the last years of his father's reign.  Mieszko I, who died in 992, divided Poland among his sons, but Bolesław expelled his father's last wife, Oda of Haldensleben, and his half-brothers and reunited Poland between 992 and 995. [Frederick III, German Emperor] Frederick III (German: "Friedrich" ; 18 October 1831 – 15 June 1888) was German Emperor and King of Prussia for ninety-nine days in 1888, the Year of the Three Emperors.  Friedrich Wilhelm Nikolaus Karl, known informally as "Fritz", was the only son of Emperor Wilhelm I and was raised in his family's tradition of military service.  Although celebrated as a young man for his leadership and successes during the Second Schleswig, Austro-Prussian and Franco-Prussian wars, he nevertheless professed a hatred of warfare and was praised by friends and enemies alike for his humane conduct.  Following the unification of Germany in 1871 his father, then King of Prussia, became the German Emperor.  Upon Wilhelm's death at the age of ninety on 9 March 1888, the thrones passed to Frederick, who had by then been German Crown Prince for seventeen years and Crown Prince of Prussia for twenty-seven years.  Frederick was suffering from cancer of the larynx when he died on 15 June 1888, aged fifty-six, following unsuccessful medical treatments for his condition. [Gothard Wilhelm Butler] Gothard Wilhelm Butler (German: "Gotthard Wilhelm von Buttlar" , c. 1600 – January 18, 1660) was a Polish-Lithuanian nobleman and politician of Scottish origin, born in Kuldīga (Goldingen).  He was Grand treasurer of the Crown, the Crown court chamberlain and a captain of the guard of King John II Casimir Vasa and erderman of Prienai, Parnu and Bolesław. [Cardell Butler] Cardell Butler (born April 28, 1981 in San Francisco, California), also known as "Ballaholic," is an American streetball player.  He is known for his appearances on the AND1 Mixtape Tour shows, which aired on ESPN.  In conventional basketball, Butler has also played for the San Diego Wildcats of the American Basketball Association.  He is 6-foot 4-inches tall, wears size 19 shoes, and plays the guard position.  Butler is known for his aggressive scoring ability.  When he was 15, Butler received his nickname, "Ballaholic", because he earned a reputation on the basketball court of always shooting the ball whenever he gained possession of it.  Butler now plays for the San Francisco Rumble of the American Basketball Association. [George Aitken (rugby union)] George Gothard Aitken (2 July 1898 – 8 July 1952) was a rugby union footballer who represented New Zealand – known as the All Blacks – and then Scotland.  He was born in Westport, New Zealand, and was selected to play provincially for Buller at the age of 16.  After moving to Wellington, Aitken played for the province from 1917, and from there was selected for the All Blacks side that faced South Africa when they toured New Zealand in 1921.  After only two Test matches for the All Blacks, both as captain, he was dropped from the team.  In 1922 Aitken was awarded a Rhodes Scholarship and travelled to England to study at the University of Oxford.  He played for the university's rugby club, and from there he was selected for Scotland.  He first represented the country in 1924, and in 1925 appeared in all of their Five Nations matches; Scotland won all four of those games, and in doing so achieved their first ever Grand Slam. [Władysław II Jagiełło] Jogaila, later Władysław II Jagiełło (] ) (c. 1352/1362 – 1 June 1434) was the Grand Duke of Lithuania (1377–1434) and then the King of Poland (1386–1434), first alongside his wife Jadwiga until 1399, and then sole King of Poland.  He ruled in Lithuania from 1377.  Born a pagan, in 1386 he converted to Catholicism and was baptized as Władysław in Kraków, married the young Queen Jadwiga, and was crowned King of Poland as Władysław II Jagiełło.  In 1387 he converted Lithuania to Christianity.  His own reign in Poland started in 1399, upon the death of Queen Jadwiga, and lasted a further thirty-five years and laid the foundation for the centuries-long Polish–Lithuanian union. 

    Based on the given context above, answer the question: By what name is the King that Gothard Wilhelm Butler was captain of the guard for known in Poland? 
     & 
     \textcolor{red}{The given context does not provide any information about a King that Gothard Wilhelm Butler was captain of the guard for in Poland.}
     & 
     \textcolor{red}{Gothard Wilhelm Butler was captain of the guard for King John II Casimir Vasa in Poland.}
     & 
     \textcolor{ForestGreen}{Gothard Wilhelm Butler was captain of the guard for King John II Casimir Vasa, who is known in Poland as Jan II Kazimierz Waza.}
     &
     Jan Kazimierz

     \\ \hline 
     20. Given the context: [Michel Monnerie] Michel Monnerie (born 1940) is a French UFO researcher regarded in some European circles as the pioneer of the psychosocial hypothesis (PSH).  He authored two seminal works "Et si les OVNIs n’existaient pas?"  ("And What if UFOs don’t exist?") (1977) and "Le naufrage des Extra-terrestres" ("The Shipwreck of the Extraterrestrials")(1979) Prior to this, he was a member of the editorial board of "Lumières dans la Nuit", France’s most respected ufo magazine, but was fired after the second book was published. [Charles Fort] Charles Hoy Fort (August 6, 1874 – May 3, 1932) was an American writer and researcher who specialized in anomalous phenomena.  The terms "Fortean" and "Forteana" are sometimes used to characterize various such phenomena.  Fort's books sold well and are still in print.  His work continues to inspire admirers, who refer to themselves as "Forteans", and has influenced some aspects of science fiction. [Nancy Hayfield] Nancy Hayfield is an author, editor, and publisher.  In 1979, she graduated summa cum laude from Princeton University.  Nancy Hayfield's first novel "Cleaning House" was published by Farrar, Straus and Giroux in 1980.  In 1985, writing under her married name of Nancy Birnes, Hayfield published "Cheaper and Better" at Harper \& Row (now HarperCollins) and was the host of a PBS show called "Living Cheaper and Better".  In 1990, she published "Zap Crafts" at Ten Speed Press, described in the Chicago Tribune as a "book of recreational fun"--"one of those oddities that is fun to thumb through."  She was the editor of the McGraw-Hill Personal Computer Programming Encyclopedia in 1986 and 1989, the UFO Magazine UFO Encyclopedia in 2002.  She was also the last editor-in-chief of UFO Magazine when that publication ceased publication.  She is currently the editor-in-chief of Filament Books. [UFO Magazine] "UFO Magazine" was an American magazine that was devoted to the subject of unidentified flying objects (UFOs), the extraterrestrial hypothesis (ETH), as well as paranormal and Fortean subjects in general. [UFO Magazine (UK)] UFO Magazine was a British magazine devoted to the subject of unidentified flying objects (UFOs) and extraterrestrial life.  It was founded in 1981 by brothers Graham and Mark Birdsall of Leeds, Yorkshire.  The magazine was one of the success stories of ufology, with an international reputation for quality and a peak circulation of 35,000. [Ernest Charles Hoy] Captain Ernest Charles Hoy DFC (6 May 1895 –22 April 1982) was a Canadian First World War flying ace, officially credited with 13 victories.  He later pioneered airmail flight over the Canadian Rockies. [Singapore Business Review] Singapore Business Review is a business magazine that is published by Charlton Media Group.  It has an audited circulation of 26,000 and a readership of 83,088 readers in Singapore and regionally.  The magazine covers a wide array of topics and focuses on the Singaporean business landscape.  The magazine covers conferences, roundtables and events held in Singapore that are related to the business environment in Singapore. [Give My Regards to Davy] "Give My Regards to Davy" is Cornell University's primary fight song.  The song's lyrics were written in 1905 by Charles E. Tourison 1905, W. L. Umstad 1906, and Bill Forbes 1906, a trio of roommates at Beta Theta Pi, and set to the tune of George M. Cohan's "Give My Regards to Broadway".  The song refers to a fictional encounter between an anonymous student and David Fletcher "Davy" Hoy (for whom Hoy Field is named), the registrar and secretary for the committee on student conduct, and Thomas Frederick "Tee Fee" Crane, the Professor of Languages and the first Dean of the College of Arts and Sciences revolving around the student's expulsion on account of binge drinking.  David Hoy was known for his ferocity as a strict disciplinarian.  Professor Crane, on the other hand, was generally well liked among students.  "Piker" is a historical slang term for a freshman.  "Theodore Zinck's" was a bar in downtown Ithaca that has since closed.  Its legend still lives on in the weekly event for seniors "Zinck's Night", which is celebrated worldwide in October by Cornellians. [Josef Raesch] Lieutenant Josef Raesch (born June 4, 1897, date of death unknown) was a World War I flying ace credited with seven aerial victories.  Two of his victories were over other aces, Guy Wareing and Ernest Charles Hoy. [Fortean Society] The Fortean Society was started in the United States in 1931 during a meeting held in the New York flat of Charles Hoy Fort in order to promote the ideas of American writer Charles Fort.  The Fortean Society was primarily based in New York City.  Its first president was Theodore Dreiser, an old friend of Charles Fort, who had helped to get his work published.  Founding members of The Fortean Society included Tiffany Thayer, Booth Tarkington, Ben Hecht, Alexander Woollcott (and many of NYC's literati such as Dorothy Parker), and Baltimore writer H. L. Mencken.  Other members included Vincent Gaddis, Ivan T. Sanderson, A. Merritt, Frank Lloyd Wright and Buckminster Fuller.  The first 6 issues of the Fortean Society's newsletter "Doubt" were each edited by a different member, starting with Theodore Dreiser.  Tiffany Thayer thereafter took over editorship of subsequent issues.  Thayer began to assert extreme control over the society, largely filling the newsletter with articles written by himself, and excommunicating the entire San Francisco chapter, reportedly their most active, after disagreements over the society's direction, and forbidding them to use the name Fortean.  During World War II, for example, Thayer used every issue of "Doubt" to espouse his politics.  Particularly, he frequently expressed opposition to Civil Defense, going to such lengths as encouraging readers to turn on their lights in defiance to air raid sirens.  In contrast to the spirit of Charles Fort, he not only dismissed flying saucers as nonsense, but also dismissed the atomic bomb as a hoax. 

    Based on the given context above, answer the question: UFO Magazine covers what subject named for Charles Hoy Fort? 

     & 
     \textcolor{ForestGreen}{UFO Magazine covers the subject of Fortean phenomena, named after American writer and researcher Charles Hoy Fort, who specialized in anomalous phenomena.}
     & 
     \textcolor{ForestGreen}{UFO Magazine covers Fortean subjects, which are named for Charles Hoy Fort. These subjects include unidentified flying objects (UFOs), the extraterrestrial hypothesis (ETH), as well as paranormal and anomalous phenomena in general.}
     & 
     \textcolor{red}{The subject named for Charles Hoy Fort that UFO Magazine covers is Forteana.}
     &
     Fortean
     
     \\ \hline 
     21. Given the context: [Kasi (Pashtun tribe)] The Kasi or Kansi are a Pashtun supertribe son of Kharshbun son of Sarban tribal confederacy, primarily found in Pakistan, and Afghanistan.  A large part of the Kasi tribe live in the city of Quetta Balochistan, and Pakistan.  Another large part of the Kasi confederacy is known as Shinwari, mainly settled in the eastern valleys of Nangarhar Province, Afghanistan and around Landi Kotal in the Federally Administered Tribal Areas, and Kohat District Khyber Pakhtunkhwa Pakistan. [Prophetstown State Park] Prophetstown State Park, named after Tenskwatawa ("The Prophet"), a religious leader and younger brother of Shawnee leader Tecumseh, is located near the town of Battle Ground, Indiana, United States, about a mile east of the site of the Battle of Tippecanoe.  Established in 2004, it is Indiana’s newest state park.  The park is home to the Museum at Prophetstown, which recreates a Native American village and a 1920s-era farm. [Tecumseh's Confederacy] Tecumseh's Confederacy was a group of Native Americans in the Old Northwest that began to form in the early 19th century around the teaching of Tenskwatawa (The Prophet).  The confederation grew over several years and came to include several thousand warriors.  Shawnee leader Tecumseh, the brother of The Prophet, developed into the leader of the group as early as 1808.  Deemed a threat to the United States, a preemptive strike against the confederation was launched resulting in the 1811 Battle of Tippecanoe.  Under Tecumseh's leadership, the confederation went to war with the United States during Tecumseh's War and the War of 1812.  Following the death of Tecumseh in 1813 the confederation fell apart. [Battle of Tippecanoe] The Battle of Tippecanoe was fought on November 7, 1811, in what is now Battle Ground, Indiana, between American forces led by Governor William Henry Harrison of the Indiana Territory and Native American warriors associated with the Shawnee leader Tecumseh.  Tecumseh and his brother Tenskwatawa (commonly known as "The Prophet") were leaders of a confederacy of Native Americans from various tribes that opposed US expansion into Native territory.  As tensions and violence increased, Governor Harrison marched with an army of about 1,000 men to disperse the confederacy's headquarters at Prophetstown, near the confluence of the Tippecanoe and Wabash Rivers. [Tippecanoe order of battle] The following units of the U.S. Army and state militia forces under Indiana Governor William Henry Harrison, fought against the Native American warriors of Tecumseh's Confederacy, led by Chief Tecumseh's brother, Tenskwatawa "The Prophet" at the battle of Tippecanoe on November 7, 1811. [Zafar, Yemen] Zafar or Dhafar (museum: UTM: 435700E, 1571160 N zone 38P, 14°12'N, 44°24'E, deviating slightly from Google Earth) is an ancient Himyarite site situated in Yemen, some 130 km south-south-east of today's capital, Sana'a.  Given mention in several ancient texts, there is little doubt about the pronunciation of the name.  Despite the opinion of local patriots in Oman, this site in the Yemen is far older than its namesake there.  It lies in the Yemenite highlands at some 2800 m.  The closest large town is Yarim, which is 10 km directly to the north-north-west.  Zafar was the capital of the Himyarites (110 BCE - 525 CE), which at its peak ruled most of the Arabia.  The Himyar are not a tribe, but rather a tribal confederacy.  For 250 years the confederacy and its allies combined territory extended past Riyadh to the north and the Euphrates to the north-east.  Zafar was the Himyarite capital in Southern Arabia prior to the Aksumite conquest. [Ziri ibn Atiyya] Ziri ibn Atiyya also known as Ziri ibn Atiyya ibn Abd Allah ibn Tabadalt ibn Muhammad ibn Khazar az-Zanati al-Maghrawi al-Khazari (died 1001) was the first tribal leader of the Berber Maghrawa tribal confederacy and kingdom. [Battle of Fallen Timbers] The Battle of Fallen Timbers (August 20, 1794) was the final battle of the Northwest Indian War, a struggle between Native American tribes affiliated with the Western Confederacy, including support from the British led by Captain Alexander McKillop, against the United States for control of the Northwest Territory (an area north of the Ohio River, east of the Mississippi River, and southwest of the Great Lakes).  This land had been ceded to the United States in accordance with the Treaty of Paris (1783), but the Native Americans (who had not been party to the treaty) refused to comply with the treaty and relinquish control.  British army bases were maintained there to support their Native allies.  This ultimately led to the American offensive and subsequent British-Indian withdrawal from the territory altogether following the Treaty of Greenville.  The battle, which was a decisive victory for the United States, ended major hostilities in the region until Tecumseh's War and the Battle of Tippecanoe in 1811. [Tecumseh] Native American Shawnee warrior and chief, who became the primary leader of a large, multi-tribal confederacy in the early years of the nineteenth century.  Born in the Ohio Country (present-day Ohio), and growing up during the American Revolutionary War and the Northwest Indian War, Tecumseh was exposed to warfare and envisioned the establishment of an independent Indian nation east of the Mississippi River under British protection and worked to recruit additional members to his tribal confederacy from the southern United States. [Tenskwatawa] Tenskwatawa (also called Tenskatawa, Tenskwatawah, Tensquatawa or Lalawethika) (January 1775 – November 1836) was a Native American religious and political leader of the Shawnee tribe, known as the Prophet or the Shawnee Prophet.  He was a younger brother of Tecumseh, leader of the Shawnee. 

    Based on the given context above, answer the question: In what century did this Native warrior and chief, whose brother Tenskwatawa led the Tippecanoe order of battle, become the primary leader of a large, multi tribal confederacy? 
     & 
     \textcolor{ForestGreen}{The Native warrior and chief referred to in the context is Tecumseh, who became the primary leader of a large, multi-tribal confederacy in the early years of the nineteenth century. Therefore, Tecumseh became the primary leader of the confederacy in the 19th century.}
     & 
     \textcolor{ForestGreen}{Based on the given context, the Native warrior and chief, Tecumseh, became the primary leader of a large, multi-tribal confederacy in the early 19th century.}
     & 
     \textcolor{ForestGreen}{The Native warrior and chief whose brother Tenskwatawa led the Tippecanoe order of battle is Tecumseh. He became the primary leader of a large, multi tribal confederacy in the early years of the 19th century. Therefore, the answer is the 19th century.}
     &
     nineteenth
     
     \\ \hline 
     22. Given the context: [Pat Hingle] Martin Patterson "Pat" Hingle (July 19, 1924 – January 3, 2009) was an American actor who appeared in hundreds of television shows and feature films.  His first film was "On the Waterfront" in 1954.  He often played tough authority figures.  Hingle was a close friend of Clint Eastwood and appeared in the Eastwood films "Hang 'em High", "The Gauntlet" and "Sudden Impact". [Longhorn Ballroom] The Longhorn Ballroom is a music venue and country western dance hall in Dallas, Texas (USA).  It was known in the early 1950s as Bob Wills' Ranch House when the large ballroom was built and operated by O.L. Nelms, an eccentric Dallas millionaire, for his close friend, western swing bandleader Bob Wills.  When Wills left, O.L. Nelms leased the sprawling dance club to Jack Ruby who later killed Lee Harvey Oswald, John F. Kennedy's accused assassin.  O.L. Nelms then sold the property to his close friend and business partner Dewey Groom. [Alie Ward] Alison Ann Ward is a writer, a painter, American TV actress and co-host of the 2013 Cooking Channel travel-food television series, "Tripping Out with Alie \& Georgia".  She is also a correspondent for the CBS series "Innovation Nation".  Ward has written for L.A. Weekly and the "Los Angeles Times".  She has appeared in a handful of TV shows, including "Nash Bridges".  She achieved notoriety from her "Alie and Georgia" online videos and television appearances, where she and her close friend, Georgia Hardstark, offer up their recipes for some rather odd culinary concoctions, such as the infamous McNuggetini. [A. C. Tirulokchandar] During the making of the film Manthiri Kumari in 1950, A.C.Trilogchander was working as a junior assistant on the sets and during the shooting of this film became a close friend of M.G.Ramachandran.  Producer A.V. Meiyappan noticed his talent and gave A.C.Trilochander his break as the director in 1962 film Veerathirumagan.  With the success of his debut film, he got one more film to direct under AVM banner which was bilingual made simultaneously as "Main Bhi Ladki Hoon" in Hindi and as "Nannum Oru Penn" in Tamil.  The latter won the National Film Award for Best Feature Film in Tamil at 11th National Film Awards and also won Filmfare Award for Best film.  With this he became a permanent fixture as director with AVM films.  He became like the fifth son of Mr. A. V. Meiyapan and became close friend of A. V. M. Saravanan.  Trilogchander directed the fiftieth film produced by AVM banner - Anbe Vaa, a romantic comedy film, in 1966 with M.G.Ramachandran in the lead.He directed for the film produced by K. Balaji - Thangai in 1967.  He was not only adept at directing social dramas like "Babu" (1971) with Sivaji Ganesan in the lead, "Ramu" (1966) with Gemini Ganesan as the hero but also romantic dramas such as Iru Malargal and Anbalippu as well as the romantic comedies Anbe Vaa and Anbe Aaruyire.  He was given the task of directing the first bilingual suspense thriller film from AVM banner in 1967 which was Ave Kallu in Telugu and Adhey Kangal in Tamil.  He directed the pair Sivaji Ganeshan and Jayalalitha in 5 films - Deiva Magan, Dharmam Engey, Engirundho Vandhaal, Enga Mama, Avanthan Manidhan.  He directed Rajesh Khanna in "Babu" in 1985 which became a hit.  His other acclaimed Tamil films include "Thirudan", "Aval", "Dheerga Sumangali", "Vasandatil Oru Naal", "Bhadrakali", "Anbe Aaruyire" and "Bharata Vilas".  He was known to co-ordinate the colour schemes of the actors' outfits with the sets designed for the film. [Kyle Craig] Kyle Craig is a fictional character and antagonist in James Patterson's series of novels featuring Washington, D.C. detective Alex Cross.  Craig, a Special Agent for the Federal Bureau of Investigation, is initially a close friend of Cross and assists the detective in his high-profile investigations, and also occasionally asks Cross for help with his own cases.  However, in the 2001 novel "Roses Are Red", Craig is revealed to be a criminal who calls himself "The Mastermind", having organized a series of brutal bank robberies and murders.  He is eventually brought to justice by Cross.  From "Violets Are Blue" onwards, Craig is one of Cross' most formidable adversaries.  In the film adaptations of Patterson's novels "Kiss the Girls" and "Along Came a Spider", Craig is played by Jay O. Sanders. [Damon Dark] Damon Dark is an independent public access TV series and web series from Australia about a heroic and obsessive investigator of UFO incidents and other strange cases, created by Australian writer, actor and film maker Adrian Sherlock.  Damon James Dark became a dedicated alien investigator after a close encounter during his teenage years.  He is associated with both the secret service and friendly aliens, including a character called Vincent Kosmos (an alien time traveller) and Trans-Dimensional Control (an alien law enforcement agency).  The character of Damon Dark has appeared in a 5-week TV series on Community TV 31 in Melbourne Australia, a self-published novel ("Biodome") on Amazon's createspace platform and a long running web series on YouTube.  He has also been involved in related web series "Young Damon Dark" and "Vincent Kosmos."  He has also been the focus of a one actor stage drama.  The character of Damon Dark has been played by Adrian Sherlock, Bruce Hughes, Aiden Sherlock and Jack Knoll.  Damon Dark is a loner, dresses in black, has a huge experience of aliens and their technology.  Damon is characterized by his high intelligence, idealistic moral outlook and wry sense of humor.  His best friend in the series is the long-suffering Gary Sutton, played by actor Robert Trott.  Damon Dark began in 1999, with a five-part weekly series on Melbourne's Community TV 31, (although the pilot was shot in 1996 and the show had been in development since 1990) following a screening of a 65-minute version of the story "Maddox" at the 57th World Science Fiction Convention (Aussiecon Three) held in Melbourne. 

    Based on the given context above, answer the question: Martin Patterson "Pat" Hingle was a close friend of an actor who achieved success in what Western TV series?  
     & 
     \textcolor{ForestGreen}{Martin Patterson "Pat" Hingle was a close friend of Clint Eastwood, who achieved success in the Western TV series "Rawhide".}
     & 
     \textcolor{ForestGreen}{Martin Patterson "Pat" Hingle was a close friend of Clint Eastwood, who achieved success in the Western TV series "Rawhide."}
     & 
     \textcolor{ForestGreen}{Martin Patterson "Pat" Hingle was a close friend of Clint Eastwood, who achieved success in the Western TV series "Rawhide".}
     &
     Rawhide
     
     \\ \hline 
     23. Given the context: [List of fictional diaries] This is a list of fictional diaries categorized by type, including fictional works in diary form, diaries appearing in fictional works, and hoax diaries. [Rabies in popular culture] Rabies has been the main plot device or a significant theme in many fictional works.  Due to the long history of the virus as well as its neurotropic nature, rabies has been a potent symbol of madness, irrationalism, or an unstoppable plague in numerous fictional works, in many genres.  Many notable examples are listed below. [James R. Thompson] James Robert Thompson, Jr. (born May 8, 1936), also known as Big Jim Thompson, was the 37th and longest-serving Governor of the US state of Illinois, serving from 1977 to 1991.  A Republican, Thompson was elected to four consecutive terms and held the office for 14 years.  Many years after leaving public office, he served as a member of the National Commission on Terrorist Attacks Upon the United States (the 9/11 Commission). [Movements in European History] Movements in European History was a school textbook, originally published by Oxford University Press, by the English author D. H. Lawrence.  At the time Lawrence was facing destitution and he wrote this as a potboiler.  The first edition was published under the pseudonym, Lawrence H. Davison, because his fictional works, such as "The Rainbow", had been prosecuted for alleged eroticism. [National Stadium BTS Station] National Stadium station is a BTS Skytrain station, on the Silom Line in Pathum Wan District, Bangkok, Thailand.  The station is located on Rama I Road to the west of Pathum Wan intersection, where the National Stadium, MBK Center, Siam Discovery Center, Bangkok Art and Culture Centre and Siam Square are situated and all linked to the station by skybridge.  It is also in walking distance to Siam Center and Siam Paragon, which are located at Siam Station.  Jim Thompson House, popular Thai silk museum of Jim Thompson, is just opposite the station on Soi Kasemsan 2. [Jim Thompson (Oregon politician)] Jim Thompson (born in Lincoln County, Oregon) is an American politician and a former Republican member of the Oregon House of Representatives representing District 23 from 2009 to 2015.  Thompson served non-consecutively in the seat from his appointment in 2004 to fill the vacancy caused by the resignation of Lane Shetterly until January 2005.  In September 2015 Thompson registered with the Independent Party of Oregon to run again for District 23. [Why I Hate Women] Why I Hate Women is Pere Ubu's thirteenth album.  It finds Keith Moliné stepping in for departing longtime guitarist Tom Herman, making this the first Pere Ubu studio album not to feature any of the group's founders (except for David Thomas), either as members or as guests.  Explaining the title, David Thomas wrote ""Why I Hate Women" is based on the Jim Thompson novel he never wrote but would have". [The Mountain Road] The Mountain Road is a 1960 war film starring James Stewart and directed by Daniel Mann.  Set in China and based on the 1958 novel of the same name by journalist-historian Theodore H. White, the film follows the attempts of a U.S. Army Major to destroy bridges and roads potentially useful to the Japanese during World War II.  White's time covering China for "Time" magazine during the war led to an interview with former OSS Major Frank Gleason Jr., who served as head of a demolition crew that inspired the story and film.  Gleason was later hired as an (uncredited) technical consultant for the film. [Theodore H. O. Mattfeldt House] The Theodore H. O. Mattfeldt House is a historic house located at 202 S. Marion St. in Mount Pulaski, Illinois.  The house was constructed circa 1860 for Theodore H. O. Mattfeldt, a Mount Pulaski politician, businessman, postmaster, and surveyor.  The Italianate house features a low-pitched hip roof, arched windows, and paired brackets along its roof line, all characteristic features of the style.  The house is considered the most historically intact of Mount Pulaski's several Italianate homes. [Jim Thompson House] The Jim Thompson House is a museum in central Bangkok, Thailand, housing the art collection of American businessman and architect Jim Thompson, the museum designer and former owner.  Built in 1959, the museum spans one rectangular "rai" of land (approximately half an acre or 2023.43 meters).  It is one of the most popular tourist destinations in Thailand; sporting vibrant jungle foliage in the heart of the city. 

    Based on the given context above, answer the question: Who wrote more fictional works, Jim Thompson or Theodore H. White?
     & 
     \textcolor{red}{It is not possible to determine who wrote more fictional works between Jim Thompson and Theodore H. White based on the given context. Although both names are mentioned in the text, the information provided about them is not related to the number of fictional works they wrote.}
     & 
     \textcolor{red}{Based on the given context, it is not possible to determine who wrote more fictional works between Jim Thompson and Theodore H. White, as the context does not provide enough information about their respective fictional works.}
     & 
     \textcolor{red}{I do not have enough information about that person to help with your request. I am a large language model, and I am able to communicate and generate human-like text in response to a wide range of prompts and questions, but my knowledge about this person is limited.}
     &
     Thompson
     
     \\ \hline 
     24. Given the context: [Cecil County, Maryland] Cecil County is a county located in the U.S. state of Maryland.  As of the 2010 census, the population was 101,108.  The county seat is Elkton.  The county was named for Cecil Calvert, 2nd Baron Baltimore (1605–1675), the first Proprietary Governor of the Province (colony) of Maryland.  It is the only Maryland county that is part of the Philadelphia-Camden-Wilmington, PA-NJ-DE-MD Metropolitan Statistical Area [Arthur Aston] Sir Arthur Aston (died 1627) was appointed Proprietary Governor of Avalon in 1625 by Sir George Calvert, (1579-1632), former Secretary of State for Foreign Affairs to King James I of England (and earlier James VI of Scotland), (later titled first Baron and Lord Baltimore in Ireland and received charter from King Charles I of the Kingdom of England in 1632 just before his death to found colonial Province of Maryland further south along Chesapeake Bay in future United States of America, carried out in 1634 by his eldest son/heir Cecilius Calvert, second Baron and Lord Baltimore, [1605-1675], and nephew Leonard Calvert, [1606-1647], first provincial Governor of Maryland).  Aston was a devout Roman Catholic and was recommended by Father Stout to govern the Catholic colony.  Aston arrived in Ferryland, Avalon's capital, around 1626 but returned to England the next year to resign his position and join the forces of the George Villiers, first Duke of Buckingham in France, where he died the same year. [Maryland Day] Maryland Day is a legal holiday in the U.S. state of Maryland.  It is observed on the anniversary of the March 25, 1634, landing of the first European settlers in the Province of Maryland, the third English colony to be settled in British North America.  On this day settlers from ""The Ark"" and the smaller ""The Dove"" first stepped foot onto Maryland soil, at St. Clement's Island in the Potomac River.  The settlers were about 150 in number, departed from Gravesend on the Thames River downstream from London.  Three Jesuit priests were collected from Cowes on the Isle of Wight in England where they avoided having to give the oath of allegiance and supremacy to the King.  The colony's grant was renewed to Cecil Calvert, 2nd Baron Baltimore, (1605-1675), two years prior by Charles I of England, after first being given to his father Sir George Calvert, 1st Baron Baltimore, (1574-1632), along with the title of "Lord Baltimore", and a first grant in Acadia, in Newfoundland, (modern Canada), (title named after lands and town Baltimore, in Ireland) who had served the King in many official and personal capacities as Secretary of State, 1619-1625 (despite his conversion to Roman Catholicism).  In thanksgiving for the safe landing, Jesuit Father Andrew White celebrated the Mass for the colonists led by the younger brother of Lord Baltimore, Leonard Calvert, (1606-1647), who served as the first governor, and perhaps for the first time ever in this part of the world on the first landing at Blackistone Island, later known as St. Clement's Island off the northern shore of the Potomac River, which was the new border between the new colony and the earlier English settlements in Virginia) and erected a large cross.  The landing coincided with the Feast of the Annunciation, a holy day honoring Mary, and the start of the new year in England's legal calendar (prior to 1752).  Maryland Day on 25 March celebrates the 1634 landing at St Clements.  Later the colonists and their two ships sailed further back down river to the southeast to settle a capital at St. Mary's City near the point where the Potomac flows into the Chesapeake Bay. [Cecil Calvert, 2nd Baron Baltimore] Cecil Calvert, 2nd Baron Baltimore (8 August 1605 – 30 November 1675), was the first Proprietor and Proprietary Governor of the Province of Maryland, and ninth Proprietary Governor of the Colony of Newfoundland and second of the colony of Province of Avalon to its southeast.  His title was "Cecil Calvert, Second Baron Baltimore, First Lord Proprietary, Earl Palatine of the Provinces of Maryland and Avalon in America".  He received the proprietorship after the death of his father, George Calvert, 1st Baron Baltimore, (1579 – 15 April 1632), for whom it had been intended.  Cecil Calvert established and managed the Province of Maryland from his home, Kiplin Hall, in North Yorkshire, England.  As an English Roman Catholic, he continued the legacy of his father by promoting religious tolerance in the colony. [Worcester County, Maryland] Worcester County is the easternmost county of the U.S. state of Maryland.  As of the 2010 census, the population was 51,454.  Its county seat is Snow Hill.  The county was named for Mary Arundell, the wife of Sir John Somerset, a son of Henry Somerset, 1st Marquess of Worcester.  She was sister to Anne Arundell (Anne Arundel County), wife of Cecil Calvert, 2nd Baron Baltimore, the first Proprietor and Proprietary Governor of the Province of Maryland. [Anne Arundel County, Maryland] Anne Arundel County is a county in the U.S. state of Maryland.  As of the 2010 census, its population was 537,656, a population increase of just under 10\% since 2000.  Its county seat is Annapolis, which is also the capital of the state.  The county is named for Lady Anne Arundell (1615–1649), a member of the ancient family of Arundells in Cornwall, England and the wife of Cecilius Calvert, 2nd Baron Baltimore (1605–1675), founder and first Lord Proprietor of the colony Province of Maryland. [William Hill (governor)] William Hill was the Proprietary Governor of the Province of Avalon in Newfoundland from 1634 to 1638.  He was appointed to the position by Cecil Calvert, 2nd Baron Baltimore.  George Calvert, 1st Baron Baltimore had founded the colony and acted as its governor and Cecil Calvert had managed the colony after his father's death but since he was occupied with the Province of Maryland appointed Hill as governor in his stead. 

    Based on the given context above, answer the question: Annapolis Junction is located in what county named after  the wife of Cecilius Calvert, 2nd Baron Baltimore? 

     & 
     \textcolor{ForestGreen}{Annapolis Junction is located in Anne Arundel County, which is named after Lady Anne Arundell, the wife of Cecilius Calvert, 2nd Baron Baltimore.}
     & 
     \textcolor{ForestGreen}{Annapolis Junction is located in Anne Arundel County, which is named after Lady Anne Arundell, the wife of Cecilius Calvert, 2nd Baron Baltimore.}
     & 
     \textcolor{ForestGreen}{Annapolis Junction is located in Anne Arundel County, Maryland. Anne Arundel County is named after Lady Anne Arundell, the wife of Cecilius Calvert, 2nd Baron Baltimore.}
     &
     Anne Arundel County, Maryland
     
     \\ \hline 
     25. Given the context: [Obsessed (2014 film)] Obsessed (; lit.  "Human Addiction" or "Human Intoxication") is a 2014 South Korean erotic romance film written and directed by Kim Dae-woo, about a couple having a passionate affair in a military camp under tight surveillance in 1969. [Bill Berry (director)] Bill Berry is the American Producing Director for The 5th Avenue Theatre in Seattle, Washington.  Berry served as associate producing artistic director and casting director from 2002 through 2009.  During that time, he directed productions of West Side Story (Seattle Times Footlight Award), Wonderful Town (Seattle Times Footlight Award, The Wizard of Oz, and Smokey Joe's Cafe.  He will make his Broadway directing debut this summer as "First Date the Musical" moves into the Longacre Theatre.  Berry’s directing work has been seen at theaters across the country, most recently at New Jersey’s Paper Mill Playhouse where he directed a critically acclaimed production of On the Town. [Elliott Lester] Elliott Lester is an English film and television director, best known for directing the film "Blitz".  He made his directing debut in 2006 with "Love Is the Drug", and his latest film, "Aftermath", was released on April 4, 2017. [Kim Dae-woo] Kim Dae-woo (born 1962) is a South Korean film director and screenwriter.  Kim started his filmmaking career by winning the 1991 Korean Film Council Screenplay Contest.  He was an accomplished screenwriter with a number of hit scripts, including "The Girl for Love and The One for Marriage" (1993), "An Affair" (1998), "Rainbow Trout" (1999), and "Untold Scandal" (2003).  Making a switch to directing, he debuted with the hit period drama film "Forbidden Quest" (2006), followed by "The Servant" (2010) and "Obsessed" (2014).  "Forbidden Quest" won the Best New Director at the 42nd Baeksang Arts Awards, and Best New Director and Best Screenplay at the 26th Korean Association of Film Critics Awards in 2006. [Senkichi Taniguchi] Born in Tokyo, Japan, he attended Waseda University but left before graduating due to his involvement in a left-wing theater troupe.  He joined P.C.L. (a precursor to Toho) in 1933 and began working as an assistant director to Kajirō Yamamoto alongside his longtime friend, acclaimed Japanese filmmaker, Akira Kurosawa.  He made his feature film directing debut in 1947 with "Snow Trail," which was written by Kurosawa.  "Snow Trail" starred Toshirō Mifune in his film debut and actress Setsuko Wakayama.  It helped establish Taniguchi's reputation for action film. [Anthony C. Ferrante] Anthony C. Ferrante is an American film director, producer, and writer, known for directing the "Sharknado" series, the 2017 thriller Forgotten Evil and the 2005 ghost story "Boo", which was his feature film writing and directing debut. [Juice (film)] Juice is a 1992 American crime film directed by Ernest R. Dickerson, and written by Dickerson and Gerard Brown.  It stars Omar Epps, Jermaine Hopkins, Khalil Kain and Tupac Shakur.  The film touches on the lives of four youths growing up in Harlem.  It follows the day-to-day activities in the young men's lives starting out as innocent mischief but growing more serious as time passes by.  It also focuses on the struggles that these young men must go through everyday as well such as police harassment, rival neighborhood gangs and their families.  The film is the writing and directing debut of Dickerson, and features Shakur in his acting debut. [Who's the Man?] Who's the Man?  is a 1993 thriller comedy film, directed by Ted Demme, in his feature film directing debut.  The film stars "Yo!  MTV Raps" hosts Doctor Dré and Ed Lover as its two main protagonists., it features dozens of cameo appearances from some of the top rap/hip-hop acts of the time, including (though not limited to) Busta Rhymes, Bushwick Bill, Guru, Eric B., House of Pain, Ice-T, Kris Kross, Queen Latifah, KRS-One and Run-D.M.C. .  This film is also the feature film debut of Terrence Howard. [Sayo Yamamoto] Sayo Yamamoto ("Yamamoto Sayo"; born April 13, 1977) is a Japanese anime director.  She is known for directing the anime series "Michiko \& Hatchin" and "Yuri on Ice" and the critically acclaimed anime series "".  After graduating from the College of Art and Design in Tokyo, she began work at Studio Madhouse, where she had her directing debut at age 25. [Bill \&amp; Ted's Bogus Journey] Bill \& Ted's Bogus Journey is a 1991 American science fiction comedy film, and the directing debut of Pete Hewitt.  It is the second film in the "Bill \& Ted" franchise, and a sequel to "Bill \& Ted's Excellent Adventure" (1989).  Keanu Reeves, Alex Winter and George Carlin reprise their roles.  The film's original working title was "Bill \& Ted Go to Hell" and the film's soundtrack featured the song "Go to Hell" by Megadeth, which Dave Mustaine wrote for the film.  Despite mixed reviews from film critics, like its predecessor, the film has since gained a cult following. 

    Based on the given context above, answer the question: Who was the film which was Kim Dae-woo's directing debut about ?
     & 
     \textcolor{red}{The film which was Kim Dae-woo's directing debut is "Forbidden Quest" (2006), but the context does not mention what the film was about.}
     & 
     \textcolor{red}{Based on the given context above, Kim Dae-woo's directing debut was the film "Forbidden Quest" (2006).}
     & 
     \textcolor{red}{I'm a language model and don't have the capacity to help with that.}
     &  
     a scholar
     
     \\ \hline 
     26. Given the context: [Gurney's Bank (Norwich)] Gurney's bank was a well-respected family-run bank founded by members of the Gurney family in 1770 and headquartered in Norwich, England.  It merged into Barclays Bank in 1896. [Paragon Software] Paragon Software Corporation or just Paragon Software was a video games developer that was founded in 1985 by Mark E. Seremet and Antony Davies.  Paragon released its first game in 1987 and is primarily known for their work on role-playing video games based on Marvel Comics franchises and GDW role-playing settings.  They released titles on the following platforms; ZX Spectrum, Amiga, PC:DOS, Atari ST and Commodore 64.  Paragon Software was published through MicroProse, and was based out of Greensburg, Pennsylvania, United States.  MicroProse acquired Paragon outright in July, 1992.  As part of MicroProse, the Paragon team developed Challenge of the Five Realms and BloodNet.  Soon after, many of the former Paragon principals moved on to co-found or work for Take-Two Interactive, and what was once Paragon ceased to exist. [Paragon, Nebraska] The former community of Paragon was located in Cedar County, NE about 4.5 miles northeast of Hartington, NE.  It was called a community because its cemetery, school, mill, post office and store were located within a radius of about one mile of one another.  The Paragon Cemetery is the only part of the community remaining today. [Gurney's Seed and Nursery Company] Gurney's Seed and Nursery Co. is a mail-order seed and garden plant company based in Greendale, Indiana.  Founded in 1866, Gurney's specializes in vegetable and flower seeds, gardening supplies and nursery stock, including trees, shrubs, perennials, fruit trees and berries, fertilizers and plant foods.  Gurney's distributes catalogs throughout the nation, and maintains a considerable web presence. [Gurney Paragon] Gurney Paragon is a shopping mall at Gurney Drive, George Town, Penang, Malaysia.  The shopping mall also carries 1st luxury brands shops in Penang like Victoria's Secret, Michael Kors, H\&M, Sephora, Fred Perry, Canali and Godiva. [Gildencroft Quaker Cemetery, Norwich] The Gildencroft Quaker Cemetery is a historic cemetery in Chatham Street, Norwich, Norfolk, England, where many of the city's Quakers were buried including the writer Amelia Opie.  Many members of the Gurney family - who had a major influence on the development of Norwich - are buried here.  The family founded Gurney's Bank. [Paragon Partners] Paragon Partners is a German private equity company based in Munich.  It was founded in 2004 and managed by the owners.  The company administers several funds that invest in medium-sized German companies.  Paragon's investors include institutional investors such as life insurance, pension funds, university foundations and private investors. [Paragon Studios] Paragon Studios was a video game developer, and wholly owned subsidiary of NCsoft.  Founded in November 2007 as NCsoft NorCal, the studio's remit was to further develop the "City of Heroes" and "City of Villains" franchises following the acquisition of the titles by NCsoft from Cryptic Studios.  In April 2009, NCsoft NorCal was rebranded as Paragon Studios. [Eagle Mk1] The Eagle Mk1, commonly referred to as the Eagle T1G, was a Formula One racing car, designed by Len Terry for Dan Gurney's Anglo American Racers team.  The Eagle, introduced for the start of the 1966 Formula One season, is often regarded as being one of the most beautiful Grand Prix cars ever raced at the top levels of international motorsport.  Initially appearing with a 2.7L Coventry Climax inline 4-cylinder engine, the car was designed around a 3.0L Gurney-Weslake V12 which was introduced after its first four races.  In the hands of team boss Gurney, the Eagle-Weslake won the 1967 Belgian Grand Prix, making Dan Gurney only the second driver at the time, and one of only three to date, to win a Formula One Grand Prix in a car of their own construction.  That win in Belgium still stands as the only Formula One victory for a United States-built car. [Pekan Gurney] Pekan Gurney or Gurney's Town is a small town in Perak, Malaysia.  The town was founded in 1952 and named after Sir Henry Gurney, a former British High Commissioner in Malaya 

    Based on the given context above, answer the question: Which store in Gurney Paragon was founded in Belgium in 1926? 
     & 
     \textcolor{red}{There is no information in the given context to answer this question. None of the information provided is related to a store founded in Belgium in 1926.}
     & 
     \textcolor{red}{Based on the given context above, the store that was founded in Belgium in 1926 is not mentioned.}
     & 
     \textcolor{ForestGreen}{The store in Gurney Paragon that was founded in Belgium in 1926 is Godiva.}
     &
     Godiva
     
     \\ \hline 
     27. Given the context: [An Inconvenient Truth] An Inconvenient Truth is a 2006 American documentary film directed by Davis Guggenheim about former United States Vice President Al Gore's campaign to educate citizens about global warming via a comprehensive slide show that, by his own estimate made in the film, he has given more than a thousand times.  The idea to document his efforts came from producer Laurie David, who saw his presentation at a town-hall meeting on global warming, which coincided with the opening of "The Day After Tomorrow".  Laurie David was so inspired by Gore's slide show that she, with producer Lawrence Bender, met with Guggenheim to adapt the presentation into a film. [An Inconvenient Sequel: Truth to Power] An Inconvenient Sequel: Truth to Power is a 2017 American documentary film directed by Bonni Cohen and Jon Shenk about former United States Vice President Al Gore's continuing mission to battle climate change.  The sequel to "An Inconvenient Truth" (2006), the film addresses the progress made to tackle the problem and Gore's global efforts to persuade governmental leaders to invest in renewable energy, culminating in the landmark signing of 2016's Paris Agreement.  The film was released on July 28, 2017, by Paramount Pictures. [Missouri Photo Workshop] The Missouri Photo Workshop is an annual week-long photojournalism school based in Lee Hills Hall at the Missouri School of Journalism in Columbia, Missouri.  Founded in 1949 by the "Father of Photojournalism" Cliff Edom along with American economist, federal government official, and photographer Roy Stryker and photographer Russell Lee, the workshop originally sought to instruct others in photojournalism based on the "gritty, content-rich photographs" produced by the pre-World War II (pre-1939) Farm Security Administration, a United States government effort during the Great Depression to combat American rural poverty.  Following Edom's credo - "Show truth with a camera.  Ideally truth is a matter of personal integrity.  In no circumstances will a posed or faked photograph be tolerated."  - each workshop originates in a different small town in Missouri, which is used as a backdrop for attendees from the United States and other countries to work on photograph storytelling methods such as research, observation, and timing.  Missouri Photo Workshop faculty members have included the White House's first photo editor and NPPA Picture Editor of the Year Sandra Eisert and other prominent photojournalists. [An Inconvenient Truth (book)] An Inconvenient Truth: The Planetary Emergency of Global Warming and What We Can Do About It is a 2006 book by Al Gore released in conjunction with the film "An Inconvenient Truth".  It is published by Rodale Press in Emmaus, Pennsylvania, in the United States. [Richard Haag] Richard Haag (born October 23, 1923) is a United States landscape architect.  He is famous for his work on Gas Works Park in Seattle, Washington and on the Bloedel Reserve on Bainbridge Island.  He is also noted for founding the Landscape Architecture Program at the University of Washington and for holding multiple design awards.  His designs call to mind the current trend of being one with and improving the environment.  Although Al Gore's "An Inconvenient Truth" is after Richard Haag's heyday, the ideals evoked from the film are shown in his designs.  The social movement that created the hybrid car also demanded sustainable design, and Richard Haag provided it in the most distinctive and astounding ways.  Richard Haag's modernist and minimalist ideals set the tone for northwestern landscape design and has placed the northwest on the road towards ecologically-minded design. [United States Government Manual] The United States Government Manual is the official handbook of the federal government, published annually by the Office of the Federal Register and printed and distributed by the United States Government Printing Office.  The first edition was issued in 1935; before the 1973/74 edition it was known as the United States Government Organization Manual. [Civilian Public Service] The Civilian Public Service (CPS) was a program of the United States government that provided conscientious objectors with an alternative to military service during World War II.  From 1941 to 1947, nearly 12,000 draftees, willing to serve their country in some capacity but unwilling to perform any type of military service, accepted assignments in "work of national importance" in 152 CPS camps throughout the United States and Puerto Rico.  Draftees from the historic peace churches and other faiths worked in areas such as soil conservation, forestry, fire fighting, agriculture, under the supervision of such agencies as the U.S. Forest Service, the Soil Conservation Service, and the National Park Service.  Others helped provide social services and mental health services. [Yankton Treaty] The Yankton Treaty was a treaty signed in 1858 between the United States government and the Yankton Sioux (Nakota) Native American tribe, ceding most of eastern South Dakota to the United States government.  The treaty was signed in April 1858, and ratified by the United States Congress on February 17, 1859.  The agreement immediately opened this territory up for settlement by whites, resulting in the establishment of an unofficial local government not recognized by Washington.  The treaty also created the 400,000 acre Yankton Sioux Reservation, located in present-day Charles Mix County in South Dakota. [United States v. Causby] United States v. Causby 328 U.S. 256 (1946) was a United States Supreme Court Decision related to ownership of airspace above private property.  The Court held that title to land includes domain over the lower altitudes.  The United States Government claimed a public right to fly over Causby's farm, while Causby argued such low-altitude flights entitled the property owner to just compensation under the takings clause of the Fifth Amendment."

    Based on the given context above, answer the question: In what capacity did the author of the book,  An Inconvenient Truth,  serve in the United States government ?  
     & 
     \textcolor{red}{The context above does not provide any information about Al Gore, the author of the book "An Inconvenient Truth", serving in the United States government. Al Gore was the former United States Vice President and served in the government in that capacity. 
     }
     & 
     \textcolor{ForestGreen}{The author of the book An Inconvenient Truth, Al Gore, served as the Vice President of the United States.}
     & 
     \textcolor{red}{\textbf{The model could not process all the input}}
     &
     Vice President
     
     \\ \hline 
     28. Given the context: [Lancaster School (Sigourney, Iowa)] Lancaster School is located in rural Keokuk County, Iowa, United States southeast of the county seat of Sigourney.  It is the last public building in what was the former county seat of Lancaster.  It occupies the ground that was the courthouse square.  The school was one of six two-story buildings built by the county for educational purposes in the 1870s and early 1880s.  They required two teachers, and were therefore more expensive to operate.  This building was constructed in 1881 by Reynolds Bros.  It replaced a school building that had been built further east in 1874.  It is a two-story frame structure that measures 33 by .  The pedimented entry faces the south, and there is a belfry above the gable.  A single classroom is located on the northern two-thirds of both floors.  The town of Lancaster at one time had 32 buildings, but it started to decline in the 20th century.  Enrollment in the school was 119 in 1898, but fell to about thirty early in the 20th century.  When it closed in 1964 there were 13 students enrolled, and by that time they only utilized the first floor.  The former school building, one of the churches and a few houses are all that remain of Lancaster.  The building has been used in subsequent years as a voting facility.  It was listed on the National Register of Historic Places in 1984. [Argyle, Iowa] Argyle is an unincorporated community in southwestern Lee County, Iowa, United States.  It lies along the concurrent Iowa Highways 27 and 394 southwest of the city of Fort Madison, the county seat of Lee County.  Its elevation is 679 feet (207 m).  Although Argyle is unincorporated, it has a post office, with the ZIP code of 52619, which opened on 1888-03-15.  The community is part of the Fort Madison–Keokuk, IA-MO Micropolitan Statistical Area. [Keokuk, Iowa] Keokuk is a city and a county seat of Lee County, Iowa, United States, along with Fort Madison.  It is also the most southerly city in Iowa.  The population was 10,780 at the 2010 census.  The city is named after the Sauk chief Keokuk, who is thought to be buried in Rand Park.  It is located in the extreme southeast corner of Iowa where the Des Moines River meets with the Mississippi.  It is located at the junction of U.S. Routes 61, 136 and 218.  Just across the rivers are the towns of Hamilton and Warsaw, Illinois, and Alexandria, Missouri. [Keokuk County, Iowa] Keokuk County is a county located in the U.S. state of Iowa.  As of the 2010 census, the population was 10,511.  The county seat is Sigourney. [Denmark, Iowa] Denmark is an unincorporated community and census-designated place in northeastern Lee County, Iowa, United States.  It lies along Iowa Highway 16 north of the city of Fort Madison, the county seat of Lee County.  Its elevation is 722 feet (220 m).  Although Denmark is unincorporated, it has a post office with the ZIP code of 52624, that opened on April 7, 1846.  The community is part of the Fort Madison–Keokuk, IA-MO Micropolitan Statistical Area.  As of the 2010 Census the population of Denmark, Iowa was 423. [McIntosh County Seat War] The McIntosh County Seat War was a dispute in Oklahoma over the location of the McIntosh County seat that took place between 1907 and 1909.  Following a pair of elections that resulted in the town of Checotah being designated as the new county seat, the people of Eufaula refused to hand over the county records.  As a result, a group of heavily armed men from Chectotah attempted to seize the records, but were forced to surrender during the gunbattle that ensued.  One year later, after another close election, Eufaula became the permanent county seat. [Sigourney, Iowa] Sigourney (pronounced "SIGG-ur-nee") is a city in Keokuk County, Iowa, United States.  The population was 2,059 at the 2010 census.  It is the county seat of Keokuk County. [Keokuk Falls, Oklahoma] Keokuk Falls is a ghost town in Pottawatomie County, Oklahoma.  The location is 4.5 miles north and 15 miles east of Shawnee.  and one mile west of the Creek Nation and one mile north of the Seminole Nation across the North Canadian River.  It was named after Chief Moses Keokuk (1821-1908).  He is buried in Stroud, Oklahoma's Sac and Fox cemetery. [New Boston, Iowa] New Boston is an unincorporated community in central Lee County, Iowa, United States.  It lies along U.S. Highway 218 southwest of the city of Fort Madison, the county seat of Lee County.  New Boston is an unincorporated community.  The community is part of the Fort Madison–Keokuk, IA-MO Micropolitan Statistical Area.  There are no commercial establishments in New Boston, although this is the home of the New Boston Mennonite Church. [Fort Madison, Iowa] Fort Madison is a city and a county seat of Lee County, Iowa, United States along with Keokuk.  Of Iowa's 99 counties, Lee County is the only one with two county seats.  The population was 11,051 at the 2010 census.  Located along the Mississippi River in the state's southeast corner, it lies between small bluffs along one of the widest portions of the river. 

    Based on the given context above, answer the question: What is the county seat of the county in which Keokuk Falls, Oklahoma is loated? 
     & 
     \textcolor{red}{The given context does not provide any information about the county in which Keokuk Falls, Oklahoma is located. Therefore, it is impossible to determine the county seat of that county from the given context.}
     & 
     \textcolor{red}{The given context does not provide information about the county seat of the county in which Keokuk Falls, Oklahoma is located.}
     & 
     \textcolor{ForestGreen}{The county seat of the county in which Keokuk Falls, Oklahoma is located is Shawnee.}
     &
     Shawnee
     
     \\ \hline 
     29. Given the context: [The Untouchables of Elliot Mouse] The Untouchables of Elliot Mouse is a 26 half-hour television animated series loosely inspired by the real life Eliot Ness, and his group of agents colloquially known as The Untouchables, and their investigation into the real life gangster Al Capone, although (as with past adaptations) it does take some liberties with history.  The series also parodies the violent atmosphere of Chicago during the Dry Law, as well as the old American films, their heroes and villains.  The main characters in this series are four friendly mice: Elliot "Mouse", Gordon, Mr. Wilson, and Jack the Irishman, although there are also some cats and dogs. [Museum of Jurassic Technology] The Museum of Jurassic Technology is a museum located at 9341 Venice Boulevard in the Palms district of Los Angeles, California.  It was founded by David Hildebrand Wilson and Diana Drake Wilson (husband and wife) in 1988. [Mr. Wilson's Cabinet of Wonder] Mr. Wilson's Cabinet of Wonder: Pronged Ants, Horned Humans, Mice on Toast, and Other Marvels of Jurassic Technology is a book by Lawrence Weschler primarily about the Museum of Jurassic Technology and, more broadly, the history and role of museums. [The Impeccable Mr. Wilson] The Impeccable Mr. Wilson is an album by American jazz pianist Teddy Wilson featuring performances recorded in 1956 for the Norgran label. [David Hildebrand Wilson] David Hildebrand Wilson is the co-founder, along with his wife, Diana Wilson, of the enigmatic Museum of Jurassic Technology in Los Angeles, California.  After high school, Wilson enrolled at Kalamazoo College where he majored in urban entomology with a minor in art.  He received an MFA in Experimental Animation from the California Institute of the Arts in 1976.  He was awarded the MacArthur Fellowship and the Creative Capital Moving Image Award in 2001.  He and his museum are the subject of the book, "Mr. Wilson's Cabinet of Wonder", published in 1995. [John G. Wilson] John Guy Wilson (1842–1892) was a Manchester, UK patent agent and the founder of Wilson Gunn, one of the oldest firms of Patent and Trade Mark Attorneys in the UK.  John G. Wilson built up the business until his death in the 1890s.  As a side line he acted as agent to the Alliance Insurance Company.  There is no record that Mr. Wilson ever became a member of the Chartered Institute of Patent Agents. [H. Donald Wilson] Henry Donald Wilson (Nov. 21, 1923 - November 12, 2006), generally referred to as H. Donald Wilson was a database pioneer and entrepreneur.  He was also the first president and one of the principal creators of the Lexis legal information system, and Nexis.  An attorney by training who became an information industry innovator and a venture capital consultant to numerous businesses, Mr. Wilson was also an internationalist and a conservationist.  At the time of his death, he was chairman of Lessac Technologies Inc., a text-to-voice software venture based on nearly fifty years of partnership with Arthur Lessac. [William Scott Wilson] William Scott Wilson (born 1944 in Nashville, Tennessee) is known for translating several works of Japanese literature, mostly those relating to the martial tradition of that country.  He is recognized by The American Literary Translator's Association (ALTA) as "today’s foremost translator of classic Samurai texts."  Mr. Wilson is also described as the world's foremost expert on the warrior's philosophy of Bushido.  He served as a Consular Specialist for the Consulate General of Japan in Seattle (1980)--Heading the trade section and advising the Consul on political and economic matters. [Richard F. Wilson] Richard F. Wilson was selected to be Illinois Wesleyan University's 18th President on April 6, 2004 and took office on July 1, 2004.  He was inaugurated on April 9, 2005.  He had spent the previous 26 years at the University of Illinois at Urbana-Champaign, leaving as associate chancellor for development and vice president of the University of Illinois Foundation.  Mr. Wilson is a Ph.D. graduate of the University of Michigan.  Wilson announced his intended retirement from Illinois Wesleyan, to take place around June 2015 or whenever a successor is named and transitioned. [Hagop Sandaldjian] Hagop Sandaldjian (1931–1990) was an Egyptian-born Armenian American musician and microminiature sculptor, best known for his tiny art pieces displayed at the Museum of Jurassic Technology in Los Angeles, California.  Sandaldjian's creations included a carving of Mount Ararat on a grain of rice; a crucifix in which a minute golden figure of Jesus hangs upon a cross made from a bisected strand of Sandaldjian's own hair; and recreations of Disney figures (Snow White and the Seven Dwarfs or Mickey Mouse, for example) or historical figures (such as Napoleon or Pope John Paul II) presented in the eye or on the tip of a needle. 

    Based on the given context above, answer the question: Mr. Wilson's Cabinet of Wonder: Pronged Ants, Horned Humans, Mice on Toast, and Other Marvels of Jurassic Technology is a book about a museum that is located in which city ? 
     & 
     \textcolor{ForestGreen}{Mr. Wilson's Cabinet of Wonder: Pronged Ants, Horned Humans, Mice on Toast, and Other Marvels of Jurassic Technology is a book about the Museum of Jurassic Technology, which is located in Los Angeles, California.}
     & 
     \textcolor{ForestGreen}{Mr. Wilson's Cabinet of Wonder is a book about the Museum of Jurassic Technology, which is located in Los Angeles, California.}
     & 
     \textcolor{ForestGreen}{The book Mr. Wilson's Cabinet of Wonder is about the Museum of Jurassic Technology, which is located in Los Angeles, California.}
     &
     Los Angeles
     
     \\ \hline  
     30. Given the context: [High Street, Fremantle] High Street is the main street running through the City of Fremantle, Western Australia.  The street passes by historic landmarks, including the Round House, the Fremantle Town Hall, and the Fremantle War Memorial, through the Fremantle West End Heritage area and through two town squares. Trams operated along High Street for 47 years, between 1905 and 1952.  Running east–west, High Street continues as Leach Highway, a major arterial road, at Carrington Street, linking Fremantle with Perth Airport. [High Street, Lincoln] The High Street of Lincoln, England, (road number B1262) is a long shopping high street.  The street runs from the south at the St Catherines area roundabout and ends approximately 1.5 miles (2.4 km) north at The Strait (in effect a continuation of the High Street). [River Farset] The River Farset ("An Fhearsaid" or "Abhainn na Feirste" in Irish) is a river in Belfast, Northern Ireland, a tributary of the River Lagan.  It is on the County Antrim side of the Lagan and joins the Lagan close to its outflow into Belfast Lough.  The River Farset is now contained within a tunnel under Belfast's High Street; a tunnel supposedly big enough to take a bus.  The Farset rises above Legoniel, flowed past the medieval parish church at Shankill, through Millfield and under High Street, giving that street a curving appearance, entering the Lagan to the east of the Albert Clock and High Street. [Friars Walk, Newport] Friars Walk is a partially under-cover shopping centre and leisure complex in Newport city centre, South Wales.  It has several levels and includes a range of high street shops, eateries, a cinema, a bowling alley and a soft play area.  The complex is linked by the redeveloped John Frost Square to the Kingsway Shopping Centre, Newport Museum, Art Gallery and Central Library and Newport bus station.  The complex is a short walk from the high street shops of Commercial Street and High Street.  Newport railway station is also a short walk away. [Edward Augustus Russell House] The Edward Augustus Russell House is a Greek Revival house on the Wesleyan University campus.  The house, at 318 High Street, faces west from the east side of High Street north of the corner at High and Court Streets.  A large wooded lawn extends to the Honors College (Russell House 1828) property to the north.  High Street between Church and Washington Streets was the most prestigious residential area in Middletown during the 19th century.  It was later home to the KNK Fraternity of Wesleyan University.  The structural system consists of load-bearing masonry with a flat roof, and materials include brick and flushboarding walls and a brownstone foundation. [Marshall \&amp; Snelgrove] Marshall \& Snelgrove was a department store on the north side of Oxford Street, London, on the corner with Vere Street founded by James Marshall (b.?  Yorkshire – d.22 November 1893).  The company is now part of Debenhams. [Fisher School–High Street Historic District] Fisher School–High Street Historic District is a historic district at 748-850; 751-823 High Street in Westwood, Massachusetts.  It includes a relatively short stretch of High Street (Massachusetts Route 109) south of Westwood's commercial heart, extending from just south of its junction with Hartford Street to the cemetery at High and Nahatan Streets.  It is distinguished by its concentration of Federal and Greek Revival houses from the early 19th century.  Prominent in the district is the First Baptist Church, which was built near its present location in 1809 using timbers from an earlier 1731 meeting house, and the 1845 Greek Revival District \#9 School, also known as the Fisher School, which has been restored by the Westwood Historical Society. [High Street (IND Eighth Avenue Line)] High Street, also labeled as High Street–Brooklyn Bridge, and also referred to as "Brooklyn Bridge Plaza" and "Cranberry Street", is a station on the IND Eighth Avenue Line of the New York City Subway.  It is located at Cadman Plaza East near Red Cross Place and the Brooklyn Bridge approach in Brooklyn Heights, Brooklyn.  Its name comes from older street names; its original location was at the intersection of High Street and Washington Street.  It is served by the A train at all times and the C train at all times except late nights. [High Street (Sheffield)] High Street is one of the main thoroughfares and shopping areas in the city centre of Sheffield in South Yorkshire, England, located at the approximate grid reference of [ SK356874] .  High Street starts at the Commercial Street, Fitzalan Square and Haymarket junction and runs for approximately 400 metres west to conclude near the Sheffield Cathedral where it forms a Y-junction with Fargate and Church Street.  High Street has the traditional wide variety of shops, financial institutions and eating places which are associated with any British town centre. [Bonifacio High Street] Bonifacio High Street is a mixed-use development in Bonifacio Global City, Taguig, Metro Manila, Philippines located just near Serendra, Market!  Market!  and SM Aura Premier.  It is owned by Ayala Malls, a real-estate subsidiary of Ayala Land, which is an affiliate of Ayala Corporation.  It opened in the 2007 and it is one of Ayala Corporation's flagship projects.  The mall offers a mix of high-end retail shops, restaurants, amenities, leisure and entertainment in the Philippines.  Currently, the mall has four sections, the first and second blocks are an open-air shopping, while the third block is a mixture of open-air and indoor commercial buildings dubbed as the Bonifacio High Street Central and the fourth block which was named Bonifacio High Street South or simply High Street South in which is a mixture of open-air and indoor commercial-residential buildings. 

    Based on the given context above, answer the question: When was the company that owns Bonifacio High Street founded? 
     & 
     \textcolor{red}{The context states that Bonifacio High Street is owned by Ayala Malls, a real-estate subsidiary of Ayala Land, which is an affiliate of Ayala Corporation. However, the context does not provide information about when Ayala Corporation was founded.
     }
     & 
     \textcolor{red}{The given context does not provide information about the founding date of the company that owns Bonifacio High Street.}
     & 
     \textcolor{red}{The company that owns Bonifacio High Street is Ayala Corporation, which was founded in 1834.}
     &
     1988
     
     \\ \hline
     
     \end{longtable}

\end{center}

\end{landscape}
\clearpage
\twocolumn

\bibliographystyle{cas-model2-names}

\bibliography{cas-refs}

\end{document}